\documentclass[final,faculty=firw,department=elt,phddegree=elt]{adsphd}

\title{ Continual Learning in Neural Networks}

\author{Rahaf}{Aljundi}

\supervisor{Prof.~dr.~ir.~T.~Tuytelaars}{}
\president{Prof.~dr.~ir.~H.~Neuckermans}
\jury{Prof.~dr.~ir.~L.~Van~Gool\\
      Prof.~dr.~ir.~H.~Van~Hamme\\
      Prof.~dr.~ir.~R.~Vogels
}
\externaljurymember{Prof.~dr.~A.~Vedaldi}{University of Oxford}

\researchgroup{ESAT - PSI}
\website{http://homes.esat.kuleuven.be/~raljundi/} 
\email{rahaf.aljundi@esat.kuleuven.be} 

\address{Kasteelpark Arenberg 10 - box 2441}

\date{September 2019}
\copyyear{2019}

\usepackage{nomencl}   
\renewcommand{\nomname}{List of Symbols}
\newcommand{\myprintnomenclature}{%
  \cleardoublepage%
  \printnomenclature%
  \chaptermark{\nomname}
  \addcontentsline{toc}{chapter}{\nomname} 
}
\makenomenclature%

\usepackage{textcomp} 
\usepackage{pifont}   
\usepackage{booktabs}
\usepackage{amssymb,amsthm}
\usepackage{amsmath}
\usepackage{paralist}
\usepackage{times}
\usepackage{epsfig}
\usepackage{graphicx}
\usepackage{amsmath}
\usepackage{amssymb}
\usepackage{graphicx}
\usepackage{caption}
\usepackage{float}
\usepackage[caption = false]{subfig}
\usepackage{amsmath,amssymb} 
\usepackage{color}
\usepackage{times}
\usepackage{epsfig}
\usepackage{graphicx}
\usepackage{amsmath}
\usepackage{amssymb}
\usepackage{booktabs}       
\usepackage{standalone}
\usepackage{xspace}
\usepackage{multirow}
\usepackage{diagbox}
\usepackage{wrapfig}
\usepackage{algorithm}
\usepackage{algpseudocode}
\usepackage{titling}
\usepackage{caption}
\usepackage{wrapfig,blindtext}
\usepackage{capt-of}
\usepackage{tabularx}
\usepackage{times}
\usepackage{epsfig}
\usepackage{amsmath}
\usepackage{amssymb}
\usepackage{booktabs}
\usepackage{amsthm}
\usepackage{paralist}
\usepackage{algorithm}
\usepackage{algpseudocode}
\usepackage{booktabs}       
\usepackage{amsfonts}       
\usepackage{nicefrac}       
\usepackage{microtype}
\usepackage{bbm}
\usepackage{mathalfa}
\usepackage{mathtools}
\usepackage[inline]{enumitem}
\usepackage[linguistics]{forest}
\usepackage{bibentry}

\def\eqref#1{equation~\ref{#1}}

\def\1{\bm{1}}

\DeclareMathAlphabet{\mathsfit}{\encodingdefault}{\sfdefault}{m}{sl}
\SetMathAlphabet{\mathsfit}{bold}{\encodingdefault}{\sfdefault}{bx}{n}

\newcolumntype{L}{>{\raggedright\arraybackslash}X}
\def \hfillx {\hspace*{-\textwidth} \hfill}

\newcommand{\myparagraph}[1]{\noindent\textbf{#1}}
\def \hfillx {\hspace*{-\textwidth} \hfill}

 \makeatletter
 \DeclareRobustCommand\onedot{\futurelet\@let@token\@onedot}
 \def\@onedot{\ifx\@let@token.\else.\null\fi\xspace}
 \def\eg{e.g\onedot} 
 \def\ie{i.e\onedot}

 \def\etal{\textit{et~al\onedot}} 
  \def\Eqn{Equation} 
 
 \makeatother

\newtheorem{proposition}{Proposition}
 \makeatletter
 \DeclareRobustCommand\onedot{\futurelet\@let@token\@onedot}
 \def\@onedot{\ifx\@let@token.\else.\null\fi\xspace}
 \def\eg{e.g\onedot} 
 \def\ie{i.e\onedot}

 \def\etal{\textit{et~al\onedot}} 
  \def\Eqn{Equation} 
 
 \makeatother
\newcommand{\SNI}{{\tt{SNID}}\xspace}
\newcommand{\SNIwoNI}{{\tt{SNI}}} 
\newcommand{\SLNIwoNI}{{\tt{SLNI}}\xspace}
\newcommand{\SLNI}{{\tt{SLNID}}\xspace}

\newcommand{\SLNIDFull}{Sparse coding through Local Neural Inhibition and Discounting (\SLNI)\xspace}

\algblockdefx{MRepeat}{EndRepeat}{\textbf{repeat}}{}
\algnotext{EndRepeat}

\begin{document}

\makefrontcoverXII

\maketitleX

\frontmatter 

\includepreface{preface}
\includeabstract{abstract}
\includeabstractnl{abstractnl}

\includeabbreviations{abbreviations}

\nomenclature{$T_{t}$}{task number $t$}
\nomenclature{$\theta_{t}$}{ neural network parameters after learning  task $T_t$}
\nomenclature{$J$}{ objective function, minimized during training}
\nomenclature{$R(\theta)$}{parameters regularizer}
\nomenclature{$R(H)$}{representation regularizer}
\nomenclature{$\Omega$}{importance weights of  neural network parameters}
\nomenclature{$\lambda$}{ weight of an importance weights regularizer}
\nomenclature{$\mathcal{N}_i$}{neuron in a neural network.}
\nomenclature{$out_i$}{output of a neuron $\mathcal{N}_i$}
\nomenclature{$h_i$}{activation of  neuron $\mathcal{N}_i$}
\nomenclature{$\theta_{ij}$}{parameter connecting two neurons, $\mathcal{N}_i$ and $\mathcal{N}_j$, in two consecutive layers }
\nomenclature{$Q_{t}$}{data generating distribution of a task $T_t$}
\nomenclature{$D_{t}$}{dataset of a task $T_t$}
\nomenclature{$x_{n}$}{input sample }
\nomenclature{$y_{n}$}{output label }
\nomenclature{$N_{t}$}{number of training samples in a task $T_t$ }
\myprintnomenclature

\tableofcontents
\listoffigures
\listoftables


\mainmatter 

\instructionschapters\cleardoublepage
  \graphicspath{{\chaptersdir/introduction/image/}}%
\newcommand{\twodots}{\mathinner {\ldotp \ldotp}}
\chapter{Introduction}\label{ch:introduction}
  \begin{figure}[!hbtp]
  \vspace{-0.4cm}
    \center{\includegraphics[width=1\textwidth]{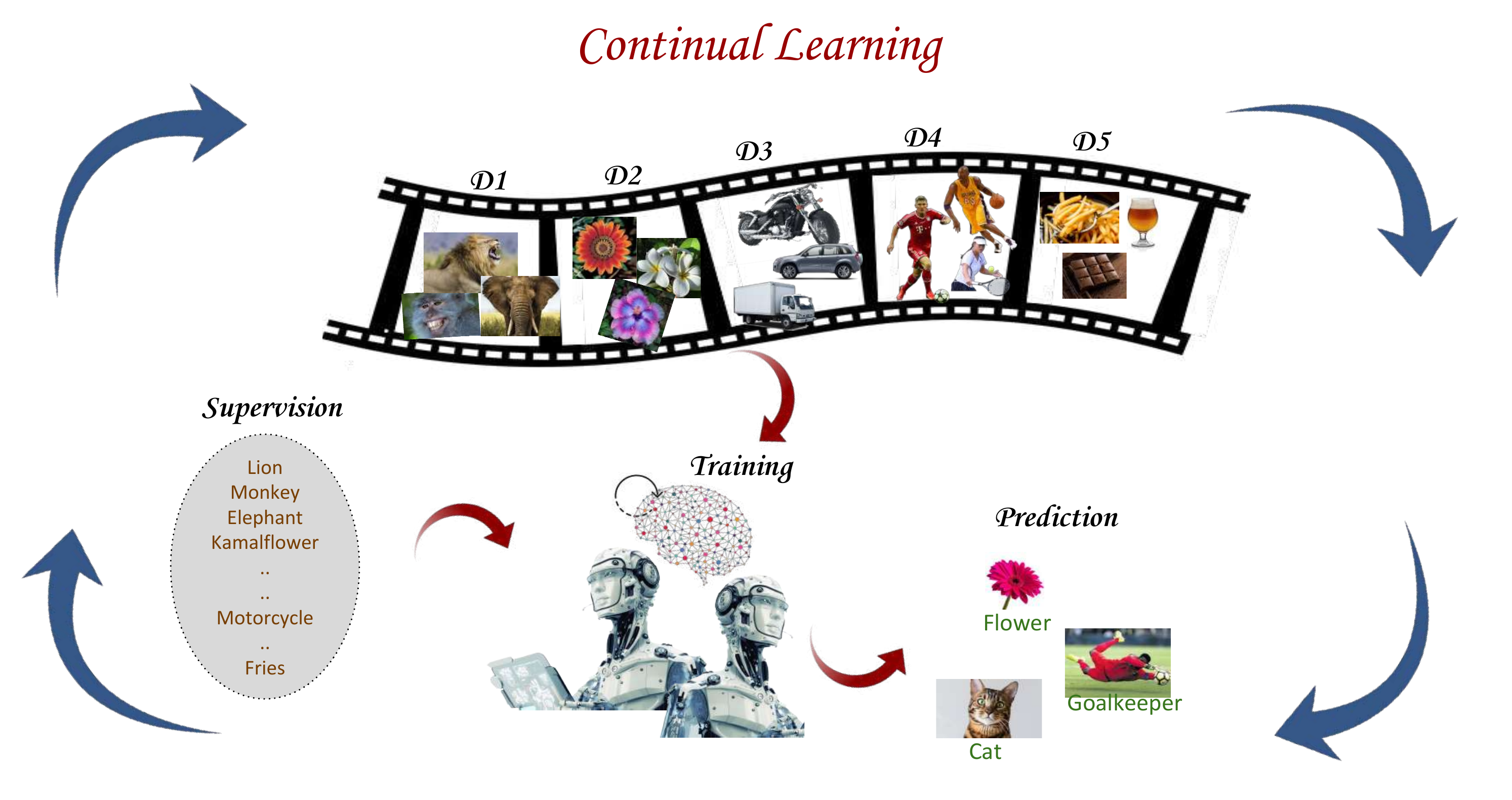}}
    \caption[An illustration of the continual machine learning cycle]{\label{fig:Continual_machine_learning}  An illustration of the continual machine learning cycle. Data are received sequentially with optional supervision. The agent alters between learning and predicting. Red arrows represent the inner cycle that occurs whenever new data arrive. }
    \vspace*{-0.2cm}
  \end{figure}
Our world is  complex, constantly changing and evolving 
and so  are our brains, continually forming new organismic states   to adapt and interact with this world. The evolution and self-organization depicted in living creatures 
 resemble the essence of the difference between   classical physics   and physiology~\cite{GrossbergStephen1982Soma}.  While classical physics studies describe a stationary world,  physiology concerns with evolutionary systems and their non-stationary world. Our artificial agents have to be deployed and behave in this same dynamic non-stationary world. Without mechanisms  allowing these agents to constantly adapt and exploit new information,  effective machine intelligence can't be realized. 
 
Current machine learning models represented by neural networks are able to learn  and even 
outperform human level  performance in individual tasks,   as  in Atari games~\cite{silver2018general} and object recognition
~\cite{russakovsky2015imagenet}. However, this learning process creates  static  neural models that are incapable of adapting and expanding their ``function''. Whenever new data are available, the training process of a neural network has to start all over again.  In a world like ours, such a practice becomes intractable when moving to real scenarios where  data are streaming, might be disappearing after a given period of time or even can't be stored at all due to storage constraints or privacy issues. Each day millions of images with new tags appear on social media. Every minute hundreds of hours of video are uploaded on Youtube. This new content contains new topics and trends that may be very different from what one has seen before - think e.g. of new emerging  topics, fashion trends, social media hypes or technical evolution. This makes it crucial for neural networks to be able to adapt and be updated over time.

The main obstacle  towards developing continually adapting systems is the ``catastrophic forgetting'' of old learned information once new knowledge is learned. Mccloskey and Ratcliff~\cite{mccloskey1989catastrophic,ratcliff1990connectionist} were the first to show  catastrophic forgetting   in neural networks, where the learning of new patterns of data results in a complete erase of the previously acquired knowledge. 
Catastrophic forgetting has also been attested in other machine learning models~\cite{hammer2016incremental,li2004heuristic}. However, the ability of neural networks to implicitly store the acquired knowledge in addition to its success and biological plausibility urge the need to study and understand  the catastrophic forgetting phenomenon. 

While  natural cognitive systems can gradually forget  old information, a complete loss of previous knowledge is rarely attested~\cite{french1999catastrophic}. 
 Humans tend to learn concepts sequentially one after another. Some concepts are revisited but this revisiting is unnecessary for a new concept to be conceived.  Given current artificial 
 neural networks,  learning cannot occur in this sequential manner due to the catastrophic forgetting of previous concepts  as new ones are learned. Typically, data of a given task are shuffled and performance largely increases with repeated revisiting over the training data.
 Since this shuffling and repeated re-visiting of all training data is clearly not the case for humans, that are able to learn and even exhibit better learning behavior when information is presented sequentially, French and Ferrara~\cite{french1999modeling} posed the question of whether this  immunity to catastrophic forgetting is present in other mammals. They showed that a learning of   two  time events sequentially in rats results in a complete wipe out of the first event once the second is learned. However, a concurrent learning of the same two events does show a forgetting but not the catastrophic forgetting shown in the sequential learning. 
 This is very similar to what would occur if a simple neural  network was used to model events presented sequentially.
It has been suggested  that overcoming of catastrophic forgetting  in higher mammals could be due to the development of a hippocampal–neocortical separation~\cite{mcclelland1995there,french1999modeling}. 

Catastrophic interference is a direct result of a more general problem in neural network,  the so-called ``stability–plasticity'' dilemma~\cite{GrossbergStephen1982Soma}. While plasticity refers to the ability of integrating new knowledge, stability indicates the preservation of previous knowledge  while new data is encoded. Hence, stability–plasticity is an essential building block in artificial and biological neural intelligent systems.
{\em The main challenge is how to build intelligent systems that are dynamic and sensitive to new information while at the same time are stable and immune to catastrophic interference with  previously acquired knowledge. Overcoming this challenge has been the driving goal of the work developed during the course of this PhD.}

\section{Continual Learning}
Continual learning, also referred to as lifelong learning, sequential learning or incremental learning, studies the problem of learning from an infinite stream of data stemmed from changing input domains and associated with different tasks, 
with the goal of using the acquired knowledge in problem solving and future learning~\cite{chen2018lifelong}. The main criterion is the sequential nature of the learning process where only a small portion of input data from one or few tasks is available at once. It is impossible to label all training examples from all tasks  before initiating the learning process and even if so, with a constantly evolving world, adaptation and continual learning is a must. 
For such a system or process to be efficient, all previously seen data should not be stored in their raw format and a full re-training at each point is simply infeasible at such a large scale. 

Since the early development of neural networks, researchers studied the catastrophic forgetting problem and proposed that  the parameters sharing which allows neural networks to generalize from seen data is the reason behind catastrophic forgetting~\cite{mccloskey1989catastrophic,ratcliff1990connectionist}. After learning one task, the network parameters correspond to one point in the parameter space. When learning a new task, the parameters will change their values to a new point  that might not correspond to a solution to the first task. It  has been shown that the parameter space of shallow networks contains cliffs  in which small moves lead to a severe change in the function output~\cite{kolen1991back}.

Early research works developed several strategies to mitigate the forgetting under the condition of not storing the training data, mostly at a small scale of few examples and considering shallow networks~\cite{kruschke1992alcove,ans1997avoiding,robins1995catastrophic}. 
Recently,  after the revival  of neural networks, the catastrophic forgetting problem and the continual learning paradigm  received increased attention~\cite{li2016learning,kirkpatrick2017overcoming,rusu2016progressive,lee2017overcoming}.  We will first define the general continual learning setting and  describe the main desired criteria of a continual learning system. We then move to point at the differences with other machine learning fields that share characteristics with continual learning.
\paragraph{\label{par:general_ocl_settings}General Continual Learning Setting.}
The general continual learning setting considers an infinite stream of data where at each time step $t$, the system receives a new sample(s)  $\{x_t, y_t\}$ drawn non i.i.d. from a current distribution $Q$ that could itself experience sudden or gradual changes. 

The main goal is to  learn a function $f$ parameterized by $\theta$ that minimizes a predefined loss\footnote{The loss itself might be learned or synthesized but this is left for future directions.} $\ell$  on the new sample(s) without interfering with and possibly improving on those that were learned previously.
\begin{align}
\label{eqn:gcls_objective}
\theta^t=&\>\underset{\theta,\xi}{\text{argmin}}\>\ell(f(x_t;\theta), y_t) +\sum \xi_i \\
\text{s.t.} & \>\ell(f(x_i;\theta), y_i) \leq \ell(f(x_i;\theta^{t-1}), y_i)+\xi_i  , \\
&\xi_i \ge 0 \quad ;\>\forall{i}\in{[0\twodots{t-1}]}  \nonumber
\end{align}
Where $\xi=\{\xi_i\}$ is a slack variable  that tolerates  a small increase in some previous samples losses, those that are hard to maintain without affecting the learning of current samples .
\subsection{Desiderata of Continual Learning \label{sec:intro-desiderata}}
To build a machine learning system that achieves the goal of continual learning described in Equation~\ref{eqn:gcls_objective}, it is important  to aim for  some if not all the desired characteristics listed below. These characteristics facilitate the realization of  a continual learning system.
\begin{enumerate}
\item \textbf{Constant memory.} The memory consumed by the continual learning paradigm should be constant w.r.t. the number of tasks or the length of the data stream. This  avoids the need to deal with unbounded systems.
\item \textbf{No task boundaries.} Being able to learn from the input data without requiring a clear task division brings great flexibility to the continual learning method and makes it applicable to any scenario where data distribution is shifting and environment slowly changing.
\item \textbf{Online learning.} A largely ignored characteristic of continual learning is being able to learn from a continuous stream of data without offline training of large batches or separate tasks. 
\item \textbf{Forward transfer.} This characteristic indicates the use of the previously acquired knowledge to aid the learning of new data/tasks.
\item \textbf{Backward transfer.} A continual learning system shouldn't  only aim at retaining previous knowledge but preferably improving the performance on previous tasks when learning future related tasks.

\item \textbf{Problem agnostic.} A continual learning method should be general and not limited to a specific setting (e.g. only classification). 
\item  \textbf{Adaptive.} Being able to learn from unlabeled data 
would increase the method applicability to cases where original training data no longer exist and  open the door to  a specific user setting adaptation.  
\item \textbf{No test time oracle.} A well designed continual learning method shouldn't rely on a task oracle to perform prediction.
\item \textbf{ Task revisiting.} When revisiting a previous task, the system should be able to successfully incorporate the new task knowledge.
\item \textbf{Graceful forgetting.} Given bounded system and infinite stream of data, a selective forgetting of unimportant information is needed to achieve a balance of stability and plasticity. 

\end{enumerate}

Due to the difficulty of the described  continual learning problem and the various challenges that have to be dealt with, in order to meet the different desiderata, methods try to overcome ``catastrophic forgetting'' with different levels of relaxations. Of all desiderata, ``online learning'' is the most commonly violated due to the difficulty of strict per-example incremental learning. 
Therefore,  a milder task incremental assumption is usually adopted.

\paragraph{Task Incremental Setting}
In this setting the data are streamed one task at a time, with different distributions for each task, while keeping the i.i.d. assumption and performing offline training within each task training phase.
For each task, we are given a dataset $D_t=\{{X}^{(t)},{Y}^{(t)}\}$ where ${X}^{(t)},{Y}^{(t)}=\{x_n^{(t)},y_n^{(t)}\}^{N_{t}}_{n=1}$  randomly drawn from a distribution $Q_t$ of a current task $T_t$.
The goal is to control the empirical risk of all seen tasks:
\begin{equation}
R= \sum_{t=1}^{\mathcal{T}}\frac{1}{N_t} \sum_{n=1}^{N_t} \ell(f(x_n^{(t)};\theta),y_n^{(t)}) 
\label{eq:StatRisk_intro}
\end{equation}

where $\mathcal{T}$ is the number of tasks seen so far. Given a limited  or no access to data from previous tasks, this can be expressed as minimizing the empirical risk of each new task with the constraints of not increasing the loss of the previous tasks:
\begin{equation}
\begin{aligned}
\theta^\mathcal{T}=\>\underset{\theta,\xi}{\text{argmin}}\>& \frac{1}{N_\mathcal{T}}  \sum_{n=1}^{N_\mathcal{T}} \ell(f(x_n^{(\mathcal{T})};\theta),y_n^{(\mathcal{T})})+\sum \xi_t\\
s.t.& \, \frac{1}{N_t} \sum_{n=1}^{N_t} \ell(f(x_n^{(t)};\theta),y_n^{(t)})\le\frac{1}{N_t} \sum_{n=1}^{N_t} \ell((f(x_n^{(t)};\theta^{\mathcal{T} -1}),y_n^{(t)}) +\xi_t ,\\
& \xi_t\ge 0 \quad ; \>\forall{t}\in{[0\twodots{\mathcal{T}-1}]}
\label{eq:EmpRiskSetting} 
\end{aligned}
\end{equation}
Where $\xi=\{\xi_t\}$ is a slack variable  that tolerates  a small increase in some previous tasks losses. The word task  here refers to an isolated training phase of a new batch of data that belongs to a new group of classes, a new domain or a different output space, e.g. scenes classification v.s. hand written digit classification. 
As such, following~\cite{hsu2018re}, a  finer categorization can be used: incremental class  learning where $P({Y}^t)=P({Y}^{t+1})$ but $\{{Y}^t\}\neq \{Y^{t+1}\}$ indicating disjoint labels in each task, incremental domain learning where $P({X}^t)\neq P({X}^{t+1})$ and $P({Y}^t)=P({Y}^{t+1})$ and the   incremental task learning indicating $P({Y}^t)\neq P({Y}^{t+1})$ and $P({X}^t)\neq P({X}^{t+1})$.
\section{Relation to Other Machine Learning Fields}~\label{sec:intro-fields}
\begin{figure}[!hbtp]
    \center{\includegraphics[width=1.\textwidth]{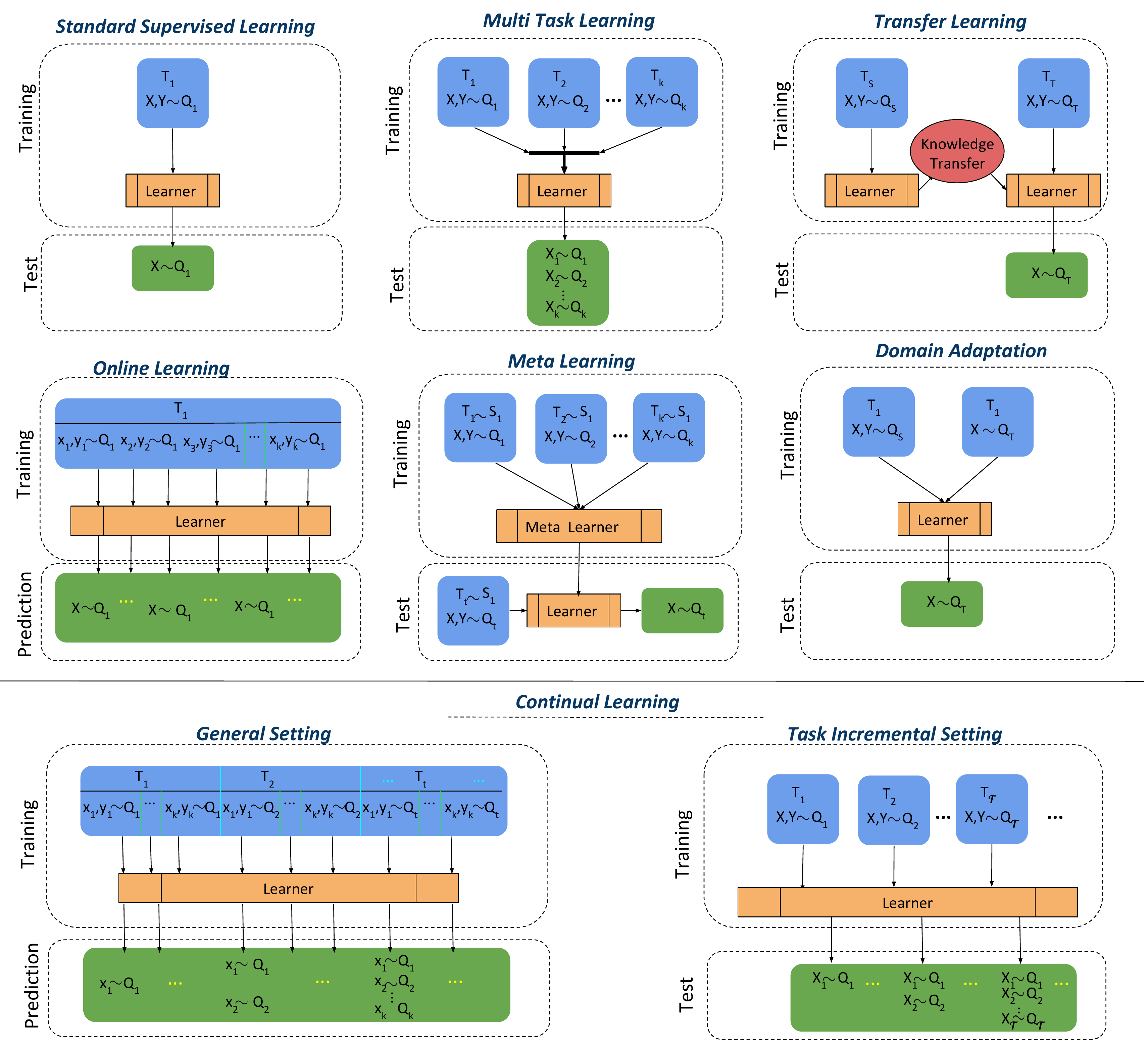}}
    \caption[The main setup of each related machine learning field.]{\label{fig:machine_learning_fields} The main setup of each related machine learning field, illustrating the differences with general continual learning setting and task incremental setting.}
  \end{figure}

The ideas of knowledge sharing, adaptation and transfer depicted in the outlined desiderata have been studied previously in machine learning and developed in isolated fields. We will describe each of them briefly and highlight the main differences with continual learning. See Figure~\ref{fig:machine_learning_fields} for an illustration of each related machine learning field setting.
\paragraph{Multi Task Learning.}
Multi-Task Learning  considers the learning of multiple related tasks simultaneously using a set or a subset of shared parameters. It aims for a better  generalization and  less overfitting using the shared knowledge extracted from the related learned tasks. We refer to~\cite{zhang2017survey} for a survey on multi-task learning.
Multi-task learning  follows the offline training of all tasks with the presence of all  tasks data  at training time. It doesn't involve any adaptation after the multi-task model has been deployed, as opposed to continual learning.
\paragraph{Transfer Learning.}
Transfer learning aims at aiding the learning process of a given task by exploiting the knowledge of another task or domain. More formally, given a source domain with data distribution $Q_S$ and  its corresponding  task $T_S$ and a  target  domain $Q_T$ with task $T_T$, transfer learning aims to support the learning of the target  task $T_T$ in $Q_T$ using the knowledge of $Q_S$ and $T_S$, where $Q_S \neq  Q_T$, or $T_S \neq T_T$.
Transfer learning is mainly concerned with the forward transfer desiderata of  continual learning. However, it  doesn't involve any continuous adaptation after learning the target task. Moreover, the performance on the source task(s) is not taken into account during transfer learning. A quite popular example of transfer learning is  finetuning, where models pre-trained on large tasks are used as  initialization for tasks with limited training data.  

\paragraph{Domain Adaptation.}
Domain adaptation is a sub-field of transfer learning  where the source and target tasks are the same but drawn from different input domains. The target domain data is unlabelled and the goal is to adapt a model trained on the source domain to perform well on the unlabelled target domain. In other words, it relaxes the classical machine learning assumption of having training and test data drawn from the same distribution~\cite{csurka2017domain}. As mentioned above for transfer learning, domain adaptation is unidirectional and doesn't involve any accumulation of knowledge~\cite{chen2018lifelong}.

\paragraph{Learning to Learn (Meta Learning).}
The old definition of learning to learn was referring to the concept of improving the learning behavior of the model with training experience. According to~\cite{thrun1998learning}, given a set of tasks, a model uses its experience from the past tasks to improve the performance on the current task. However, more recently, the common interpretation is the ability for a faster adaptation on a task with few examples given a large number of training tasks.
While these ideas seem quite close to continual learning, meta learning  follows the  assumption of offline training (i.e. all tainting data are available at the same time) but with training data being tasks randomly drawn from a task training distribution and test data being test tasks with few examples. Hence, it is not  capable, alone, of preventing forgetting on those previous tasks. An integration of the concept of meta learning with continual learning is an interesting area of   research~\cite{javed2019meta,he2019task}.
\\
\\
\textbf{Online Learning.}
Whereas in traditional offline learning, the entire training data has to be made available prior to learning the task, on the contrary online learning studies learning algorithms that learn to optimize predictive models over a stream of data instances sequentially. We refer to~\cite{shalev2012online,bottou1998online}
for surveys and overviews on the topic. In spite of considering a stream of data, online learning still assumes the i.i.d. data sampling procedure  and just one task/domain, in contrast to  continual learning.  

\paragraph{Open World Learning.}
Open world  learning~\cite{DBLP:journals/corr/abs-1809-06004,bendale2015towards} deals with the problem of detecting new classes at test time, hence avoiding wrong assignments to known classes. When those new classes are then integrated in the model, it meets the problem of continual learning.
As such, open world learning can be seen as a sub task of continual learning.
\section{Main Contributions}
The goal of this thesis is to study the stability/plasticity dilemma in neural networks and to propose solutions to provide artificial agents with the ability to learn continually and accumulate the acquired knowledge over their lifetime. This is  approached while aiming at achieving a reasonable trade-off between the characteristics described in Section~\ref{sec:intro-desiderata}.

We first consider the task incremental  setting  where tasks associated with their training and test data are received sequentially. Only access to the current task data is permitted, data from previous tasks are not retained. 
Under this setting, we start by proposing a system that allows for an expert level performance on each seen task while tolerating a linear increase in storage memory. A model is learned for each task and a gate is sequentially learned to assign the test samples to their corresponding models.  This work was published as an article in the Computer Vision and Pattern Recognition conference, CVPR 2017.
We then study the learning of the tasks sequence using a  shared  neural network model with fixed capacity. We propose a solution to learn a low dimensional   manifold that captures the  features needed for performing each task. Using the data of the new task as a proxy, features changes  in the learned manifold are constrained while changes orthogonal to the manifold are free of charge. This work was presented as a conference paper at the International Conference on Computer Vision, ICCV 2017.

As to mitigate the effect of tasks being drawn from very different distributions  and to avoid the need for a separate training phase to learn a lower dimensional feature space, we  instead  identify the important parameters in the neural network for a given task.  The changes on the important parameters are penalized during the learning process. Important parameters can be estimated on any data opening the door for adaptive continual learning method. This work was published as a conference paper in the European Conference on Computer Vision, ECCV 2018. 
To account for future tasks when the capacity of the shared model is fixed, we study the role of sparsity in continual learning and propose a new regularizer to encourage the learning of a sparse representation. This work was published as a conference paper in the International Conference on Learning Representations, ICLR 2019.

We then move to study the more challenging online continual learning setting where data are streaming with no given task boundaries or offline training on big batches of data. We first propose a protocol that brings our important parameters regularizer to the online continual learning setting. The work was published in the  Computer Vision and Patter Recognition conference, CVPR 2019. 
Next and as a solution to online continual learning  when the data generating distribution is largely shifting, we rely on a buffer of historical samples to prevent forgetting. We present a method to select representative samples from the data stream using as a criterion, maximizing the samples diversity in the gradient space of the model parameters. This work published in the   2019 Conference on Neural Information Processing Systems, NeurIPS.

The rest of the manuscript is organized as follows: We present the background materials closely related to our contributions in Chapter~\ref{ch:background}. We then discuss from a broad perspective works on continual learning that emerged during the past few years in Chapter~\ref{ch:relatedwork}.  Our work on sequential learning of experts is detailed in Chapter~\ref{ch:expertgate} while we discuss how to constrain the features space to reduce forgetting in Chapter~\ref{ch:ebll}. 
We move to present our method that estimates importance weights to the parameters  of a neural network  and penalizes their changes in Chapter~\ref{ch:MAS}. A study on the role of sparsity  and how to  impose it in continual learning is described in Chapter~\ref{ch:ssl}. 
Under the online continual learning setting, we first present  in Chapter~\ref{ch:reg_ocl}  a protocol for deploying our importance weight based method. Then, our gradient based sample selection method is detailed in Chapter~\ref{ch:reh_ocl}.
We conclude and discuss the state of the art of continual learning, limitations and future directions in Chapter~\ref{ch:conclusion}.

\cleardoublepage%

  \graphicspath{{\chaptersdir/Background/image/}}%
  \chapter{Background}\label{ch:background}
In this chapter we explain briefly  background materials related to the works presented in this manuscript. We give a very short introduction to neural networks in Section~\ref{sec:background_neural-network} and to autoencoders in Section~\ref{sec:background_autoencoders}. We then remind the reader of parameters estimators and some popular regularizers  in Section~\ref{sec:background_estimators}. We present a general Bayesian view on continual learning in Section~\ref{sec:background_bayesian}. Knowledge distillation is explained in Section~\ref{sec:background_distillation}. We shortly define common  continual learning terms in Section~\ref{sec:background-terms}. Finally, we describe  learning sequences and evaluation metrics used frequently in the experiments of this thesis (Section~\ref{sec:background_evaluation}).



 
\section{Neural Networks\label{sec:background_neural-network}}
\begin{figure}[!hbtp]
    \center{\includegraphics[width=0.45\textwidth]{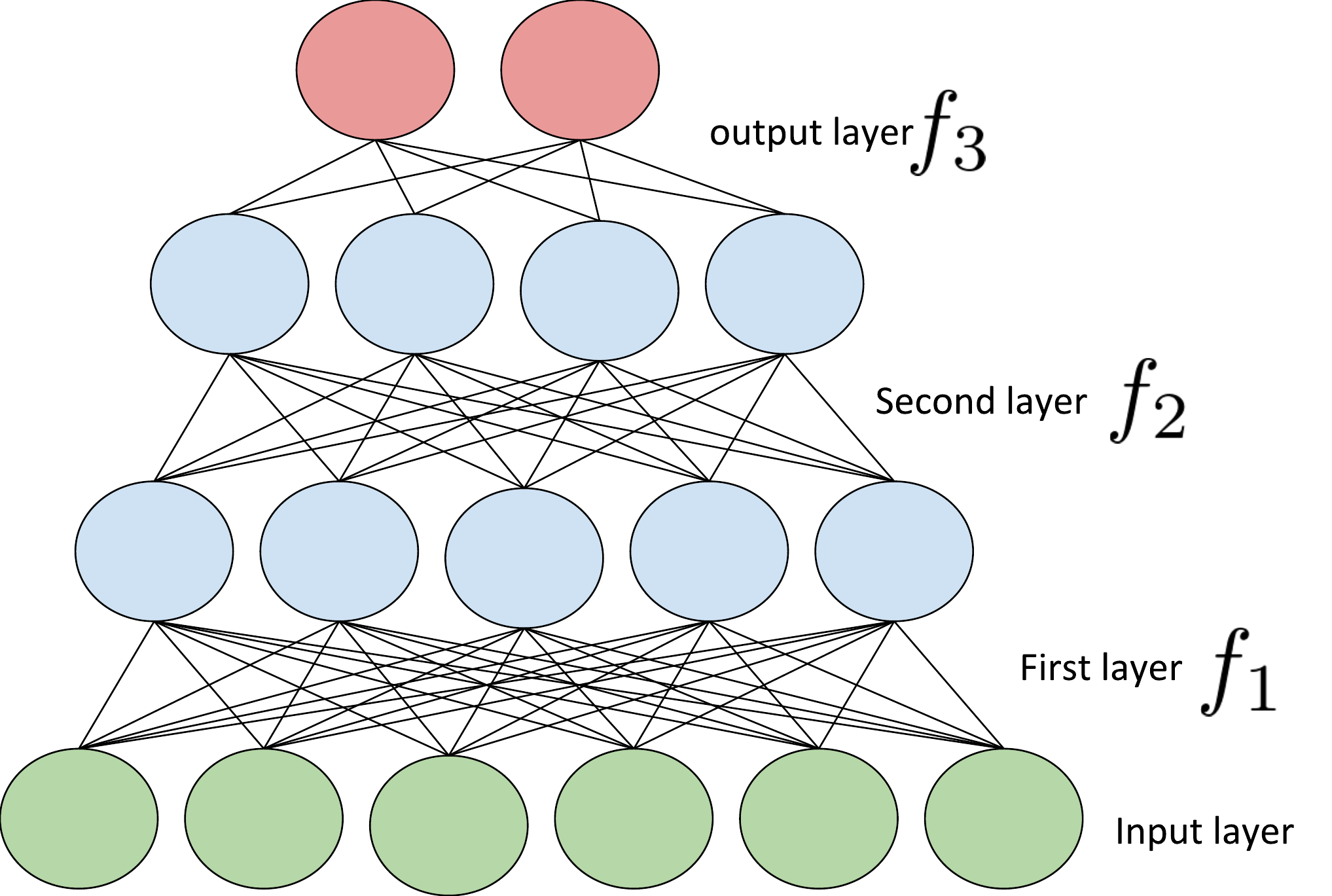}}
    \caption{\label{fig:neural_network} An example of a 3 layers neural network.}
     \vspace*{-10pt}
  \end{figure}

Given a training dataset $D=\{x,y\}^N_{n=1}$ sampled i.i.d. from a target distribution $Q_{x,y}$,  a neural network  is employed to learn a function $f$ that maps a given input $x_n$ to a target output $y_n$. For example in handwritten digit recognition, $x_n$ represents an image  while  $y_n$  corresponds to the drawn digit.  They are called networks because the learned function is composed of multiple functions, e.g. $f(x)=f_3(f_2(f_1(x)))$ where $f_1$ is the first layer, $f_2$ is the second layer and $f_3$ is the output layer. While the target ${y_n}$ represents the desired output for the output layer, there is no target output for the other layers and they are called hidden layers. Each layer, represented by  $f_l$, is parametrized by a weight vector $\theta_l=\{\theta_{ij}\}$~\footnote{Note that we don't mention explicitly  the bias, but it can be incorporated by padding the input to each layer by one. } with a vector input and a vector output, $f_l(x)=\phi(\theta_lx )$  where $\phi$ is the activation function. The layer can be  viewed as a group of  neurons, also called  units or perceptrons, where each neuron maps a vector input to a scalar output, hence the name multi layer perceptron, see Figure~\ref{fig:neural_network} for an example. $\phi$ is   usually a non linear function except maybe from the output layer. By stacking   multiple layers of neurons, one can approximate various non linear functions. In fact neural networks can be seen as  universal approximators if provided with enough neurons in a hidden layer. The learning occurs through minimizing a loss function that represents the divergence between the output of the neural network and the target output. Due to the non-linearity of the neural networks, most cost functions become non-convex; and they are usually optimized by iterative gradient descent steps. The gradients of the loss function are computed through the backpropagation~\cite{rumelhart1988learning} of the estimated loss to the preceding layers of the network. Typically, the optimizer performs gradients steps on batches randomly sampled from the training set.  This process is repeated and samples are re-visited until convergence. After a local minimum is reached, the neural network is deployed and used for performing predictions on new samples. We refer to~\cite{Goodfellow-et-al-2016-Book} for a detailed introduction to neural networks.

\section{Autoencoders\label{sec:background_autoencoders}}

\begin{wrapfigure}{r}{0.35\textwidth}
    \vspace*{-25pt}
    \includegraphics[width=0.35\textwidth]{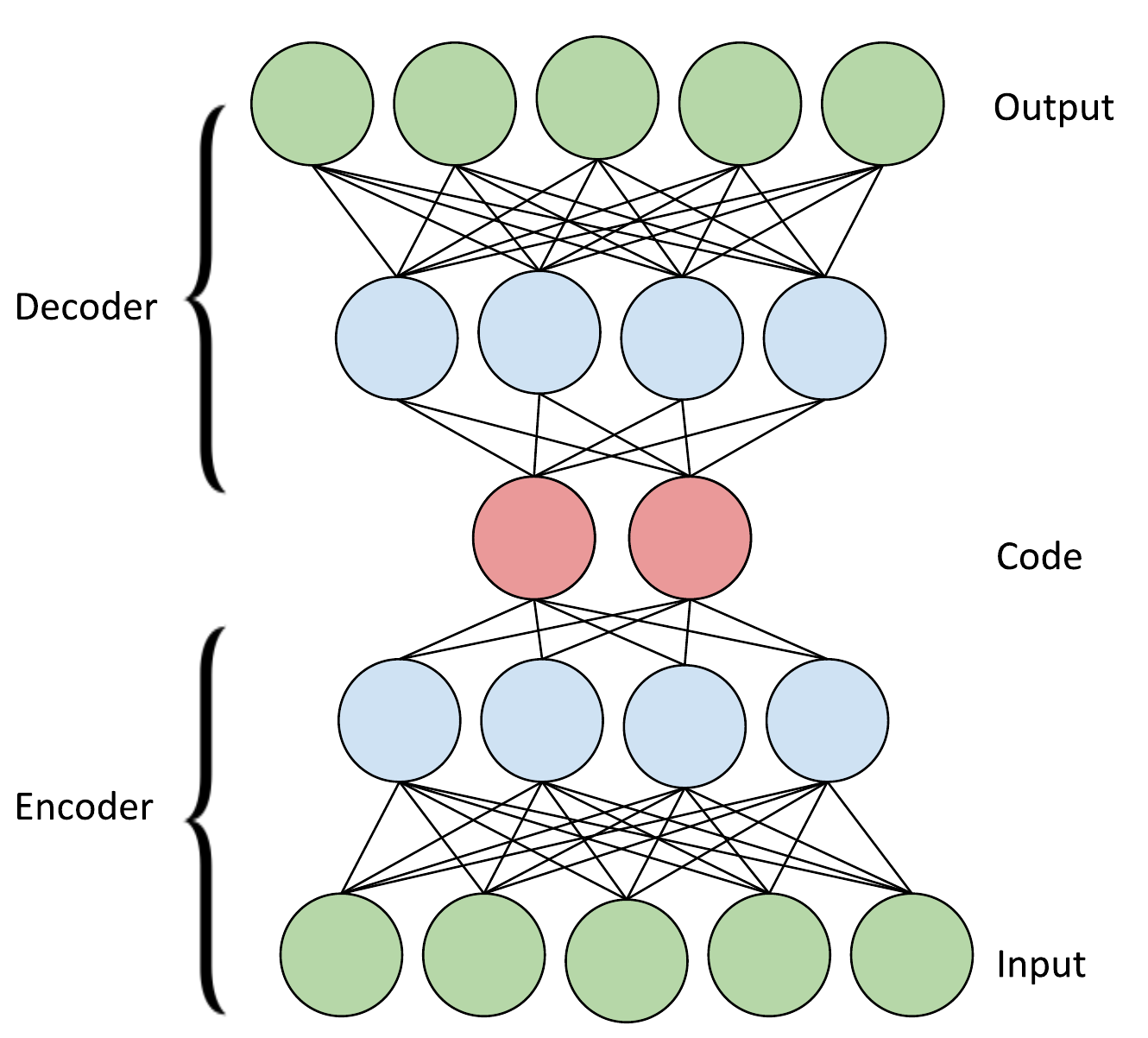}
  \caption[An example of under-complete autoencoder]{ An under-complete autoencoder with code size 2. }
    \label{fig:autoencder_example}
 \vspace*{-15pt}
\end{wrapfigure}
An autoencoder \cite{bourlard1988auto} is a neural network that learns to produce an output similar to its input \cite{Goodfellow-et-al-2016-Book}. The network is composed of two parts: an encoder $h(x)$ which maps the input $x$ to a code $c=h(x)$, and a decoder $g(c)$ that maps the code to a reconstruction of the input. 
The loss function $\ell(x,g(c))$ is simply the reconstruction error, e.g.  the mean $\ell_2$ distance between the inputs and their reconstructions. 
The encoder learns, through one or more hidden layers, a lower dimensonional representation (undercomplete autoencoder)  or a higher dimensional representation (overcomplete autoencoder) of the input data. A regularization term is necessary in overcomplete autoencders  to prevent the autoencoders from copying their input.  Figure~\ref{fig:autoencder_example} shows an  example of undercomplete autoencoder.  
A linear autoencoder with a Euclidean loss function learns the same subspace as principal component analysis~(PCA). 
However, autoencoders with non-linear functions yield better dimensionality reduction compared to PCA~ \cite{hinton2006reducing}.
Autoencoders are usually used to learn feature representations in an unsupervised manner or for  dimensionality reduction.  

The low dimensional manifold learned by  an undercomplete autoencoder  represents mostly the variations that are needed to reconstruct relevant samples. It is  maximally sensitive to variations  in the  data generating distribution but insensitive to changes orthogonal to the manifold. In Chapter~\ref{ch:expertgate} and Chapter~\ref{ch:ebll}, we  exploit this characteristic of autoencoders to approach the continual learning problem.

\section{Parameters Estimators \& Popular Regularizers \label{sec:background_estimators}}
\myparagraph{Maximum likelihood Estimation}

Given a dataset $D=\{x,y\}^N_{n=1}$ sampled i.i.d. from a target distribution $Q_{x,y}$  and a model with parameters $\theta=\{\theta_{ij}\}$, the likelihood of the data given the model parameters is defined as $L(\theta)=p(D|\theta)$.\\
A maximum likelihood estimator of the model parameters $\theta$ is:
\begin{equation}
    \theta^{MLE*}=\underset{\theta}{max} \quad  p(D|\theta)
\end{equation}
$\theta^{MLE*}$ is the model parameters that maximize the likelihood of the data $D$.
Usually in optimization, we minimize an objective function, e.g. loss function. As such instead of maximizing the likelihood, we minimize the negative log likelihood.
\begin{equation}
    \theta^{MLE*}=\underset{\theta}{min} \quad  -log(p(D|\theta))
\end{equation}

\myparagraph{Regularization}

The dataset used for training can be small and   noisy \footnote{By noise we mean that the data points might not represent the true properties of the  data generating distribution.} while the model is typically overparameterized.  Minimizing the log-likelihood of the training data could result in overfitting, i.e. small training errors but higher test errors. To reduce overfitting, regularization is imposed to control the model variance without a significant increase in the bias. Many regularization techniques aim at limiting the model capacity by adding a penalty that corresponds to the norm of the model parameters. As  examples we mention:

\textbf{L2 regularization.} An $\ell_2$ norm penalty which is usually referred to as weight decay. It pulls the parameters towards  the origin, keeping them with small magnitude. L2 regularization is also known as ridge regression
or Tikhonov regularization. Minimizing the penalized objective would lead to:
\begin{equation}
    \theta^{MLE*}=\underset{\theta}{min} \quad  -log(p(D|\theta)) +\frac{1}{2}\parallel\theta\parallel_2^2
\end{equation}
\textbf{L1 regularization.}  Instead of $\ell_2$ penalty, $\ell_1$ norm is also a popular norm for regularization. It is defined as 
$\parallel \theta \parallel_1= \sum_{i,j} |\theta_{ij}| $. $\ell_1$ norm results in  more sparse parameters than $\ell_2$ norm and it has been used for feature selection as in the LASSO~\cite{tibshirani1996regression}~(least absolute shrinkage and selection operator).

Note that usually a hyper parameter $\lambda$ is used to weight the contribution of the regularizer. We  explore the regularization effect on continual learning in Chapter~\ref{ch:ssl}.

\myparagraph{Maximum a posteriori estimation}

Bayes theorem estimates a  distribution over the model parameters.
\begin{equation}
    p(\theta|D)=\frac{p(\theta)p(D|\theta)}{p(D)}
\end{equation}
where $p(\theta)$ is the prior that we want to impose when optimizing the parameters $(\theta)$ on $D$,  $p(D|\theta)$ is the likelihood term, and $p(D)$ represents the  marginal likelihood. 
For a given dataset, it  is usually assumed that the marginal likelihood is constant.
\begin{equation}
    \quad p(\theta|D) \propto  \quad p(\theta)p(D|\theta)
\end{equation}
In  a maximum a posteriori probability (MAP) estimate, we maximize the posterior:
\begin{align}
\theta^{MAP*}&= \underset{\theta}{max} \quad p(\theta|D)\\
&=\underset{\theta}{max}  \quad log\,  p(\theta|D)\quad \\
 &=\underset{\theta}{max}\quad log\, p(D|\theta) + log \, p(\theta)
 \end{align}
The prior imposed in the MAP represents our beliefs about the problem we want to solve.
\paragraph{Gaussian prior}
A popular choice is to assume a Gaussian prior. It presumes that each parameter $\theta_{ij}$  is an independent random variable with 0 mean and $\sigma_{ij}$ standard deviation:
\begin{equation}
    \theta_{ij} \sim \mathcal{N}(0,\sigma_{ij}^2)
\end{equation}
\begin{equation}
    p(\theta) = \prod_{i,j} p(\theta_{ij})
\end{equation}
\begin{equation}
    log\,(p(\theta)) = \sum_{i,j} log(p(\theta_{ij}))
\end{equation}
\begin{equation}
    log\,(p(\theta)) = \sum_{i,j} log \frac{1}{\sqrt{2\pi\sigma_{ij}^2}} -\sum_{ij}\frac{1}{2\sigma_{ij}^2} | \theta_{ij} |^2
\end{equation}
\begin{equation}
    log\,(p(\theta)) = C -\sum_{i,j}\frac{1}{2\sigma_{ij}^2} | \theta_{ij} |^2
\end{equation}
As such a MAP estimate with a Gaussian prior over the model parameters would be:
\begin{align}
    \theta^{MAP*}&=\underset{\theta}{max}  \quad log \,  p(D|\theta) + log \, p(\theta)\\
    &=\underset{\theta}{max}  \quad log \, p(D|\theta)  -\sum_{i,
    j}\frac{1}{2\sigma_{ij}^2} | \theta_{ij} |^2
\end{align}
If $\theta \sim \mathcal{N}(0,I)$ then the MAP estimator  corresponds to the maximum likelihood estimator with L2 regularization.
\begin{align}
    \theta^{MAP*}&=\underset{\theta}{max}  \quad log \, p(D|\theta)  -\frac{1}{2} \parallel \theta\parallel_2^2\\
    &=\underset{\theta}{min}  \quad -log \, p(D|\theta)  +\frac{1}{2} \parallel \theta\parallel_2^2
\end{align}

\section{Continual Learning from a Bayesian Point of View\label{sec:background_bayesian}}
Suppose that we have a first dataset $D_1$ that corresponds to a given task $T_1$. According to Bayes rule, the posterior of the parameters is as follows:
\begin{equation}
   log\,p(\theta|D_1)\propto\log \,p(D_1|\theta) + log\,p(\theta)
\end{equation}
Suppose that after learning task $T_1$ and estimating its optimal parameters $\theta^1$, e.g. using MAP estimator, we have received a new task $T_2$ with its dataset $D_2$ then:
\begin{equation}
   log\,p(\theta|D_1,D_2)=\log \,p(D_2|\theta) + log\,p(\theta|D_1) -log(p(D_2|D_1))
\end{equation}
The first term is the log likelihood of the new task. Note that $ log\,p(D_2|D_1)$ is a constant and can be ignored in the estimation of the parameters. Now the information of the previous task is encoded in the posterior $p(\theta|D_1)$.  The posterior represents the uncertainty of the parameters  for that task.  Estimating the true posterior is intractable. It can be approximated as a Gaussian distribution~\cite{nguyen2017variational,goodfellow2013empirical}:
\begin{equation}
     p(\theta|D_1)=\mathcal{N}(\theta^1,\Sigma)
\end{equation}
with the mean equal to the optimal parameters for the previous task and a  diagonal covariance matrix assuming independent parameters. %

The objective function of the new task is then:

\begin{align}
   \underset{\theta}{max}& \quad \log \,p(D_2|\theta) -\frac{1}{2}(\theta - \theta^1)^\intercal\Sigma^{-1}(\theta - \theta^1)\\
   \underset{\theta}{max}& \quad \log \,p(D_2|\theta) -\sum_{i,j}\frac{1}{2\sigma_{ij}^2}(\theta_{ij} - \theta_{ij}^1)^2
\end{align}
While $\sigma_{ij}^2$ corresponds to the uncertainty of the parameter $\theta_{ij}$, $\frac{1}{\sigma_{ij}^2}$ represents  the parameter's importance for the previous task. Assuming an equal parameters importance, $\Sigma$ can be set to the identity matrix $I$. As such, optimizing the previous objective would encourage all the parameters to stay close to their optimal values at the previous task $\theta^1$. However, it has been shown to be vulnerable to catastrophic forgetting~\cite{kirkpatrick2017overcoming,goodfellow2013empirical}.
As proposed by~\cite{kirkpatrick2017overcoming},  Laplace approximation can be made assuming the diagonal precision matrix equal to the diagonal Fisher information matrix $F$ where  $\frac{1}{\sigma_{ij}^2}=F_{ij}$. Other methods have proposed different ways of estimating $\Sigma^{-1}$ or, as we shall call it the importance weights $\Omega$. The methods that assume a Gaussian prior to prevent forgetting are called prior focused methods as we show in the next chapter. 

\subsection{ Fisher Information Matrix}

The Fisher information matrix represents 
the covariance of the gradient of the model’s log likelihood function with respect to points
sampled from the model's distribution~\cite{martens2014new}. 
Given a dataset $D$ sampled i.i.d. from a target distribution $Q_{x,y}$, we want to learn a model with distribution $P_{x,y}(\theta)$ that maps the observations $\{x_n\}$ to their corresponding labels $\{y_n\}$.
The Fisher information matrix of $P_{x,y}(\theta)$ is then:
\begin{equation}
\mathcal{F}=E_{P_{xy}} \Big[\nabla \log \,p({x},{y}|\theta)\nabla \log \, p({x},{y}|\theta)^\intercal\Big]
\end{equation}
Estimating the Fisher information matrix requires sampling from the model's distribution.  To avoid extra computation cost, an approximation of the Fisher information matrix called empirical Fisher information is usually used.

\myparagraph{Empirical Fisher information}

The empirical Fisher employs cases sampled from the dataset instead of the model's distribution.
As such, the expectation is computed over the target distribution $Q_{xy}$ 
instead of the model's distribution $P_{xy}(\theta)$~\cite{martens2014new}.

\begin{align}
\bar{\mathcal{F}}&=E_{Q_{xy}} \Big[\nabla \log \,p(x,y|\theta)\nabla \log \, p(x,y|\theta)^\intercal\Big]\\
&=E_Q{_{x}} \Big[E_{Q_{y}}\Big[\nabla \log \,p(y|x,\theta)\nabla \log \, p(y|x,\theta)^\intercal\Big]\Big]\\
&=\frac{1}{N}\sum_n\Big[\nabla \log \,p(y_n|x_n,\theta)\nabla \log \, p(y_n|x_n,\theta)^\intercal\Big]
\end{align}

In~\cite{kirkpatrick2017overcoming} the inverse of the diagonal of the Fisher information or its empirical approximation is used to estimate the covariance of the prior distribution, i.e. the inverse of the previous task importance weights. In Chapter~\ref{ch:MAS}, we propose an alternative of estimating the importance weights.

\section{Knowledge Distillation}\label{sec:background_distillation}
Knowledge distillation~\cite{hinton2015distilling} is a technique that aims at transferring the  knowledge embedded in one model to another. Usually, a big model or an ensemble of models is used to learn a function from a big training dataset, which is quite computationally demanding. This technique aims at distilling the knowledge from the big/ensemble to a smaller model that is much lighter to deploy. The  probabilities of each class in the output layer of the big model convey richer information than the label of the image. It also captures the similarities/dissimilarities between the different classes that are learned by the large model. Suppose that $\hat{y}$ is the output of the light model and ${y}^*$ is the output of the large model. Then the knowledge distillation loss is defined as: 
\begin{equation}
\ell_{dist}(\hat{{y}} ,{y}^*) = -\langle {z}^*,\log \hat{{z}}\rangle
\end{equation}
where $\log$ is operated entry-wise and
\begin{equation}
z^*_i = \frac{{y}_i^{*1/\tau}}{\sum_j{y}_j^{*1/\tau}} \text{ and } \hat{{z}}_i = \frac{\hat{{y}}_i^{1/\tau}}{\sum_j\hat{{y}}_j^{1/\tau}} 
\end{equation}
where $i,j \in \{1,..,J\}$ running over the  output layer units, $\tau$ is the temperature. The application of a high temperature $\tau$ softens the probability distribution over the classes. Knowledge distillation can  be used to reduce  forgetting by distilling the knowledge from a previous model to a  model being trained on a new task, as we  show  in Chapter~\ref{ch:ebll}. 

\section{Continual Learning Terminology}\label{sec:background-terms}
In  this manuscript, we use terms  that have become common in the continual learning community.  Table~\ref{tab:terminology} gives a short definition of each frequently used terms.
\begin{table*}[h!]

\centering
\resizebox{\textwidth}{!}{
\begin{tabular}{ | l || l|}
\hline
Term & Description \\  \hline \hline  
{\tt Problem}&  Describes  what a machine learning model is
  employed to solve, e.g. the problem \\& of image classification.  \\ & \\  \hline  \hline  
{\tt Task}&  A more specific term than ``problem'' with a predefined input and output. \\  & For example, scene classification is a task of the image classification problem. \\ &\\  \hline  \hline  
{\tt Domain}&  Indicates a unique data generating distribution  of a given task input. \\ &\\  \hline  \hline  
{\tt Environment}& A  physical world in which a machine learning agent operates;  it is mostly used \\
& in reinforcement learning.\\ & \\ \hline  \hline 
{\tt Joint Training} & Indicates the training of a shared model on multiple tasks using all their data. \\ &\\   \hline  \hline  
{\tt Fine-tuning}&  Indicates training  a given task using as initialization a model pre-trained on \\ &  another task, typically with large amount of labelled data. \\ & It is the most typical form of  transfer learning. \\ &\\   \hline  \hline  
{\tt Catastrophic} & The complete erase of the previously
acquired knowledge  once new knowledge \\
{\tt Forgetting} &   is learned~\cite{french1999catastrophic}.  \\ &\\   \hline  \hline  
{\tt Catastrophic }& A similar terminology to catastrophic forgetting, indicating  that the new          \\
{\tt Interference}& learning  process has destructively affected the previous learning and caused \\& catastrophic forgetting.  \\& \\   \hline  \hline  
{\tt Replay}&   When learning  a new task or new samples, samples from previous history \\ & are re-intorduced  to the learner.\\ &\\ \hline  \hline  
{\tt Rehearsal}&  Performing learning steps on the replayed samples from previous history \\ & when learning a new task or new samples. \\ &\\ \hline  \hline  
{\tt Pseudo Rehearsal}&  Performing learning steps on  generated samples, mimicking previous history,  \\ &  when learning a new task or new samples. \\ &\\ \hline  \hline  
{\tt Revisiting}&   When a learner encounters samples from previous tasks or previous  distributions\\ & again.  This doesn't indicate that the exact same samples are reintroduced. \\& \\ \hline

\end{tabular}}
\caption{\label{tab:terminology} Terminology: list of the main terms used in this manuscript with a brief description each.}
\end{table*}

\section{Continual Learning Evaluation}~\label{sec:background_evaluation} 
\begin{figure}[!hbtp]
  \vspace{-20pt}
    \center{\includegraphics[width=1\textwidth]{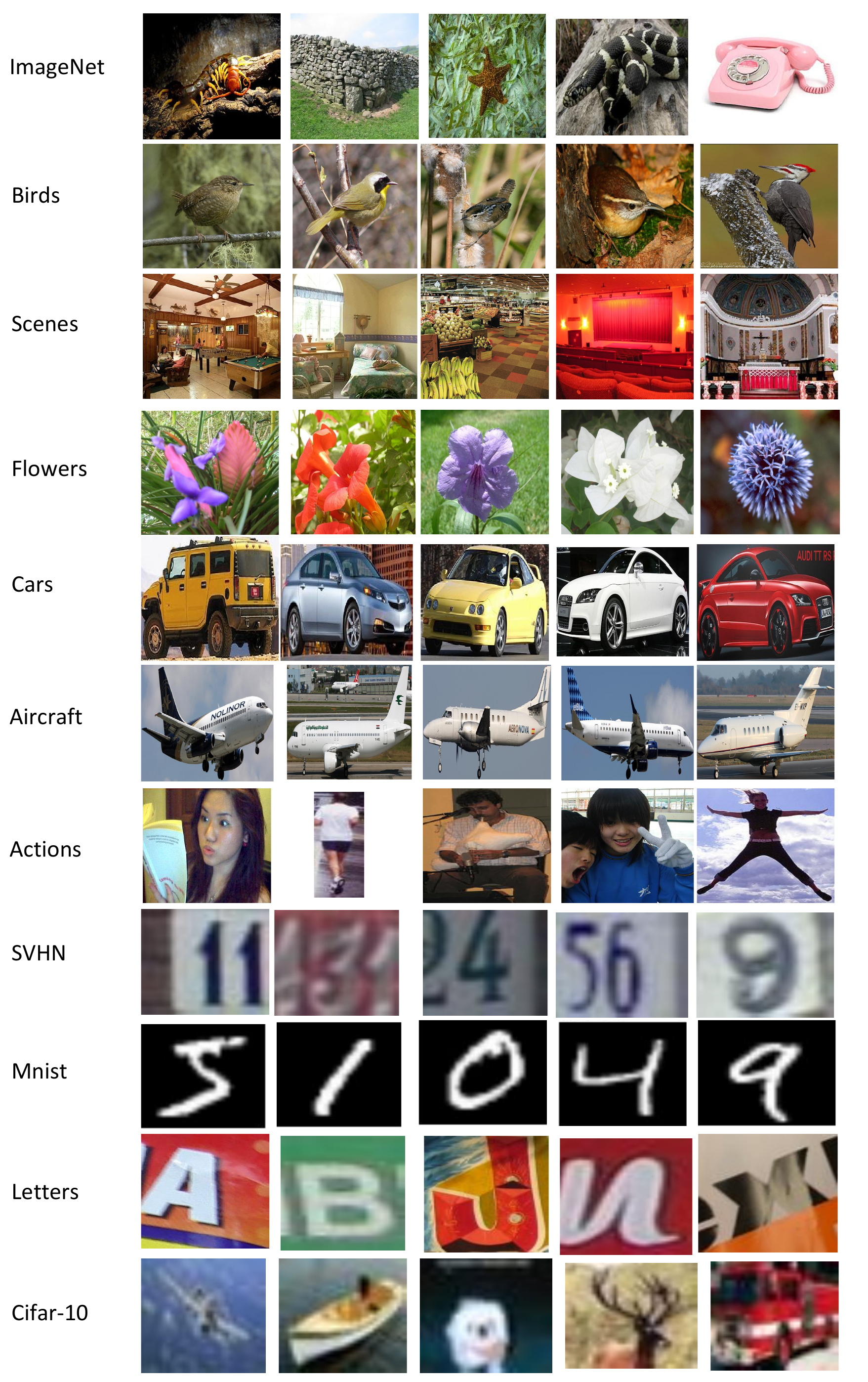}}
    \caption{\label{fig:background_dataset_examples}  Sample images from datasets used in this manuscript. }
    
  \end{figure}

\vspace{-40pt}
\paragraph{Datasets}
We provide a short description of  datasets repeatedly used in the experiments of this thesis and in the continual learning literature. See Figure~\ref{fig:background_dataset_examples} for example images.
\begin{enumerate}
\item  \textit{ImageNet} (LSVRC 2012 subset)~\cite{russakovsky2015imagenet}, is an image database organized according to the WordNet hierarchy which has more than 1 million training images and 1000 classes of different objects.  It is a big dataset. Models usually are pre-trained on ImageNet and then finetuned on other tasks.

 \item MIT \textit{Scenes} \cite{quattoni2009recognizing} for indoor scene classification. It is composed of 67 Indoor categories with 100 images per category divided into 80 images for training and 20 images for test. 
 
 \item Caltech-UCSD  \textit{Birds}~\cite{WelinderEtal2010} for fine-grained bird classification  with 200 bird species (mostly North American) and a total number of 5,994 samples.

 \item Oxford \textit{Flowers}~\cite{Nilsback08} for fine-grained flower classification. It is composed of 102 different categories of flowers with 2040 training samples and more than 6000 test samples.
\item Stanford  \textit{Cars} dataset~\cite{krause20133d} for fine-grained car classification. It contains 16185 images of 196 classes of cars split into a half for training and half for testing. 

\item FGVC-\textit{Aircraft} dataset~\cite{maji13fine-grained} for fine-grained classification of aircraft, this dataset contains 10200 images of aircraft belonging to about 100 different model variants.  

\item VOC \textit{Actions}, the human  action classification subset of VOC  challenge 2012~\cite{pascal-voc-2012}. It has  over 6000 train and test images.  
It is composed of 10 action classes and one bin class for other non classified actions. 
This dataset has multi-label annotations. For sake of consistency, we only use the actions with single label.
\item \textit{SVHN}~\cite{netzer2011reading} the  Street View House Numbers, a dataset for digit recognition. SVHN is obtained from house numbers in Google Street View images. It has over 600,000 digit images.  Each is a 32-by-32 image centered around a single character.
\item \textit{Letters}~\cite{deCampos09} the Chars74K dataset  for character recognition in natural images. We use the English set and exclude the digits, considering only the characters.  It has 52 classes both lower and upper case letters, and over 10000 training images with 1151 test samples.
\item  \textit{MNIST} dataset \cite{lecun1998gradient} for handwritten digit classification. It has a training set of 60,000 examples, and a test set of 10,000 examples. Images are 28-by-28 with black background and white digits.

\item \textit{Cifar-10} dataset~\cite{krizhevsky2009learning}  consists of 60,000 32-by-32 colour images in 10 classes of different objects, with 6000 images per class. There are 50,000 training images and 10,000 test images.

\item \textit{Cifar-100} dataset~\cite{krizhevsky2009learning}  is similar to Cifar-10 but with 100 classes grouped into 20 super categories. Each class has 600 images split into 500 for training and 100 for test.

\end{enumerate}
 For \textit{Cars}, \textit{Aircraft} and \textit{Actions}, we use the bounding boxes instead of the full images as the images might contain more than one object.
\paragraph{Benchmarks} 
Continual learning aims at developing methods that are able to accumulate knowledge over time. To benchmark a continual learning method, datasets mentioned above are usually grouped or split to form a sequence of tasks. There are three main typical ways of creating tasks sequences:
\begin{itemize}
    \item Pixel perumutation:  Given a  dataset, a first task is formed from this dataset $T_1$.  Then,  a sequence of $T$ tasks is created with the same dataset images but with unique pixel permutations  each. The goal is to perform the task  on all the different permutations that are learned sequentially. One of the most popular examples is permuted MNIST based on MNIST dataset.
    \item Group of datasets: By arranging a group of different datasets,   a sequence of tasks is formed. The goal is to learn the different tasks sequentially and be able to perform all the tasks. 
 One of the frequently used sequences  is an 8 tasks sequence, we proposed in Chapter~\ref{ch:MAS}. The sequence is as follows:
 Flowers$\rightarrow$Scenes$\rightarrow$Birds$\rightarrow$Cars$\rightarrow$Aircraft$\rightarrow$Actions$\rightarrow$Letters$\rightarrow$SVHN.
    \item A split of one dataset: Another scenario is to split one dataset into multiple groups of classes forming a sequence of tasks. For example, splitting Cifar-100 into 10 tasks where each task is composed of 10 classes. 
\end{itemize}
\paragraph{Evaluation metrics}
Continual learning is an emerging filed; recently multiple evaluation metrics have been proposed to measure different aspects of continual learning. However, throughout this manuscript, we focus on two simple, yet  important measures:
\begin{itemize}
    \item Accuracy: At the end of a learned sequence, the accuracy of each learned task is computed in addition to the  the average accuracy on all the tasks. 
    \item Forgetting: The difference between the accuracy of a given task once learned and its accuracy at the end of the learned sequence. An average forgetting can also be reported.  The performance of a previous task could be better than its performance once learned and a negative forgetting is then measured, usually referred to as backward transfer. However, this is rarely the case since a degradation in performance is typically attested.
    
\end{itemize}

\cleardoublepage
  \graphicspath{{\chaptersdir/Related_Work/image/}}%
  \chapter{Related Work}\label{ch:relatedwork}
As we explained in the introduction, Chapter~\ref{ch:introduction}, continual learning studies the problem of learning from an infinite stream of data drawn non i.i.d. from a distribution $Q_t$ that itself experiences sudden or gradual changes. These changes correspond to different input domains, classes or tasks switching from $T_t$ to $T_{t+1}$.

Early works  approaching continual learning focused on the catastrophic interference problem when  learning sequentially examples  of different input patterns (e.g. of different categories).  Several directions have been explored such as reducing representation overlap~\cite{french1992semi,french1994dynamically,kruschke1992alcove,kruschke1993human,sloman1992reducing}, replay samples or virtual samples from previous history~\cite{robins1995catastrophic,silver2002task} or  dual architectures~\cite{FRENCH1997recurrent,Rueckl1993Jumpnet,ans1997avoiding}. However, these works  mainly considered few examples (in the order of tens)  based on specific shallow architectures. 

After the success of deep learning thanks to the powerful GPUs allowing training on large datasets and deep architectures, continual learning and catastrophic forgetting have received increased attention.~\cite{goodfellow2013empirical} studied empirically the forgetting when learning two tasks sequentially  using different activation functions and dropout regularization~\cite{Srivastava2014Dropout}. It was concluded that dropout mitigates forgetting with possible advantages to activation functions that encourage neurons competition  such as Maxout~\cite{goodfellow2013maxout} and LWTA~\cite{NIPS2013_5059}.  
~\cite{pentina2014pac} studied incremental task learning from a theoretical perspective with the goal of transferring knowledge to future tasks.

More recent works have addressed continual learning with longer sequences of tasks and large number of examples. The various developed methods can be categorized into three major families based on how the information of previous data is stored and used in future learning and prediction. A first family are the \textit{replay} based methods. They store the information in the example space either directly in a replay buffer, or compressed in a generative model. When learning from new data, old examples are reproduced from the replay buffer or the generative model, and used for rehearsal/retraining or to constrain the current learning step. We call this category replay  instead of rehearsal since conventionally rehearsal implies retraining, yet the old examples can also be used to provide constraints \cite{lopez2017gradient}.
A second family are the \textit{regularization} based methods. They  consolidate the knowledge stored in the learned model through imposing a functional or parameter regularization term. We term the last major family as \textit{parameter isolation}.  They use isolated parameters for different tasks to prevent interference. Dynamic architectures that freeze/grow the network belong to this family. 
Below, we describe generally the core  concepts and the main methods of each family while leaving to each chapter the detailed description of its closely related works. Figure~\ref{fig:related-tree} illustrates the different families of continual learning methods and their sub-groups with example methods each.

\definecolor{level1}{HTML}{f48e82}
\definecolor{level2}{HTML}{63aff3}
\definecolor{level3}{HTML}{74df7b}
\definecolor{level4}{HTML}{ffe080}
\definecolor{level5}{HTML}{cafcda}

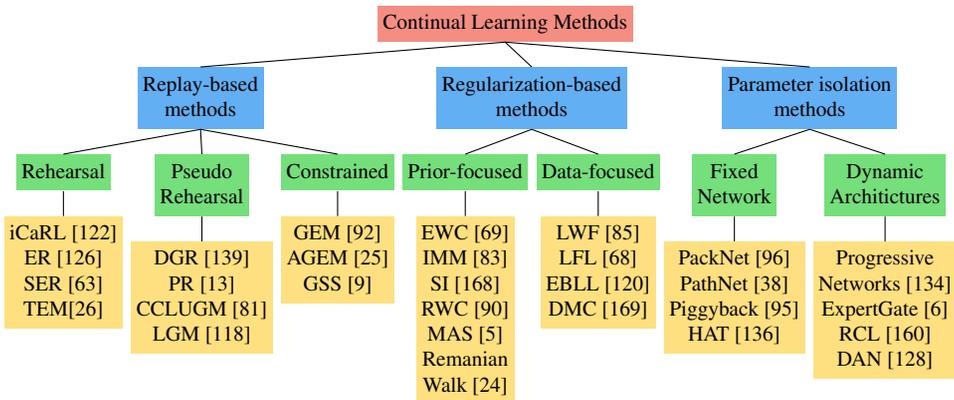
\begin{figure}
\centering
\caption{\label{fig:related-tree}A tree diagram illustrating the different continual learning families of methods and the different branches in each family. Leaves list example methods. }
\makebox[.95\textwidth]{
\medskip
\footnotesize
 \begin{forest}
for tree={s sep=1mm, inner sep=2, l=1}
[
{Continual Learning Methods },fill=level1
[Replay-based\\ methods,fill=level2 
[Rehearsal,fill=level3 
[{iCaRL~\cite{rebuffi2016icarl}}\\ ER~\cite{David2018Experience}\\SER~\cite{isele2018selective}\\TEM\cite{chaudhry2019continual},fill=level4]]
[Pseudo\\Rehearsal,fill=level3
[DGR~\cite{shin2017continual}\\PR~\cite{Craig2018Recursal}\\CCLUGM~\cite{lavda2018continual}\\LGM~\cite{ramapuram2017lifelong} ,fill=level4]]
[Constrained,fill=level3 
[GEM~\cite{lopez2017gradient}\\AGEM~\cite{chaudhry2018efficient}\\GSS~\cite{Aljundi2019continualNoT},fill=level4]
]
]
[{Regularization-based \\ methods},fill=level2 
[{Prior-focused},fill=level3 
[EWC~\cite{kirkpatrick2017overcoming}\\IMM~\cite{lee2017overcoming}\\SI~\cite{Zenke2017improved}\\RWC~\cite{liu2018rotate}\\MAS~\cite{aljundi2018memory}\\Remanian\\Walk~\cite{chaudhry2018riemannian},fill=level4
]
]
[{Data-focused},fill=level3 
[LWF~\cite{li2016learning}\\LFL~\cite{Jung_LFL}\\EBLL~\cite{rannen2017encoder}\\DMC~\cite{Junting2019Class-incremental},fill=level4]
]
]
[{Parameter isolation\\ methods},fill=level2 
[Fixed\\Network ,fill=level3
[PackNet~\cite{mallya2018packnet}\\PathNet~\cite{fernando2017pathnet}\\Piggyback~\cite{mallya2018piggyback}\\HAT~\cite{Serr2018Hard},fill=level4]]
[Dynamic\\Architictures,fill=level3
[Progressive\\ Networks~\cite{rusu2016progressive}\\ExpertGate~\cite{aljundi2016expert}\\RCL~\cite{xu2018reinforced}\\DAN~\cite{rosenfeld2018incremental},fill=level4]
]
]
]
\end{forest}
}
\end{figure}

\section{Replay-based Methods}
\label{sec:related_replay-based}

As the name indicates the replay based methods replay memories from previous history during the learning from new data. Although storage of the original examples in memory for rehearsal dates back to the 1990s \cite{robins1995catastrophic}, to date it is still a rule of thumb to overcome catastrophic forgetting in practical problems. For example, experience replay is widely used in reinforcement learning where the training distribution is usually non-stationary and learning is apt to  forgetting \cite{lin1993reinforcement,mnih2013playing}. 

In the context of incremental classification, iCaRL\cite{rebuffi2016icarl} maintains a subset of  exemplars per class that are rehearsed when new classes are learned. To ensure a bounded system, the number of stored exemplars is set to a fixed budget through  a feature mean matching selection criterion.
In settings where data is streaming with no clear task boundaries, ~\cite{David2018Experience} suggests the use of reservoir sampling to limit the number of stored samples to a fixed budget assuming an overall i.i.d. distributed data stream.

While rehearsal might be prone to overfitting and seems  to be bounded by joint training  when all previous data is available, constrained optimization is an alternative solution that leaves  more room for forward/backward transfer. As proposed in GEM~\cite{lopez2017gradient} under the task incremental setting, the key idea is to  constrain the parameters update on the new task  to not interfere with the previous tasks. This is achieved through projecting the estimated gradient direction  in the feasible region outlined by previous tasks gradients through a first order Taylor series approximation. AGEM~\cite{chaudhry2018efficient} has relaxed the problem to projection on one  direction estimated by randomly selected samples from a buffer of previous tasks data. We have recently extended this solution to a pure online continual learning setting where no task boundaries are known and have proposed to select a subset of samples that maximally approximate  the feasible region of the  historical data, as we will show in Chapter~\ref{ch:reh_ocl}.

In the absence of previous samples, pseudo rehearsal is an alternative strategy used in the early works with shallow neural networks. Random inputs and the outputs of previous model(s) given these inputs are used to approximate the previous tasks  samples~\cite{robins1995catastrophic}. With deep networks and large input vectors (e.g. full resolution images) random input cannot cover the input space~\cite{Craig2018Recursal}. Recently,  generative models have shown the ability to generate high quality images~\cite{goodfellow2014generative,DBLP:conf/iclr/2014} which opened up the possibility to model the data generating distribution and retrain on the generated examples \cite{shin2017continual}. However, this also adds to the complexity of training the generative model continually, with extra care needed to balance the retrieved examples and avoid the mode collapse problem.



\section{Regularization-based  Methods}
\label{sec:related_regularization}
When no storage of raw input is possible, an important line of works  proposes an extra regularization term to consolidate the previous knowledge when learning from new data. The constraint of not storing any historical sample is mainly motivated  by privacy reasons,  as in the case of medical applications, in addition to being memory efficient. We can further divide these methods into data focused and prior focused methods.
\subsubsection{Data-focused Methods}
\label{sec:related_datafocused}
The basic building  block in data-focused methods is the knowledge distillation from a previous model (trained on a previous task) to the model being trained on the new data. It was first proposed by ~\cite{silver2002task} to use the output of previous tasks models given new task input images mainly for improving the new task performance. It has been re-introduced in LwF~\cite{li2016learning} to mitigate the forgetting and for knowledge transfer. This is done by using  the outputs of the previous model  as soft labels for previous tasks. Other works~\cite{Jung_LFL,Junting2019Class-incremental} have been introduced with related ideas, however, as we will show in Chapter~\ref{ch:expertgate}; this strategy might be vulnerable to domain shift between tasks. Therefore, we have suggested an additional term to constrain the features of each task in their own learned low dimensional manifold ( see Chapter~\ref{ch:ebll}).

\subsubsection{Prior-focused methods}
\label{sec:related_priorfocused}
To mitigate the forgetting, prior focused methods estimate a distribution over the model parameters  which is  used as the prior distribution when learning from new data, following the Bayesian view on continual learning described in Chapter~\ref{ch:background}, Section~\ref{sec:background_bayesian}.
As this quickly becomes infeasible w.r.t. the  number of parameters in deep neural networks, parameters are usually assumed independent and an importance weight is estimated for each parameter in the neural network. During the training of later tasks, changes to important parameters are penalized.
Elastic weight consolidation (EWC) \cite{kirkpatrick2017overcoming} was the first to establish this approach.  Variational Continual Learning (VCL) has introduced a variational framework for this family \cite{nguyen2017variational}. Synaptic Intelligence (SI)~\cite{Zenke2017improved} estimates the parameters importance during the training of a task based on the contribution of their update to the decrease of the loss.
 In Chapter~\ref{ch:MAS} we explain our approach to estimate online the importance weights  based on unlabelled data, which allows for user adaptive settings and adds  great flexibility in deploying the method. While the prior focused family relies on tasks boundaries to estimate the prior distribution, we  further extend our work to task free settings, as we shall explain in Chapter~\ref{ch:reg_ocl}. 


Overall, the soft penalty introduced in the regularization family might not be sufficient to restrict the optimization process to stay in the feasible region of the  previous tasks, especially with long sequences~\cite{farquhar2018towards}, which might result in an increased forgetting of earlier tasks. 

\section{Parameter Isolation Methods}
\label{sec:related_parameterisolation}
To prevent any possible forgetting of  the previous tasks, in parameter isolation methods, different subsets of the model parameters are dedicated to each task. When there is no constraints on the size of the architecture,  this can be done by freezing the set of parameters learned after each previous task and growing new branches for new tasks~\cite{rusu2016progressive,xu2018reinforced}. Our Expert Gate~\cite{aljundi2016expert}   is a representative work that ensures no forgetting by design and aims at optimal performance for future tasks through  knowledge transfer, as we will show in Chapter~\ref{ch:expertgate}. 

Alternatively, under a fixed architecture, methods proceed by identifying the parts that are used for the previous tasks and masking them out during the training of the new task. This can be  either imposed at the parameters level \cite{mallya2018packnet,fernando2017pathnet} or at the neurons level as proposed in \cite{Serr2018Hard}.
These methods usually require a task oracle to activate the corresponding masks or task branch at prediction time. Our Expert Gate~\cite{aljundi2016expert} avoids this problem through learning an auto-encoder gate.  In general, this family is restricted to the task incremental setting and  better suited for learning long sequences of tasks when models capacity is not constrained and optimal performance is a priority.




\cleardoublepage

  \graphicspath{{\chaptersdir/Expert_Gate/image/}}%
  \chapter{Sequentially Learning a Network of Experts}\label{ch:expertgate}
In this chapter, we consider the task incremental  setting and set the target of achieving  an optimal performance on each seen task.  We argue that specialized models are needed for  expert level performance and  develop a gate, learned sequentially, to forward the test samples to the corresponding specialized model at test time. We evaluate our system on image classification and video prediction problems, and demonstrate the desired expert level performance on all the learned tasks.  This work was published as an article in CVPR 2017~\cite{aljundi2016expert}.
\section{Introduction}
\label{sec:Into}

In the age of deep learning and big data, we face a situation
where we train ever more complicated models with ever increasing amounts of data. We have different models for different tasks trained on different datasets, each of which is an expert on its own domain, but not on others. 
In a typical setting, each new task comes with its own dataset.
Learning a new task, say scene classification based on a pre-existing object recognition network trained on ImageNet, requires adapting the model to the new set of classes and
fine-tuning it with the new data. The newly trained network performs well on the new task, but has a degraded performance on the old ones due to the catastrophic forgetting problem. %
\begin{figure}[h]
\centering
\includegraphics[width=0.7\textwidth]{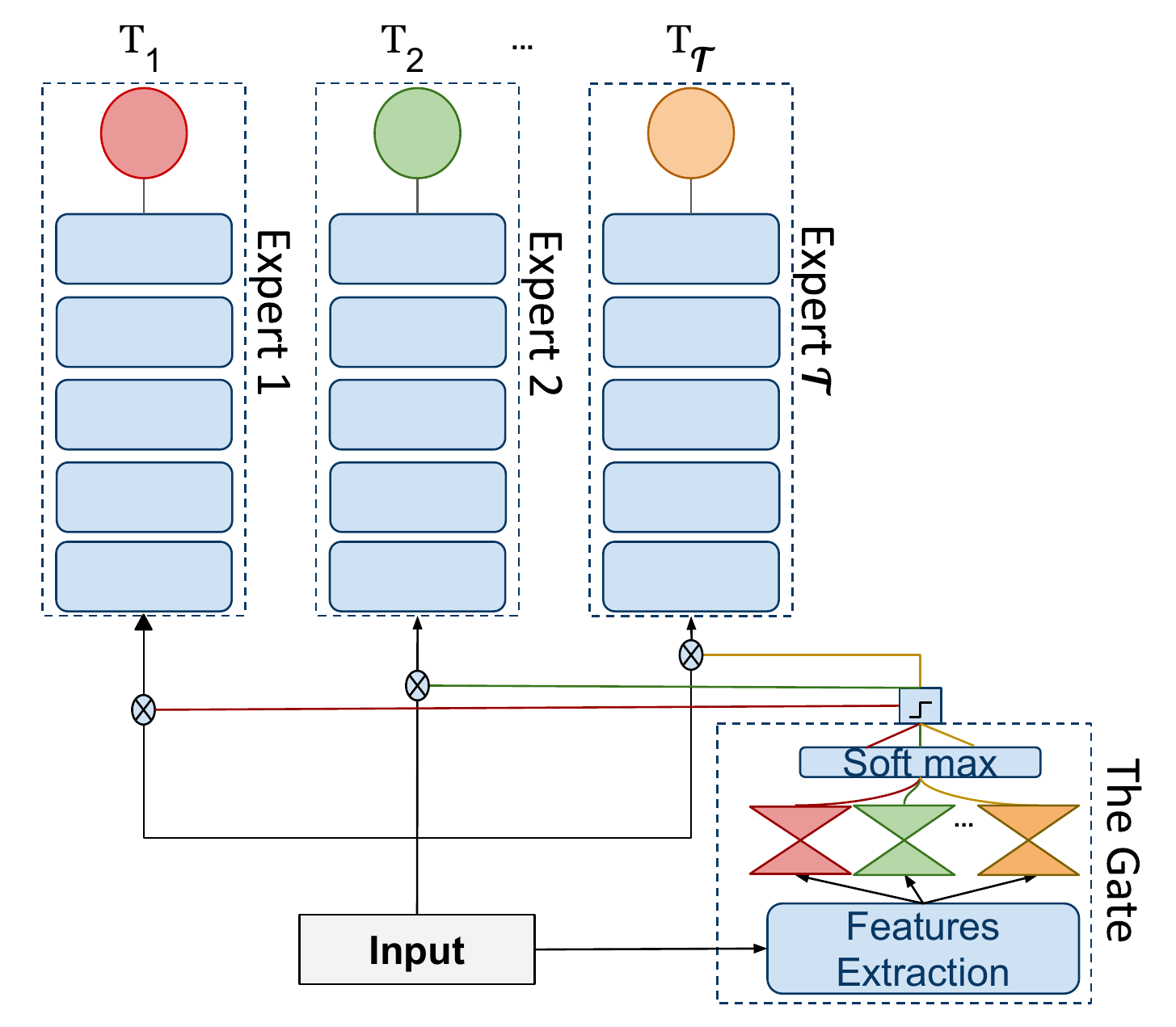}
  \caption{The architecture of our Expert Gate system.}
   
  \label{fig:expertgate_system}

\end{figure}

Ideally, a system should be able to operate on different tasks and domains and give the best performance on each of them. For example, an image classification system that is able to operate on generic as well as fine-grained classes, and in addition performs action and scene classification. If all previous training data were available, a direct solution would be to jointly train a model on all the different tasks or domains. Each time a new task arrives along with its own training data, new layers/neurons are added, if needed, and the model is retrained on all the tasks.
Such a solution has three main drawbacks. The first is the risk of the negative inductive bias when the tasks are not related or simply adversarial. Second, a shared model might fail to capture specialist information for particular tasks as joint training will encourage a hidden representation beneficial for all tasks. Third, each time a new task is to be learned, the whole network needs to be re-trained.
Apart from the above drawbacks, the biggest constraint with joint training is that of keeping all the data from the previous tasks. This is a difficult requirement to be met for a continual learning system, especially in the era of big data.

For example, ILSVRC~\cite{ILSVRC15} has 1000 classes, with over a million images, amounting to 200 GB of data. Yet the AlexNet
model trained on the same dataset, is only 200 MB, a difference in size of three orders of magnitude. With increasing amounts of data collected, it becomes less and less feasible to store all the training data, and more practical to just store the models learned from the data.

Without storing the data, one can consider regularization strategies for continual learning. While this can  mitigate the forgetting of old tasks, it is unlikely to maintain the achieved performance when learning scales  to long sequences
causing a bias towards the new tasks and a possible  buildup of errors on the older ones,
as we show in our experiments, Section~\ref{sec:expert_gate-expres}.
Moreover, regularization strategies suffer from the same drawbacks as the joint training described above. 
Instead of having a  network that is jack of all trades and master of none, to obtain the best performance for each task we propose having different specialist or expert models for different tasks, as also advocated
in~\cite{hinton2015distilling,jacobs1991adaptive,rusu2016progressive}. 
Therefore we build a Network of Experts, where a new expert model is added whenever a new task
arrives 
and knowledge is transferred from previous models. 

With an increasing number of task specializations, the number of expert models increases. Modern GPUs, used to speed up training and testing of neural nets, have limited memory (compared to CPUs), and can only
load a relatively small number of models at a time. We obviate the need for loading all the models by
learning a gating mechanism that uses the test sample to decide which expert to activate ( see Figure~\ref{fig:expertgate_system}).
For this reason, we call our method {\em Expert Gate}.

Unlike \cite{kokkinos2016ubernet}, who train one Uber network for performing vision tasks as diverse as semantic segmentation, object detection and human body part detection, our work focuses on tasks with a similar objective.
For example, imagine a drone trained to fly through an environment using its frontal camera. For optimal performance, it needs to deploy different models for different environments such as indoor, outdoor or forest. 
Our gating mechanism then selects a model on the fly based on the input video.
Another application could be a visual question answering system, that has multiple models trained using images from different domains.
Here too, our gating mechanism could use the data itself to select the associated task model. 

Even if we could deploy all the models simultaneously, %
selecting the right expert model %
is not straightforward.
Just using the output of the highest scoring expert is no guarantee for success
as neural networks can erroneously give high
confidence scores, as shown in \cite{nguyen2015deep}.  
We also demonstrate this in our experiments (Section~\ref{sec:expert_gate-expres}).
Training a discriminative classifier to distinguish between tasks
is also not an option since that would again require storing all training data. %
What we need is a task recognizer that can tell the relevance of its associated task model for a given test sample. This is exactly what our gating mechanism provides. In fact, also the prefrontal cortex of the primate brain is considered to have neural representations of task context that act as a gating in different brain functions~\cite{mante2013context}. %

We propose to implement such task recognizer  using an undercomplete autoencoder as a gating mechanism.
We learn for each new task or domain, a gating function that captures the shared characteristics among the training samples and can recognize similar samples at test time. We do so using a one layer undercomplete autoencoder. Each autoencoder is trained along with the corresponding expert model and maps the training data to its own lower dimensional subspace. At test time, each task autoencoder projects the sample to its learned subspace and measures the reconstruction error due to the projection. The autoencoder with the lowest reconstruction error is used like a switch, selecting the corresponding expert model (see Figure \ref{fig:expertgate_system}).

Interestingly, such autoencoders can 
also be used to evaluate {\em task relatedness} at training time, which in turn can be used  to determine which prior model is more relevant to a new task. 
We show how, based on this information, Expert Gate can decide which specialist model to transfer knowledge from when learning a new task and whether to only use fine-tuning or a dedicated regularization strategy~\cite{li2016learning}.

To summarize, the contributions of this chapter are the following.
We develop Expert Gate, a continual learning system that can sequentially deal with new tasks without storing all previous data. It automatically selects the most related prior task to aid  learning  the new task. At test time, the appropriate model is loaded automatically to deal with the task at hand.
We evaluate our gating network on image classification and video prediction problems. 

The rest of the chapter is organized as follows. We discuss related work in Section \ref{relWork}. Expert Gate is detailed in Section~\ref{method}, followed by experiments 
in Section \ref{sec:expert_gate-expres}. We finish with concluding remarks and a summary of the chapter in Section~\ref{conclusions}.

\section{Related Work}
\label{relWork}
Our end goal is to develop a system that can reach expert level performance on multiple tasks, with tasks learned sequentially. As such, it lies at the intersection between multi-task learning and incremental task learning. 
\paragraph{Multi-task learning.}

In multi-task learning
often one shared model is used for all tasks. This has the benefit of relaxing the number of required samples per task but could lead to suboptimal performance on the individual tasks.
On the other hand, multiple models can be learned, that are each optimal for their own task, but utilize inductive bias / knowledge from other models \cite{caruana1998multitask}.

To determine which related tasks to utilize,  ~\cite{thrun1998clustering}  clusters the  tasks  based on the mutual information gain when using the information from one task while learning another. 
This is an exhaustive process. As an alternative, 
\cite{jacob2009clustered,xue2007multi,kumar2012learning}  assume that the parameters of related tasks models lie close by in the original space or in a lower dimensional subspace and thus cluster the tasks parameters. They first learn tasks models independently, then use the tasks within the same cluster to help improving or relearning their models. 
This requires learning individual tasks models first.
Alternatively, we use our tasks autoencoders, that are fast to train, to identify related tasks.

\paragraph{Multiple models for multiple tasks.}
One of the first examples of using multiple models, each one handling a subset of tasks, was by Jacobs et al.~\cite{jacobs1991adaptive}. They trained an adaptive mixture of experts
(each a neural network) for multi-speaker vowel recognition and used a separate gating network to determine which network to use for each sample.
They showed that this setup
outperformed a single shared model. A downside, however, was that each training sample needed to pass through each  expert, for the gating function to be learned.
To avoid this issue, 
a mixture of one generalist model and many specialist models has been proposed~\cite{ahmed2016network,hinton2015distilling}. %
At test time, the generalist model acts as a gate, forwarding the sample to the correct network. 
However, unlike our model, these approaches require all the data to be available for learning the generalist model, which needs to be retrained each time a new task arrives.

\paragraph{Incremental task learning.} 

all previous knowledge  is used,
regardless of task relatedness.

Our system  is immune to forgetting by design since each task has its own specialist model and we obviate the need for storing all the training data collected during the lifetime of an agent, by learning task autoencoders that learn the distribution of the task data, and hence, also capture the meta-knowledge of the task. 

Our expert gate belongs to  the parameter isolation family (see Chapter~\ref{ch:relatedwork}, Section~\ref{sec:related_parameterisolation}). In this family, two  architectures, namely the progressive network \cite{rusu2016progressive} and the modular block network \cite{terekhov2015knowledge},
also use multiple networks, a new one for each new task. They add new networks as additional columns with lateral connections to the previous nets. These lateral connections mean that each layer in the new network is connected to not only its previous layer in the same column, but also to previous layers from all previous columns. This allows the networks to transfer knowledge from older to newer tasks. However, in these works, choosing which column to use for a particular task at test time is done manually, and the authors leave its automation as future work. 
Here, we propose to use an autoencoder to determine which model, and consequently column, is to be selected for a particular test sample. 

\section{The Proposed  Method}
\label{method}
We consider the case of incremental task learning where tasks and their corresponding training data arrive sequentially. For each task, we learn
a specialized model (expert) by transferring knowledge from previous tasks -- in particular, we build on the {\em most related} previous task. Simultaneously we learn a gating function that captures the characteristics of each task.
This gate forwards the test data to the corresponding expert resulting in a high performance over all learned tasks. 

The main question then is: how to learn such a gate function to differentiate between tasks, without having access to the training data of previous tasks? To this end, we learn a low dimensional subspace for each task/domain. At test time we then select the representation (subspace) that best fits the test sample. We do that using an undercomplete autoencoder per task. 
Below, we first describe this autoencoder in more detail (Section~\ref{sec:autoencoder}). Next, we explain how to use them for selecting the most relevant expert (Section~\ref{sec:selection}) and for estimating task relatedness (Section~\ref{sec:relatedness}).
  \subsection{The Autoencoder Gate}
  \label{sec:autoencoder}
As we explained in Chapter~\ref{ch:background} Section~\ref{sec:background_autoencoders}, an autoencoder \cite{bourlard1988auto} is a neural network that learns to produce an output similar to its input \cite{Goodfellow-et-al-2016-Book}. 

Autoencoders are usually used to learn feature representations in an unsupervised manner or for  dimensionality reduction. Here, we use them for a different goal.
The lower dimensional manifold learned by one of our undercomplete autoencoders will be maximally sensitive to variations observed in the task data but insensitive to changes orthogonal to the manifold. In other words, it represents only the variations that are needed to reconstruct relevant samples.
%
\begin{figure}[t]
\centering
\includegraphics[width=0.5\textwidth,height=0.3\textheight]{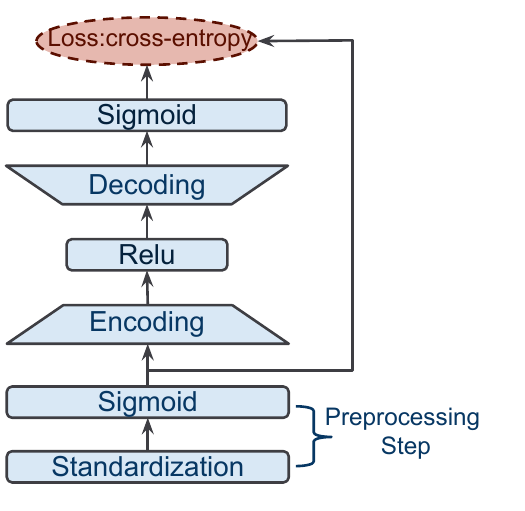}
  \caption{The deployed autoencoder gate structure.}    
  \label{fig:structure}

  \end{figure}
Our main hypothesis is that the autoencoder of one domain/task should thus be better at reconstructing the data of that task than the other autoencoders. Comparing the reconstruction errors of the different tasks autoencoders then allows to successfully forward a test sample to the most relevant expert network. 
It has been stated by~\cite{alain2014regularized} that in regularized (over-complete) autoencoders, the opposing forces between the risk and the regularization term result in a score like behavior for the reconstruction error. As a result, a zero reconstruction loss means a zero derivative which could be a local minimum or a local maximum. However, we use an unregularized one-layer undercomplete autoencoder and for these, it has been shown~\cite{marc2007unified} that the mean squared error criterion we use as reconstruction error estimates an energy function that outputs low values for input samples   similar to  training data, and high energy values otherwise. 
In fact, there is no need in such a one-layer autoencoder to add a regularization term to pull up the energy on unseen data because the narrowness of the code already acts as an implicit regularizer that prevents the autoencoder from learning the identity function.

\textbf{Preprocessing} 
We start from a robust image representation $F(x)$, namely the activations of the last convolutional layer of AlexNet pretrained on ImageNet. Note that other feature extraction methods can be used. Before the encoding layer, we pass this input through a preprocessing step, where the input data is standardized, followed by a sigmoid function. The standardization of the data, i.e. subtracting the mean and dividing the result by the standard deviation, is essential as it increases the robustness of the hidden representation to input variations. Normally, standardization is done using the statistics of the data that a network is trained on, but in this case, this is not a good strategy since at test time, we compare the relative reconstruction errors of the different autoencoders.
Different standardization regimes lead to non-comparable reconstruction errors. Instead, we use the statistics of ImageNet for the standardization of each autoencoder. Since this is a large dataset, it gives a good approximation of the distribution of natural images.
After standardization, we apply the sigmoid function to map the input to a range of $[0 \, 1]$. 

\textbf{Network architecture} 
We design a simple autoencoder that is no more complex than one layer in a deep model, with a one layer encoder/decoder (see Figure~\ref{fig:structure}). 
The encoding step consists of one fully connected layer followed by ReLU~\cite{zeiler2013rectified}. We make use of ReLU activation units as they are fast and easy to optimize. ReLU also introduces sparsity in the hidden units which leads to better generalization. 
For decoding, we use again one fully connected layer, but now followed by a sigmoid.
The sigmoid yields values between $ [0 \, 1]$, which allows us to use binary cross entropy as the loss function. At test time, we use the Euclidean distance to compute the reconstruction error.


\subsection{Selecting the Most Relevant Expert}
\label{sec:selection}
At test time, and after learning the autoencoders $\{A_t\}_{t=1}^{\mathcal{T}}$ for the seen tasks, we add a softmax layer that takes as input the reconstruction errors $\{er_t\}$ from the different tasks autoencoders $\{A_t\} $ given a test sample $x$. 
The reconstruction error $er_t$ of the $t$-th autoencoder is the Euclidean distance between output of the autoencoder and  the  representation of the input sample $F(x)$.
The softmax layer gives a probability  for each task autoencoder indicating its confidence:
\begin{equation}
p_a=\frac{ e^{(-er_a/{{\tau}})}}{\sum_t e^{ (-er_t/{\tau})}}
\end{equation}
where $\tau$ is the temperature. We use a temperature value of 2 as in  \cite{hinton2015distilling,li2016learning} leading to soft probability values. 
Given these confidence values, we load the expert model associated with the most confident autoencoder. For tasks that  have some overlap, it may be convenient to activate more than one expert model instead of taking the max score only.
This can be done by setting a threshold on the confidence values, see section \ref{sec:expres-gate}. 

\subsection{Measuring Task Relatedness}
 \label{sec:relatedness} 
 Given a new task $T_{\mathcal{T}}$ associated with its data $D_\mathcal{T}$, we first learn an autoencoder for this task $A_\mathcal{T}$. Let $T_t$ be a previous task with associated autoencoder $A_t$. We want to measure the task relatedness between task $T_\mathcal{T}$  and task $T_t$. Since we do not have access to the data of task $T_t$, we use the validation data from the current task $T_\mathcal{T}$. We compute the average reconstruction error $Er_\mathcal{T}$ on the current task data made by the current task autoencoder $A_\mathcal{T}$ and, likewise, the average reconstruction error $Er_t$ made by the previous task autoencoder $A_t$ on the current task data.
 The relatedness between the two tasks is then computed:
 \begin{equation}
 Rel(T_\mathcal{T},T_t)=1-(\frac {Er_t-Er_\mathcal{T}}{Er_\mathcal{T}})
 \end{equation}
 Note that the relatedness value is not symmetric. Applying this to every previous task, we get a relatedness value to each previous task. 
 
We exploit task relatedness in two ways. First, we use it to select the most related task to be used as a prior model for learning the new task. Second, we exploit the level of task relatedness to determine which transfer method to use: fine-tuning or learning-without-forgetting (LwF)~\cite{li2016learning}. LwF applies additional regularization term insuring that the output of the model trained on the new task remains close to the previous model output. This is done by keeping a separate output layer for the initial task and adding a new one for the new task.
Before training on the new task,  the outputs of the previous task layer are recorded given the new task data. During training, these outputs are preserved through the use of the knowledge distillation loss~\cite{hinton2015distilling} (see Chapter~\ref{ch:background}, Section \ref{sec:background_distillation}).
We found in our experiments that LwF only outperforms fine-tuning when the two tasks are sufficiently related.
When this is not the case, 
enforcing the new model to give similar outputs for the old task
may actually hurt performance.
Fine-tuning, on the other hand, only uses the previous task parameters as a starting point and is less sensitive to the level of task relatedness.
Therefore, 
we apply a threshold on the task relatedness value to decide when to use LwF and when to fine-tune.  
Algorithm  \ref{expertgate:algo} shows the main steps of our Expert Gate in both  training and test phase.
%
 \begin{algorithm}[t]
\caption{Expert Gate }\label{expertgate:algo}
  \begin{algorithmic}[1]
  \Statex \textbf{\textit{Training Phase}} input: expert-models $(E_1,.,E_{\mathcal{T}-1})$, tasks-autoencoders $(A_1,.,A_{\mathcal{T}-1})$, new task ($T_\mathcal{T}$), data ($D_\mathcal{T}$) ; output: $E_\mathcal{T}$, $A_\mathcal{T}$
      \State  $A_\mathcal{T}$ =train-task-autoencoder $(D_\mathcal{T})$
      \State (\textit{rel,rel-val})=select-most-related-task($D_\mathcal{T}$,$A_\mathcal{T},\{A_t\}$)
      
      \If{  \textit{rel-val} $>$\textit{rel-th}}
      \State $E_\mathcal{T}$=LwF$(E_{rel},D_\mathcal{T})$
    \Else
    \State $E_\mathcal{T}$=fine-tune$(E_{rel},D_\mathcal{T})$
  \EndIf
  \Statex  \textbf{\textit{ Test Phase}}  input: $x$ ; output:  prediction
  \State $a$=select-expert$(\{A_t\},x)$
  \State  prediction = activate-expert$(E_{a},x)$
 
  \end{algorithmic}
  \end{algorithm}

\section{Experiments}
\label{sec:expert_gate-expres}
First, 
we compare our method against various baselines on a set of three image classification tasks (Section~\ref{sec:expres-baselines}).
Next, we analyze our gate behavior in more detail on a bigger set of tasks (Section~\ref{sec:expres-gate}), followed by an analysis of our task relatedness measure
(Section~\ref{sec:expres-related}).
Finally, we test Expert Gate on a video prediction problem (Section~\ref{sec:expres-video}).

\noindent
\textbf{Implementation details}
We use the activations of the last convolutional layer of an AlexNet  pre-trained on ImageNet as image representation for our autoencoders. 
We experimented with the size of the hidden layer in the autoencoder, trying sizes of 10, 50, 100, 200 and 500, and found an optimal value of 100 neurons. This is a good compromise between complexity and performance. If the task relatedness is higher than $0.85$, we use LwF; otherwise, we use fine-tuning. 
We use MatConvNet framework \cite{vedaldi2015matconvnet} in all our experiments in this chapter.

\subsection{Comparison with Baselines}
\label{sec:expres-baselines}
\begin{wrapfigure}{r!}{0.4\textwidth}
\centering
\vspace*{-0.6cm}
   \includegraphics[width=0.35\textwidth]{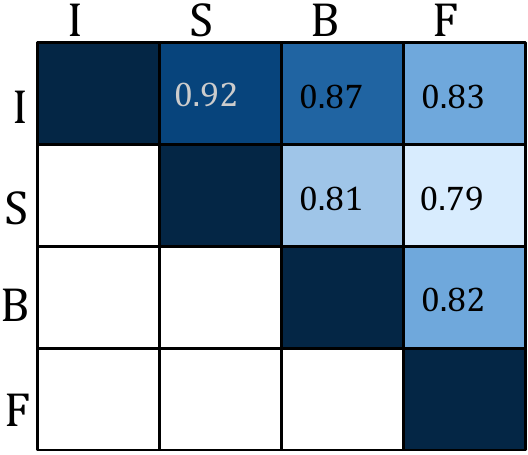}
   \caption[Task relatedness.]{Task relatedness. First letters indicate tasks. } 
   \label{fig:rel3}
  \vspace*{-0.1cm}
\end{wrapfigure}

We start with the sequential learning of three image classification tasks: in order, we train on MIT \textit{Scenes} \cite{quattoni2009recognizing}, Caltech-UCSD \textit{ Birds}~\cite{WelinderEtal2010} and Oxford \textit{Flowers}~\cite{Nilsback08}.
To simulate a scenario in which an agent or robot has some prior knowledge, and is then exposed to datasets in a sequential manner, we start off with an AlexNet model pre-trained on ImageNet. We compare against the following baselines:\\
\begin{enumerate}

    \item {\tt Joint Training.} Assuming all training data are always available, a model is jointly trained (by finetuning an AlexNet model pretrained on ImageNet) on all three tasks together.
    \item  {\tt Multiple fine-tuned models.} Distinct AlexNet models (pretrained on ImageNet) are finetuned separately, one for each task. At test time, an oracle gate is used, i.e. a test sample is always evaluated by the correct model.
    \item {\tt Multiple LwF  models.} Distinct models are learned with learning-without-forgetting~\cite{li2016learning}, one model per new task, always using AlexNet pre-trained on ImageNet as previous task. This is again combined with an oracle gate.
    \item  {\tt A single fine-tuned model.} one AlexNet model (pre-trained on ImageNet) sequentially fine-tuned on each task. 
    \item  {\tt   A single LwF model.} LwF sequentially applied to multiple tasks. Each new task is learned with all the outputs of the previous network as soft targets for the new training samples. So, a network (pre-trained on ImageNet) is first trained for  $T_1$ data without forgetting ImageNet (i.e. using the pretrained AlexNet predictions as soft targets). Then, this network is trained with $T_2$ data, now using ImageNet and $T_1$ specific layers outputs as soft targets; and so on.  While preserving ImageNet  outputs might seem unfair, it acts as a great source for knowledge transfer and  benefits all the later tasks. 
\end{enumerate}
For baselines with multiple models (2 and 3), we rely on an oracle gate to select the right model at test time. So reported numbers for these are upper bounds of what can be achieved in practice. The same holds for baseline 1, as it assumes all previous training data are stored and available.
Table \ref{tab:sequential} shows the  classification accuracy achieved on the test sets of the different tasks.
For our Expert Gate system and for each new task, we first select the most related previous task (including ImageNet) and then learn the new task expert model by transferring knowledge from the most related task model, using LwF or finetuning.
\begin{table}[h]

\centering
\begin{tabular}{ | l | c| c | c ||c |}
\hline
Method & Scenes &Birds & Flowers&Avg. \\  \hline 
{\tt Joint Training}* & 63.1\%  & 58.5\%  & 85.3\% & 68.9\%  \\  \hline  
{\tt Multiple fine-tuned models}** &  63.4\%  &  56.8\%  &85.4\% &68.5\%   \\ 

{\tt Multiple LwF models}** &  63.9\%  &  58.0\%  &84.4\%  &68.7\%  \\ \hline 

{\tt Single fine-tuned model} & 63.4\%  & - & -  &-\\ 
             & 50.3\%  & 57.3\%  & - &- \\ 
			 & 46.0\%  & 43.9\%  &84.9\%  &58.2\%  \\ \hline
{\tt Single LwF model} &  63.9\%  &  - &- &- \\ 
            &  61.8\%  &  53.9\%  &-&-  \\ 
			&61.2\% &53.5\% & 83.8\% &66.1\% \\   \hline
{\tt Expert Gate (ours)} &  63.5\%  & 57.6\%  &84.8\% &68.6\%   \\ \hline

\end{tabular}
\caption[Classification accuracy for the sequential learning of 3 image classification tasks.]{\label{tab:sequential}Classification accuracy for the sequential learning of 3 image classification tasks. Methods with * assume all previous training data is still available, while methods with ** use an oracle gate to select the proper model at test time.}
  \vspace*{-0.5cm}
\end{table}
For the \textit{Single fine-tuned model} and \textit{Single LwF model}, we also report intermediate results in the sequential learning. 
When learning multiple models (one per new task), LwF improves over vanilla fine-tuning for Scenes and Birds, as also reported  by \cite{li2016learning}\footnote{Note these numbers are not identical to \cite{li2016learning} but show similar trends. At the time of experimentation, the code for LwF was not available, so we implemented this ourselves in consultation with the authors of~\cite{li2016learning}, and used parameters provided by them.}. However, for Flowers, performance degrades compared to   fine-tuning. 
We measure a lower degree of task relatedness to ImageNet for Flowers than for Birds or Scenes (see Figure \ref{fig:rel3}) which might explain this effect.
Comparing the {\tt Single fine-tuned model} (learned sequentially) with the {\tt Multiple fine-tuned models}, we observe an increasing drop in performance on older tasks: sequentially fine-tuning a single model for new tasks shows catastrophic forgetting and is not a good strategy for continual learning.
%
The {\tt Single LwF model} is less sensitive to forgetting on previous tasks.
However, it is still inferior to training exclusive models for those tasks ({\tt Multiple fine-tuned / LwF models}), both for older as well as newer tasks. Lower performance on previous tasks is because 
of a buildup of errors and degradation of the soft targets of the older tasks. This results in LwF failing to compensate for forgetting in a sequence involving more than 2 tasks. This also adds noise in the learning process of the new task. Further, the previous tasks have varying degree of task relatedness. 

On these datasets, we systematically observed the largest task relatedness values for ImageNet (see Figure \ref{fig:rel3}). 
Having to remember other tasks prevents the new task from getting the same benefit of ImageNet as in the {\tt Multiple LwF models} setting.
Our {\tt Expert Gate} always correctly identifies the most related task, i.e. ImageNet. Based on the relatedness degree, it used LwF for Birds and Scenes, while fine-tuning was used for Flowers. As a result, the best expert models were learned for each task. At test time, our gate mechanism succeeds to select the correct model for $99.2 \%$ of the test samples.
This leads to superior results to those achieved by the other two sequential learning strategies ({\tt Single fine-tuned model} and {\tt Single LwF model}). We achieve comparable performance on average to the {\tt Joint Training} that has access to all the tasks data. Also, performance is on par with  {\tt Multiple fine-tuned models} or {\tt Multiple LwF models} that both assume having the task label for activating the associated model.   

\begin{table*}

\centering
\resizebox{\textwidth}{!}{\begin{tabular}{ | l | c| c | c | c | c | c ||c|}
\hline
Method & Scenes &Birds & Flowers & Cars & Aircraft & Actions &Avg.\\  \hline
{\tt Joint Training}* & 59.5\%  & 56.0\%  & 85.2 \% &77.4\% & 73.4\%  &47.6\%  &66.5\% \\  \hline  
{\tt Most confident model} & 40.4\%  & 43.0\% & 69.2\%  &78.2\% &54.2\% &8.2\%  &48.7\% \\  \hline  


{\tt Expert Gate}  &60.4\%  &57.0\%  &84.4\%  & 80.3\%  &72.2\% &49.5\% &67.3\% \\ \hline

\end{tabular}}
\caption[Classification accuracy for the sequential learning of 6 tasks.]{\label{tab:six} Classification accuracy for the sequential learning of 6 tasks. Method with * assumes all the training data is available.}

\end{table*}

\subsection{Gate Analysis}
\label{sec:expres-gate}
The goal of this experiment is to further evaluate our Expert Gate ability in successfully selecting the relevant network(s) for a given test image. 
For this experiment, we add 3 more tasks: Stanford  \textit{Cars} dataset~\cite{krause20133d}, 
FGVC-\textit{Aircraft} dataset~\cite{maji13fine-grained}, and
VOC \textit{Actions}. 
 So in total we deal with 6 different tasks: Scenes$\rightarrow$Birds$\rightarrow$Flowers$\rightarrow$Cars$\rightarrow$Aircraft $\rightarrow$Actions, along with ImageNet that is considered as a generalist model or initial pre-existing model.

We compare again with {\tt Joint Training}, where we fine-tune the ImageNet pre-trained AlexNet jointly on the six tasks assuming all the data is available. We also compare with a setting with multiple fine-tuned models where the model with the maximum score is selected ({\tt Most confident model}).
For our Expert Gate, we follow the same regime as in the previous experiment. The most related task is always ImageNet. Based on our task relatedness threshold, LwF was selected for Actions, while Aircraft and Cars were fine-tuned. Table \ref{tab:six} shows the results.


Even though the jointly trained model has been trained on  all the previous tasks data simultaneously, its average performance is inferior to our Expert Gate system.
This can be explained by the negative inductive bias where some tasks negatively affect others, as is the case for Scenes and Cars.

As we explained in the introduction of this chapter, Section ~\ref{sec:Into}, deploying all  models and taking the max score ({\tt Most confident model}) is not an option: for many test samples the most confident model is not the correct one, resulting in poor performance. Additionally, with the size of each expert model around 220 MB and the size of each  autoencoder  around 28 MB, there is almost an order of magnitude difference in memory requirements.

\subsubsection{Comparison with a discriminative classifier.} 
\begin{figure}
\centering
\includegraphics[width=0.8\textwidth]{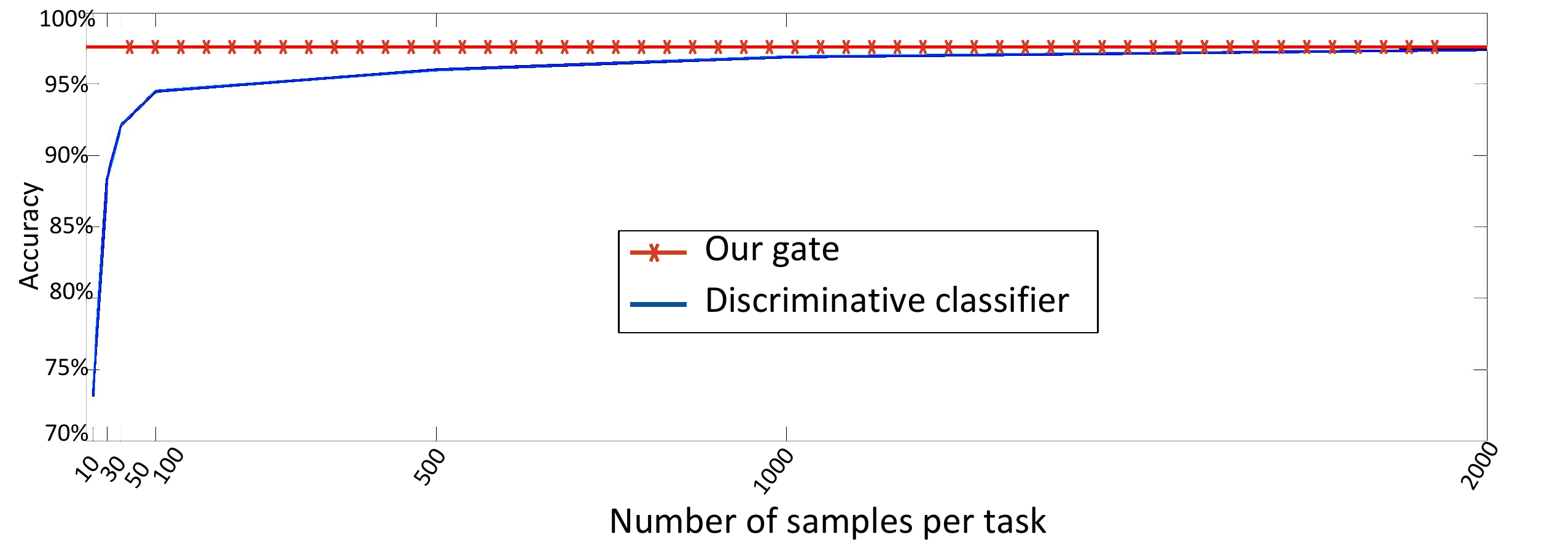}
  \caption[Comparison between our gate and the discriminative classifier with varying number of stored samples per task]{Comparison between our gate and the discriminative classifier with varying number of stored samples per task.  }
   
  \label{fig:dis_class}

\end{figure}

We compare with a discriminative classifier trained to predict the task considering the six tasks sequence. For this classifier, we first  assume that all data from the previous tasks are stored, even though this is not in line with the task incremental    setting. 
Thus, it serves as an upper bound.
For this classifier ({\tt Discriminative Task Classifier}) we use a neural net with one hidden layer composed of 100 neurons (same as our autoencoder code size). It takes as input the same data representation as our autoencoder gate and its output is the different tasks labels.
Table \ref{tab:gate} compares the performance of our gate on recognizing each task data to that of the discriminative classifier.
Further, we test the scenario of a discriminative classifier with the number of stored samples per task varying from 10-2000 (Figure \ref{fig:dis_class}). It approaches the accuracy of our gate with 2000 samples per task. Note that this is  $\frac{1}{2}$ to $\frac{1}{3}$ of the size of the used datasets. For larger datasets, an even higher number of samples would probably be needed to match performance.
In spite of not having access to any of the previous tasks data, our Expert Gate achieves similar performance to the discriminative classifier. In fact, our Expert Gate can be seen as a sequential classifier with new classes arriving one after another. This is one of the most important results from this chapter:  {\em without ever having simultaneous access to the data of  different tasks, our Expert Gate based on autoencoders manages to assign test samples to the relevant tasks equally accurately as a discriminative classifier}.
\begin{table*}[t!]
\centering
\resizebox{\textwidth}{!}{
\begin{tabular}{ | l | c| c | c | c | c | c || c|}
\hline
Method & Scenes &Birds & Flowers & Cars & Aircraft & Actions &Avg.\\  \hline
 {\tt Task Classifier } &97.0 \%  & 98.6\% & 97.9\% & 99.3\%& 98.8\%&95.5\%&97.8\%\\  
 \small {\em using all the tasks data} &  & &  &  &  &  &\\ \hline 
{\tt Expert Gate} \  &94.6\% &97.9\%& 98.6\% & 99.3\%&97.6\%&98.1\%&97.6\%\\ 
small {\em no access to the previous tasks data} &  & &  &  &  &  &\\ \hline
\end{tabular}}
\caption{\label{tab:gate}Results on discriminating between the 6 tasks (classification accuracy)}

\end{table*}

\subsubsection{Analysis of confusion cases}
As to have a better  understanding of the different kinds of mistakes that we are likely to expect from  Expert Gate, we show    diverse  qualitative examples from the 6 tasks learning sequence.
 
\begin{figure}[h]
\centering
\includegraphics[width=0.8\textwidth]{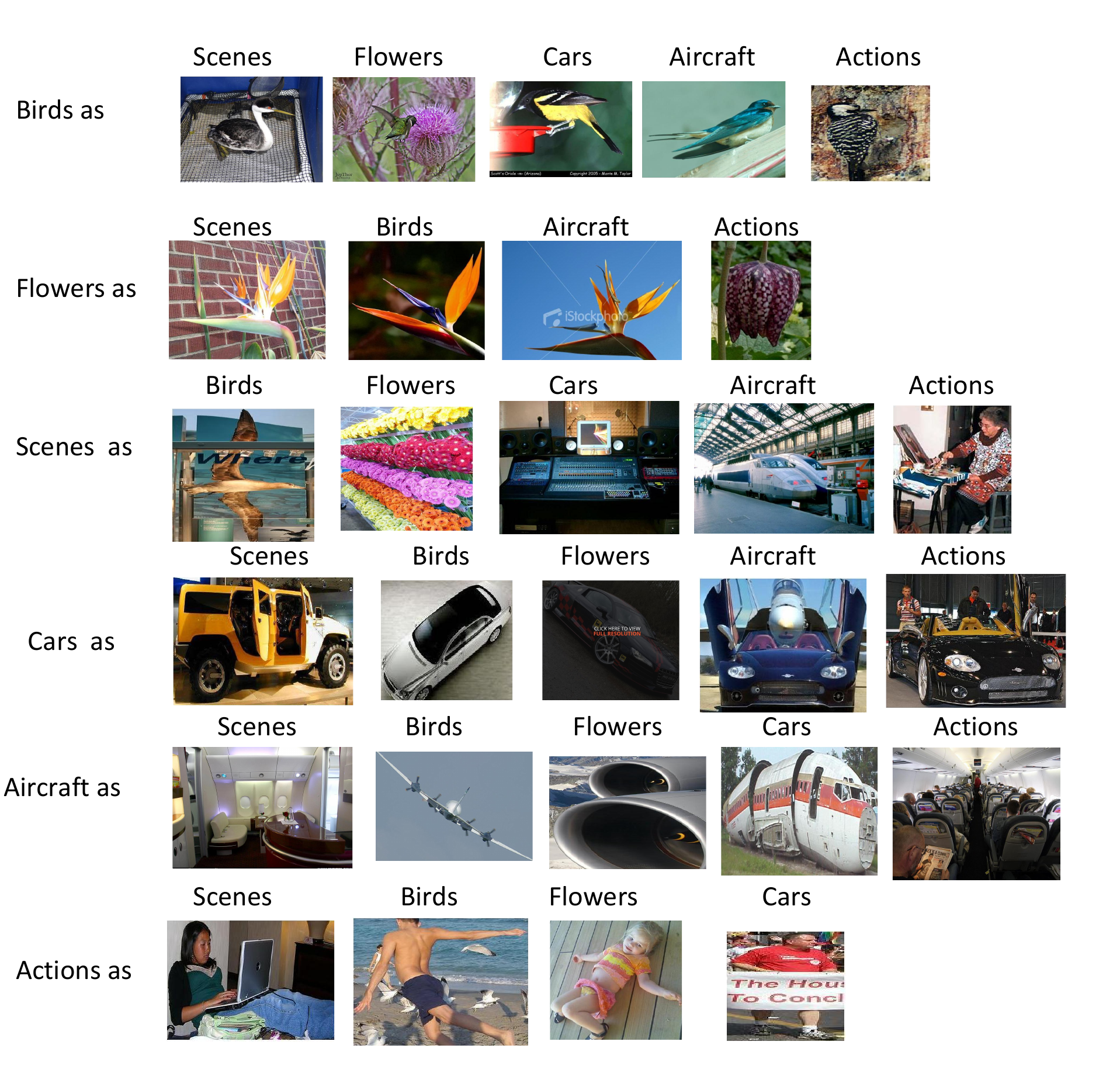}
  \caption{Detailed confusion cases that occurred using Expert Gate in the six tasks sequence.}
   
  \label{fig:confusion}

\end{figure}

In Figure \ref{fig:confusion}, you can find for each task: an example that has been assigned mistakenly to one of the other tasks. Note  that for some cases, the examples shown are the only mistakes made by our Expert Gate.
For some test samples even humans have a hard time telling which expert should be activated.
Most of these mistakes  are explainable and due to one of the following reasons:
\begin{itemize}
\item the image  contains objects from two different tasks and our Expert Gate has to choose one of them.   For example, Scenes images containing humans can also be classified as Actions. To deal with such cases, it may be preferable in some settings to allow more than one expert to be activated. This can be done by setting a threshold on the probabilities for the different tasks. We tested this scenario with a threshold of 0.1 and  observed 
$3.7\%$ of the test samples being analyzed by multiple expert models. Note that in this case we can only evaluate the label given by the corresponding task as we are missing the ground truth for the other possible tasks appearing in the image. This leads to an average accuracy of $68.2\%$, i.e. a further increase of $0.9\%$.
\item the  image  is an outlier w.r.t.  its own task dataset. For example, only a small part of the object appears which means it is not a useful  example or it could even harm the classifier if used in the training phase. This sheds light on another potential use of our  gate, i.e. to detect outliers.  In fact, the autoencoder represents the distribution of  each task data. Thus an outlier that is in the long tail of the task distribution will have a higher reconstruction error. This has the potential of being used in cleaning the annotations of each new task data. 
\item objects from one task that look similar to the images of another task.
\end{itemize}

\subsection{Task Relatedness Analysis}
\label{sec:expres-related}
 \begin{figure}[t]
\centering
\includegraphics[width=0.8\textwidth]{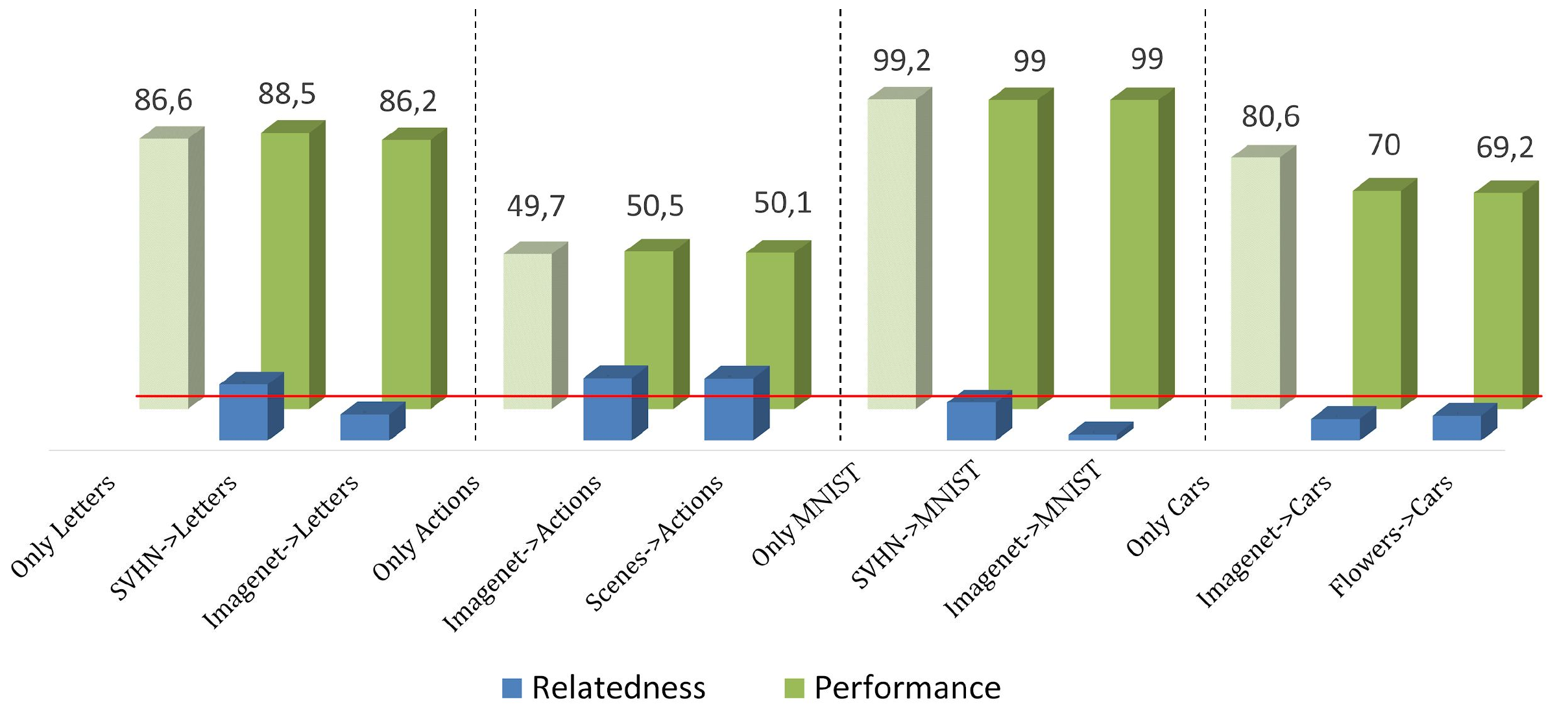}

  \caption[Relatedness analysis.]{Relatedness analysis. The relatedness values are normalized for the sake of better visualization. The red line indicates our relatedness threshold value.}
  \label{fig:relatedness}

  \end{figure}

In the previous cases, the most related task was always ImageNet. This is due to the  similarity between the images of these different tasks and those of ImageNet. Also, the wide diversity of ImageNet classes enables it to cover a good range of these tasks. 
Does this mean that ImageNet should be the only task to transfer knowledge from, regardless of the current task nature? To answer this question, we add three more different tasks to our previous basket: \textit{SVHN},  \textit{Letters}, and \textit{MNIST}.  From the previous set, we pick the two most related tasks, Actions and Scenes, and the two most unrelated tasks, Cars and Flowers. We focus on LwF \cite{li2016learning} as a method for knowledge transfer.  We also consider ImageNet as a possible source.  We consider the following knowledge transfer cases: Scenes $\rightarrow$ Actions, ImageNet $\rightarrow$ Actions, SVHN $\rightarrow$ Letters, ImageNet $\rightarrow$ Letters, SVHN $\rightarrow$ MNIST, ImageNet $\rightarrow$ MNIST, Flowers $\rightarrow$ Cars and ImageNet $\rightarrow$ Cars. 
Figure~\ref{fig:relatedness} shows the  performance of LwF compared to  fine-tuning  the tasks with pre-trained AlexNet (indicated by "Only X")  along with the degree of task relatedness. The red line indicates the threshold of $0.85$ task relatedness used in our previous experiments. 

In the case of a high score for task relatedness,  LwF  uses the knowledge from the previous task and improves  performance on the target task -- see e.g. (SVHN$\rightarrow$Letter, Scenes$\rightarrow$Actions, ImageNet$\rightarrow$Actions). When the tasks are less related, the method fails to improve and starts to degrade its performance, as in (ImageNet$\rightarrow$Letters, SVHN $\rightarrow$ MNIST). When the tasks are highly unrelated, LwF can even fail to reach a good performance for the new task, as in the case of (ImageNet$\rightarrow$ Cars, Flowers$\rightarrow$ Cars). This can be explained by the fact that each task is pushing the shared parameters in a different direction and thus the model fails to reach a good local minimum. We conclude that {\em our gate autoencoder succeeds to predict when a task could help another in the LwF framework and when it cannot}. 
\subsection{Autoencoder Design Choices}

We show the effect of using different design choices for our autoencoder gate. As a test case, we use the 3 sequential learning tasks, namely, Scenes, Birds and Flowers.
We consider the following  alternatives:
\begin{itemize}
\item {\tt Linear autoencoder:} no use of nonlinear activation functions. The loss function is the Euclidean distance. This setup learns the same subspace as PCA.
\item {\tt No standardization:} the same structure of our autoencoder gate but without the standardization  of the input. Also, no fine-tuning is performed here.
\item {\tt Sigmoid activation function:} in our design choice, we use a  ReLU activation function for the  encoding layer. In this baseline, we use the  sigmoid activation function for the encoding as well as decoding layer.
\item {\tt ReLU activation function:} our design choice of using  a  ReLU activation function, here without the initialization of the ImageNet pretrained autoencoder.
\item {\tt Expert Gate:} our autoencoder gate with the full design: standardization step, ReLU hidden activation and ImageNet fine-tuning.
\end{itemize}
 For all the different alternatives, we use the AdaGrad optimizer~\cite{duchi2011adaptive} for the training of the autoencoder.
\begin{table}[h]

\centering
\begin{tabular}{ | l | c| c | c |c|}
\hline
Method & Scenes &Birds & Flowers&Avg. \\  \hline \hline

{\tt Linear Autoencoder}&  93.6\%  & 98.3\%  &35.4\% &75.8\%   \\  \hline
{\tt No standardization} &  97.4\%  &  97.7\%  &97.4\% &97.5\% \\ \hline 

{\tt Sigmoid activation function} &  97.3\%  &  98.4\%  &97.4\%  &97.7\% \\   \hline
{\tt ReLU activation function} & 97.6\%  &98.4\% &97.4\%  &97.8\%  \\ \hline

{\tt Expert Gate}  &  99.4\%  & 99.2\%  &99.2\%  &99.3\%  \\ \hline
\end{tabular}
\caption{\label{tab:ablation} Comparison of different autoencoder designs: classification accuracy of the autoencoders for the sequential learning of 3 image classification tasks.}
\end{table}

Table \ref{tab:ablation} shows the classification accuracy for the task labeling problem achieved by each of the different choices and our Expert Gate autoencoder. It can be noticed that the linear gate ({\tt  Linear Autoencoder}) fails to recognize the examples from the Flowers dataset -- the linearly learned subspace might not be different from the subspace learned for Birds (due to the visual similarity). There is a slight relative improvement between the gate with the standardization ({\tt ReLU activation function} and {\tt Sigmoid activation function}) and without ({\tt No standardization}).
Lastly, our design choice along with finetuning after the autoencoder learned on ImageNet achieves the highest accuracy in recognizing the three different tasks.

\subsection{Video Prediction}
\label{sec:expres-video}
Next, we evaluate our Expert Gate for video prediction in the context of autonomous driving. 
For video prediction, we use the Dynamic Filter Network (DFN) \cite{de2016dynamic}. Given a sequence of 3 images, the task for the network is to predict the next 3 images. This is quite a structured task, where the task environment and training data affect the prediction results quite significantly. 
An autonomous vehicle that uses video prediction needs to be able to load the correct model for the current environment. It might not have all the data from the beginning, and so it becomes important to learn specialists for each type of environment, without the need for storing all the training data. Even when all data are available, joint training does not give the best results on each domain, as we show below. 

We show experiments conducted on three domains/tasks:
for \textit{Highway}, we use the data from DFN \cite{de2016dynamic}, with the same train/test split;  for \textit{Residential} data, we use the two longest sequences 
from the KITTI dataset \cite{geiger2013vision}; and for  \textit{City} data, we use the Stuttgart sequence from the CityScapes dataset \cite{Cordts2016Cityscapes}, i.e. the only sequence in that dataset with densely sampled frames. We use a 90/10 train/test split on both residential and city datasets. 
We train the 3 tasks using 3 different regimes: sequential training using a {\tt Single Fine-tuned Model}, {\tt Joint Training} and {\tt Expert Gate}. 
For video prediction, LwF does not seem applicable as the previous and current task share the same output space. In this experiment, we use the autoencoders only as  gating function. We do not use task relatedness.
Video prediction results are expressed as the average pixel-wise $\ell_1$-distance between predicted
and ground truth images (lower is better), and shown in table \ref{tab:JntSeqTrainingVideoPred}. 

Similar trends are observed as for the image classification problem: sequential fine-tuning (~{\tt Single Fine-tuned Model}) results in catastrophic forgetting, where a model fine-tuned on a new dataset deteriorates on the original dataset after fine-tuning. {\tt Joint Training} leads to better results on each domain, but requires all the data for training. Our Expert Gate system gives better results compared to both {\tt Single Fine-tuned Model} and {\tt Joint Training}. 
These experiments show the potential of our Expert Gate system for video prediction tasks in autonomous driving applications.
%
\begin{table}
\footnotesize
\centering

\begin{tabular}{|l|c |c |c||c|}
    \hline
    Method &Highway&Residential&City &Avg.\\ \hline 
     {\tt Joint Training*} &14.0& 40.7&16.9& 23.8\\ \hline
    {\tt Single Fine-tuned Model} & 13.4&-&-&- \\ 
    					& 25.7&45.2 &-&- \\
    & 26.2& 50.0&17.3&31.1\\ \hline
    
     {\tt Expert Gate (ours)} &\textbf{ 13.4}& \textbf{40.3}&\textbf{16.5}&\textbf{23.4}\\ \hline
\end{tabular}
\caption[Video prediction results.]{ Video prediction results (average pixel L1 distance, lower is better). For methods with * all the previous data needs to be available.}
\label{tab:JntSeqTrainingVideoPred}
\end{table}

%

\paragraph{Video Prediction - Qualitative Results}
\begin{figure*}[h!]
\centering
\includegraphics[width=0.9\textwidth]{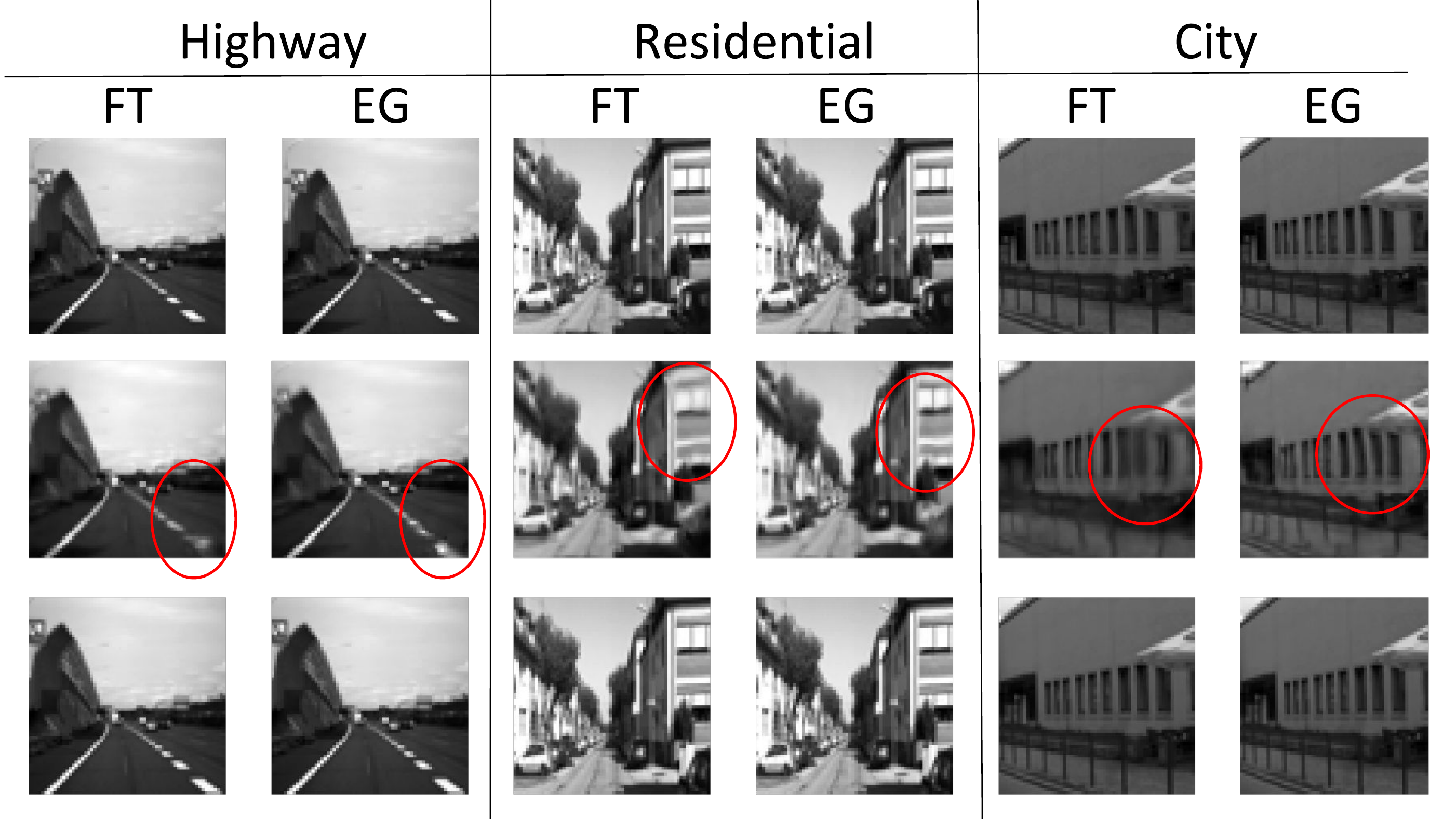}
\caption[Video prediction qualitative results.]{Video prediction on the Highway, Residential and City datasets from left to right (two columns for each dataset) using {\tt Single Fine-tuned Model}~(first column) and our {\tt Expert Gate}~(second column).}
   \label{fig:qualResultsVideoPred1}
\end{figure*}

Figure \ref{fig:qualResultsVideoPred1} shows qualitative results  using sequential fine-tuning and Expert Gate. The first row shows the first in a sequence of 3 images input from the highway, residential and city datasets, to {\tt Single Fine-tuned Model} and {\tt Expert Gate}   respectively. 
The second row contains the last of the three images predicted by the 2 systems, again for the 3 datasets. Ground truth predictions are shown on the last row. It can be seen that {\tt Expert Gate} gives consistently superior qualitative results in the 3 datasets.
\section{Summary}
\label{conclusions}
In the context of incremental task  learning, most work has focused on reducing forgetting and on how to exploit knowledge from previous tasks and transfer it to a new task. Little attention has gone to the related and equally important problem of how to select the proper (i.e. most relevant) model at test time. 
To the best of our knowledge, we have been the first to propose a solution that does not require storing data from previous tasks. Surprisingly, Expert Gate autoencoders can distinguish different tasks equally well as a discriminative classifier trained on all data. 
Moreover, they can be used to select the most related task and the most appropriate transfer method during training. 
Combined, this gives us a powerful method for incremental task learning, that outperforms not only the state-of-the-art 
but also joint training of all tasks simultaneously. 

Later research works on continual learning still rarely consider this problem and rely on an imaginary  task oracle to decide at test time the correct model/head. 
In industry, we have been informed through confidential personal communication that our system has been deployed in a
 big industrial  project with over 80 domains of different machine settings.
 
Our Expert Gate allows for an optimal performance on each learned task without having  access to previous data. However, it still learns a new task model and autoencoder per task which leads to a linear increase in storage  requirements. In the next chapters, we discuss how to sequentially learn multiple tasks using one model.
%


\cleardoublepage%

  \graphicspath{{\chaptersdir/Encoder_Based_Lifelong_Learning/image/}}%
  \chapter{Continual Learning with a Fixed Model Capacity based on Autoencoders}\label{ch:ebll}
In the previous chapter, we opted for an optimal performance in each seen task while tolerating a linear increase in the number of trained models. We  argued that storing the trained models is much cheaper than storing the training datasets and more secure. In this chapter, we consider  a different trade-off. We approach continual learning using one fixed capacity training model, again without assuming  access to previous tasks data. This requires  developing a mechanism that balances the stability-plasticity trade-off explained in Chapter~\ref{ch:introduction}.
Close to our  work described in the previous chapter,  we  rely on autoencoders. However, here we use them to  preserve the knowledge acquired from  previous tasks while learning a new one. After each task, an under-complete autoencoder is learned, capturing the features that are crucial for this task  achievement. When a new task is presented to the system, we prevent the reconstructions of the features with these autoencoders from changing, which has the effect of preserving the information on which the previous tasks are mainly relying. At the same time, the features are given space to adjust to the most recent environment as only their projection into a low dimensional submanifold is controlled. The proposed system is evaluated on image classification tasks and shows a reduction of forgetting over state of the art at the time of development.

This is a joint work with Amal Rannen. I  worked on the initial ideas starting from Expert Gate (see Chapter~\ref{ch:expertgate}). Then Amal joined and we started working closely together on further developing the approach and deploying it under the task incremental setting. In the last weeks of this work, I focused on empirical validation while Amal worked on  providing theoretical  justification for the proposed approach. This work was published as an article in ICCV 2017~\cite{rannen2017encoder}.

\section{Introduction}
In this chapter, we consider the task incremental setting where tasks associated with their training data are received sequentially and, importantly, no access to previous tasks training data can be gained. In spite of the strict task incremental assumption, this scenario occurs frequently in real applications.  For instance, imagine an agent that was trained to localize the defects on a set of factory products. Then, when new products are introduced and the agent has to learn to detect the anomalies in these new products, training images of the previous products might not be available any more and a fine-tuning on the new products images would cause  a forgetting of the previous products. Another example is medical imaging applications where user-data is important to be analysed privately. For example, a tumor recognition model might be trained on  MRI-images data from one clinic and deployed in different clinics. Now if one clinic wants to improve the model or extend it to other modalities, this must be done without access to the previous data.  
%

The  main challenge is to make the learned model adapt to new data from a similar or a different environment~\cite{pentina2015lifelong}, without losing knowledge on the previously seen task(s). Most of the classical solutions for this challenge suffer from important drawbacks. Feature extraction (as in~\cite{donahue2014decaf}), where the model / representation learned for the old task is re-used to extract features from the new data without adapting the model parameters, preserves all previous tasks information but fails to optimally learn the new ones unless tasks are very similar. Fine-tuning (as in~\cite{girshick2014rich}),  adapts the model to the new task using the optimal parameters of the old task as initialization. As a result, the model is driven towards the newly seen data but forgets what was learned previously. Joint training 
 as we showed in  Chapter~\ref{ch:expertgate}, converges to the best compromise between tasks, but requires the presence of all the data at the same time.

To overcome these drawbacks without the constraint of storing data from the previously seen tasks, yet with a constant model capacity, methods usually impose  an additional regularizer to preserve previous tasks knowledge while learning a new one.  Such a regularizer is exploited to 
\begin{inparaenum}[(i)]
\item preserve the performance on the previously seen data, 
\item improve this knowledge using inductive bias~\cite{mitchell1980need} from the new task data, and
\item regularize the training of the new task, which can be beneficial for the performance.
\end{inparaenum}

{\em In this chapter we aim at preserving the knowledge of the previous tasks and possibly benefiting from this knowledge while learning a new task, without storing data from previous tasks using a fixed model capacity. }

\textbf{Learning without forgetting} (LwF)~\cite{li2016learning},  was the first work to re-introduce the data focused regularization approach after the recent success of neural networks. As explained in Chapter ~\ref{ch:expertgate}, Section~\ref{sec:expres-related}, it
proposes to preserve the previous performance through the knowledge distillation loss introduced in~\cite{hinton2015distilling}. It considers a shared convolutional network between the different tasks in which only the last classification layer is task specific. When encountering a new task, the outputs of the existing classification layers given the new task data are recorded. During training, these outputs are preserved through a modified cross-entropy loss that softens the class probabilities in order to give a higher weight to the small outputs. More details about this loss and the method can be found in Section~\ref{sec:ebll_LWF}. This method reduces the forgetting, especially when the datasets come from related manifolds. Nevertheless, it has been shown by { iCaRL}~\cite{rebuffi2016icarl} that LwF suffers from a build up of errors in a sequential scenario where the data comes from the same environment. Similarly, we showed in the previous chapter that   LwF performance drops when the model is exposed to a sequence of tasks drawn from different distributions. 
{ iCaRL}~\cite{rebuffi2016icarl} proposes to store a selection of the previous tasks data to overcome this issue -- something we try to avoid. 

In the work presented in this chapter, we propose a compromise between these two methods. Rather than heavily relying on the new task data or requiring to store samples from the previous tasks, we introduce the use of autoencoders as a tool to preserve the knowledge from one task while learning another. For each task, an undercomplete autoencoder is trained after training the task model. It captures the most important features for the task objective. When facing a new task, this autoencoder is used to ensure the preservation of those important features. This is achieved by defining a loss on the reconstructions made by the autoencoder, as we will explain in the following sections. In this manner, we only restrict a subset of the features to be unchanged while we give the model the freedom to adapt itself to the new task using the remaining capacity.

Below, in Section~\ref{sec:ebll_Main}, we describe how to use the autoencoders to avoid catastrophic forgetting and motivate our choice by a short analysis that relates the proposed objective to the joint training  scheme. In Section~\ref{sec:ebll_Exp}, we describe the experiments that we conducted, report and discuss their results before concluding in Section~\ref{sec:ebll_Conclusion}.

\section{Overcoming Forgetting with Autoencoders}\label{sec:ebll_Main}

When training one model on a sequence of tasks, the best compromise performance between  all the tasks simultaneously is achieved when the network is trained on the data from all the considered tasks at the same time (as in joint training). This performance is of course limited by the capacity of the used model, and can be considered an upper bound to what can be achieved in an task incremental  setting, where the data of previous tasks are no longer accessible when learning a new one.
\subsection{Joint Training}
In the following, we will use $Q_t$ to denote  the distribution from which the dataset $D_t$ of the task $T_t$ is sampled, and $x_n^{(t)}$ and $y_n^{(t)}$ for the data samples (input and output labels).
When we have access to the data from all $\mathcal{T}$ tasks jointly, the network training aims to control the statistical risk:
\begin{equation}
\sum_{t=1}^\mathcal{T} \mathbb{E}_{Q_t(x^{(t)},y^{(t)})}[\ell(f_t(x^{(t)}),y^{(t)})],
\label{eq:StatRisk}
\end{equation}
by minimizing the empirical risk:
\begin{equation}
\sum_{t=1}^\mathcal{T} \frac{1}{N_t} \sum_{n=1}^{N_t} \ell(f_t(x_n^{(t)}),y_n^{(t)}),
\label{eq:EmpRisk}
\end{equation}
where $N_t$ is the number of samples and $f_t$ the function implemented by the network for task $t$. For most of the commonly used models, we can decompose $f_t$ as $O_t\circ O \circ F$ where:
\begin{itemize}
\item $F$ is a feature extraction function (e.g.\ Convolutional layers in ConvNets)
\item $O_t\circ O$ is a task operator. It can be for example a classifier or a segmentation operator. $O$ is shared among all tasks, while $O_t$ is task specific. (e.g.\ in AlexNet, $O_t$ could be the last fully-connected layer, and $O$ the remaining fully-connected layers.)
\end{itemize}

\begin{figure}[t]
\centering
\includegraphics[width =\columnwidth]{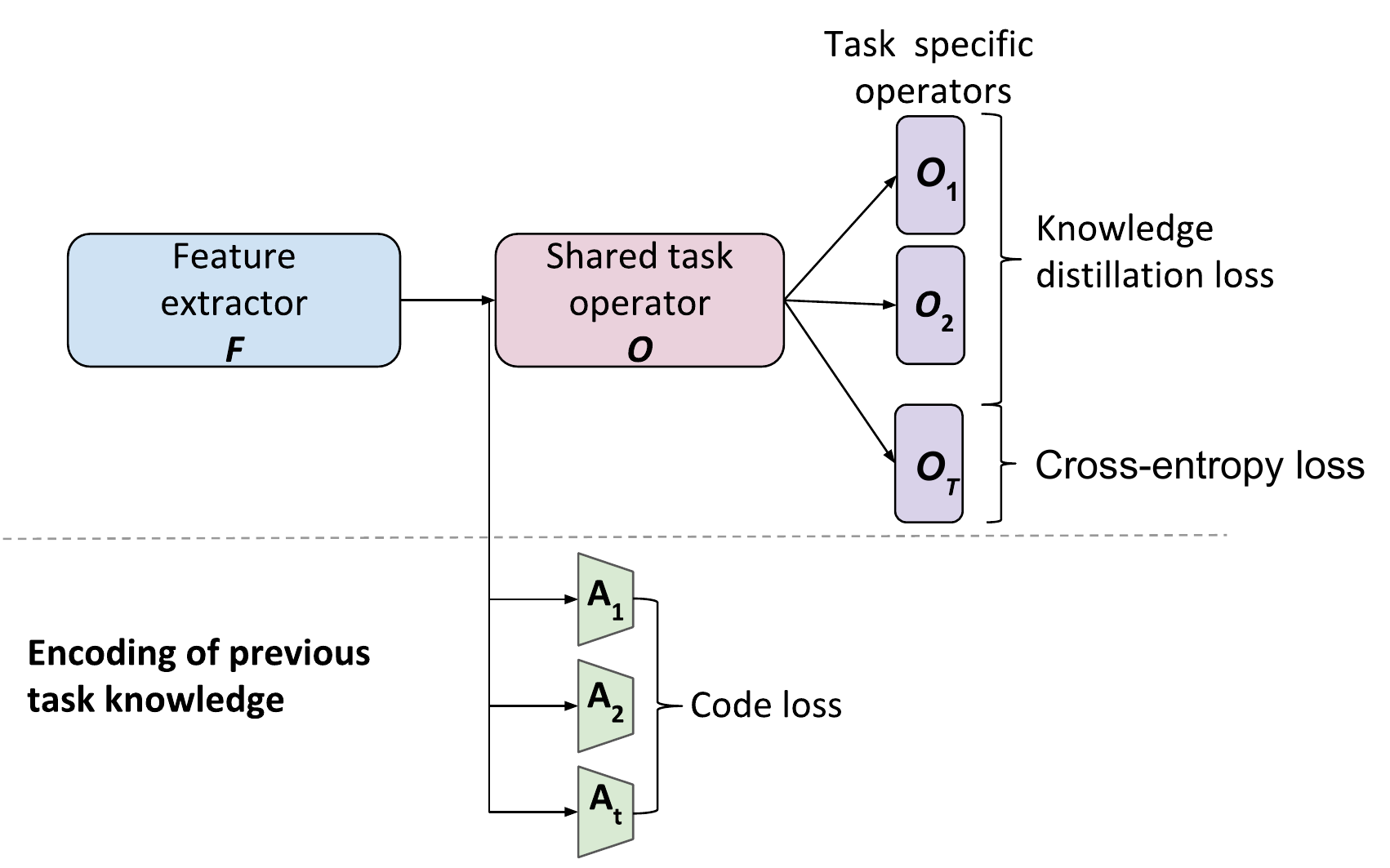}
\caption[Diagram of our encoder based lifelong learning model.]{Diagram of the proposed model.  Above the dotted line are the model components that are retained during test time, while below the dashed line are components necessary to our improved training scheme.}
\label{fig:ebll_Global_model}
\end{figure}

The upper part of Figure~\ref{fig:ebll_Global_model} gives a scheme of this general model. For simplicity, we will focus below on two-task training before generalizing to a multiple task scenario in section~\ref{sec:ebll_Proc}. 


\subsection{Shortcomings of Learning without Forgetting}\label{sec:ebll_LWF}

As a first step, we want to understand the limitations of LwF~\cite{li2016learning}. In that work, it is suggested to replace in~\Eqn~\ref{eq:StatRisk} $ \ell(O_1\circ\,O\circ\,F(x^{(1)}),~y^{(1)}) $ with $ \ell(O_1\circ\,O\circ\,F(x^{(2)}), O^*_1\circ\,O^* \circ\,F^*(x^{(2)})) $, where $O^*_1\circ\,O^* \circ\,F^*$ is obtained from training the network on the first task.
If we suppose that the model has enough capacity to integrate the knowledge of the first task with a small generalization error, then we can consider that  
\begin{equation}
\mathbb{E}_{Q_1(x)}[\ell(O_1\circ\,O\circ\,F(x^{(1)}),O^*_1\circ\,O^* \circ\,F^*(x^{(1)}))]
\label{eq:LossApprox}
\end{equation} is a reasonable approximation of $\mathbb{E}_{Q_1(x^{(1)},y^{(1)})}[\ell(O_1\circ\,O\circ\,F(x^{(1)}),y^{(1)})]$. However, in order to be able to compute the measure in~\Eqn~\ref{eq:LossApprox} using samples from $x^{(2)}$, further conditions need to be satisfied. 

If we consider that $O_1\circ\,O\circ\,F$ tries to learn an encoding of the data in the target distribution $Q_1(x^{(1)})$, then one can say that the loss of information generated by the use of $Q_2(x^{(2)})$ instead of $Q_1(x^{(1)})$  is a function of the Kullback-Leibler divergence of the two related probability distributions.

In this work, we build on top of the LwF method. In order to make the used approximation less sensitive to the data distributions, we see an opportunity in controlling $\|O_1\circ\,O\circ\,F(x^{(1)})-O_1\circ\,O\circ\,F(x^{(2)})\|$. Under mild conditions about the model functions, namely Lipschitz continuity, this control allows us to use $O_1\circ\,O\circ\,F(x^{(2)})$ instead of $O_1\circ\,O\circ\,F(x^{(1)})$ to better approximate the first task loss in~\Eqn~\ref{eq:StatRisk}.  Note that the condition of continuity on which this observation is based is not restrictive in practice. Indeed, most of the commonly used functions in deep models satisfy this condition (e.g.\ sigmoid, ReLU). 

\begin{figure}[t]
\centering
\includegraphics[width = 0.9\columnwidth]{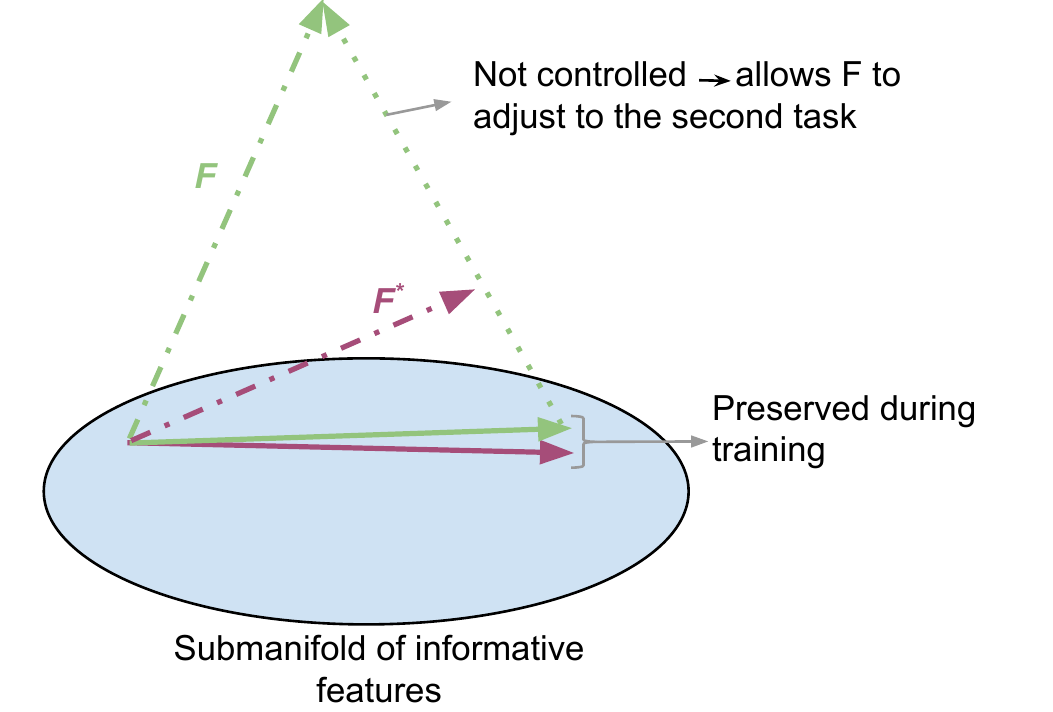}
\caption[Preservation of the features that are important for task $T_1$ while training on task $T_2$.]{Preservation of the features that are important for task 1 while training on task 2. During training, we enforce the projection of $F$ into the submanifold that captures these important features to stay close to the projection of $F^*$, the optimal features for the first task. The part of $F$ that is not meaningful for the first task is allowed to adjust to the variations of the second task.}
\label{fig:ebll_preserve_feat}
\end{figure}

Our main idea is to learn a submanifold of the representation space $F(x^{(1)})$ that contains the most informative features for the first task. Once this submanifold is identified, if the projections of the features  $F(x^{(2)})$ onto this submanifold do not change much during the training of a second task,
then two consequences follow: 
\begin{inparaenum}[(i)]
\item $F(x^{(2)})$ will stay informative for the first task 
during the training, and  
\item at the same time there is room to adjust to the second task 
as only its projection in the learned submanifold is controlled.
\end{inparaenum}
Figure~\ref{fig:ebll_preserve_feat} gives a simplified visualization of this mechanism. In the next section, we propose a method to learn the submanifold of informative features for a given task using autoencoders.





\subsection{Informative Features Preservation}

When beginning to train the second task, the feature extractor $F^*$ of the model is optimized for the first task. A feature extraction type of approach would keep this operator unchanged in order to preserve the performance on the previous task. This is, however, overly conservative, and usually suboptimal for the new task. Rather than preserving \emph{all} the features during training, our main idea is to preserve only the features that are the most informative for the first task while giving more flexibility for the other features in order to improve the performance on the second task. An autoencoder~\cite{bourlard1988auto} trained on the representation of the first task data obtained from an optimized model can be used to capture the most important features for this task. 
\subsubsection{Learning the informative submanifold with Autoencoders} \label{sec:ebll_AE}
\begin{figure}[t]
\centering
\includegraphics[width = \columnwidth]{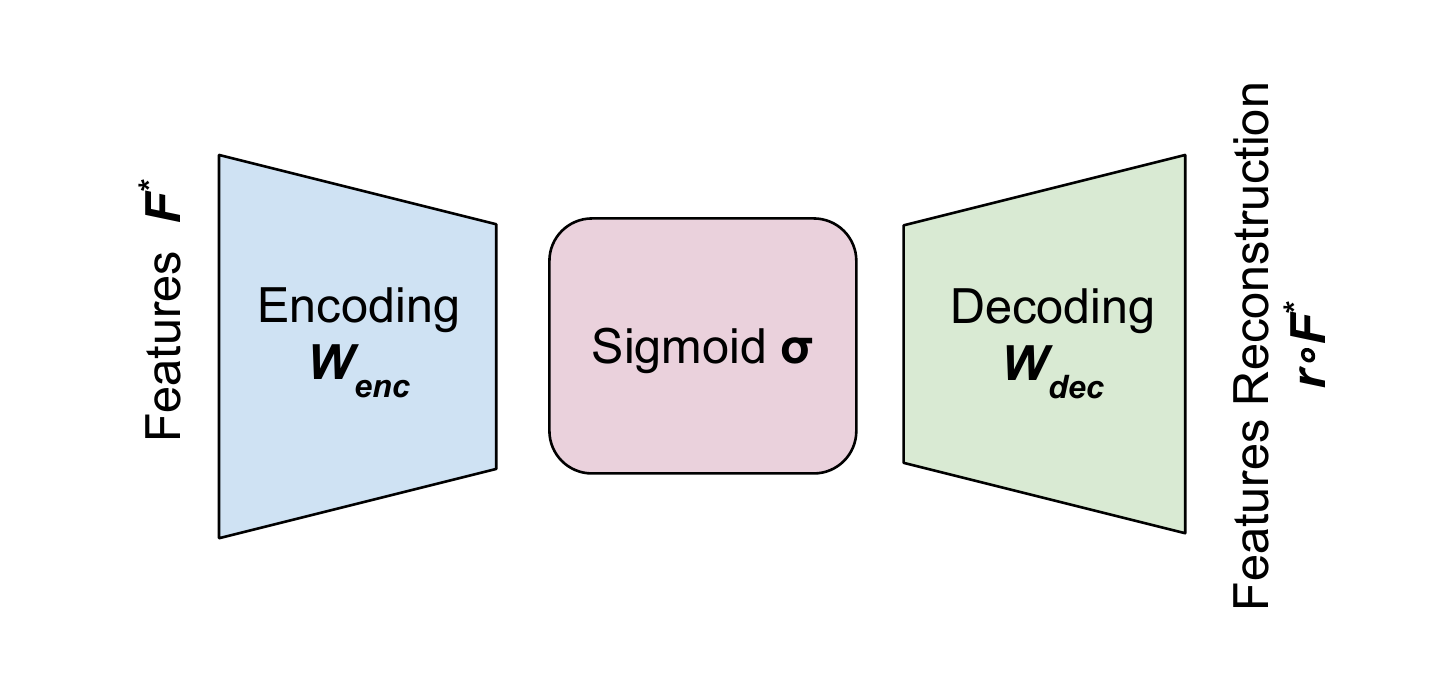}
\caption{Scheme of an undercomplete autoencoder trained to capture the important features submanifold.}
\label{fig:ebll_AE}
\end{figure}

An under-complete autoencoder, where the dimension of the code is smaller than the dimension of the input,  learns a lower dimensional projection of its input (see also Chapter~\ref{ch:background}, Section~\ref{sec:background_autoencoders}  ). The learned  submanifold  represents  best the structure of the input data.  More precisely, we choose to use a two-layer network with a sigmoid activation in the hidden layer: $r(x) = W_{dec}\sigma(W_{enc}x)$. We want the encoding layer to act as a switch where non informative features have a close to zero projection and informative ones are preserved with close to one projection, hence we favored the use of a sigmoid activation function over the ReLU activation function used in Chapter~\ref{ch:expertgate}.  Figure~\ref{fig:ebll_AE} shows a general scheme of such an  autoencoder.

Here, our aim is to obtain through the autoencoder a submanifold that captures the information that is not only important to reconstruct the features (output of the feature extraction operator $F^*$) of the first task, but also important for the task operator ($O^*_1\circ O^*$). We therefore select our training objective as:
\begin{align}
\arg\min_{r} \mathbb{E}_{Q_1(x^{(1)},y^{(1)})}[ &\beta\|r(F^*(x^{(1)}))- F^*(x^{(1)})\|_2  \label{eq:LossAE}\\ 
&+ \ell(O_1^*\circ O^*(r(F^*(x^{(1)}))),y^{(1)}) ],
\nonumber
\end{align}
where $\ell$ is the loss function used to train the model on the first task data.  $\beta$ is a hyper-parameter that controls the compromise between the two terms in this loss. 
In this manner, the autoencoder represents the variations that are needed to reconstruct the input and at the same time contain the information that is required by the task operator.

\subsubsection{Representation control with separate task operators} \label{sec:ebll_intuition}




To explain how we use these autoencoders,
we start with the simple case where there is no task operator shared among all tasks (i.e., $O = \emptyset$). The model is then composed of a common feature extractor $F$, and a task specific operator for each task $O_t$. 
In a two tasks scenario, after training the first task, we have $O_1^*$ and $F^*$  optimized for that task. Then, we train an undercomplete autoencoder using $F^*(x_n^{(1)})$ minimizing the empirical risk corresponding to~\Eqn~\ref{eq:LossAE}. The optimal performance for the first task, knowing that the operator $O_1$ is kept equal to $O_1^*$, is obtained with $F$ equal to $F^*$.

Nevertheless, preventing $F$ from changing will lead to suboptimal performance on the second task. The idea here is to keep only the projection of $F$ into the manifold represented by the autoencoder ($r\circ~F$) unchanged. 
The second term of~\Eqn~\ref{eq:LossAE} explicitly enforces $r$ to represent the submanifold needed for good performance on $T_1$. Thus, controlling the distance between $r\circ F$ and $r\circ F^*$ will preserve the necessary information for $T_1$.
 Since this autoencoder captures the submanifold of the features that is crucial to the task operator $T_1$, preserving the   projection of $F$   would prevent the performance of the model on the first task from dropping.
From the undercompleteness of the encoder, $r$ projects the features into a lower dimensional manifold, and by controlling only the distance between the reconstructions, we give the features flexibility to adapt to the second task variations. Thus, the autoencoder can be used to realize the mechanism described in Figure~\ref{fig:ebll_preserve_feat}.

\subsubsection{Representation control with shared task operator}\label{sec:ebll_Shared}

We now consider the model presented in Figure~\ref{fig:ebll_Global_model} where a part of the task operator is shared among the tasks as in the setting used in LwF~\cite{li2016learning}. 
Our main idea is to start from the loss used in the LwF method and add an additional term coming from the idea presented previously in this section. Thus, in a two tasks scenario, in addition to the loss used for the second task, we propose to use two constraints:
\begin{enumerate}
\item The first constraint is the knowledge distillation loss ($\ell_{dist}$) used in~\cite{li2016learning}. If 
$
\hat{y} = O_1\circ O\circ F(x^{(2)}) 
$
and 
$
y^* = O^*_1\circ O^*\circ F^*(x^{(2)}) 
$
then:
\begin{equation}
\ell_{dist}(\hat{y} ,y^*) = -\langle z^*,\log \hat{z}\rangle
\end{equation}
where $\log$ is operated entry-wise and
\begin{equation}
z^*_i = \frac{y_i^{*1/\tau}}{\sum_jy_j^{*1/\tau}} \text{ and } \hat{z}_i = \frac{\hat{y}_i^{1/\tau}}{\sum_j\hat{y}_j^{1/\tau}} 
\end{equation}
The application of a high temperature $\tau$ increases the small values of the output and reduces the weight of the high values. This mitigates the influence of the use of different data distributions.
\item The second constraint is related to the preservation of the reconstructions of the second task features ($r\circ F^*(x^{(2)})  $). The goal of this constraint is to keep $r\circ F$ close to $r\circ F^*$ as explained previously.
\end{enumerate}
For the second constraint, rather than controlling the distance between the reconstructions, we will here constrain the codes $\sigma(W_{enc}\cdot )$. From sub-multiplicity of the Frobenius norm, we have:
\begin{equation}
\|r(x_1) - r(x_2)\|_2 \le \|W_{dec}\|_F\|\sigma(W_{enc}x_1) - \sigma(W_{enc}x_2)\|_2. \nonumber
\end{equation}
The advantage of using the codes is their lower dimension. As the codes or reconstructions need to be recorded before beginning the training on the second task, using the codes will result in a better usage of the memory.

Finally, this results in minimizing the following  objective function for the training of the second task: 
\begin{align}
J &=\mathbb{E}_{Q_2(x)}\Big[\ell(O_2\circ O\circ  F(x^{(2)}), y^{(2)}) )\nonumber \\
&+  \ell_{dist}(O_1\circ O\circ F(x^{(2)}), O_1^*\circ O^*\circ F^*(x^{(2)})  ) \nonumber  \\
& + \frac{\alpha}{2} \| \sigma(W_{enc}F(x^{(2)})) - \sigma(W_{enc}F^*(x^{(2)})) \|_2^2\Big] \label{eq:CodeLoss}.
\end{align}
The choice of the parameter $\alpha$ will be done through model selection.
 \begin{algorithm}[t]
\caption{Encoder based Lifelong Learning~(\tt EBLL)}\label{encoderbased:algo}
  \begin{algorithmic}[1]
  \Statex \textbf{{Input}} :
  \Statex $F^*$ shared feature extractor; $O^*$ shared task operator; $\left\{O_t\right\}_{t=1..\mathcal{T}-1}$ previous task operators; 
  \Statex $\left\{W_{enc,t}\right\}_{t=1..\mathcal{T}-1}$ previous task encoders;
  \Statex $(X^{(\mathcal{T})}, Y^{(\mathcal{T})})$ training data and ground truth of the new task $\mathcal{T}$;
  \Statex $\alpha_t$ and $\beta$ hyper parameters
  \Statex \textbf{{Initialization}} :
   \For{$ t \in \{1,\mathcal{T}-1 \}$} 
	\State  $Y_t^* = O_t^*\circ O^*\circ F^*(X^{(\mathcal{T})})$	   //record task targets  
    \State $C_t^* = \sigma(W_{enc,t}F^*(X^{(\mathcal{T})}))$   //record new data codes
    \EndFor
  \State $O_\mathcal{T} \leftarrow$ Init($|Y^{(\mathcal{T})}|$) // initialize new task operator
  
  \Statex \textbf{{Training}} :
\State $ O_t^*, O^*, F^* \leftarrow \arg\min_{O_t, O, F}\Big[\ell(O_\mathcal{T}\circ O\circ F(X^{(\mathcal{T})}), Y^{(\mathcal{T})}) $
\Statex $+  \sum_{t=1}^{\mathcal{T}-1} \ell_{dist}(O_t\circ O\circ F(X^{(\mathcal{T})}), Y_t^* ) +\sum_{t=1}^{\mathcal{T}-1}\frac{\alpha_t}{2} \|  \sigma(W_{enc,t}F(X^{(\mathcal{T})})) -  C_t^*\|_2^2\Big]$
   \State $(W_{enc,\mathcal{T}},W_{dec,\mathcal{T}}) \leftarrow$ autoencoder( $O_\mathcal{T}^*, O^*, F^*,X^{(\mathcal{T})}, Y^{(\mathcal{T})}; \beta$ ) // minimizes~\Eqn~\ref{eq:LossAE}

  \end{algorithmic}
 
  \end{algorithm}
\subsection{Training Procedure}\label{sec:ebll_Proc}
The proposed method in Section~\ref{sec:ebll_Shared} generalizes easily to a sequence of tasks. An autoencoder is then trained after each task. Even if the needed memory will grow linearly with the number of tasks, the memory required by an autoencoder is a small fraction of that required by the global model. For example, in the case of AlexNet as a base model, an autoencoder comprises only around 1.5\% of the memory. Figure~\ref{fig:ebll_Global_model} displays the model we propose to use. 

In practice, at task $T_\mathcal{T}$ the empirical risk defined by the  objective function in Equation~\ref{eq:ebll-MultiEmp} is minimized. 
\begin{align}
J&= \frac{1}{N_\mathcal{T}} \sum_{n=1}^{N_\mathcal{T}}\Biggl( \ell(O_\mathcal{T}\circ O\circ F(x_n^{(\mathcal{T})}), y_n^{(\mathcal{T})}) \nonumber \\
&+ \sum_{t=1}^{\mathcal{T}-1} \ell_{dist}(O_t\circ O\circ F(x_n^{(\mathcal{T})}), O_t^*\circ O^*\circ  F^*(x_n^{(\mathcal{T})})  ) \nonumber  \\
& +\sum_{t=1}^{\mathcal{T}-1}\frac{\alpha_t}{2} \| \sigma(W_{enc,t}F(x_n^{(\mathcal{T})})) - \sigma(W_{enc,t}F^*(x_n^{(\mathcal{T})})) \|_2^2\Biggr). \label{eq:ebll-MultiEmp}
\end{align}
The training is done using stochastic gradient descent (SGD)~\cite{bottou2010large}.
The autoencoder training is also done by SGD but with an adaptive gradient method, AdaDelta~\cite{zeiler2012adadelta} which alleviates the need for setting the learning rates and has nice optimization properties.
Algorithm \ref{encoderbased:algo} shows the main steps of the proposed method.

\section{Experiments}\label{sec:ebll_Exp}
We  compare our method against several baselines on image classification tasks.
%
We consider sets of 2, 3 and 5 tasks learned sequentially, in two settings: 1) when the first task is a large dataset, and 2) when the first task is a small dataset.

\paragraph{Architecture}
Similarly to the experiments performed in the previous chapter, we deploy AlexNet \cite{krizhevsky2012imagenet} as a backbone architecture due to its widespread use and similarity to other popular architectures. The feature extraction block $F$ corresponds to the convolutional layers. By default, the shared task operator $O$ corresponds to all but the last fully connected layers (i.e., $fc6$ and $fc7$), while the task-specific part $O_t$ contains the last classification layer ($fc8$). Other choices for $F$ and $O$ are possible. 
During the training, we used an $\alpha$ of $10^{-3}$ for ImageNet and $10^{-2}$ for the rest of the tasks. Note that this parameter sets the trade off between the allowed forgetting on the previous task and the performance on the new task.

For the autoencoders, we use a very shallow architecture, to keep their memory footprint low. Both the encoding as well as the decoding consist of a single fully connected layer, with a sigmoid as non-linearity in between. The dimensionality of the codes is 100 for all datasets, except for ImageNet where we use a code size of 300. The size of the autoencoder is 3MB, compared to 250MB for the size of the network model. The training of the autoencoders is done using AdaDelta as explained in Section~\ref{sec:ebll_Proc}. During training of the autoencoders, we use a hyperparameter $\beta$ (cf.~\Eqn~\ref{eq:LossAE}) to find a compromise between the reconstruction error and the classification error. This parameter is tuned manually and is set to $10^{-6}$ in all cases.

\paragraph{Datasets}\label{sec:ebll_dataset}
We use multiple datasets of moderate size: MIT \textit{Scenes} \cite{quattoni2009recognizing}, Caltech-UCSD \textit{Birds}~\cite{WelinderEtal2010}, Oxford \textit{Flowers}~\cite{Nilsback08} and VOC \textit{Actions}~\cite{pascal-voc-2012}. 

For the scenario based on a large initial dataset, we start from ImageNet~\cite{russakovsky2015imagenet}, which has more than 1 million training images.  For the small dataset scenario, we start from Oxford \textit{Flowers}, which has only 2,040 training and validation samples. 

The reported results are obtained with respect to the test sets of Scenes, Birds, Flowers and Actions, and on the validation set of ImageNet. 
As in LwF~\cite{li2016learning}, we need to record the targets corresponding to the old tasks before starting the training procedure for a new task. Here, we perform an offline augmentation with 10 variants of each sample (different crops and flips).
\paragraph{Compared Methods}
We compare our method 
 with Learning without Forgetting ({\tt LwF})~\cite{li2016learning}, which represents the  state-of-the-art at the time of development~(2017). 
Additionally, we consider two baselines: {\tt Finetune}, where each model (incl. $F$ and $O$) is learned for the new task using the previous task model as initialization, and {\tt Feature extraction}, where the weights of the previous task model ($F$ and $O$) are fixed and only the classification layer ($O_t$) is learned for each new task. We coin our method  ({\tt EBLL}) as short for Encoder Based Lifelong Learning which is the name of the published article. We also report results for a variant of our method, 
{\tt EBLL-separateFCs} where we only share the representation layers ($F$) while each task has its own fully connected layers (i.e., $O = \emptyset$ and $O_t = \{fc6-fc7-fc8\}$). This variant aims at finding a universal representation for the current sequence of tasks while allowing each task to have its own fully connected layers. With less sharing, the risk of forgetting is reduced, at the cost of a higher memory consumption and less regularization for new tasks. Note that in this case the fully connected layers of the previous tasks are not retrained, and thus there is no need to use the knowledge distillation loss. Moreover, task autoencoders can be used at test time to activate only the fully connected layers of the task that a test sample belongs to, in a similar manner to what we did in the previous chapter.

\paragraph{Setup}\label{sec:ebll_setup}
We consider sequences of 2, 3 and 5 tasks. In the {\bf Two Tasks} setup, we are given a model trained on one previously seen task and then add a second task to learn. This follows the experimental setup of LwF~\cite{li2016learning}. In their work, all the tested scenarios start from a large dataset, ImageNet. Here we also study the effect of starting from a small dataset, Flowers.\footnote{Due to the small size of the Flowers dataset, we use a network pretrained on ImageNet as initialization for training the first task model. The main difference hence lies in the fact that in this case we do not care about forgetting ImageNet.}

Further, we also consider a setup involving {\bf Three  Tasks}. First, we use a sequence of tasks starting from ImageNet, i.e.~ImageNet $\rightarrow$ Scenes $\rightarrow$ Birds. Additionally, we consider Flowers as a first task in the sequence Flowers $\rightarrow$ Scenes $\rightarrow$ Birds. Note that this is different from what was conducted  in \cite{li2016learning} where the sequences  were only composed of  splits of  one dataset  i.e.~one task overall. Finally, for a stronger validation of our method, we also test on a longer sequence:  ImageNet $\rightarrow$ Scenes $\rightarrow$ Birds $\rightarrow$ Flowers $\rightarrow$ Actions.

\begin{table*}[t!]
\centering
\resizebox{\textwidth}{!}{
\tabcolsep=0.11cm
\small
\begin{tabular}{|l|ll|ll||ll|ll||ll|ll||l|l|}
\hline
& \multicolumn{4}{|c|}{ImageNet $\rightarrow$ Scenes} &   \multicolumn{4}{c|}{ImageNet $\rightarrow$ Birds} & \multicolumn{4}{c|}{ImageNet $\rightarrow$ Flowers} & \multicolumn{2}{c|}{Average} \\ \hline
& \multicolumn{2}{c|}{Acc. on $T_1$} & \multicolumn{2}{c|}{Acc. on $T_2$} & \multicolumn{2}{c|}{Acc. on $T_1$} & \multicolumn{2}{c|}{Acc. on $T_2$} & \multicolumn{2}{c|}{Acc. on $T_1$} & \multicolumn{2}{c|}{Acc. on $T_2$} & Forg. on $T_1$ & Drop on $T_2$ \\ \hline
{\tt Finetune}& 48.0 &(-9) & 65.0 &(ref)	&  41.3 & (-15.7) 	&  59.0 & (ref)	& 50.8 & (-6.2) & 86.4 & (ref) &10.3 & 0 \\ 
&& &  &	&  &  	&  & 	& &  &  &  & &  \\ \hline
{\tt Feature} & 57.0 & (ref) & 60.6 &(-4.4)	&   57.0 & (ref) 		& 51.6 & (-7.4) & 57.0 & (ref)	& 84.6 & (-1.8) & 0 & 4.5\\ 
{\tt extraction}&& &  &	&  &  	&  & 	& &  &  &  & &  \\ \hline
{\tt LwF}  & 55.4 &(-1.6)              & 65.0 &(-0)	&  54.4 & (-2.6) 	&  58.9 & (-0.1)& 55.6 & (-1.4) & 85.9 & (-0.5) & 1.9 & 0.2\\
&& &  &	&  &  	&  & 	& &  &  &  & &  \\ \hline
{\tt EBLL} & 56.3 & (-0.7)       & 64.9 &(-0.1) &  55.3 & (-1.7) 	&  58.2 & (-0.8)& 56.5 & (-0.5) & 86.2 & (-0.2) & 1.0 & 0.4  \\ 
&& &  &	&  &  	&  & 	& &  &  &  & &  \\ \hline
{\tt EBLL}  & 57.0 & (-0) & 65.9 &(+0.9)&  57.0 & (-0)	&  57.7 & (-1.3)& 56.5 & (-0.5) & 86.4 & (-0)   & 0.2 & 0.1  \\
{\tt separate FCs}&& &  &	&  &  	&  & 	& &  &  &  & &  \\ \hline
\end{tabular}}
\caption[Classification accuracy (\%)for the Two Tasks scenario starting from ImageNet.]{Classification accuracy (\%)   for the Two Tasks scenario starting from ImageNet. For the first task, the reference performance is given by Feature extraction. For the second task, we consider {\tt Finetune} as the reference as it is the best that can be achieved by one task alone.}
\label{tab:res_twotasks_imagenet}
\end{table*}
\begin{table*}[t!]
\centering
\resizebox{\textwidth}{!}{
\tabcolsep=0.11cm
\small
\begin{tabular}{|l|ll|ll||ll|ll||l|l|}
\hline
& \multicolumn{4}{|c|}{Flowers $\rightarrow$ Scenes} &   \multicolumn{4}{c|}{Flowers $\rightarrow$ Birds}  & \multicolumn{2}{c|}{Average}   \\ \hline
& \multicolumn{2}{c|}{Acc. on $T_1$} & \multicolumn{2}{c|}{Acc. on $T_2$} & \multicolumn{2}{c|}{Acc. on $T_1$} & \multicolumn{2}{c|}{Acc. on $T_2$} & Forg. on $T_1$ & Drop on $T_2$\\ \hline
{\tt Finetune}		& 61.6 & (-24.8)	& 63.9 & (ref) 	& 66.6      & (-19.8)		 & 57.5 & (ref) &22.3 &0\\ 
{\tt Feature extraction}  & 86.4 & (ref) 		& 59.6 & (-4.3) &  86.4 & 0 & 48.6 & (-8.9) &0 &6.6\\ 
{\tt LwF } 				& 83.7 & (-2.7) 	& 62.2 & (-1.7)	&  82.0 & (-4.4) &  52.2 & (-5.3)  & 3.6&3.5\\ 
{\tt EBLL} 			& 84.9 & (-1.5) 	& 62.3 & (-1.6) &  83.0 & (-3.4) & 52.0  & (-5.5) & 2.4&3.5\\ 
{\tt EBLL-separateFCs} & 86.4 & (-0) 	& 63.0 & (-0.9) &  85.4 & (-1.0) & 55.1 & (-2.4) &0.5&1.6\\ \hline
\end{tabular}}
\caption[Classification accuracy (\%)   for the Two Tasks scenario starting from Flowers.]{Classification accuracy (\%)   for the Two Tasks scenario starting from Flowers. For the first task, the reference performance is given by Feature extraction. For the second task, we consider {\tt Finetune} as reference as it is the best that can be achieved by one task alone.}
\label{tab:res_twotasks_flower}
\end{table*}

\paragraph{Results}
Table \ref{tab:res_twotasks_imagenet} shows, for the different compared methods, the achieved performance on the Two Tasks scenario with ImageNet as the first task.  While {\tt Finetune} is optimal for the second task, it shows the most forgetting of the first task. The performance on the second task is on average comparable for all methods except for {\tt  Feature extraction}. Since the {\tt Feature extraction} baseline doesn't allow the weights of the model to change and only optimizes the last fully connected layers, its performance on the second task is suboptimal and significantly lower than the other methods. Naturally, the performance of the previous task is kept unchanged in this case.
{\tt EBLL-separateFCs} shows  high performance on both tasks, with 
the performance of the second task being comparable or better to the methods with shared FCs. This variant of our  method has a higher capacity as it allocates separate fully connected layers for each task, yet its memory consumption increases more rapidly as tasks are added, a severe drawback. Our method with a complete shared model {\tt EBLL} systematically outperforms the {\tt LwF} method on the previous task and on average achieves a similar performance on the second task. 
\begin{table}[t!]
\centering
\tabcolsep=0.11cm
\small
\begin{tabular}{|l|lll|c|c|}
\hline
& ImageNet & Scenes & Birds & Avg. Accuracy & Avg. Forgetting   \\

  \hline
{\tt Finetune}& 37.5\% &45.6\% & \textbf{ 58.1}\% &47.2\%&13.0\%\\  
{\tt LwF} & 53.3\% & 63.5\% & 57.2 \%& 58.0\% & 1.7\% \\
{\tt EBLL} & \textbf{54.9}\% &\textbf{64.7}\% & 56.9\% &\textbf{58.8}\%  &\textbf{1.3}\% \\ 
\hline 
\end{tabular}
\caption[Classification accuracy for the Three Tasks scenario starting from ImageNet.]{Classification accuracy for the Three Tasks scenario starting from ImageNet. {\tt EBLL} achieves the best trade off between the tasks in the sequence with less forgetting to the previous tasks.}
\label{tab:res_seq_imgnet}

\end{table}
\begin{table}[h!]
\centering
{
\tabcolsep=0.11cm
\small
\begin{tabular}{|l|lll|c|c|}
\hline
& Flowers & Scenes & Birds& Avg. Accuracy & Avg. Forgetting\\
\hline
{\tt Finetune} &51.2\% &48.1\% &{\bf 58.5\%} &51.6\% &17.8\%\\ 
{\tt LwF }   & 81.1\% &59.1\% & 52.3\% &64.1\% &2.8\% \\
{\tt EBLL} & {\bf 82.8\%} &{\bf 61.2 \%} &  51.2\% & {\bf 65.0\%} & {\bf 1.6\%} \\ \hline 

\end{tabular}
}
\caption[Classification accuracy for the Three Tasks scenario starting from Flowers. ]{Classification accuracy for the Three Tasks scenario starting from Flowers. {\tt EBLL} achieves the best trade off between the tasks in the sequence with less forgetting to the previous tasks.}
\label{tab:res_seq_flower}

\end{table}

When we start from a smaller dataset, Flowers, the same trends can be observed, but with larger differences in accuracy 
(Table~\ref{tab:res_twotasks_flower}).
The performance on the second task is lower than that achieved with ImageNet as a starting point for all the compared methods. This is explained by the fact that the representation obtained from ImageNet is more meaningful for the different tasks than what has been finetuned for Flowers. Differently from the ImageNet starting case, {\tt EBLL-separateFCs}  achieves a considerably better performance on the second task than {\tt EBLL} and  {\tt LwF} while preserving the previous task performance.  {\tt Finetune} shows the best performance on the second task while suffering from severe forgetting on the previous task. The pair of tasks here is of a different distribution and finding a compromise between the two tasks is a challenging problem.
As in the previous case, {\tt EBLL} reduces the forgetting of  {\tt LwF} while achieving a similar average performance on the second task.

Overall,  {\tt EBLL-separateFCs} achieves the best performance on the different pairs of tasks. However, it requires allocating separate fully connected layers for each task which requires a lot of memory. Thus, for the sequential experiments we focus on the shared model scenario.
   

In Table \ref{tab:res_seq_imgnet} we report the performance achieved by {\tt EBLL}, {\tt LwF} and {\tt Finetune} for the sequence of  ImageNet $\rightarrow$ Scenes $\rightarrow$ Birds.  As expected, the {\tt Finetune} baseline suffers from severe forgetting on the previous tasks. The performance on  ImageNet (the first task) drops from 57\% to 37.9\% after finetuning on the third task. As this baseline does not consider the previous tasks in its training procedure, it has the advantage of achieving the best performance on the last task in the sequence.  

{\tt EBLL}  continually  reduces forgetting compared to  {\tt LwF} on the previous tasks while showing a comparable performance on the new task in the sequence. For example, {\tt EBLL} achieves 54.9\% on ImageNet compared to 53.3\% by {\tt LwF}. Similar conclusions can be drawn regarding the sequential scenario starting from Flowers as reported in Table~\ref{tab:res_seq_flower}.

\begin{figure}[h]
\centering
\includegraphics[width=0.6\textwidth]{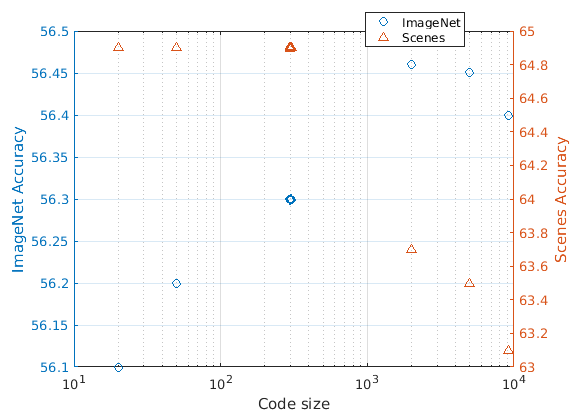}
\caption[Classification accuracy for the Two Tasks scenario ImageNet $\rightarrow$ Scenes with different code sizes.]{Classification accuracy for the Two Tasks scenario ImageNet $\rightarrow$ Scenes with different code sizes. The points in bold mark the code size corresponding to the setting of the experiments reported in Table~\ref{tab:res_twotasks_imagenet}. }
\vspace*{-15pt}
\label{fig:ebll_CodeSize}
\end{figure}

\begin{figure}
\vspace*{-10pt}
\centering
\includegraphics[width=.8\textwidth]{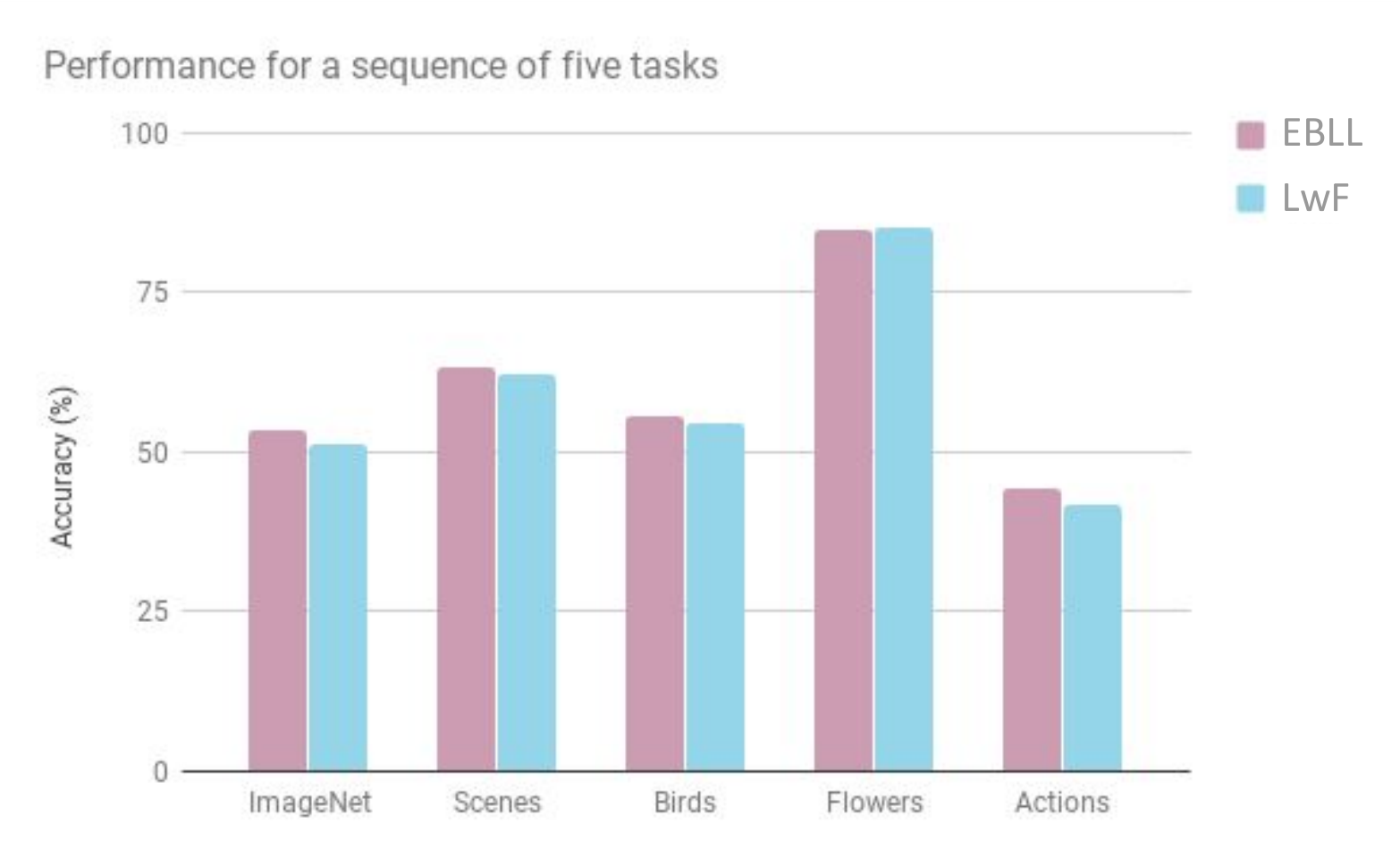}
\caption[Classification accuracy for the Five Tasks scenario.]{Classification accuracy for the Five Tasks scenario. The tasks (horizontal axis) are presented sequentially from left to right. Figure compares the  accuracy of each task at the end  the sequence for both ours~({\tt EBll}) and {\tt LwF}.  }
\label{fig:ebll_FiveTask}
\end{figure}

\paragraph{Effect of the code length}\label{sec:ebll_CodeSize}
In order to examine the effect of varying the code size, we apply our method to the two-task experiment ImageNet $\rightarrow$ Scenes using AlexNet and autoencoders with code size varying from 20 to 9216 (full feature dimension) trained on the output of $conv5$. Figure~\ref{fig:ebll_CodeSize} shows the classification accuracies for both of the datasets for different code sizes.  The results indicate that the smaller code sizes favor the performance on the new task, while the larger code sizes lean towards a better preservation of the old task knowledge but a lower performance on the new task. This experiment confirms the motivation behind using autoencoders to encode the knowledge of previous tasks: a larger code means then a better preservation (see Figure~\ref{fig:ebll_preserve_feat} and Section~\ref{sec:ebll_AE}), but a very large code preserves a noisy information, which explains the drop in the first task performance for the largest codes.
\paragraph{Performance on longer sequences} \label{sec:ebll_LongSeq}
We compare the performance of our method 
to {\tt LwF} on a longer sequence of five tasks: ImageNet $\rightarrow$ Scenes $\rightarrow$ Birds $\rightarrow$ Flowers $\rightarrow$ Actions. Figure~\ref{fig:ebll_FiveTask} shows the obtained accuracies for the five datasets with both methods. It shows that our method outperforms {\tt LwF} by 1.32\% on average over all the tasks. The performance of our method on ImageNet after five tasks (53.6\%) is higher than the performance of {\tt  LwF}  (51.2\%) . It is worth noticing that our performance after 5 tasks is better than {\tt LwF} performance after only 3 tasks (53.3\%, Table \ref{tab:res_seq_imgnet}).

\section{Summary} \label{sec:ebll_Conclusion}
Strategies for efficient continual learning are still an open research problem. In this chapter, we tackled the problem of learning a sequence of tasks using only the data from the most recent environment, based on a fixed capacity training model while aiming at obtaining a reasonable performance on the whole sequence. 

The solution presented here reduces forgetting of earlier tasks by controlling the distance between the representations of the different tasks. We suggest to preserve the features that are crucial for the performance in the corresponding environments. Undercomplete autoencoders are used to learn the submanifold that represents these important features. The method is tested on image classification problems, in sequences of two, three or five tasks, starting either from a small or a large dataset. An improvement in performance over the state-of-the-art at time of development is achieved in all the tested scenarios. Especially, we showed a better preservation of the old tasks.

Despite the demonstrated improvements using a fixed capacity training model, the solution proposed in this chapter still requires for a small but linear increase in memory.  After each task, we need a post processing phase to train the autoencoder and a further preprocessing phase for recording the targets on the new task data. In the next chapter, we present an alternative strategy that moves a step forward towards  a smoother one model continual learning solution.

\cleardoublepage
  \graphicspath{{\chaptersdir/Importance_Weight_Regularization/image/}}%
  \chapter{Importance Weight Regularization}\label{ch:MAS}
Humans can learn in a continuous manner. Old rarely utilized  knowledge can be overwritten by new incoming information while important, frequently used knowledge is prevented from being erased. Continual learning so far has focused mainly on accumulating knowledge over tasks and overcoming forgetting of all seen tasks. In this chapter, we argue that, given the limited model capacity and the unlimited 
new information to be learned, knowledge has to be preserved or erased selectively. 
Inspired by neuroplasticity, we propose a novel approach for continual learning, coined Memory Aware Synapses ({\tt MAS}). It computes the importance of the parameters of a neural network in an unsupervised and online manner. Given a new sample which is fed to the network, {\tt MAS} accumulates an importance measure for each parameter of the network, based on how sensitive the predicted output 
is to a change in this parameter.
When learning a new task, changes to important parameters can then be penalized, effectively preventing important knowledge related to previous tasks from being overwritten.  Further, we show an interesting connection between a local version of our method and
Hebb's rule,
which is a model 
for the learning process in the  brain. We test our method on a sequence of object recognition tasks and on the challenging problem of learning an embedding
for predicting $<$subject, predicate, object$>$ triplets. We show state-of-the-art performance at the time of development~(2018) and, for the first time, the ability to adapt the importance of the parameters based on unlabeled data towards what the network needs (not) to forget, which may vary depending on  test conditions. 
 
 This work was the  fruit of collaboration with Francisca Babiloni and researchers from Facebook AI Research~(FAIR), Mohamed Elhoseiny and Marcus Rohrbach. It was published as an article in  ECCV 2018~\cite{aljundi2018memory}.

\section{Introduction}\label{introduction}

\begin{figure*}[t]
\centering
\includegraphics[width=0.95\textwidth]{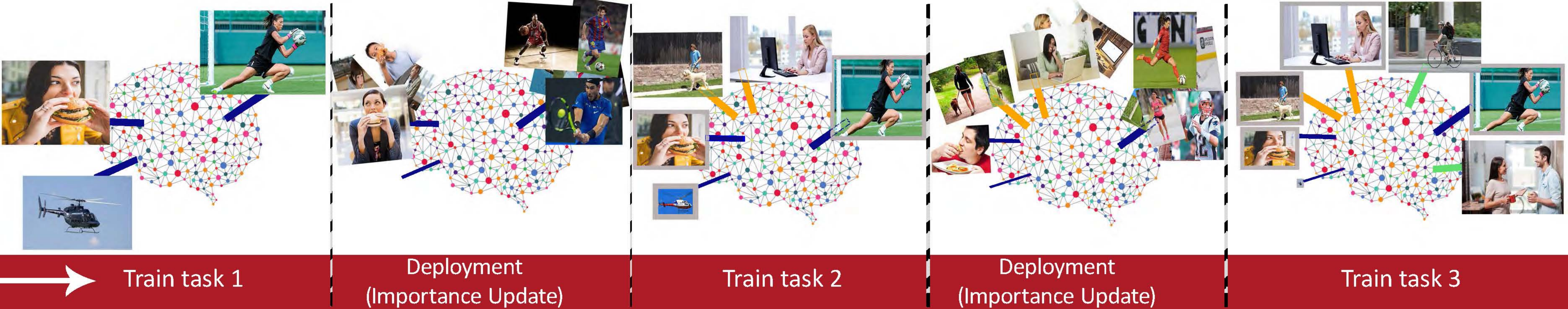} ...
  \caption[An illustration of the considered continual learning   setup. The agent is active and performs the learned tasks. 
Data that appears frequently, will have a bigger contribution. This way, the agent learns 
what is
important and should not be forgotten.  ]{
Our considered learning setup. As common in the task incremental setting, tasks are learned in sequence, one after the other. 
If, in between learning tasks, the agent is active and performs the learned tasks, we can use these unlabeled samples 
to update importance weights for the model parameters. 
Data that appears frequently, will have a bigger contribution. This way, the agent learns 
what is
important and should not be forgotten. 
}  

  \label{fig:mas-teaser}
\end{figure*}

When we started looking into this problem~(2017),  continual learning methods had mostly (albeit not exclusively) 
been applied to relatively short sequences -- often consisting of no more than two tasks (\eg~\cite{lee2017overcoming,li2016learning,rannen2017encoder}), and using relatively large networks with plenty of capacity (\eg~\cite{aljundi2016expert,fernando2017pathnet,rusu2016progressive}). 
However, in a true continual learning setting with a never-ending list of tasks, the capacity of the model sooner or later reaches its limits and compromises need to be made. 
Instead of aiming for no forgetting at all, figuring out what can possibly be forgotten becomes at least as important.
In particular, exploiting context-specific test conditions may pay off in this case. Consider for instance a surveillance camera. Depending on how or where it is mounted, it always captures images under particular viewing conditions. Knowing how to cope with other conditions is no longer relevant and can be forgotten, freeing capacity for other tasks. This calls for a continual learning method that can learn what (not) to forget using unlabeled  test data. We illustrate this setup in Figure~\ref{fig:mas-teaser}.

Such adaptation and memory organization is what we also observe in  biological neurosystems. Our ability to preserve what we have learned before is largely dependent on how frequently we make use of it. Skills that we practice often, appear to be unforgettable, unlike those that we have not used for a long time.
Remarkably, this flexibility and adaptation occur
in the absence of any form of supervision.
According to Hebbian theory~\cite{hebb2002organization}, the process at the basis of this phenomenon is the strengthening of synapses connecting neurons that fire synchronously, compared to those connecting neurons with unrelated firing behavior.

In this work, we propose a new method for continual learning, coined {\em Memory Aware Synapses}, or {\tt MAS} for short, inspired by the model of Hebbian learning in biological systems. Unlike previous works, {\em our  method can learn 
what parts of the model are important 
using unlabelled data}. This allows for  
adaptation to specific test conditions and continuous updating of importance weights. 
This is achieved by estimating  importance weights for the network parameters without relying on the loss~( as in EWC~\cite{kirkpatrick2017overcoming} and SI~\cite{Zenke2017improved}), but by looking at the sensitivity of the output function instead. This way, our method not only avoids the need for labeled data, but importantly it also avoids complications due to the loss being in a local minimum, resulting in gradients being close to zero.
This makes our method not only more versatile, but also simpler, and, as it turns out, more effective in learning what not to forget, compared to other regularization based continual learning approaches.

\textbf{Contributions} of the   work  described in this chapter are threefold:\ 
{\em First}, we propose a new continual learning method, {\em Memory Aware Synapses} (MAS). It estimates importance weights for all the network parameters in an unsupervised and online manner,
allowing adaptation to unlabeled data, \eg in the actual test environment. \ {\em Second}, we show how a local variant of MAS 
is linked to the Hebbian learning scheme.\ 
{\em Third}, we achieve better performance   than state-of-the-art at the time of development~(2018), both when using the standard incremental learning setup and when adapting to specific test conditions, both for object recognition and for predicting $<$subject, predicate, object$>$ triplets, where an embedding is used instead of a softmax output.

In the following we discuss closely related work 
in Section \ref{sec:mas-related_work} and give some background information
in Section~\ref{sec:mas-background}. Section \ref{sec:mas-method} describes our method  and its connection with Hebbian learning.
Experimental results are given in Section \ref{sec:mas-MAS_experiments} and Section \ref{sec:mas-conculosion} concludes the chapter.



\section{Related Work}\label{sec:mas-related_work}
The work presented in this chapter  doesn't store any explicit knowledge from the previous tasks but assumes that the acquired knowledge is embedded in the model parameters.  When learning a new task, the training objective is regularized to not erase important parts of previous knowledge. Our work  belongs to the regularization based family~(see Chapter~\ref{ch:relatedwork}, Section~\ref{sec:related_regularization}).
Unlike the data-focused methods~\cite{li2016learning,rannen2017encoder,shmelkov17iccv} that rely on distilling the knowledge from the previous task model, the prior focused methods, including ours, estimate a prior distribution over the model parameters   with a mean equal to the optimal model parameters learned so far and a variance proportional to the inverse of the parameters importance weights. 
Most similar to our work are \cite{kirkpatrick2017overcoming,Zenke2017improved}. Like them, we estimate an importance weight for each model parameter and add a regularizer when training a new task that penalizes any changes to important parameters. The difference lies in the way the importance weights are computed. In the \textit{Elastic Weight Consolidation} work~\cite{kirkpatrick2017overcoming}, this is done based on an approximation of the diagonal of the Fisher information matrix. 
In the \textit{Synaptic Intelligence} work~\cite{Zenke2017improved}, importance weights are computed during training in an online manner. To this end, they record how much the loss would change due to a change in a specific parameter and accumulate this information over the training trajectory. 
However,  also this method has some drawbacks: 1) Relying on the weight changes in a batch gradient descent might overestimate the importance of the weights, as noted by the authors. %
2) When starting from a pretrained network, as in most practical computer vision applications, some weights might be used without big changes. As a result, their importance will be underestimated. %
3) The computation of the importance is done during training and fixed later.  In contrast, we believe the importance of the weights should be able to adapt to the test data which the system is applied to. 
In contrast to the above two methods, we propose to look at the sensitivity of the learned function, rather than the loss. This simplifies the setup considerably since, unlike the loss, the learned function is not in a local minimum, so complications with gradients being close to zero are avoided. 

In this work, we propose a regularization-based method that computes the importance of the network parameters not only in an online manner but also adaptive to the data that the network is tested on  in an unsupervised manner. While previous works \cite{quadrianto2009distribution,royer2015classifier} adapt the learning system  at prediction time in a transductive setting, our goal here is to build a continual system that can adapt the importance of the weights to what the system needs to remember.  
Our method enjoys more desired characteristics of continual learning~(see Chapter~\ref{ch:introduction}, Section~\ref{sec:intro-desiderata}), namely, constant memory, problem agnostic, adaptive, and graceful forgetting while at the same time achieving  competitive performance.

\section{Background}\label{sec:mas-background}

%
\noindent
{\bf Notations.} 
As in the previous chapter, we train a single, shared neural network over a sequence of tasks. The parameters $\{\theta_{ij}\}$ of the model are the weights of the connections between pairs of neurons $\mathcal{N}_i$ and $\mathcal{N}_j$ in two consecutive layers\footnote{In convolutional layers, parameters are shared by multiple pairs of neurons. For the sake of clarity, yet without loss of generality, we focus here on fully connected layers.}. %
As in other prior-focused methods, our goal is then to compute an importance value $\Omega_{ij}$ 
for each parameter $\theta_{ij}$,
indicating its importance with respect to the previous tasks. 
Similarly to the settings followed in the previous chapters,  we receive a sequence of tasks $\{T_t\}$ to be learned, each with its training data $D_t=(X_t,{Y}_t)$, with $X_t=\{x_n\}_{n=1}^{N_t}$ the input data and ${Y}_t=\{y_n\}_{n=1}^{N_t}$
the corresponding ground truth output data (labels). Note that we drop the task superscript from the  individual samples to avoid cluttering the notations. 
Each task loss $\ell(f(X_t),Y_t)$,  will be combined with an extra loss term to avoid forgetting.
When the training procedure converges to a local minimum, the model has learned an approximation $f$ of the true function $\bar{f}$. 
$f$  maps new input samples $X$ to the corresponding outputs ${\hat{Y_1}, ... ,\hat{Y_{\mathcal{T}}}}$ for tasks $T_1 ... T_\mathcal{T}$ learned so far.
%
\section{Our Approach}
\label{sec:mas-method}

In the following, we introduce our approach.
Like other prior-focused methods~\cite{kirkpatrick2017overcoming,Zenke2017improved}, we estimate an importance weight for each parameter in the network.
Yet in our case, these importance weights approximate the {\em sensitivity of the learned function} to a parameter change  rather than a measure of the (inverse of) parameter uncertainty, as in~\cite{kirkpatrick2017overcoming}, or the sensitivity of the loss to a parameter change, as in~\cite{Zenke2017improved}. As it does not depend on the ground truth labels, our approach allows computing the importance using any available data (unlabeled) which in  turn allows for an adaptation to user-specific settings.
In a learning sequence, we start with task $T_1$, training the model to minimize the task loss  on the training data $\ell(X_1, Y_1)$ -- or simply using a pretrained model for that task.
\subsection{ Estimating Parameter Importance}
After convergence,  the model has learned an approximation $f$ of the true function $\bar{f}$. $f$  maps the input $X_1$ to the output $\hat{Y_1}$.  This mapping $f$ is the target we want to preserve while learning additional tasks. %
To this end, we measure how sensitive the  function $f$ output is to changes in the network parameters. 
For a given data point 
$x_n$,
the output of the network is $f(x_n; \theta)$. A small perturbation $\delta = \{\delta_{ij}\}$ in the parameters $\theta = \{\theta_{ij}\}$ results  in a change in the function output that can be approximated by: 
\begin{equation}\label{eq:perturbation_1}
   f(x_n; \theta+ \delta) -f(x_n; \theta)\approx \sum_{i,j} g_{ij}(x_n)  \delta_{ij}
\end{equation}
\begin{figure}[t]
\centering
\includegraphics[width=0.8\linewidth]{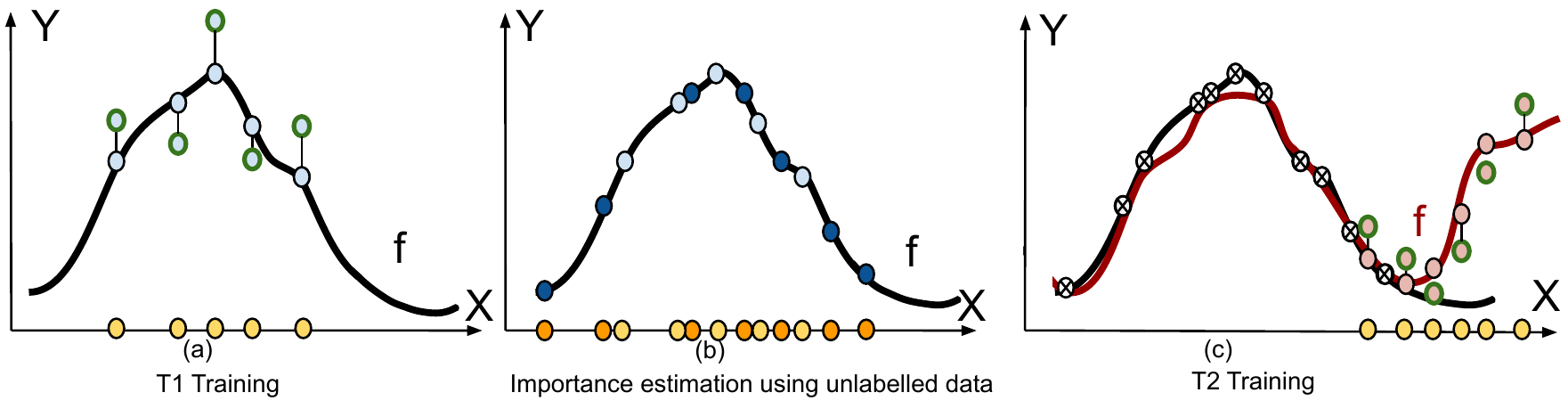}
  \caption[An illustration of the estimation of the importance  weights based on the sensitivity of the loss compared to the sensitivity of the learned function, as we propose.]{ \cite{Zenke2017improved,kirkpatrick2017overcoming} estimate the parameters importance based on the loss, comparing the network output (light blue) with the ground truth labels (green) using training data (in yellow) (a). In contrast, we estimate the parameters importance, after convergence, based on the sensitivity of the learned function to their changes (b). This allows using additional unlabeled data points (in orange). When learning a new task, changes to important parameters are penalized,  the function is preserved over the domain densely sampled in (b), while adjusting not important parameters to ensure good performance on the new task (c).} 
  \label{fig:mas-Function_preserve}
\end{figure}
where $g_{ij}(x_n)=\frac{\partial (f(x_n; \theta))}{\partial \theta_{ij}}$ is the gradient of the learned function  with respect to the parameter $ \theta_{ij}$  evaluated at the data point $x_n$ and $\delta_{ij}$ is the change in parameter $\theta_{ij}$.
Our goal is to preserve the prediction of the network (the learned function) 
at each observed data point and prevent changes to parameters that are important for this prediction~(see Figure~\ref{fig:mas-Function_preserve}).

Based on Equation~\ref{eq:perturbation_1} and assuming a  small constant change $\delta_{ij}$, we can measure the importance of a parameter by the magnitude of the gradient $g_{ij}$, i.e.~how much  does a small perturbation to that parameter  change the output of the learned function for data point $x_n$. 
We then accumulate the gradients over the given data points %
to obtain importance weight $\Omega_{ij}$ for parameter $\theta_{ij}$: 
\begin{equation}\label{eq:importance_estimation_2}
\Omega_{ij} =\frac{1}{N}\sum_{n=1}^N  \mid\mid g_{ij}(x_n) \mid\mid
\end{equation}
This equation can be updated in an online fashion whenever a new data point is fed to the network. $N$ is the total number of data points at a given phase, note that it is doesn't have to be equal to $N_t$. 
Parameters with small importance weights do not affect the output much, and can, therefore, be changed to minimize the loss for subsequent tasks, while parameters with large weights should ideally be left unchanged.

When the output function $f$ is multi-dimensional, as is the case for most neural networks, \Eqn~\ref{eq:importance_estimation_2} involves computing the gradients for each output, which requires as many backward passes as the dimensionality of the output. As a more efficient alternative, we propose to use the gradients of the squared $\ell_2$ norm of the learned function output\footnote{We   square the $\ell_2$ norm as it simplifies the math and the link with the Hebbian method, see section~\ref{subsec:mas-hebb}. }, \ie, $g_{ij}(x_n)=\frac{\partial [\ell^2_2(f(x_n; \theta))]}{\partial \theta_{ij}}$.
The importance of the parameters is then measured by the sensitivity of the squared $\ell_2$ norm of the function output to their changes. This way, we get one scalar value for each sample instead of a vector output. Hence, we only need to compute one backward pass and can use the resulting gradients for estimating the parameters importance.
Using our method,  for regions in the input space that are sampled densely,
the function will be preserved and catastrophic forgetting is avoided. However, parameters not affecting those regions will be given low importance weights, and can be used to optimize the function for other tasks, affecting the function over other regions of the input space.  

\subsection{ Learning a New Task}
When a new task $T_{\mathcal{T}}$ needs to be learned, we have in addition to the new task loss $\ell(f(x;\theta),y)$, a regularizer that penalizes changes to parameters that are deemed important for previous tasks. As such, the training procedure of the new task minimizes the following objective function:

\begin{equation}\label{eq:MAS_objective_3}
{J(\theta)}=\frac{1}{N_\mathcal{T}}\sum_{n=1}^{N_\mathcal{T}} \ell(f(x_n;\theta),y_n)+{\lambda} \sum_{i,j} \Omega_{ij}(\theta_{ij}-\theta^{\mathcal{T}-1}_{ij})^2
\end{equation}
with $N_\mathcal{T}$ the number of samples of the current task, $\lambda$ a hyperparameter for the regularizer and $\theta^{\mathcal{T}-1}_{ij}$ the ``old'' network parameters (as determined by the optimization for the previous task in the sequence, $T_{\mathcal{T}-1}$), $\theta$ is initialized with $\theta^{\mathcal{T}-1}$.  Hence, we allow the new task to change parameters that are not important for the previous task (low $\Omega_{ij}$).  The important parameters (high $\Omega_{ij}$) can also be reused, via model sharing, but with a penalty when changing them.

Finally, the importance matrix $\Omega$  is to be updated after training a new task, by accumulating over the previously computed $\Omega$. Since we don't  use the loss function, $\Omega$ can be computed on any available data considered most representative for test conditions, be it on the last training epoch, during the validation phase or at test time. 
In the experimental section~\ref{sec:mas-MAS_experiments}, we show how this allows our method to adapt and specialize to any set, be it from the training or from the test.

\subsection{Connection to Hebbian Learning}\label{subsec:mas-hebb}
\begin{wrapfigure}{r}{0.4\textwidth}
    \includegraphics[width=0.4\textwidth]{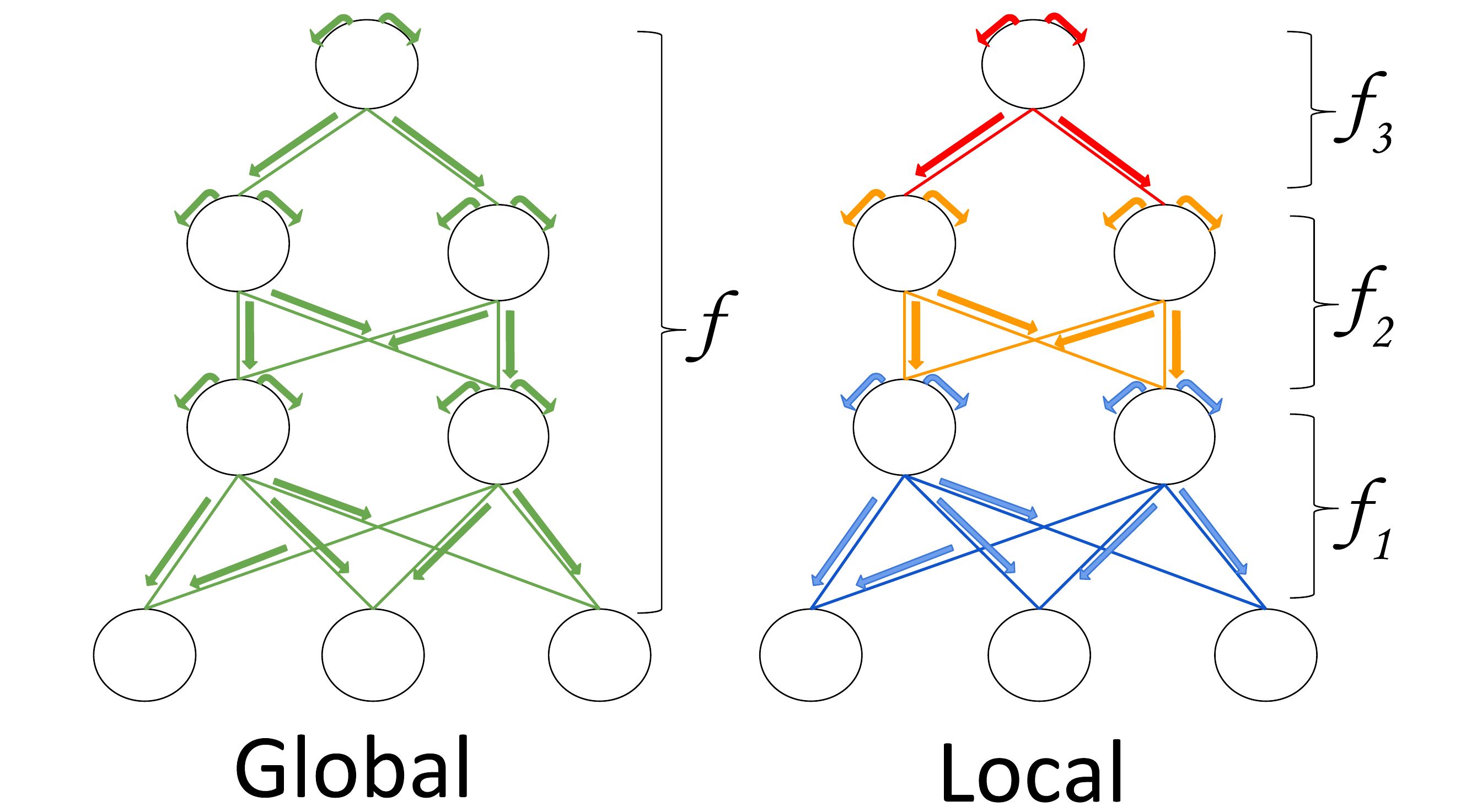}
  \caption{ Gradients flow for computing the importance weight. Local considers the gradients of each layer independently. }
    \label{fig:mas-local_vs_global}
  \vspace*{-10pt}
\end{wrapfigure}
In this section, we propose a local version of our method, by applying it to a single layer of the network rather than to the network as a whole. Next, we show an interesting connection between this local version and 
Hebbian learning~\cite{hebb2002organization}.

\noindent
{\bf A local version of our method.} 
Instead of considering the function $f$ that is learned by the network as a whole, we decompose it in a sequence of functions $f_l$ each corresponding to one layer of the network, i.e.,~$f(x) = f_L(f_{L-1}(...(f_1(x))))$, with $L$ the total number of layers.
By locally preserving the output of each layer given its input, we can preserve the global function $f$. This is further illustrated in Figure~\ref{fig:mas-local_vs_global}. Note how ``local'' and ``global'' in this context relate to the number of layers over which the gradients are computed.

We use $h^l_{i,n}$ to denote the activation of neuron $\mathcal{N}_i$  at layer $l$
for a given input $x_n$.
 Note that, we explicitly add a superscript $l$ referring to the layer and for clarity we omit the sample index $n$ when it is not needed.
Hence $H_l= \{h^l_j\}$ is the output of layer $l$ and $H_{l-1} = \{h^{l-1}_i\}$ is its input. $H_l$ is a vector with elements $\{h^l_j\}$.  
$\theta_{ij}$ is then a network parameter representing the connection between neuron $\mathcal{N}_i$  in layer $l-1$ and neuron $\mathcal{N}_j$  in layer $l$.
Analogous to the procedure followed previously, we consider the squared $\ell_2$ norm of each layer after the activation function. %
An infinitesimal change $\delta_l = \{\delta_{ij}\}$ in the parameters $\theta_l= \{\theta_{ij}\}$ of layer $l$ %
results in a change to the squared $\ell_2$ norm  of the local function $f_l$ %
for a given input to that layer $H_{l-1} = \{h^{l-1}_i\} = f_{l-1}(...(f_1(x)))$ given by:
\begin{equation}
\ell^2_2( f_l(H_{l-1};\theta_l+ \delta_l)) -\ell^2_2 (f_l(H_{l-1};\theta_l))\approx \sum_{i,j} g_{ij}(x) \delta_{ij}
\end{equation}
 where $g_{ij}(x)=\frac{\partial[\ell^2_2( f_l(H_{l-1};\theta_l))]}{\partial \theta_{ij}}$. 

In the case of a ReLU activation function 
and considering a fully connected layer with $I*J$ parameters, i.e.
~$h^{l}_j=ReLU(out^{l}_j)$ with $out^{l}_j= \sum_{o=1}^{I}\theta_{oj}*h_o^{l-1}$, we can write
\begin{equation}
g_{ij}(x)=\frac{\partial[\ell^2_2( f_l(H_{l-1};\theta_l))]}{\partial \theta_{ij}}=\frac{\partial[\sum_{k=1}^{J} (h^{l}_k)^2]}{\partial \theta_{ij}}=\sum_{k=1}^{J} \frac{\partial(  (h^{l}_k)^2)}{\partial \theta_{ij}}
\end{equation}
Since 
$\frac{\partial ( (h^{l}_k)^2)}{\partial \theta_{ij}}=0$
when  $k \neq j$, we get
\begin{equation}
g_{ij}(x)= \frac{ \partial(  (h^{l}_j)^2)}{\partial \theta_{ij}}
= 2 *h^l_j *\frac{\partial (h^{l}_j) }{\partial \theta_{ij}}
= 2 *h^l_j *\frac{\partial (ReLU(out^{l}_j))}{\partial \theta_{ij}}
\end{equation}
ReLU is non smooth since it is not differentiable at 0. We show the subgradients in the two cases (also using the fact that
$out^{l}_j= \sum_{o=1}^{I}\theta_{oj}*h_o^{l-1}$): 
\begin{enumerate}

\item  if $out^{l}_j>0$, $ReLU(out^{l}_j)=out^{l}_j$ and $(ReLU)^\prime$ =1 :
\begin{align}
\frac{\partial (ReLU(out^{l}_j))}{\partial \theta_{ij}} & = \frac{\partial(ReLU(out^{l}_j))}{\partial out^{l}_j}
\frac{\partial(out^{l}_j)}{\partial \theta_{ij}}
=\frac{\partial(out^{l}_j)}{\partial \theta_{ij}}  \nonumber \\
& =  \frac{\partial( \sum_{o=1}^{I}\theta_{oj}*h_o^{l-1})}{\partial \theta_{ij}} =\frac{\partial( \theta_{ij}*h_i^{l-1})}{\partial \theta_{ij}}=h_i^{l-1}  \nonumber \\
\Rightarrow g_{ij}(x) &= 2*h_j^l*h_i^{l-1}\label{eq:local_g1}
\end{align}

\item in the other case, $ReLU(out^{l}_j)=0$ and $(ReLU)^\prime=0$, so
\begin{equation}
 g_{ij}(x)=0
\end{equation}
At the same time,
\begin{equation}
 ReLU(out^{l}_j)=0 \Rightarrow h_j^l=0 \,\textrm{and} \, \, 2*h_j^l*h_i^{l-1}=0
\end{equation}
Hence
\begin{equation}\label{eq:local_g0}
g_{ij}(x)=2*h_j^l*h_i^{l-1}=0 
\end{equation}
\end{enumerate}
Based on equations \ref{eq:local_g1} and \ref{eq:local_g0}, we have shown
\begin{equation}
  g_{ij}(x)=2*h_j^l*h_i^{l-1}
\end{equation}
Again we consider the accumulation of the gradients evaluated at different data points $\{x_n\}$ as a measure for the importance of the parameter $\theta_{ij}$ in a layer $l$:
\begin{equation}
\Omega_{ij}=\frac{1}{N}\sum_{n=1}^N g_{ij}(x_n) = 2*\frac{1}{N}\sum_{n=1}^N h^l_{j,n}*h^{l-1}_{i,n}
\label{eq:local}
\end{equation}
%
%

\noindent
{\bf Link with Hebbian theory.} 
In neuroscience, Hebbian learning theory~\cite{hebb2002organization} provides an explanation for the phenomenon of synaptic plasticity. It postulates that
``cells that fire together, wire together'': the synapses (connections) between neurons that fire synchronously for a given input
are strengthened over time to maintain and possibly improve the corresponding outputs. 
Here we reconsider this theory from the perspective of an artificial neural network after it has been trained successfully with backpropagation. 
Following Hebb's rule, parameters connecting neurons that often fire together (high activations for both, i.e. highly correlated outputs) %
are more important for the given task than those that fire asynchronously or with low activations. 
As such, the importance weight $\Omega_{ij}$ for the parameter $\theta_{ij}$ in a layer $l$ %
can be measured purely locally in terms of the correlation between the neurons activations, i.e. 
\begin{equation}
\Omega_{ij}=\frac{1}{N}\sum_{n=1}^N h^l_{j,n}*h^{l-1}_{i,n}
\label{eq:hebb}
\end{equation}
The similarity with equation~\ref{eq:local} is striking. We can conclude that applying Hebb's rule to measure the importance of the parameters in a neural network can be seen as a local variant of our method that considers 
only one layer at a time instead of 
the global function learned by the network.
Since only the relative importance weights really matter, the scale factor $2$ can be ignored. \\ %

\noindent
\subsection{Discussion} \label{subsec:mas-discussion}
Our global and local methods both have the advantage of computing the importance of the parameters on any given data point without the need to access the labels or the condition of being computed while training the model.
The global version needs to compute the gradients of the output function while the local variant (Hebbian based) can be computed locally by multiplying the input with the output of the connecting neurons. 
{\it Our proposed method (both the local and global version) resembles an implicit memory included for each parameter of the network. 
We, therefore, refer to it as Memory Aware Synapses. }
It keeps updating its value based on the activations of the network when applied to new data points. It can adapt and specialize to a given subset of data points rather than preserving every functionality in the network. Further, the method can be added after the network is trained. It can be applied on top of any pretrained network and compute the importance on any set of data without the need to have the labels. This is an important criterion that differentiates our work from methods that rely on the loss function to compute the importance of the parameters.

\section{Experiments}\label{sec:mas-MAS_experiments}
We start by comparing our method to different existing continual learning methods in the standard  incremental learning setup of object recognition tasks.
Next, we move to the more challenging problem of continual learning of $<$subject, predicate, object$>$ triplets in an embedding space, Section~\ref{sec:mas-facts}. We  further analyze the behavior and some design choices of our method, Section~\ref{sec:mas-analysis}.

\subsection{Object Recognition}
\label{sec:mas-objrec} 

We follow the standard setup commonly used in  computer vision  to evaluate continual learning methods~\cite{lee2017overcoming,li2016learning,liu2018rotate}. It consists of a sequence of supervised classification tasks each from a particular dataset. 
Note that this  assumes having different classification layers for each task (different ``heads'') that remain unshared.
Moreover, an oracle is used at test time to decide on the task (\ie, which classification layer to use). 

\noindent
{\bf Compared Methods.} 

\noindent
- {\em Finetuning} ({\tt Finetune}).  After learning the first task and when receiving a new task to learn,  the parameters of the network are finetuned on the new task data. This baseline is expected to suffer from forgetting the old tasks while being advantageous for the new task. 

\noindent
- {\em Learning without Forgetting~\cite{li2016learning}} ({\tt LwF}). A data-focused regularization based method. As explained in previous chapters~(\ref{ch:expertgate},\ref{ch:ebll}) knowledge distillation  is deployed to mitigate forgetting.  The method relies on first training the new task head while freezing the shared parameters as a warmup phase and then training all the parameters until convergence.

\noindent
- {\em Encoder Based Lifelong Learning~\cite{rannen2017encoder}} ({\tt EBLL}). Our work detailed in Chapter~\ref{ch:ebll} in which we learn a  shallow encoder  of the features of each task. A penalty on the changes to the encoded features  accompanied with the distillation loss is  applied to reduce the forgetting of the previous tasks. 

\noindent
- {\em Incremental Moment Matching ~\cite{lee2017overcoming}} ({\tt IMM}). A new task is learned with an $\ell_2$ penalty equally applied  to the changes to all shared parameters. At the end of the sequence, the obtained models are merged through a  first or second moment matching. In our experiments, mean {\tt IMM} gives better results on the two tasks experiments while mode {\tt IMM} wins on the longer sequence. Thus, we report the best alternative in each experiment.

\noindent
- {\em Elastic Weight Consolidation~\cite{kirkpatrick2017overcoming}} ({\tt EWC}). 
It  is the first work introducing the prior-focused approach of the regularization based family  using as importance measure the diagonal of the Fisher information matrix. EWC uses individual penalty for each  previous task,  however, to make it computationally feasible  we  apply a single penalty as pointed out by~\cite{huszar2017quadratic}. Hence, we use a running sum of the Fishers in the 8 tasks sequence.

\noindent
- {\em Synaptic Intelligence~\cite{Zenke2017improved}} ({\tt SI}). 
This  comes closest to our approach as it estimates the importance weights in an online manner while training for a new task. Similar to {\tt EWC} and our method, changes to  parameters important for previous tasks are penalized during training of later tasks.

\noindent
- {\em Memory Aware Synapses } ({\tt MAS}). Unless stated otherwise, we use the global version of our method and with the importance weights estimated only on training data. We use a regularization parameter $\lambda$ of 1. Note that no tuning of $\lambda$ was performed as we assume no access to previous task data.

\noindent
{\bf Experimental setup.}  
We use the AlexNet~\cite{krizhevsky2012imagenet} architecture pretrained on ImageNet~\cite{ILSVRC15} from~\cite{krizhevsky2014one}\footnote{We use the pretrained model available in Pytorch. Note that it differs slightly from other implementations used \eg in~\cite{li2016learning}. }.
All the training of the different tasks has been done with stochastic gradient descent for 100 epochs and a batch size of 200 using the same learning rate as in the previous chapters.  Performance is measured in terms of classification accuracy. 

\noindent
{\bf Two tasks experiments.}  
We first consider sequences of two tasks based on three datasets:   \textit{Scenes}~\cite{quattoni2009recognizing} for indoor scene classification, \textit{Birds}~\cite{WelinderEtal2010} for fine-grained bird classification, and  \textit{Flowers}~\cite{Nilsback08} for fine-grained flower classification. We consider: Scene$\rightarrow$ Birds, Birds$\rightarrow$ Scenes, Flowers$\rightarrow$Scenes and Flowers$\rightarrow$  Birds. 
We didn't consider ImageNet as a task in the sequence as this would require retraining the network from scratch to get the importance weights for {\tt SI}.
\begin{table*}[t]
\centering
\tabcolsep=0.11cm
\footnotesize
\resizebox{\textwidth}{!}{
\begin{tabular}{|l||c|c||c|c||c|c||c|c||c|c|}
\hline
Method&\multicolumn{2}{c||}{Birds $\rightarrow$ Scenes} &\multicolumn{2}{c||}{Scenes $\rightarrow$ Birds}&\multicolumn{2}{c||}{Flowers $\rightarrow$ Birds}&\multicolumn{2}{c||}{Flowers $\rightarrow$ Scenes}\\ \hline
{\tt Finetune} &{45.20~(8.0)} &\bf{57.8}& {49.7~(9.3)} &\bf{52.8}& {64.87~(13.2)} &\bf{53.8}& {70.17~(7.9)} &{57.31}\\ 
{\tt LwF}~\cite{li2016learning} 
&{51.65~(2.0)} &{55.59}& {55.89~(3.1)} &{49.46}& {73.97~(4.1)} & {53.64}& {76.20~(1.9)} &{58.05} \\ 
{\tt EBLL}~\cite{rannen2017encoder} 
&{52.79~(0.8)} &{55.67}& {56.34~(2.7)} &{49.41}& {75.45~(2.6)} & {50.51}& {76.20~(1.9)} &\bf{58.35} \\ 
{\tt IMM}~\cite{lee2017overcoming} 
&{51.51~(2.1)} &{52.62}& {54.76~(4.2)} &{52.20}& {75.68~(2.4)} & {48.32}& {76.28~(1.8)} &{55.64} \\ 
{\tt EWC}~\cite{kirkpatrick2017overcoming} 
&{52.19~(1.4)} &{55.74}& \bf{58.28~(0.8)} &{49.65}& {76.46~(1.6)} & {50.7}& {77.0~(1.1)} &{57.53} \\ 
{\tt SI}~\cite{Zenke2017improved} %
&{52.64~(1.0)} &{55.89}& {57.46~(1.5)} &{49.70}& {75.19~(2.9)} & {51.20}& {76.61~(1.5)} &{57.53} \\ 
{\tt MAS} (ours) &\bf{53.24~(0.4)} &{55.0}& {57.61~(1.4)} &{49.62}& \bf{77.33~(0.7)} &{50.39}& \bf{77.24~(0.8)} &{57.38} \\ \hline
\end{tabular}}
\caption{Classification accuracy (\%), forgetting on the first task (\%) for various sequences of 2 tasks using the object recognition setup. 
}%
\label{tab:mas-res_twotasks_classification}
\end{table*}

As shown in Table~\ref{tab:mas-res_twotasks_classification},
{\tt Finetune} %
clearly 
suffers from catastrophic forgetting with a drop in performance from $8\%$ to $13\%$. All the considered methods manage to reduce the forgetting over fine-tuning significantly while having performance close to fine-tuning on the new task.  
On average, our method  achieves the lowest forgetting rates (around $1\%$) while performance on the new task is almost similar ($0 - 3\%$ lower). 

{\bf Local vs. global MAS on training/test data.} 
Next we analyze the performance of our method when preserving the global function learned by the network after each task ({\tt MAS}) and its local Hebbian-inspired variant described in section~\ref{subsec:mas-hebb} ({\tt l-MAS}). We  evaluate our methods, {\tt MAS} and {\tt l-MAS}, when using unlabeled test data and/or labeled training data.
Table~\ref{tab:mas-res_twotasks_classification_train_test} shows 
  for both {\tt l-MAS} and {\tt MAS},  the preservation of the previous task and the performance on the current task are quite similar given the different used sets of data. This illustrates  our methods ability to estimate the parameters importance of a given task given any set of points, without the need for labeled data. Further, computing the gradients locally at each layer for {\tt l-MAS} allows for faster computations but less accurate estimations. As such,   {\tt l-MAS} shows an average forgetting of $3\%$ compared to $1\%$ by {\tt MAS}.%

{ \bf $\ell^2_2$ vs. vector output.} We explained in section \ref{sec:mas-method} that considering the gradients of the learned function to estimate the parameters importance would require as many backward passes as the length of the output vector. To avoid this complexity, we suggest using the square of the $\ell_2$ norm of the function to get a scalar output. We run two experiments, Flowers$\rightarrow$ Scenes and Flowers$\rightarrow$  Birds once with computing the gradients with respect to the vector output and once with respect to the $\ell^2_2$ norm. We observe no significant difference  on forgetting over 3 random trials  where we get a mean, over 6 numbers, of $ 0.51\% \pm 0.18$  for the forgetting on the first task in the vector output case compared to $0.50\% \pm 0.19$ for the $\ell^2_2$ norm case.  No significant difference is observed on the second task either. As such, we can conclude that using $\ell^2_2$  is $m$ times faster (where $m$ is the length of the output vector)  without loss in performance.

\begin{table*}[t]
\centering
\footnotesize
\resizebox{\textwidth}{!}{
\begin{tabular}{|l|l||c|c||c|c||c|c||c|c||c|c|}
\hline
Method&$\Omega$ computed. on&\multicolumn{2}{c||}{Birds $\rightarrow$ Scenes} &\multicolumn{2}{c||}{Scenes $\rightarrow$ Birds}&\multicolumn{2}{c||}{Flowers $\rightarrow$ Birds}&\multicolumn{2}{c||}{Flowers $\rightarrow$ Scenes}\\ \hline
{\tt MAS} & Train&{53.24~(0.4)} &{55.0}& {57.61~(1.4)} &{49.62}& {77.33~(0.7)} &{50.39}& {77.24~(0.8)} &{57.38} \\ 
{\tt MAS} & Test& {53.43~(0.2)} &{55.07}& {57.31~(1.7)} &{49.01} &{77.62~(0.5)} &{50.29}& {77.45~(0.6)} &{57.45} \\ 
{\tt MAS}& Train + Test& {53.29~(0.3)} &{56.04}& {57.83~(1.2)} &{49.56}& {77.52~(0.6)} &{49.70}& { 77.54~(0.5)} &{57.39} \\ \hline
{\tt l-MAS} & Train& {51.36~(2.3)} &{55.67}& {57.61~(1.4)} &{49.86}& {73.96~(4.1)} &{50.5}& {76.20~(1.9)} &{56.68} \\ 
{\tt l-MAS} & Test& {51.62~(2.0)} &{53.95}& {55.74~(3.3)} &{50.43}& {74.48~(3.6)} &{50.32}& {76.56~(1.5)} &{57.83} \\ 
{\tt l-MAS} & Train + Test& {52.15~(1.5)} &{ 54.40}& {56.79~(2.2)} &{48.92}& {73.73~(4.3)} &{50.5}& { 76.41~(1.7)} &{57.91} \\ \hline
\end{tabular}}
\caption{ Classification accuracies (\%) for the object recognition setup - comparison between using Train and Test data (unlabeled) to compute the  parameter importance $\Omega$.}
\label{tab:mas-res_twotasks_classification_train_test}
\end{table*}

\myparagraph{Longer Sequence }
While the two tasks setup gives a detailed look at the average expected forgetting when learning a new task, it remains easy. Thus, we next consider a sequence of 8 tasks.
To do so, we add five more datasets:  {\tt Cars}~\cite{krause20133d} for fine-grained car classification; {\tt Aircraft}~\cite{maji13fine-grained} for fine-grained aircraft classification;  {\tt Actions}, for human  action classification~\cite{pascal-voc-2012}; {\tt Letters}~\cite{deCampos09} for character recognition in natural images; and {\tt SVHN} dataset~\cite{netzer2011reading} for digit recognition. 
We have also used those datasets  in the previous chapters.
We run the different methods on the following sequence: 

Flowers$\rightarrow$Scenes$\rightarrow$Birds$\rightarrow$Cars$\rightarrow$Aircraft$\rightarrow$Actions$\rightarrow$Letters$\rightarrow$SVHN.

\begin{figure}[t]
  \centering
  \subfloat[]{
    \includegraphics[width=0.5\textwidth]{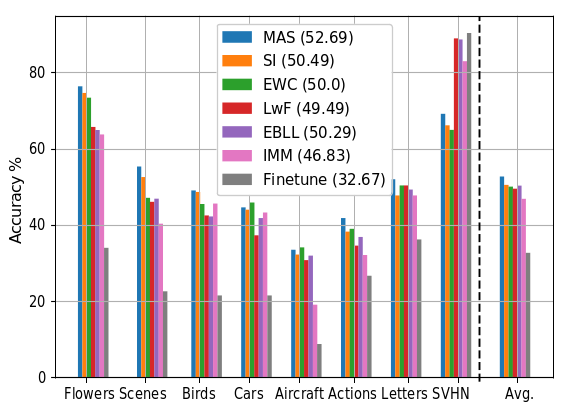}
    \label{fig:mas-8task_acc}}
    \hfillx
  \subfloat[]{
    \includegraphics[width=0.49\textwidth]{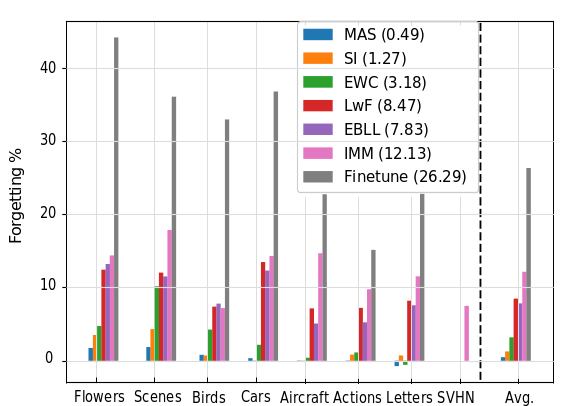}
    \label{fig:mas-8task_drop}}
 
  \caption[Performance and forgetting, at the end of the 8 tasks object recognition sequence.]{  (a) Performance  on each task, in accuracy, at the end of the 8 tasks object recognition sequence. (b) Forgetting on each task measured at the end of the sequence, lower is better.  }
  
\end{figure}

\begin{figure}
\centering
\vspace{-10pt}
  \includegraphics[width=0.55\textwidth]{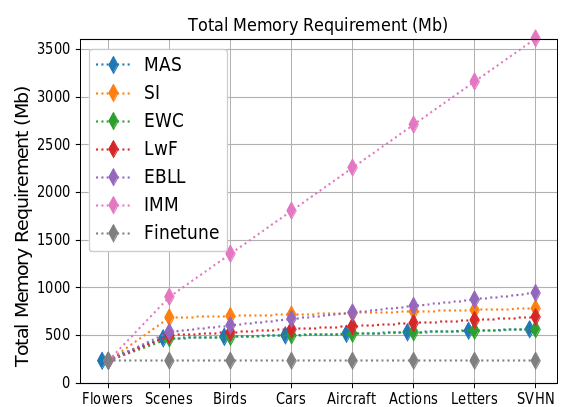}
  \caption{  Overall memory requirement for each method at each step of the sequence.}
   \label{fig:mas-memreq}
\end{figure}
While Figure~\ref{fig:mas-8task_acc} shows the performance on each task at the end of the sequence,~\ref{fig:mas-8task_drop} shows the observed forgetting on each task at the end of the sequence (relative to the performance right after training that task). 
The differences between the compared methods become more outspoken.
{\tt Finetune} suffers from a severe forgetting on the previous tasks while being advantageous for the last task, as expected. {\tt LwF}~\cite{li2016learning} suffers from a buildup of errors when facing a long sequence while {\tt EBLL}~\cite{rannen2017encoder} reduces this effect. {\tt IMM}~\cite{lee2017overcoming} merges the models at the end of the sequence and the drop in performance differs between tasks.
More importantly,  the method performance on the last task is highly affected by the moment matching. {\tt SI}~\cite{Zenke2017improved} followed by {\tt EWC}~\cite{kirkpatrick2017overcoming} has the least forgetting among our methods competitors. {\tt MAS}, our method, shows a minimal or no forgetting on the different tasks in the sequence with an average forgetting of $0.49\%$. It is worth noting that our method's absolute performance on average including the last task is $2\%$ better than {\tt SI} which indicates our method ability to accurately estimate the importance weights and to allow the new tasks to adjust accordingly.  
Apart from evaluating forgetting, we analyze the memory requirements of each of the compared methods. Figure~\ref{fig:mas-memreq} illustrates the memory usage of each method at each learning step in the sequence. After {\tt Finetune} that doesn't treat forgetting, our method has the lowest  memory consumption. Note that {\tt IMM} grows linearly in storage, but at inference time it only uses the obtained model.
%

\subsection{Fact Learning}
\label{sec:mas-facts}
Next, we move to a more challenging setup where all the layers of the network are shared, including the last layer. Instead of learning a classifier, we learn an embedding space. For this setup, we pick the problem of Fact Learning from natural images~\cite{elhoseiny2017sherlock}. For example, a fact could be ``person eating pizza''. 
We design different experimental settings to show the ability of our method to learn what (not) to forget.

\noindent
{\bf Experimental setup.} 
We use the 6DS mid scale dataset presented in~\cite{elhoseiny2017sherlock}.
It consists of  $28,624$  images, divided equally in training and test samples belonging to  $186$ unique facts. Facts are structured into 3 units: Subject (S), Object (O) and Predicate (P). We use a CNN model based on the VGG-16 architecture~\cite{simonyan2014very} pretrained on ImageNet. The last fully connected layer forks in three final layers enabling the model to have three separated and structured outputs for Subject, Predicate and Object as in~\cite{elhoseiny2017sherlock}. The loss minimizes the pairwise distance between the visual and the language embedding. For the language embedding, the Word2vec~\cite{mikolov2013efficient} representation of the fact units is used.
To study fact learning from a continual perspective, we divided randomly the dataset into tasks belonging to  different groups  of facts. 
SGD optimizer is used with a mini-batch of size 35 for 300 epochs and we use a $\lambda=5$ for our method. For evaluation, we report the fact  to image retrieval  scenario. We follow the evaluation protocol  proposed in~\cite{elhoseiny2017sherlock} and report the mean average precision (MAP). For each task, we consider retrieving the images belonging to facts from this task only. We also report the mean average precision on the whole dataset which differs from the average of the performance achieved on each task. 
We focus on the comparison between {\tt MAS}, our method, and {\tt SI}~\cite{Zenke2017improved}, the best performing method among the different competitors as shown in Figure~\ref{fig:mas-8task_acc}.

  \begin{table}
\centering 
  \tabcolsep=0.11cm   
  \footnotesize
\begin{tabular}{ | l| l | c| c | c | c ||c|}
\hline
&   & \multicolumn{5}{c|}{Method evaluated on}\\
\textbf{Method} & Split & \boldmath{$T_1$} &\boldmath{$T_2$}&\boldmath{$T_3$} & \boldmath{$T_4$} & all\\  \hline
\tt{Finetune} & 1 &0.19 & 0.19 &0.28&{\bf0.71}& 0.18\\
\tt{SI}\cite{Zenke2017improved}
& 1 & 0.36 & 0.32 &0.38 &{ 0.68}&0.25\\ 
							
{\tt MAS (ours)} & 1 & {\bf 0.42}& {\bf 0.37} &{\bf 0.41}&0.65&{\bf 0.29}\\  \hline  
\tt{Finetune}& 2 & {0.20} & 0.27 &0.18 &{ 0.66}&0.18\\
\tt{SI}\cite{Zenke2017improved}
& 2 & 0.37 & 0.39 &0.38 &{ 0.46}&0.24\\
{\tt MAS (ours)} & 2 & {\bf 0.42}& {\bf 0.42} &{\bf 0.46}&{\bf 0.65}&{\bf 0.28}\\  \hline  

\tt{Finetune}& 3 & {   0.21}& 0.25&0.24&{0.46}&0.14\\
\tt{SI }\cite{Zenke2017improved}
& 3 & {\bf   0.30}& 0.31&0.36&{0.61}&0.24\\
{\tt MAS (ours)} & 3 & {\bf  0.30}& {\bf 0.36} &{\bf 0.38}&{\bf 0.66}&{\bf  0.27}\\  \hline  

\end{tabular}
\caption{\label{tab:mas-fact_4_tasks}  MAP for fact learning on the  4 tasks  random split, from the 6DS dataset, at the end of the sequence. } 
    \end{table}

{\bf Four tasks experiments\ }
We consider a sequence of 4 tasks obtained from  splitting randomly the facts of the same dataset into 4 groups. %
Table~\ref{tab:mas-fact_4_tasks} presents the achieved performance on each set of the 4 tasks at the end of the learned sequence based on  3 different random splits.  Similar to previous experiments, {\tt Finetune} is only advantageous on the last task while drastically suffering on the previous tasks.  However, here, our method differentiates itself clearly, showing $6$ better MAP on the first two tasks compared to {\tt SI}. Overall, {\tt MAS} achieves a MAP of $0.29$ compared to $0.25$ by {\tt SI} and only $0.18$ by {\tt Finetune}. When {\tt MAS} importance weights are computed on both training and test data, a further improvement is achieved with  $0.30$  overall performance. This highlights our method ability to benefit from extra unlabeled data to further enhance the importance estimation.

\subsection{Behavior Analysis}{\label{sec:mas-analysis}}

\myparagraph{Sensitivity to the hyper parameter.}
Our method needs one extra hyper parameter, $\lambda$, that weights the penalty on the parameters changes as shown in Eq \ref{eq:MAS_objective_3}.
\begin{figure}
\begin{center}
 \includegraphics[width=0.7\textwidth]{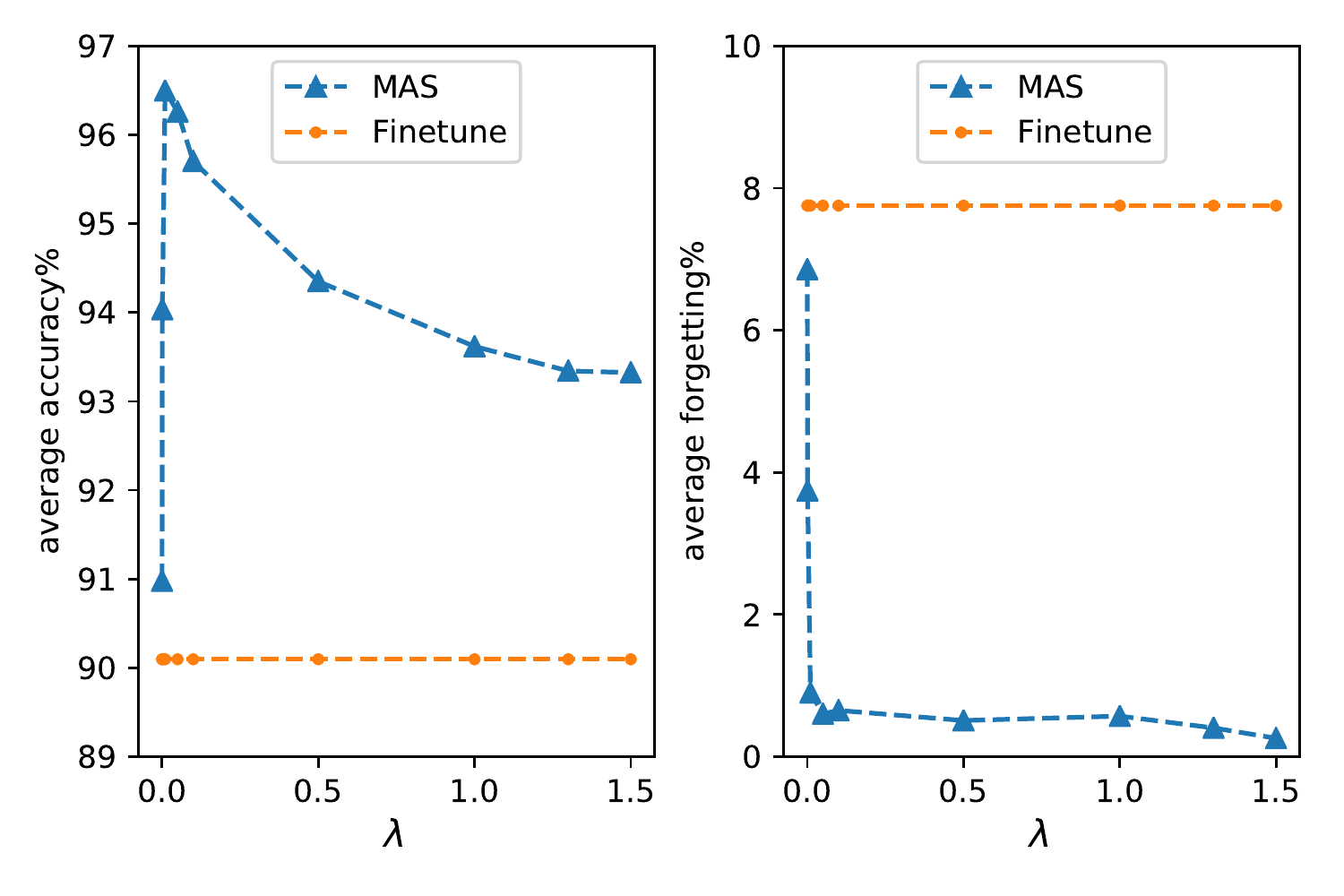}
  \caption{  Avg. performance, left, and Avg. forgetting, right, on permuted MNIST sequence.}
   \label{fig:mas-permuted}
   \end{center}
   \end{figure}
$\lambda$ is a trade-off between the allowed forgetting and the new task loss. We set $\lambda$ to the largest value that allows an acceptable performance on the new task. 
for {\tt SI}\cite{Zenke2017improved} and {\tt EWC}\cite{kirkpatrick2017overcoming} we had to vary  $\lambda$.
Figure~\ref{fig:mas-permuted} shows the effect of $\lambda$ on the Avg. performance and the Avg. forgetting in   a sequence of 5 permuted MNIST tasks with a 2 layer perceptron~(512 units).  We see the sensitivity around $\lambda=1$ is very low with low forgetting. Note that a better trade-off   could be achieved by setting $\lambda$ to the smallest value that allows an acceptable low forgetting on the previous tasks. However, when moving 
 to a new task we assume no access to the previous tasks data and hence  such a strategy can't be followed.

\myparagraph{ Object recognition adaptation test.}{\label{par-objadapt}}
As we have previously explained, {\tt MAS} has the ability to adapt the importance weights to a specific subset that has been encountered at test time in an unsupervised and online manner. To test this claim, we have selected one class from the {\tt Flowers} dataset, {\tt Krishna Kamal} flower. We learn the 8 tasks sequence as above while assuming {\tt Krishna Kamal} as the only encountered class. %
Hence, importance weights are computed on that subset only. At the end of the sequence, we observe a minimal forgetting on that subset of $2\%$ compared to $8\%$ forgetting on the {\tt Flowers} dataset as a whole. We also observe higher accuracies on later tasks as only changes to important parameters for that class are penalized, leaving more free capacity for remaining tasks (e.g. accuracy of $84\%$ on the last task, instead of $69\%$ without adaptation). We repeat the experiment with two other classes and obtain similar results. This clearly indicates our method ability to adapt to user specific settings and to learn what (not) to forget. 

\begin{figure}[t!]
    \centering
    
    \includegraphics[width=0.55\textwidth]{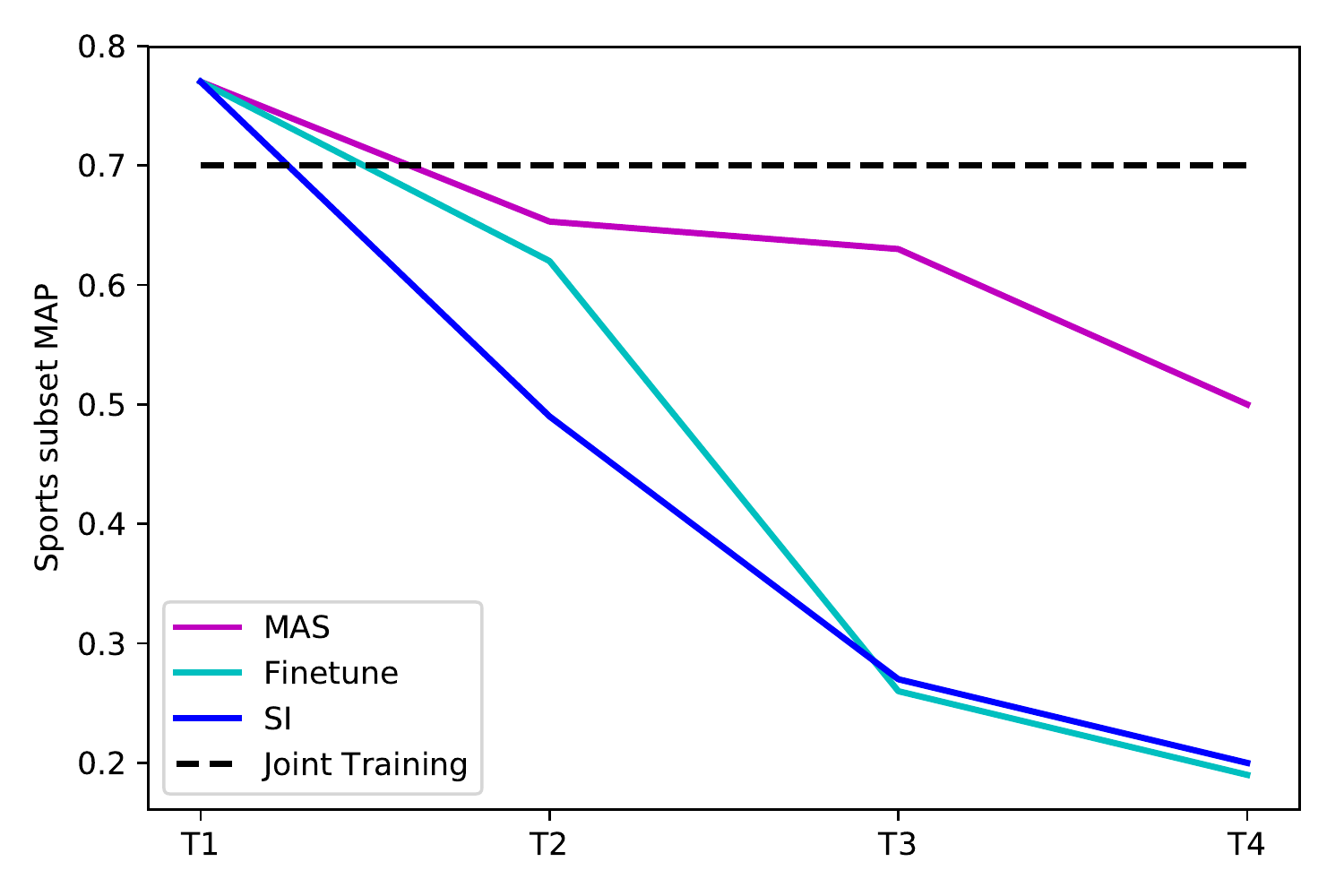}
    
    \caption[ MAP on the sport subset of the 6DS dataset after each task in a 4 tasks sequence.]{\label{fig:mas-adaptationx}    MAP on the sport subset of the 6DS dataset after each task in a 4 tasks sequence. {\tt MAS} managed to learn that the sport subset is important to preserve and prevents significantly the forgetting on this subset.}

 \end{figure}
 
 \begin{figure}[h!] 
  \centering
  \subfloat[]{
    \includegraphics[width=0.34\textwidth]{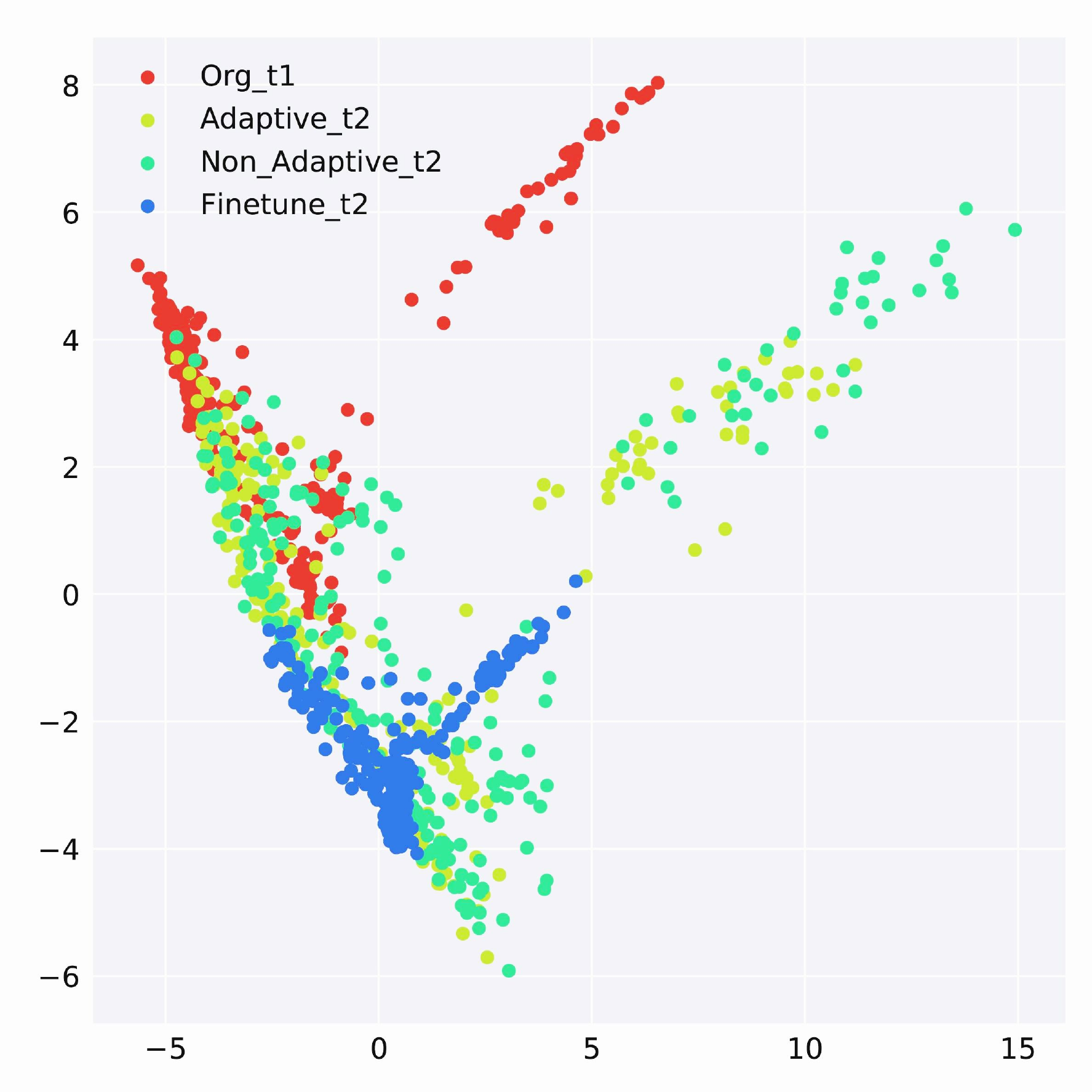}
    \label{fig:mas-t2}}
    \hfillx
  \subfloat[]{
    \includegraphics[width=0.34\textwidth]{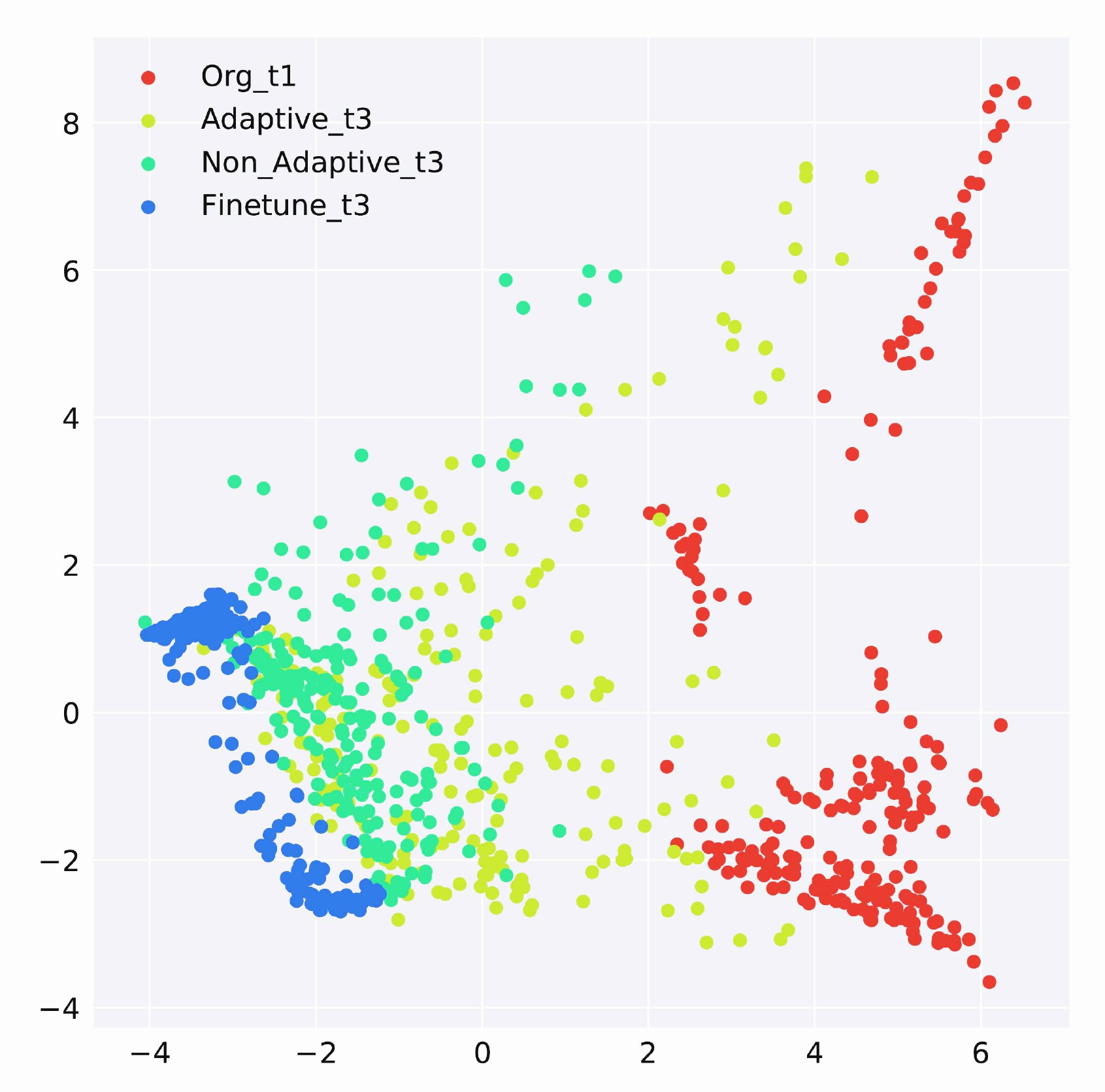}
    \label{fig:mas-t3}}
     \hfillx
  \subfloat[]{
    \includegraphics[width=0.34\textwidth]{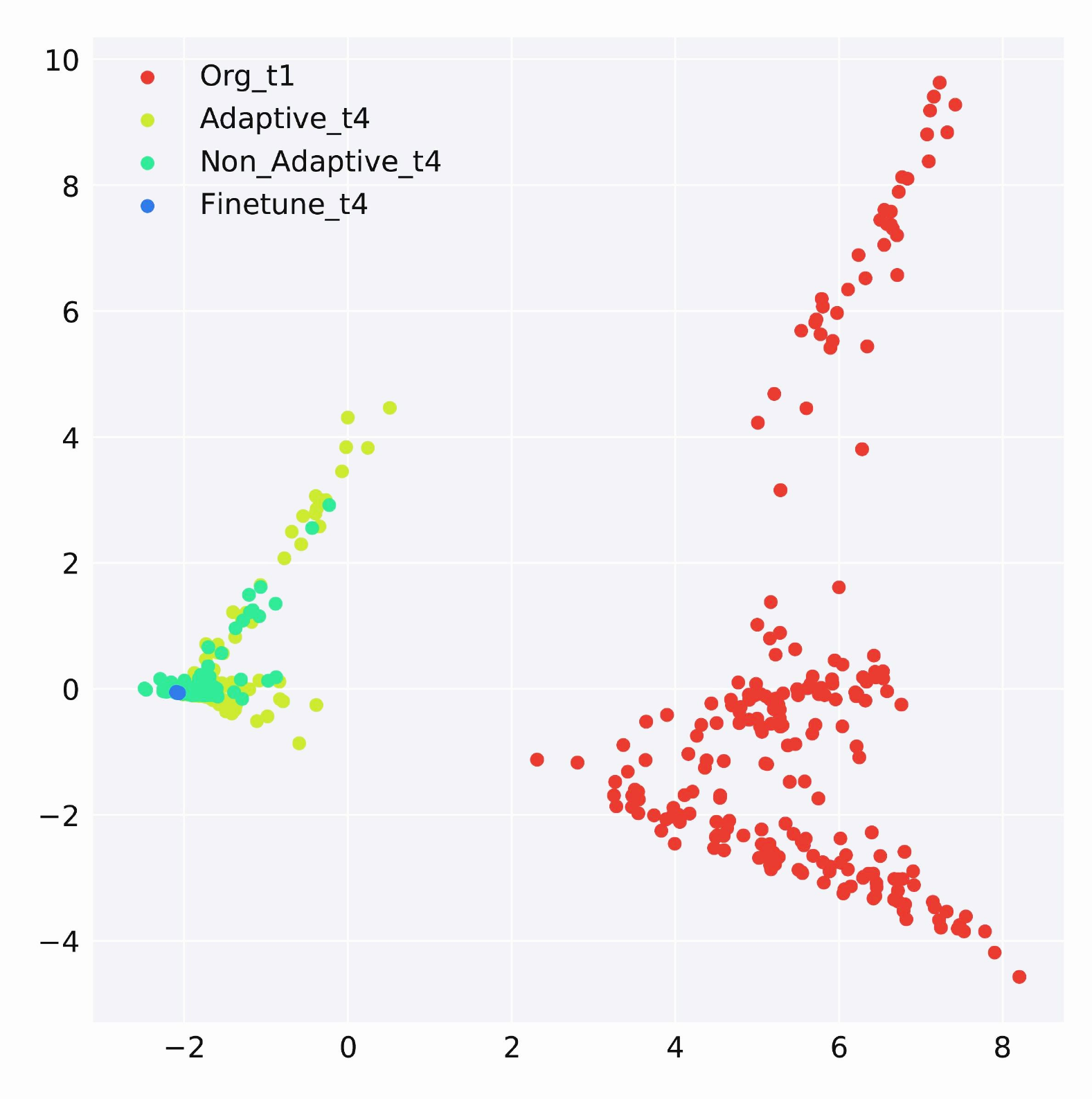}
    \label{fig:mas-t4}}
  \caption{  Projections onto a 2D embedding, after training the second task (a), after training the third task (b) and after training the fourth task (c).  }
\end{figure}
\myparagraph{ Fact learning adaptation test.}{\label{par-factadapt}}
We want to test the adaptation ability of our method, here on a specific subset of facts assumed to be frequently encountered at test time.
To proceed, we clustered the dataset into 4 disjoint groups of facts, representing 4 tasks, and then selected a specialized subset of {$T_1$}, namely 7 facts of person playing sports. 
We run our method with the importance parameters computed only over the examples from this set along the 4 tasks sequence.
Figure~\ref{fig:mas-adaptationx}  shows the achieved  performance on this sport subset by each method at each step of the learning sequence. Joint Training (black dashed) is shown as reference. It violates the continual learning setting as it trains on all data jointly. Note that {\tt SI} can only learn importance weights during training, and therefore cannot adapt to a particular subset. 
Our {\tt MAS} (pink) succeeds to learn that this set is important to preserve and achieves  a performance of 0.50 at the end of the sequence, 
while the performance of {\tt Finetune} and {\tt SI} on this set was close to 0.20.\\



\myparagraph{Visualizing the learned embedding. }\label{sec:mas-sport}
We have showed that our method reduces the forgetting on a specific subset of a task the most among the competitors that do not have this specialization capabilities, Fig.~\ref{fig:mas-adaptationx}. Now, we want to  know what happens in the learned embedding space, i.e.  how the projections of the samples that belong to this subset change along the sequence compared to how they were right after training the first task. For that purpose, we extracted tSNE $2D$ projection of the learned embedding after each task in the sequence. This was done for our method ({\tt MAS}) when adapting to sport subset ({\tt Adaptive}) and our method ({\tt MAS}) when preserving the performance on all facts of the first task ({\tt Non Adaptive}). We also show the projections of the points in the embedding learned by the finetune baseline ({\tt Finetune}, where no regularizer is used). To have a point of reference, we also show the projections of the originally learned representation after the first task ({\tt Org}).   Figure \ref{fig:mas-t2} shows the projections from the different variants after learning the second task compared to the original projections. It can be seen that the Adaptive and Non Adaptive variants of our method try to preserve the projections from this subset. The adaptive projections are closer to the original one, if we look closely, while {\tt Finetune} projections start drifting away from where they were. After the third task, as shown in figure \ref{fig:mas-t3}, the Adaptive projections are closer to the original ones than the Non Adaptive that considers this subset as part of the full task being preserved and tries to prevent forgetting them as well. {\tt Finetune} started destroying the learned topology of this subset and lies further apart. However, when it comes to the fourth task, we see that it is a quite challenging and hard task. The forgetting appears more severe than before and preservation of the projections become even harder. Nevertheless,  the Adaptive {\tt MAS} and Non Adaptive {\tt MAS} still preserve the topology of the learned projections. The Adaptive projections lie closer and look more similar to the originals than the Non Adaptive MAS. {\tt Finetune}, on the other hand, forgets completely about this subset and all the samples get projected in one point where it becomes quite hard to recognize their corresponding facts.

\myparagraph{Correlation between the parameters importance computed  on different sets.}\label{sec:mas-corr}
We want to investigate the correlation or the difference between the importance assigned to the parameters computed on different sets. 

\begin{figure*}[h!]

\minipage{0.45\textwidth}
  \includegraphics[width=\linewidth]{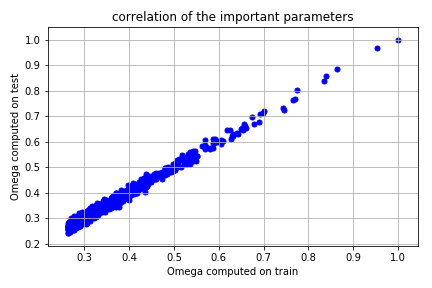}
  \caption[Top most important parameters from $\Omega$  computed on training data.]{Top most important parameters from $\Omega$  computed on training data.  The X-axis represents the values from $\Omega$ computed on training data while the Y-axis represents the values from  $\Omega$ computed on test data.  Importance shown for the last fully connected layer based on  object recognition experiment Birds$\rightarrow$Scenes}\label{fig:mas-cl6_train}
\endminipage\hfill
\minipage{0.45\textwidth}
  \includegraphics[width=\linewidth]{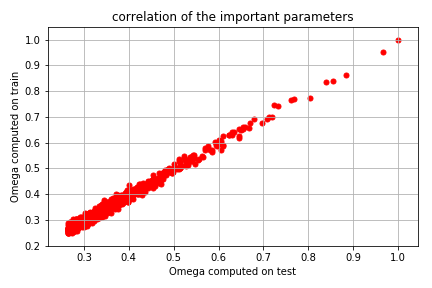}
  \caption[Top important parameters from $\Omega$ computed on test data.]{Top important parameters from $\Omega$ computed on test data. The X-axis represents the values from $\Omega$ computed on test data while the Y-axis represents the values from  $\Omega$ computed on training data. Importance shown for the last fully connected layer based on object recognition experiment Birds$\rightarrow$Scenes}\label{fig:mas-cl6_test}
\endminipage\hfill

\end{figure*}
\begin{figure*}[h!]

\minipage{0.45\textwidth}
  \includegraphics[width=\linewidth]{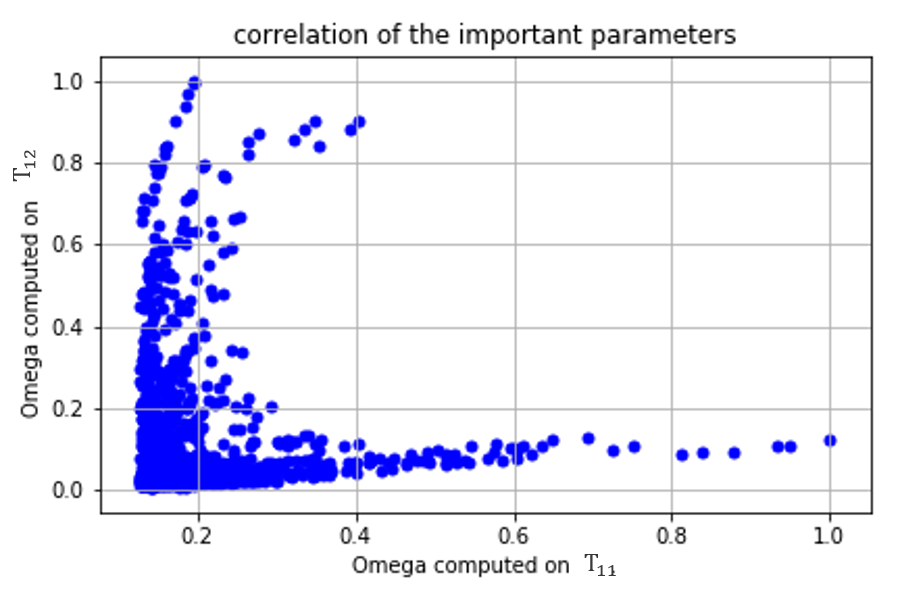}
  \caption[Top most important parameters from $\Omega$ computed on $T_{11}$]{Top most important parameters from $\Omega$ computed on $T_{11}$. The X-axis represents the values from $\Omega$ computed on $T_{11}$ while the Y-axis represents the values from  $\Omega$ computed on $T_{12}$. Importance shown for the last convolutional layer based on two tasks split of the fact learning dataset, 6DS.}\label{fig:mas-b11_b12_s_1}
\endminipage\hfill
\minipage{0.45\textwidth}
  \includegraphics[width=\linewidth]{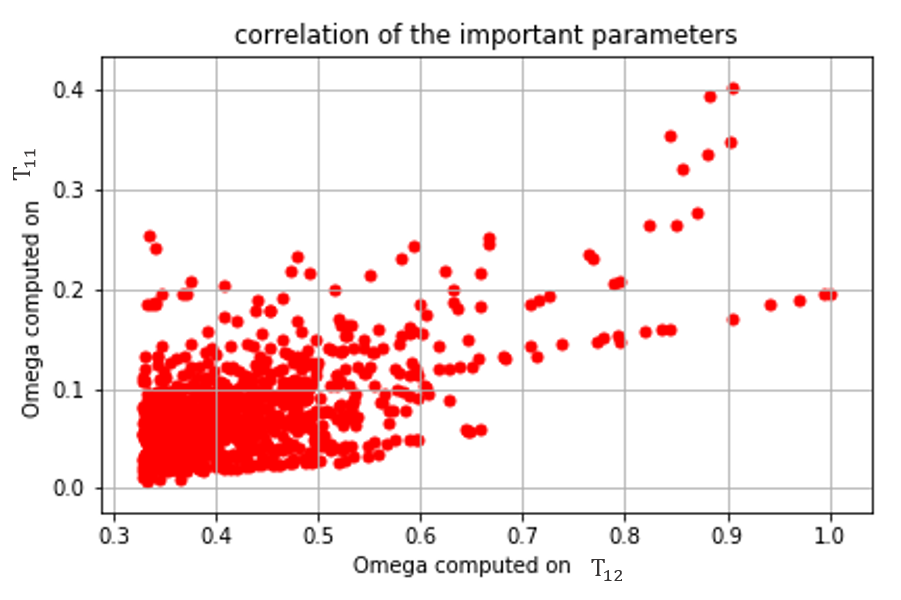}
  \caption[Top most important parameters from $\Omega$ computed on $T_{12}$]{Top most important parameters from $\Omega$ computed on $T_{12}$. The X-axis represents the values from $\Omega$ computed on $T_{12}$ while the Y-axis represents the values from  $\Omega$ computed on $T_{11}$. Importance shown for the last convolutional layer based on two tasks split of the fact learning dataset, 6DS.}\label{fig:mas-b11_b12_s_2}
\endminipage\hfill
\end{figure*} 
First, we compare the estimated parameters importance ($\Omega$)  using the training data and  $\Omega$ computed using the test data. For that, we used a model from the object recognition experiment, namely Birds$\rightarrow$Scenes, the results of which are shown in Table~\ref{tab:mas-res_twotasks_classification}. Figure \ref{fig:mas-cl6_train} shows a scatter plot for the top $1000$ most important parameters according to   the $\Omega$ computed on the training data (blue). The X-axis represents the values from $\Omega$ computed on training data while the Y-axis represents the values from  $\Omega$ computed on test data.
Figure \ref{fig:mas-cl6_test} shows a similar scatter plot for the top $1000$ important parameters according to   the $\Omega$ computed on the test data~(red). Here, the X-axis represents the values from $\Omega$ computed on test data while the Y-axis represents the values from  $\Omega$ computed on training data.
A plot where the points are closely lying around a straight line indicates that the parameters from the two $\Omega$s have similar importance values. A plot where the points are spread further from  such a line and scattered among the plotted area  indicates a lower correlation between the $\Omega$s. 
It can be seen how similar are the importance values computed on test data to those computed on training data where they form a tight grouping of points around a straight line where the values would be identical. This demonstrates our method's ability to correctly identify the important parameters in an unsupervised manner, regardless of what set is used for that purpose as long as it covers the different classes or concepts of the task at hand.  \\

How about using different subsets that cover a partial set of classes or concepts from a task? We have shown that computing the importance on one subset results in a better preservation of performance compared with the other subset that was not used for computing the importance. This suggests that the importance of the parameters differs while using different subsets. To further investigate  this claim, we split the Fact learning dataset, 6DS, into two groups of randomly selected facts resulting in two tasks. We then  split the test set of the first task data into two random subsets of facts, $T_{11}$ and $T_{12}$. After learning the task $T_1$, the importance of the parameters is computed using one subset only ($T_{11}$ or $T_{12}$).
We plotted the values of $\Omega$ for the $1000$ top most important parameters estimated on the $T_{11}$ (in blue) subset of the training data  from the first task $T_1$ along with the same parameters but with their importance computed using the other subset $T_{12}$, Figures  \ref{fig:mas-b11_b12_s_1} and \ref{fig:mas-b11_b12_s_2}.

This suggests that the method identifies the important parameters needed for each subset and when those parameters are shared the parameters importance is correlated between the two subsets while when those are different, different parameters receive different importance values based on the used subset. 

\section{Summary}  \label{sec:mas-conculosion}
In this chapter, we argued that given a limited model capacity and unlimited evolving tasks, it is not possible to  preserve all the previous knowledge. Instead, agents should learn what (not) to forget. Forgetting should relate to the rate at which a specific piece of knowledge is used. This is similar to how biological systems are learning. In the absence of error signals, synapses connecting biological neurons strengthen or weaken based on the
concurrence of the connected neurons activations. Inspired by the synaptic plasticity, we proposed a method that is able to learn the importance of network parameters from the input data that the system is active on, in an unsupervised manner. We showed that a local variant of our method can be seen  as  an application of Hebb's rule in learning the importance of parameters. We first tested our method on a sequence of object recognition problems in a traditional incremental task learning setting. We then moved to a more challenging test case where we learn facts from images in a continuous manner. We showed i) the ability of our method to better learn the importance of the parameters using training data, test data or both;   ii) at time of development~(2018)  state-of-the-art performance on all the designed experiments and iii) the ability of our method to adapt the importance of the parameters towards a frequent set of data. We believe that this is a step forward in developing systems that can always learn and adapt in a flexible manner.

\cleardoublepage
  \graphicspath{{\chaptersdir/SSL/image/}}%
  \chapter{Sparsity in Continual Learning}\label{ch:ssl}
In this chapter, we study the role of sparsity and representation decorrelation in the context of continual learning. Given a scenario with fixed model capacity, we postulate that the learning process should not be selfish, i.e. it should account for future tasks to be added and 
thus leave enough capacity for them. 
To achieve {\emph{Selfless Continual Learning}} we study 
different regularization strategies and activation functions.
We find that 
imposing sparsity at the level of the representation (i.e.~neuron activations)
is more beneficial for continual learning than encouraging parameter sparsity.

In particular, we propose a novel regularizer,
that encourages representation sparsity by means of neural inhibition. It results in few active neurons  which in turn leaves more free neurons to be utilized by upcoming tasks. 
As
neural inhibition over an entire layer
can be too drastic, especially for 
complex tasks requiring strong representations,
our regularizer only inhibits other neurons in a local neighbourhood, inspired by lateral inhibition processes in the brain.
We combine our novel regularizer
with  MAS ( see Chapter~\ref{ch:MAS}), our method that penalizes  changes to important previously learned parts of the network. We show that our new regularizer leads to increased sparsity which translates in consistent performance improvement 
on diverse datasets.

This work was also a collaboration with Marcus Rohrbach from FAIR. It was published as an article in ICLR 2019~\cite{aljundi2018selfless}.
\section{Introduction}\label{sec:scl-intro}
The key challenge of continual learning is how to 
avoid 
catastrophic interference with 
the tasks learned previously \cite{french1999catastrophic}. 
 Similar to the previous chapters~(\ref{ch:ebll},~\ref{ch:MAS}),  we are interested in learning a sequence of tasks \emph{without access to any previous or future task data} and \emph{restricted to a fixed  model capacity},
     as also studied in \cite{kirkpatrick2017overcoming,fernando2017pathnet,mallya2018packnet,Serr2018Hard}. 
     This scenario not only has many practical benefits, including privacy and scalability, but also resembles more closely how the mammalian brain learns tasks over time.

The mammalian brain is composed of billions of neurons.
Yet at any given time,
information is represented by only a few active neurons~\cite{lennie2003cost,Olshausen2004Sparse}. 
In neural biology, \emph{lateral inhibition} describes the 
process where
an activated neuron 
reduces the activity of its weaker neighbors. This creates a powerful decorrelated and compact 
representation with minimum interference between different input patterns in the brain~\cite{yu2014sparse,Lie2012Lateral}. 
This is in stark contrast with artificial neural networks, which typically learn dense representations that are highly entangled~\cite{bengio2009learning}. 
Such an entangled representation is quite sensitive to changes in the input patterns, in that it responds differently to input patterns with only small variations. 
\cite{french1999catastrophic} suggests that an overlapped 
representation 
plays a crucial role 
in catastrophic forgetting and reducing this overlap would result in  reduced  interference.
\cite{cogswell2015reducing} shows that when the amount of overfitting  in a neural network is reduced, the  representation correlation is also reduced.  As such, learning a disentangled representation is more powerful and less vulnerable to catastrophic interference. However, if the learned  disentangled representation at a given task is  not sparse, only little capacity is left for learning new tasks. 
This would in turn result in  either an underfitting to the new tasks or again a forgetting of previous tasks. In contrast,  a sparse and decorrelated representation would lead to a powerful representation and at the same time enough free neurons that can be changed without interference with the neural activations learned for the previous tasks. 

\begin{figure*}[t]
\centering
\includegraphics[width=1\textwidth]{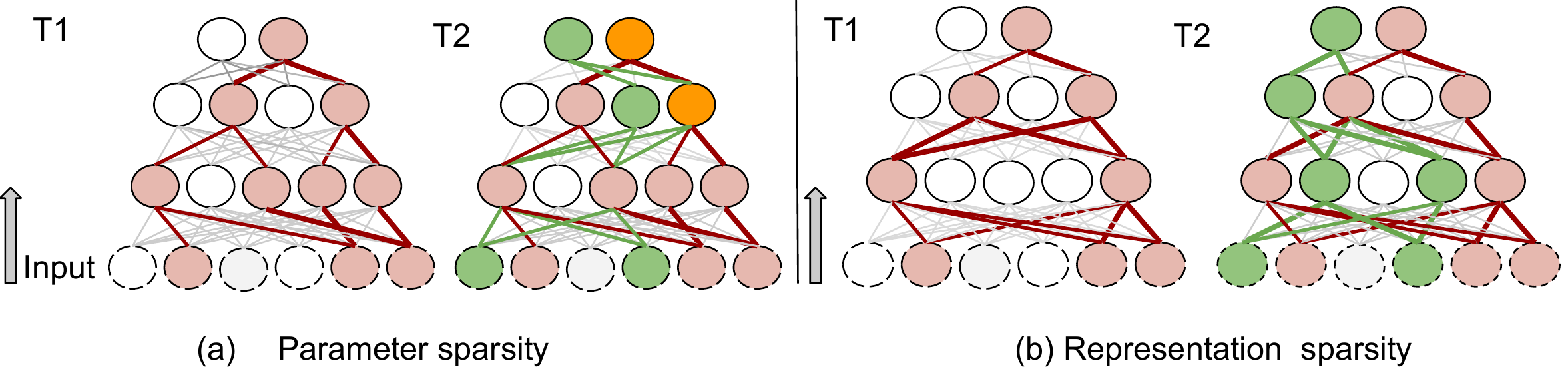} 
  \caption[The difference between parameter sparsity (a) and representation sparsity (b) in a simple two tasks case.]{ The difference between parameter sparsity (a) and representation sparsity (b) in a simple two tasks case. First layer indicates input patterns. Learning the first task utilizes parts indicated in red. Task 2 has different input patterns and 
  {uses}
  parts shown in green. Orange indicates changed neurons activations as a result of the second task. 
  In (a), when an example from the first task is encountered again, the activations of the first layer will not be affected by the changes, however, the second and later layer activations are changed.
  Such interference is 
    {largely reduced}
  when
  {imposing}
  sparsity on the representation (b). }  
  \label{fig:scl-param_vs_rep}

\end{figure*}

In general, sparsity in neural networks can be thought of either in terms of the network parameters or in terms of the representation (i.e. the activations). In this chapter we postulate, and confirm 
experimentally, that a sparse and decorrelated
representation is preferable over parameter sparsity 
in a sequential learning scenario. 
There are two arguments for this: first, a
sparse
representation is less sensitive to new and different patterns (such as data from new tasks) and second,  the training procedure of the new tasks can use the free neurons leading to less interference with the previous tasks, hence reducing forgetting.
In contrast, when the effective parameters are spread among different neurons,  changing the ineffective ones would change the function of their corresponding neurons and hence interfere with previous tasks (see also Figure~\ref{fig:scl-param_vs_rep}).
Based on these observations, we  propose a new regularizer that exhibits a behavior similar to the lateral inhibition in  biological neurons. 
The main idea of our regularizer is to penalize 
neurons that are active at the same time. This leads to more sparsity
 and a decorrelated representation. 
However, complex tasks
may actually require multiple active neurons in a layer at the same time to learn a strong representation.  Therefore, our regularizer, \textbf{\SLNIDFull},  only penalizes neurons locally. Furthermore, we don't want inhibition to affect previously learned tasks, even if later tasks use neurons from earlier tasks. An important component of \SLNI is thus to discount inhibition from/to neurons which have high \emph{neuron importance} --  a new concept that we introduce in analogy to parameter importance~\cite{kirkpatrick2017overcoming,Zenke2017improved,aljundi2018memory}.
 When combined
 with a prior focused  method \cite{aljundi2018memory,kirkpatrick2017overcoming}, 
 our proposed regularizer leads to  sparse and decorrelated representations which improve the continual learning performance.

The contribution of the work presented in this chapter is threefold. First, we direct attention to { Selfless Continual Learning } and study a diverse set of representation based regularizers, parameter based regularizers, as well as 
sparsity inducing activation functions. These 
have not been  studied extensively in the continual learning  literature before. Second, we propose a novel regularizer, \SLNI, which is inspired by lateral inhibition in the brain. Third, we show that our proposed regularizer  consistently outperforms alternatives on three diverse datasets (Permuted MNIST, CIFAR, Tiny ImageNet) and we compare to and outperform  state-of-the-art continual learning approaches at the time of development~(2018) on the  8 tasks object classification sequence~(see Chapter~\ref{ch:MAS}, Section~\ref{sec:mas-objrec}) .
 \SLNI can be applied to different regularization based continual learning methods, and we show experiments with MAS \cite{aljundi2018memory} and EWC \cite{kirkpatrick2017overcoming}.

In the following, we first 
discuss 
closely related continual learning methods 
and 
different regularization criteria from a 
continual learning
perspective  (Section~\ref{sec:scl-related}). We proceed by introducing Selfless Continual Learning and detailing our novel regularizer (Section~\ref{sec:scl-method:ssl}). %
Section~\ref{sec:scl-experiments} describes our experimental evaluation, while Section~\ref{sec:scl-conclusion} concludes the chapter.

\section{Related Work}\label{sec:scl-related}

In this chapter, we focus on learning a sequence of tasks using a fixed model capacity, i.e.~with a fixed architecture and fixed number of parameters. Under this setting, we consider the prior-focused branch of the regularization based family~\cite{kirkpatrick2017overcoming,Zenke2017improved,chaudhry2018riemannian,liu2018rotate}~( see Chapter~\ref{ch:relatedwork}, Section~\ref{sec:related_regularization}). 

A common drawback of all   methods in this family is that 
learning a task could  utilize a good portion of the network capacity, leaving few  ``free'' neurons to be adapted by the new task. This in turn leads to inferior performance on the newly learned tasks or forgetting the previously learned ones, as we will show in the experiments.

While in the previous chapter we presented our method, MAS, that learns what is important to remember leaving the future tasks with more freedom to adjust the network parameters, here we study the role of sparsity and representation decorrelation in continual learning.  
This aspect has not received much attention in the literature yet.
Recently,
\cite{Serr2018Hard} 
proposed to overcome catastrophic forgetting through learned hard attention masks for each task with $\ell_1$ regularization imposed on the accumulated hard attention masks. This comes closer to our approach although we  study and propose a regularization scheme on the learned representation.

The concept of reducing the representation overlap was suggested before in early attempts towards overcoming catastrophic forgetting in neural networks~\cite{french1999catastrophic}. This  led to several methods with the goal of orthogonalizing  the activations~\cite{french1992semi,french1994dynamically,kruschke1992alcove,kruschke1993human,sloman1992reducing}. However, these approaches are mainly designed for specific architectures and activation functions, which makes it hard to integrate them in recent neural network structures. 

The sparsification of neural networks has mostly been studied for compression. SVD decomposition can be applied 
to reduce the number of effective parameters~\cite{xue2013restructuring}. However, there is no guarantee that the training procedure converges to a low rank weight matrix. Other works iterate between pruning and retraining of a neural network as a post processing step \cite{liu2015sparse,sun2016sparsifying,aghasi2017net,louizos2017bayesian}. While compressing a neural network by removing parameters  leads to a sparser neural network, this does not necessarily lead to a sparser representation. 
Indeed, a weight vector can be highly sparse but spread among the different neurons. This reduces the effective size of a neural network, from a compression point of view, but it would not be beneficial for later tasks as most of the neurons are already occupied by the current set of tasks. In our experiments, we show the difference between using a sparse penalty on the representation versus applying it to  the weights.
\section{Selfless Continual Learning}
\label{sec:scl-method:ssl}
One of the main challenges in single model incremental  learning is to have capacity to learn new tasks and at the same time avoid 
catastrophic forgetting of previous tasks as a result of learning new tasks. In order to prevent catastrophic forgetting, as we have shown in the previous chapter, importance weight based  methods like our
MAS \cite{aljundi2018memory} or EWC \cite{kirkpatrick2017overcoming} 
introduce
an importance weight $\Omega_{ij}$ for each parameter $\theta_{ij}$ in the network.
While each method differs in how to estimate the important parameters, 
 changes to important parameters  when learning a new task $T_{\mathcal{T}}$ are typically discouraged using an $\ell_2$ penalty: 
\begin{equation} 
{J(\theta)}=\quad \frac{1}{N_{\mathcal{T}}} \sum_{n=1}^{N_{\mathcal{T}}}  \ell(f(x_n;\theta),y_n)+ \lambda\sum_{i,j} \Omega_{ij}(\theta_{ij}-\theta^{\mathcal{T}-1}_{ij})^2
\end{equation}
where $\theta^{\mathcal{T}-1}=\{ \theta^{\mathcal{T}-1}_{ij}\}$ are the optimal parameters learned so far, i.e. before the current task.  $\{x_n\}$ is the set of $N_{\mathcal{T}}$ training inputs,  with $\{f(x_n;\theta)\}$ and $\{y_n\}$ the corresponding predicted and desired outputs, respectively. 
$\lambda$ is a trade-off parameter between the new  task objective {$\ell$} and the changes on the important parameters, i.e. the amount of forgetting.

In this work, we introduce an additional regularizer $R_{\textrm{SCL}}$ (SCL short for Selfless Continual Learning ). It encourages sparsity in the activations $H_l=\{h^n_{j}\}$ for each layer $l$. The training of the new task minimizes the following objective function:
\begin{equation}
\label{eq:appraoch:loss:final}
{J(\theta)}= \quad  \frac{1}{N_{\mathcal{T}}} \sum_{n=1}^{N_{\mathcal{T}}} \ell(f(x_n;\theta),y_n)+ \lambda\sum_{i,j} \Omega_{ij}(\theta_{ij}-\theta^{\mathcal{T}-1}_{ij})^2+\lambda_{\textrm{SCL}} \sum_{l} R_{\textrm{SCL}}(H_l)
\end{equation} 
 $\lambda_{\textrm{SCL}}$ and $\lambda$ are trade-off parameters that control the contribution of each term.
When training the first task (${\mathcal{T}}=1$),   $\Omega_{ij}=0$. %

\subsection{Sparse Coding through Neural Inhibition}
\label{sec:scl-method:sni}
We consider  a hidden layer $l$ with activations $H_l=\{h^n_{j}\}$ for 
a set of input samples $X=\{x_n\}_{n=1}^{N_{\mathcal{T}}}$ in a task $T_{\mathcal{T}}$, 
and $j \in {1,..,J}$ 
running over all $J$ neurons in the hidden layer.
Here, we describe how we obtain a sparse and decorrelated representation. In the literature, sparsity has been proposed by~\cite{pmlr-v15-glorot11a} to be combined with the rectifier activation function (ReLU) to control unbounded activations and to increase sparsity.
They minimize the  $\ell_1$ norm of the activations (since minimizing the  $\ell_0$ norm is an NP hard problem).  However,  $\ell_1$ norm imposes an equal penalty on all the active neurons leading to small activation magnitude across the network.

Learning a decorrelated representation, on the other hand, has been explored before  with the goal of reducing overfitting. This is usually done by minimizing the Frobenius norm of the covariance matrix corrected by the diagonal, as in \cite{cogswell2015reducing} or \cite{xiong2016regularizing}. More formally,  decorrelating the representation following~\cite{cogswell2015reducing} can be imposed by Equation~\ref{eq:scl-forbnorm}:
\begin{equation}\label{eq:scl-forbnorm}
   R(H_l)=\sum_{j,k} \Big(\frac{1}{N_{\mathcal{T}}}  \sum_n  (h_j^n-\mu_j)(h_k^n-\mu_k)\Big)^2  ,  \quad k\neq j 
\end{equation}
where $j,k \in {1,..,J}$. Such a penalty results in a decorrelated representation but with activations that are mostly close to a non zero mean value.
We merge the two objectives of sparse and decorrelated representation  in
the following objective:
\begin{equation}
\label{eq:appraoch:regularizer:simple}
R_{\SNIwoNI}(H_l)=\frac{1}{N_{\mathcal{T}}}  \sum_{j,k}\sum_n  h_j^nh_k^n  ,  \quad k\neq j
\end{equation}

 This formula differs from minimizing the Frobenius norm of the covariance matrix  in two simple yet important aspects:\\
 (1) In the case of a ReLU activation function, used in most modern architectures, a neuron is active  if its output is larger than zero, and zero otherwise. By assuming a close to zero mean of the 
 activations, $ \mu_j \simeq 0 \, \forall j \in {1,..,J}$, we minimize the 
 correlation between any two active neurons.\\
(2) We don't use the Frobenius norm and therefor by evaluating the derivative of the presented regularizer w.r.t. the activation, we get:
\begin{equation}
\frac {\partial R_{\SNIwoNI}(H_l)}{\partial h_j^n}=\frac{1}{N_{\mathcal{T}}}   \sum_{k\neq j} h_k^n
\end{equation}
i.e., each active neuron receives a penalty from every other active neuron that corresponds to that other neuron's activation magnitude. In other  words, if a neuron fires, with a high activation value, for a given example, it will suppress firing of other neurons for that same example. Hence, this results in a decorrelated sparse representation. It is worth noting that we impose  sparsity per example, allowing our method to be extended to online setting. We hypothesize that neurons  firing strongly for a given example will fire for examples with similar patterns. As such imposing per example sparsity would lead to  neuron activation sparsity   across different examples. We empirically validate  our hypothesis in the  experiments Section~\ref{sec:scl-experiments:analysis}.
\subsection{Sparse Coding through Local Neural Inhibition}
\label{sec:scl-method:SLNI}
The loss imposed by the %
$\SNIwoNI$
objective will only be zero when there is 
at most
one active neuron per example. This seems to be too harsh for complex 
tasks
that need a richer representation. Thus, we suggest to relax the objective by imposing a spatial weighting to the correlation penalty. 
In other words, an active neuron penalizes mostly its close neighbours and this effect vanishes for 
neurons further away.
Instead of uniformly penalizing all the correlated neurons, 
we weight the correlation penalty between  two neurons  with locations $k$ and $j$ %
using a Gaussian weighting. 
 This gives
\begin{equation}
\label{eq:appraoch:regularizer:first}
R_{\SLNIwoNI}(H_l)=
\frac{1}{N_{\mathcal{T}}} \sum_{j,k} e^{-\frac{(j-k)^2}{2\sigma^{2}}}\sum_n  h_j^nh_k^n  ,  \quad j\neq k
\end{equation}
As such, each active neuron inhibits its neighbours,
introducing a locality in the network inspired by biological neurons. Note that by adding the local weighting we impose a topology on each layer of the network.
 While the notion of neighbouring neurons is not  well established in a fully connected network, our aim is to allow few  neurons to be active and not only one, thus those few activations don't have to be small to compensate for the penalty.
$\sigma^2$ is a hyper parameter representing the scale at which neurons can affect each other. Note that this is somewhat more flexible than decorrelating neurons in fixed groups as used in~\cite{xiong2016regularizing}.
Our regularizer inhibits locally the  active neurons leading to a sparse coding through local neural inhibition. 
\subsection{Neuron Importance for Discounting Inhibition}
Our regularizer is to be applied for each task in the learning sequence. In the case of  tasks with completely different input patterns, the active neurons of the previous tasks will not be activated given the new tasks input patterns. However, when the new tasks are of similar
or shared patterns, neurons used for previous tasks will be active. In that case, our penalty would discourage other neurons from being active and encourage the new task to adapt the already active neurons instead. This would interfere with the previous tasks and could increase forgetting which is exactly what we want to overcome. To avoid such interference, we  add a weight factor taking into account the importance of the neurons with respect to the previous tasks. To estimate the importance of the neurons, we use as a measure the sensitivity of the loss at the end of the training to their changes. This is approximated by the gradients of the loss w.r.t. the neurons outputs (before the activation function) evaluated at each data point.
To get an importance value, we then accumulate the absolute value of the gradients over the given data points %
obtaining importance weight $\alpha_{j}$ for neuron $\mathcal{N}_{j}$: 
\begin{equation}\label{eq:2}
\alpha_{j} =\frac{1}{N_{{t}}}\sum_{n=1}^{N_{t}}  \mid g_{j}(x_n) \mid \quad,\quad g_{j}(x_n)=\frac{\partial (\ell(f(x_n;\theta_{{t}}),y_n))}{\partial out^{n}_{j}}
\end{equation} 
 where $out^{n}_{j}$ is the output of  neuron $\mathcal{N}_j$ for a given input example $x_n$, and $\theta^{{t}}$ are the parameters after learning task $T_{{t}}$. This is in line with the  estimation of the parameters importance in  EWC~\cite{kirkpatrick2017overcoming} 
 but considering the derivation variables to be the neurons outputs instead of the parameters.\\
Instead of relying on the gradient of the loss, we can also use the gradient of the learned function, i.e. the output layer, as we have proposed in Chapter~\ref{ch:MAS} for estimating the parameters importance. During the early phases of this work, we experimented with both and observed a similar behaviour. For sake of consistency and computational efficiency 
we utilize the gradient of the function when using our MAS~\cite{aljundi2018memory} as continual learning method and the gradient of the loss when experimenting with EWC~\cite{kirkpatrick2017overcoming}.
Then, we can weight our regularizer as follows:
\begin{equation} 
\label{eq:appraoch:regularizer:final}
R_{\SLNI}(H_l)=\frac{1}{N_{\mathcal{T}}}   \sum_{j,k}e^{-(\alpha_{j}+\alpha_{k})} e^{-\frac{(j-k)^2}{2\sigma^{2}}} \sum_n h_j^nh_k^n ,  \quad k\neq j
\end{equation}
which can be read as: if  an important neuron for a previous task is active given  an input pattern from the current task, it will not suppress the other neurons from being active neither be affected by other active neurons.
For all other active neurons, local inhibition is deployed.
The final objective for training is given in Equation \ref{eq:appraoch:loss:final}, setting $R_{SCL}:=R_{\SLNI}$ and $\lambda_{\textrm{SCL}}:=\lambda_{\SLNI}$. We refer to our full method as \SLNIDFull.

\section{Experiments}\label{sec:scl-experiments}
In this section we study the 
 role of 
 standard
 regularization techniques with a focus on sparsity and decorrelation of the representation 
in a task incremental  scenario.  We first 
compare different activation functions and regularization techniques, including our proposed \SLNI,  on permuted MNIST (Section \ref{sec:scl-experiments:analysis}). 
Then, we compare the top competing techniques and our proposed method in the case of incrementally learning  Cifar-100 classes and 
Tiny ImageNet classes (Section \ref{sec:scl-experiments:cifarimagenet}).  
Our \SLNI regularizer can be integrated in any regularization-based continual learning approach such as \cite{kirkpatrick2017overcoming,Zenke2017improved,aljundi2018memory}. Here we focus on Memory Aware Synapses~\cite{aljundi2018memory} ({\tt  MAS}), our method proposed in Chapter~\ref{ch:MAS}, which is easy to integrate and  has shown superior performance. However, we also show results with Elastic weight consolidation~\cite{kirkpatrick2017overcoming}({\tt  EWC}) in Section \ref{sec:scl-experiments:EWC}. Further, we ablate the components of our regularizer, {both in the standard setting (}Section \ref{sec:scl-experiments:ablation})
{as in a setting without hard task boundaries (Section~\ref{sec:scl-experiments:noboundaries})}.
Finally, we show how our regularizer improves the state-of-the-art performance on a sequence of object recognition tasks
(Section \ref{sec:scl-experiments:object}).

\subsection{An In-depth Comparison of Regularizers and Activation Functions for Selfless Continual Learning}
\label{sec:scl-experiments:analysis}
We study possible regularization techniques that could lead to less interference between the different tasks in a continual learning scenario either by enforcing sparsity or decorrelation. Additionally, we examine the use of  activation functions that are inspired by lateral inhibition in biological neurons that could be advantageous in continual learning.  MAS \cite{aljundi2018memory} is used in all cases as continual learning method.\\

\myparagraph{Representation Based methods:} \\
-~{\tt L1-Rep}: to promote representational sparsity, an $\ell_1$ penalty on the activations is used.\\
-~{\tt Decov}~\cite{cogswell2015reducing}   aims at reducing overfitting by decorrelating  neuron activations. To do so, it minimizes the Frobenius norm of the  covariance matrix computed on the activations of the current  batch after subtracting the diagonal to avoid  penalizing independent  neuron activations. \\

\textbf{Activation functions:} \\
- {\tt Maxout} network~\cite{goodfellow2013maxout} utilizes the maxout activation function. For each group of neurons, based on a fixed window size, only the maximum activation is forwarded to the next layer. The activation function guarantees a minimum sparsity rate defined by the window size.  \\
- {\tt LWTA}~\cite{NIPS2013_5059}: similar idea to the Maxout network except that the non-maximum activations are set to zero while maintaining their connections. In contrast to Maxout, LWTA keeps the connections of the inactive neurons which can be occupied later once they are activated without changing the previously active neuron connections.\\
- {\tt ReLU}~\cite{pmlr-v15-glorot11a} the rectifier activation function (ReLU)  used as a baseline here and indicated in later experiments as {\tt No-Reg} as it represents the standard setting of continual learning on networks with ReLU.  All the studied regularizers use ReLU as activation function.\\

\myparagraph{Parameters based regularizers:} \\
- {\tt OrthReg}~\cite{rodriguez2016regularizing}: regularizing CNNs with locally constrained decorrelations. It aims at decorrelating the feature detectors  by minimizing the cosine of the angle between the weight vectors resulting eventually in orthogonal weight vectors.\\
- {\tt L2-WD}: weight decay with $\ell_2$ norm ~\cite{krogh1992simple} controls the complexity of the learned function by minimizing the magnitude of the weights.\\
- {\tt L1-Param}:  $\ell_1$ penalty on the parameters  to encourage a solution with sparse parameters.

Dropout is not considered as its role contradicts our goal. While dropout can improve each task performance and reduce overfitting, it acts as a model averaging technique.   By randomly masking neurons, dropout forces the different neurons to work independently. As such it encourages a redundant representation. As shown by~\cite{goodfellow2013empirical} the best network size for classifying  MNIST digits when using dropout was  about $50\%$ more than without it. Dropout steers the learning of a task towards occupying a good portion of the network capacity, if not all of it, which contradicts the continual learning needs.

\myparagraph{Experimental setup.}
We use the MNIST dataset \cite{lecun1998gradient} as a first task in a sequence of 5  tasks, where we randomly permute  all the input pixels differently for  tasks 2 to 5. The goal is to classify MNIST digits from all the different permutations. 
The complete random permutation of the pixels in each task requires the neural network to instantiate a new neural representation for each pattern.
A similar setup has been used by~\cite{kirkpatrick2017overcoming,Zenke2017improved,goodfellow2013empirical} with different percentage of permutations or different number of tasks.  \\
As a base network, we employ a multi layer perceptron with two hidden layers and a Softmax loss.  We experiment with  different number of neurons in the hidden layers $\{128,64\}$. 
For \SLNI we evaluate the effect of  $\lambda_{\SLNI}$ on the performance and the obtained sparsity
in Figure~\ref{fig:scl-importance_sparsity}.
In general, the best $\lambda_{\SLNI}$ is the minimum value that maintains similar or better accuracy on the first task compared to the unregularized case, and we suggest to use this as a rule-of-thumb to set  $\lambda_{\SLNI}$.  
For $\lambda$, we have used a high $\lambda$ value that ensures the least forgetting which allows us to examine the degradation in the performance  on the later tasks compared to those learned previously as a result of lacking capacity. Note that better average accuracies can be obtained with tuned $\lambda$ although it is not clear how to do so in continual learning settings.

 All tasks are trained for 10 epochs with a learning rate $10^{-2}$ using SGD optimizer. ReLU is used as an activation function unless mentioned otherwise.
Throughout the experiment, we used a scale $\sigma$ for the Gaussian function used for the local inhibition equal to $1/6$ of the hidden layer size. 
For all competing
regularizers, we tested different hyper parameters from $10^{-2}$ to $10^{-9}$ and report the best one.

\begin{figure*}[h!]
\centering
\includegraphics[width=0.92\textwidth]{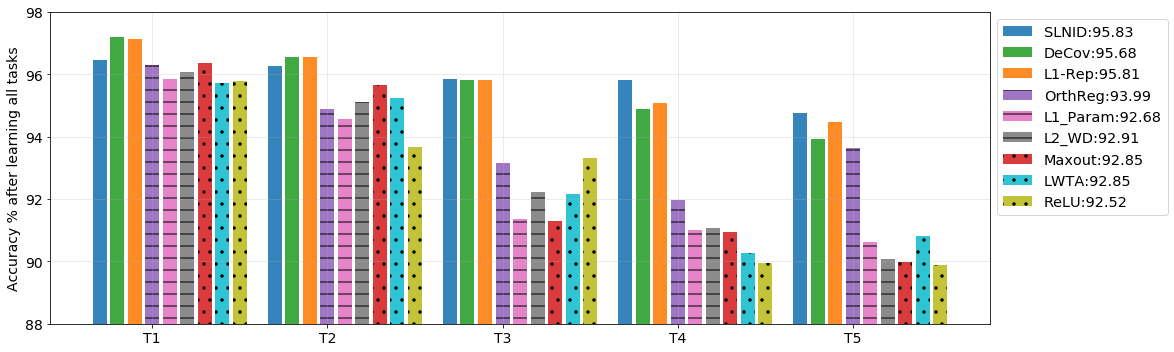}

     \caption[Comparison of different regularization techniques on 5 permuted MNIST sequence, hidden size=128.]{ Comparison of different regularization techniques on 5 permuted MNIST sequence, hidden size=128. Representation based regularizers are solid bars, bars with lines represent parameters regularizers, dotted bars represent activation functions. Average test accuracy over all tasks is given in the legend. Representation based regularizers achieve  higher performance than other compared methods including parameters based regularizers. Our regularizer, \SLNI, performs the best on the last two tasks indicating that more capacity is left to learn these tasks.}%
    \label{fig:scl-perm_comp}%
\end{figure*}

\begin{figure*}[h!]
\centering
\includegraphics[width=0.92\textwidth]{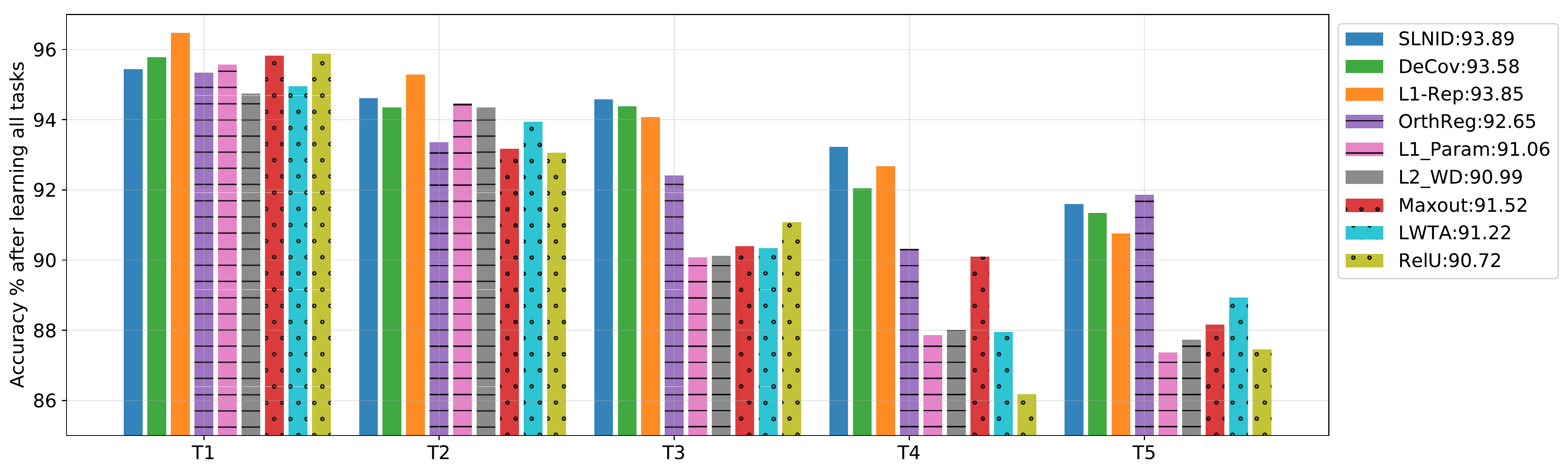} 
     \caption[Comparison of different regularization techniques on 5 permuted MNIST sequence of tasks, hidden size=64.]{ Comparison of different regularization techniques on 5 permuted MNIST sequence of tasks, hidden size=64. Representation based regularizers are solid bars, bars with lines represent parameters regularizers, dotted bars represent activation functions. See Figure~\ref{fig:scl-perm_comp} for size $128$.}%
    \label{fig:scl-perm_comp_64}%
\end{figure*}

\myparagraph{Results:}
Figures~\ref{fig:scl-perm_comp} and ~\ref{fig:scl-perm_comp_64} present the test accuracy on each task at the end of the sequence, achieved by the different regularizers and activation functions on the network with hidden layer of size $128$ and $64$ respectively.
Clearly, in all the different tasks, the representational regularizers show a superior performance to the other studied techniques. For the regularizers applied to the parameters,  {\tt L2-WD} and {\tt L1-Param}
do not exhibit a clear trend and do not systematically show an improvement over the use of the different activation functions only. While {\tt OrthReg} shows a consistently good performance, this performance is lower than what can be achieved by the representational regularizers.
It is worth noting the {\tt L1-Rep} yields superior performance over {\tt L1-Param}. This observation is consistent across different sizes of the hidden layers  and shows the advantage of encouraging sparsity in the activations compared to that in the parameters.
Regarding the activation functions, {\tt Maxout} and {\tt LWTA} achieve a slightly higher performance than {\tt ReLU}. We did not observe a significant difference between the two activation functions. However, the improvement over {\tt ReLU} is only moderate
and does not justify the use of
a fixed window size and special architecture design.
Our proposed regularizer \SLNI achieves high if not the highest performance in all the tasks and succeeds in having a stable performance. This indicates the ability of \SLNI to direct the learning process towards using minimum amount of neurons and hence more flexibility for upcoming tasks.

\subsection{Representation sparsity \& important parameter sparsity.} 

 \begin{figure}[t]
 \vspace*{-1cm} 
     \centering
    
      \subfloat[]{{\includegraphics[width=0.5\textwidth]{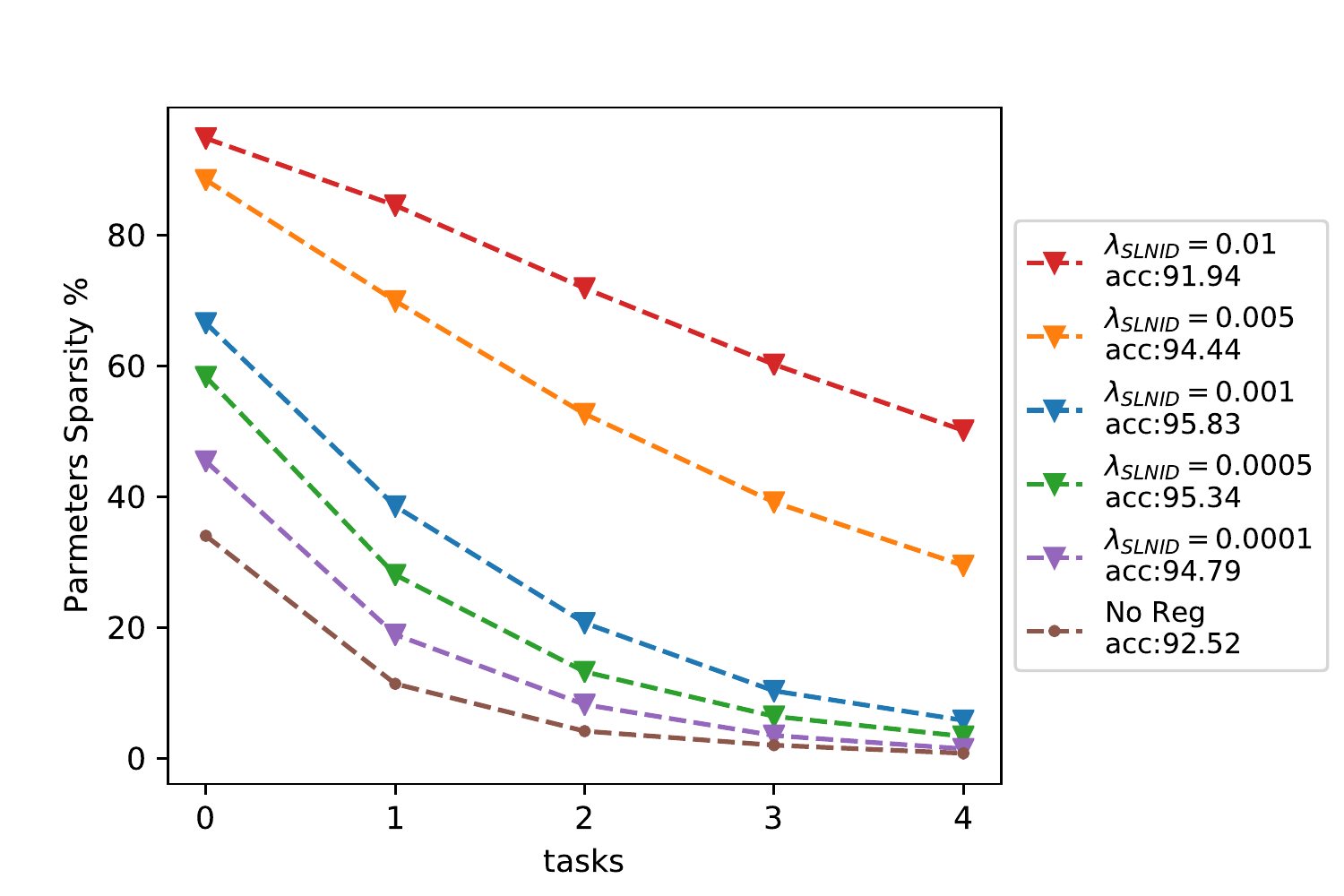} }\label{fig:scl-importance_sparsity} }%
       \hfillx
     \subfloat[]{{\includegraphics[width=0.49\textwidth]{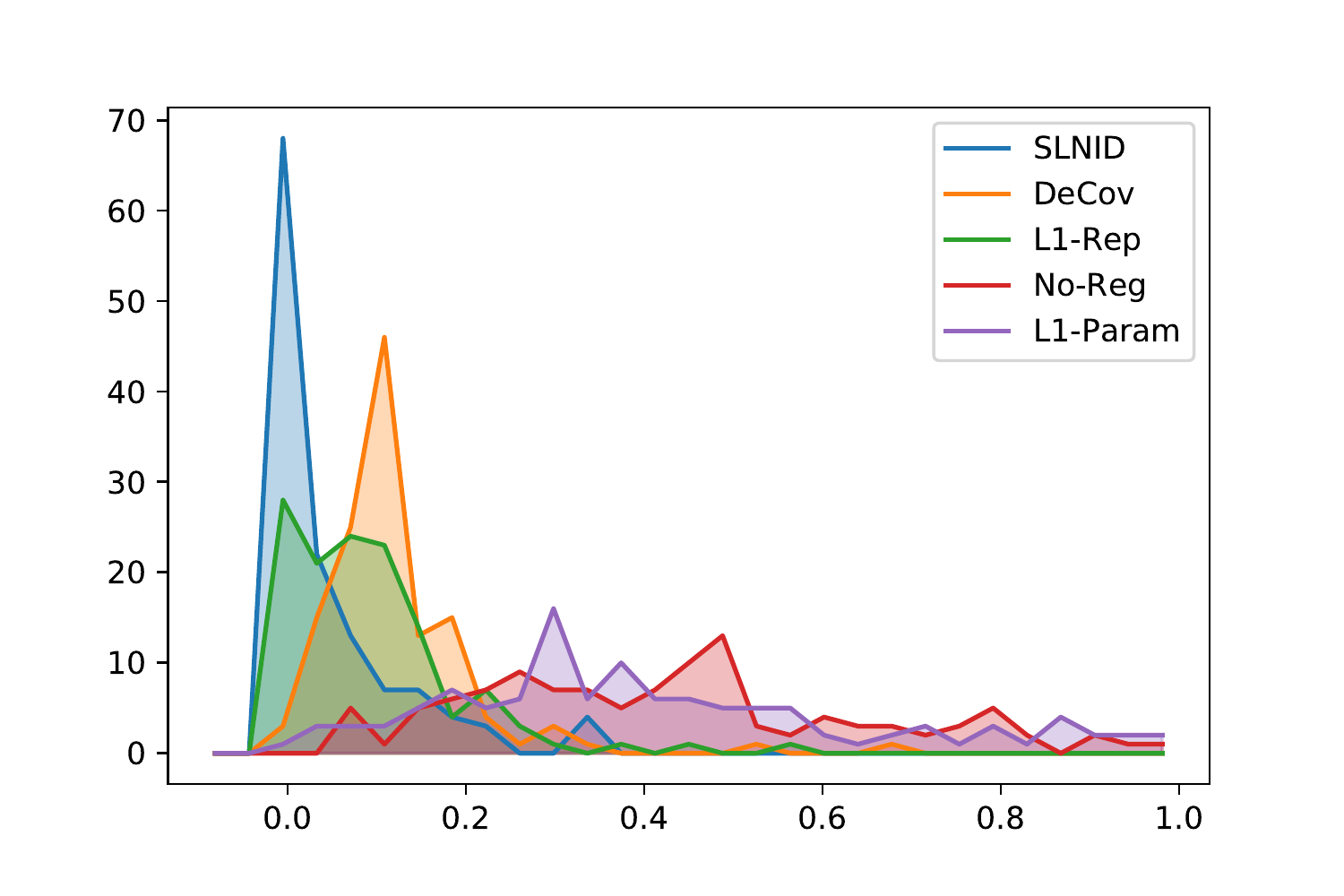} }\label{fig:scl-hist_sparsity}}%
    \caption{   On the 5 permuted MNIST sequence, hidden layer=128, (a): percentage of unused parameters in the 1st layer using different $\lambda_{\SLNI}$; 
    (b):  histogram of neural activations on the first task. }
   
\end{figure}
Here we want to examine the effect of our 
regularizer on the percentage of parameters that are utilized after each 
task and hence the
 capacity left for the later tasks. 
On the network with hidden layer size $128$, we compute the percentage of parameters with 
$\Omega_{ij} < 10^{-2}$, with $\Omega_{ij}$, 
the importance weights  estimated and accumulated over tasks.   We consider $\Omega_{ij} < 10^{-2}$ of negligible importance as in a network trained without a sparsity regularizer, $\Omega_{ij} < 10^{-2}$ covers the first 10 percentiles. 
Those parameters can be seen as unimportant and ``free'' for later tasks.  Figure~\ref{fig:scl-importance_sparsity}
shows the percentage of the unimportant (free) parameters in the first layer after each task for different $\lambda_{\SLNI}$ values along with the achieved average test accuracy at the end of the sequence.
It is clear that the larger $\lambda_{\SLNI}$, i.e. the more neural inhibition, the smaller the percentage of important parameters. Apart from the highest $\lambda_{\SLNI}$ where  tasks couldn't reach their top performance due to too strong inhibition,  improvement over the {\tt No-Reg} is always observed.

The optimal value for $\lambda_{\SLNI}$ seems to be the one that remains close to the optimal performance on the current task, while utilizing the minimum capacity feasible.
 Next, we compute the average activation per neuron, in the first layer, over all the examples and plot the corresponding histogram for \SLNI, {\tt DeCov}, {\tt L1-Rep}, {\tt L1-Param} and {\tt No-Reg} in Figure~\ref{fig:scl-hist_sparsity} at their setting that yielded the results shown in Figure~\ref{fig:scl-perm_comp}. \SLNI has a peak at zero indicating  representation sparsity while the  other methods values are  spread along the line. 
This   seems to hint at the effectiveness of our  approach~\SLNI in learning a sparse yet powerful representation and in turn maintaining  a minimal interference 
 between tasks.

  \begin{figure}[t!]%
    \centering
    \includegraphics[width=0.8\textwidth]{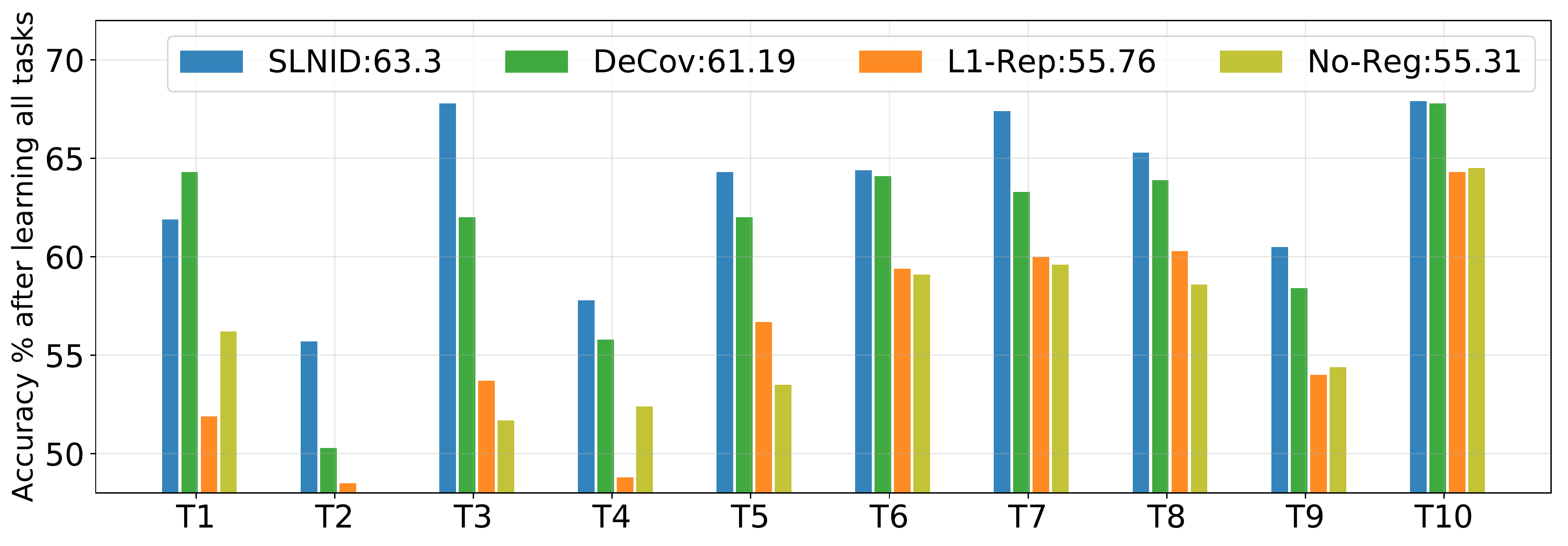}
  
  \caption[Comparison of different regularization techniques on a sequence of ten tasks from Cifar split.]{  Comparison of different regularization techniques on a sequence of ten tasks from Cifar split. The legend shows average test accuracy over all tasks. Simple L1-norm regularizer (L1-Rep) doesn't help in such more complex tasks. Our regularizer \SLNI achieves an improvement of $2\%$ over Decov and $8\% $ compared to No-Reg. }
    \label{fig:scl-cifar}%
\end{figure}  
 \begin{figure}[h!]%
    \centering
    \subfloat{{\includegraphics[width=0.8\textwidth]{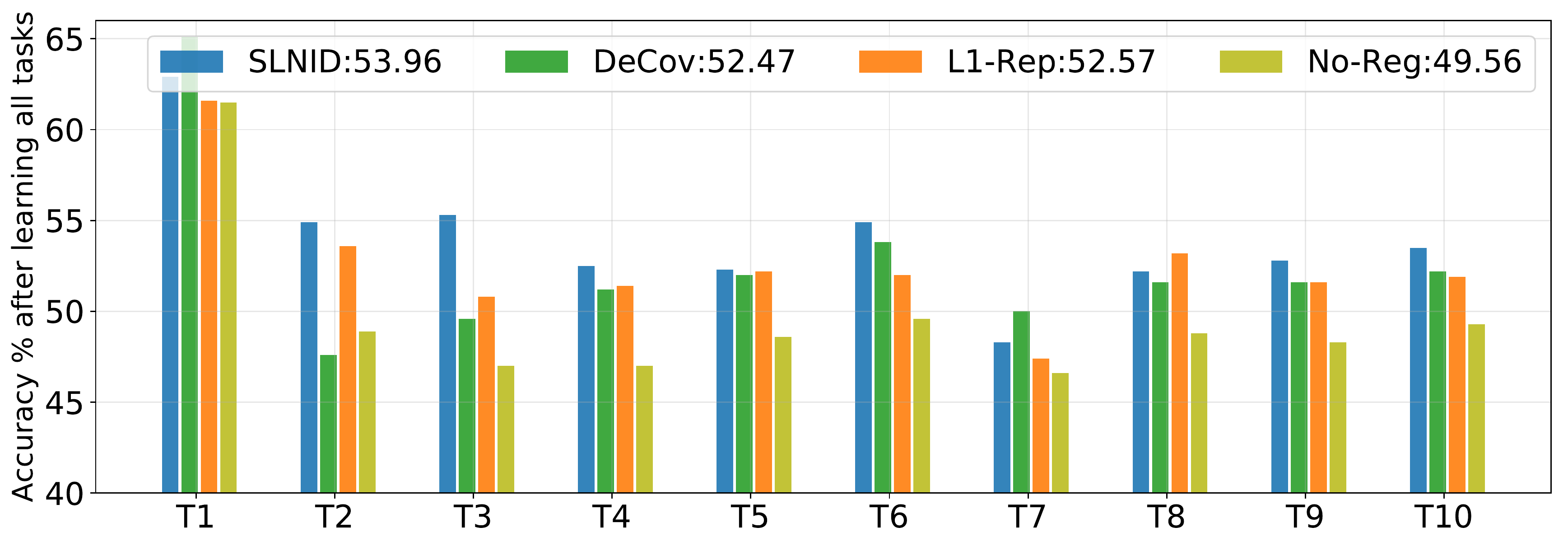} }}%
    \caption[Comparison of different regularization techniques on a sequence of ten tasks from   Tiny ImageNet split.]{  Comparison of different regularization techniques on a sequence of ten tasks from   Tiny ImageNet split. The legend shows average test accuracy over all tasks.  Our regularizer \SLNI achieves the best performance. }
    \label{fig:scl-imagenet}%
     \vspace*{-0.3cm}
\end{figure}

\subsection{10 Task Sequences on Cifar-100 and Tiny ImageNet}
\label{sec:scl-experiments:cifarimagenet}
 
While the previous section focused on learning a sequence of tasks with completely different input patterns and same objective, we now study the case of learning different categories of one dataset. \\
For this we split the  Cifar-100 and the Tiny ImageNet~\cite{yao2015tiny} dataset into ten tasks, 
respectively. We have 10 and 20 categories per task for Cifar-100 and Tiny ImagNet, respectively.
For Cifar-100, as a base network, we use 
a network similar to the one used by~\cite{Zenke2017improved} but without dropout. We evaluate two variants with hidden size $N=\{256,128\}$.

  \begin{wraptable}{r}{5cm}
 \vspace*{-0.8cm}
\centering
\scriptsize
 \begin{tabular}{|l|l|}
\hline
Layer& \# filters/neurons\\
\hline
Convolution&64\\
Max Pooling&-\\
Convolution&128\\
Max Pooling&-\\
Convolution& 256\\
Max Pooling&-\\
Convolution& 256\\
Max Pooling&-\\
Convolution& 512\\
Convolution& 512\\
Fully connected & 500\\
Fully connected & 500\\
Fully connected & 20\\
\hline
\end{tabular}
\vspace*{-0.3cm}
\caption{The network architecture  used in Tiny ImageNet experiment.}
\vspace*{-1cm}
\label{tab:archit_imagenet}
 \end{wraptable}

Throughout the experiment, we again used a scale $\sigma$ for the Gaussian function  equal to $1/6$ of the hidden layer size. 
We train the different tasks  for 50 epochs with a learning rate of $10^{-2}$ using SGD optimizer.
 For   Tiny ImageNet dataset~\cite{yao2015tiny}, as a base network, we use a variant of VGG~\cite{simonyan2014very}.
 For architecture details, please refer to Table~\ref{tab:archit_imagenet}.

We compare  the top competing methods from the previous experiments, {\tt L1-Rep}, {\tt DeCov}  and our \SLNI, and  {\tt No-Reg} as a baseline, {\tt ReLU} in previous experiment. Similarly, MAS \cite{aljundi2018memory} is used in all cases as continual learning method.
Figures~\ref{fig:scl-cifar} and \ref{fig:scl-imagenet} show the  performance on each of the ten tasks at the end of the sequence.
For both datasets, we observe that our \SLNI  performs overall best. {\tt L1-Rep} and {\tt DeCov} continue to improve over the non regularized case {\tt No-Reg}. These results confirm our proposal on the importance of sparsity and decorrelation in continual learning.

\subsection{\SLNI with EWC \cite{kirkpatrick2017overcoming} } 
\label{sec:scl-experiments:EWC}
\begin{figure}[h!]%
    \centering
\includegraphics[width=0.7\textwidth]{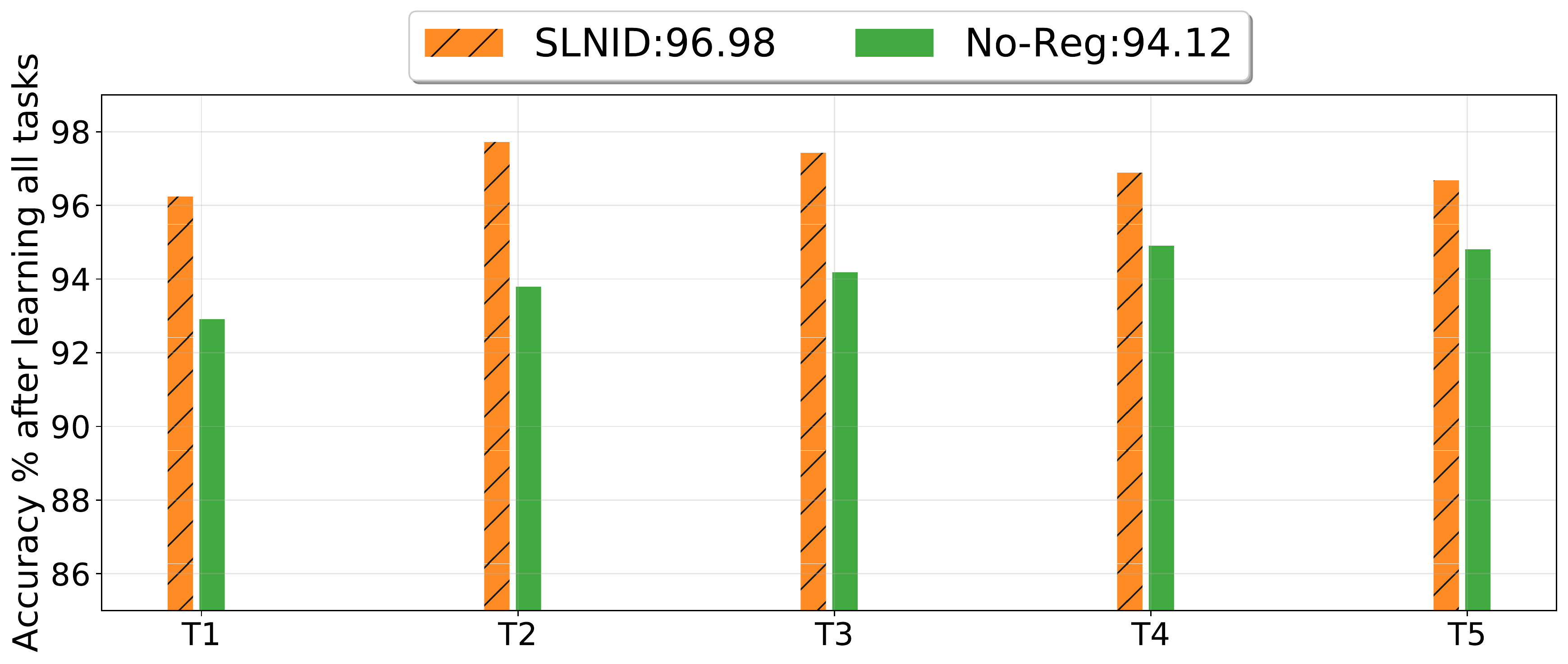} 
    \caption[Comparison of {\tt SLNID}, with EWC~\cite{kirkpatrick2017overcoming}, and ~{\tt No-Reg}, EWC alone  with no sparsity regularizer, hidden size 128. ]{  Comparison of {\tt SLNID}, with EWC~\cite{kirkpatrick2017overcoming}, and ~{\tt No-Reg}, EWC alone  with no sparsity regularizer. Figure shows at the end of  permuted MNIST sequence, each task accuracy on a network with 128 neurons in the hidden layer.}
    \label{fig:scl-ewc-128}%
     \end{figure}

\begin{figure}[h!]%
    \centering
\includegraphics[width=0.7\textwidth]{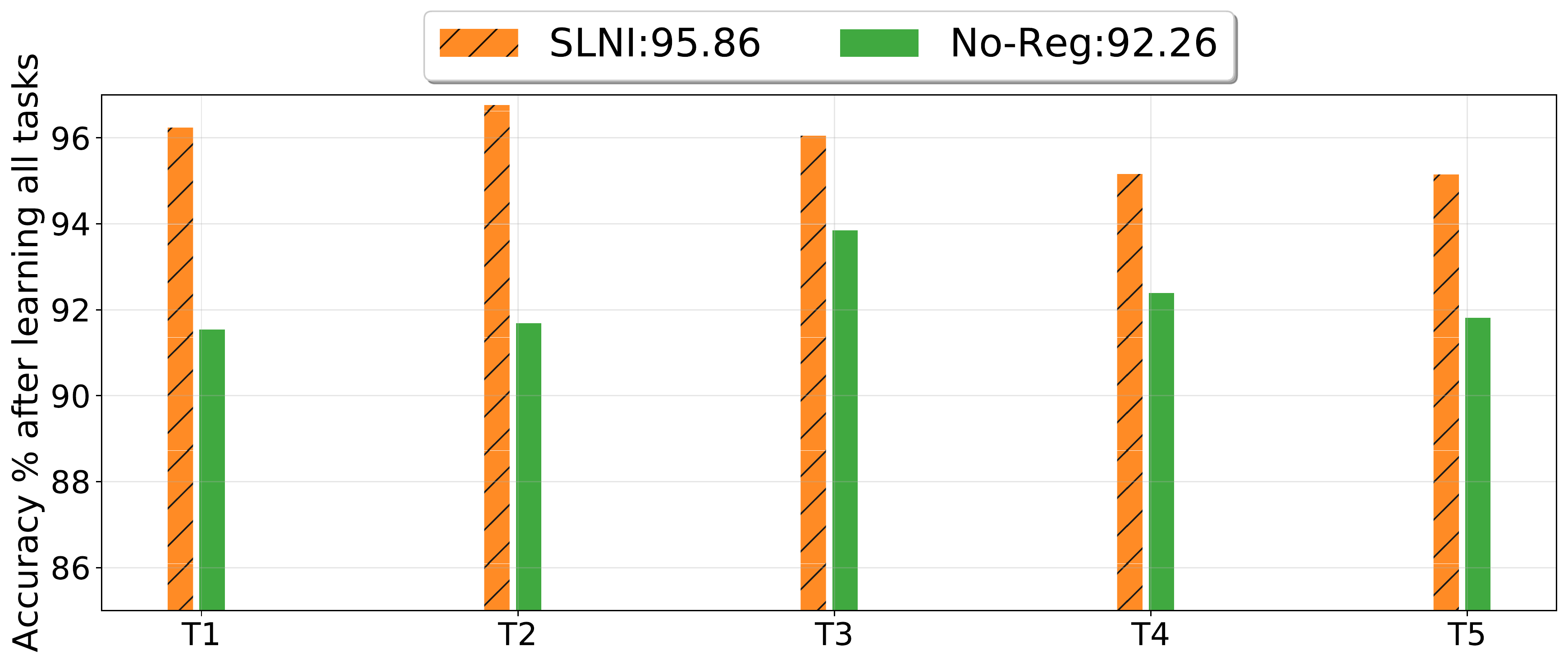} 
  \vspace*{-0.3cm} 
    \caption[Comparison of {\tt SLNID}, with EWC~\cite{kirkpatrick2017overcoming}, and ~{\tt No-Reg},  EWC alone  with no sparsity regularizer, hidden size 64.]{  Comparison of {\tt SLNID}, with EWC~\cite{kirkpatrick2017overcoming}, and ~{\tt No-Reg},  EWC alone  with no sparsity regularizer. Figure shows at the end of  permuted MNIST sequence, each task accuracy on a network with 64 neurons in the hidden layer.  }
    \label{fig:scl-ewc-64}%
     \end{figure}

We have shown that our proposed regularizer~\SLNI exhibits stable and superior performance on the different tested networks when using {\tt MAS} as continual learning method. 
To prove the effectiveness of our regularizer regardless of the used regularization  based method, we have tested~\SLNI on the 5 tasks permuted MNIST sequence in combination with Elastic Weight Consolidation ({\tt EWC}~\cite{kirkpatrick2017overcoming}). Figures ~\ref{fig:scl-ewc-128} and ~\ref{fig:scl-ewc-64} compare the test accuracy of each task  achieved by our \SLNI with  EWC and {\tt No-Reg } (here indicating {\tt EWC} without regularization) on a network of size 128 and 64 respectively. We obtained a boost in the average performance at the end of the learned sequence equal to $2.8\%$ on the network with hidden layer size 128 and a boost of  $3.6\%$ with hidden layer size 64. 
It is worth noting that with both~{\tt MAS}  and~{\tt EWC}, \SLNI was able to obtain better accuracy using a network with a $64$-dimensional hidden size than when training without regularization {\tt No-Reg} on a network of double that size ($128$), indicating that~\SLNI allows to use neurons much more efficiently. 
 
\subsection{Ablation Study}
\label{sec:scl-experiments:ablation}
Our method can be seen as composed of three components: the neural inhibition, the locality relaxation and the neuron importance integration.  To study how these components perform individually, we consider the following variants:
\begin{itemize}
    \item {\tt SNI} : our regularizer without spatial locality and neuron importance.
    \item \SNI: our regularizer without spatial locality.
    \item {\tt SLNI}: our regularizer without neuron importance.
    \item \SLNI: our full  regularizer.
\end{itemize}
Table \ref{tbl:ablation} reports the average accuracy at the end of the Cifar-100 and permuted MNIST sequences for each variant.  As we explained in Section~\ref{sec:scl-method:ssl}, when tasks have completely different input patterns, the neurons that were activated on the previous task examples will not fire for new task samples and exclusion of important neurons is not mandatory. However, when sharing is present between the different tasks, a term to prevent our regularizer from causing any interference is required. This is manifested in the reported results: for permuted MNIST, all the variants work nicely alone, as a result of the simplicity and the disjoint nature of this sequence. However, in the Cifar-100 sequence, the integration of the neuron importance in the \SNI and \SLNI regularizers exclude important neurons from the inhibition, resulting in a clearly better performance. The locality in \SLNI  improves the performance in the Cifar sequence, which suggests that a richer representation is needed and  multiple active neurons should be tolerated.
\clearpage
\begin{table}[h!]
 \vspace*{-0.7cm} 
\centering
\begin{tabular}{|l|l|l|l|l|}
\hline
 & \multicolumn{2}{|c|}{Permuted  MNIST} & \multicolumn{2}{|c|}{Cifar}  \\ \hline
 h-layer dim. &   128 &   64 &   256 &  128  \\ \hline
  {\tt Joint Training*}&  97.30  & 96.80  &70.99  & 71.95\\ \hline
 {\tt No-Reg} &92.67&90.72&55.06&55.3 \\ \hline
 {\tt SNI}& 95.79 & {\bf 94.89} &  55.30 & 55.75 \\ 
 \SNI&  95.90&  93.82& 61.00& 60.90 \\ 
{\tt SLNI} & {\bf 95.95}  & 94.87  & 56.06  & 55.79 \\ 
 \SLNI&   95.83   & 93.89  &{\bf 63.30} & {\bf 61.16}\\ \hline

\end{tabular}
\caption[\SLNI ablation.]{\label{tbl:ablation}   \SLNI ablation. Average test accuracy per task after training the last task in \%. * denotes that Multi-Task Joint Training violates the continual learning scenario as it has access to all tasks at once and thus can be seen as an upper bound.  }
 \vspace*{-0.5cm} 
\end{table}

 %
\subsection{Continual Learning without Hard Task Boundaries}
\label{sec:scl-experiments:noboundaries}

\begin{wraptable}{r}{5cm}
\vspace*{-0.2cm}
\begin{scriptsize}
\begin{tabular}{|l|l|}
\hline
 Method& Avg.acc -tasks models\\
\hline
{\tt No-Reg} \texttt{w/o MAS}&69.20\%\\
\SLNIwoNI \texttt{w/o MAS} &72.14\% \\
\SLNI \texttt{w/o MAS} &73.03\% \\
{\tt No-Reg} &66.88\%\\
\SLNIwoNI &71.32\% \\
\SLNI  &72.33\% \\
\hline
\hline
 Method& Avg.acc-last model \\
 \hline
{\tt No-Reg} \texttt{w/o MAS}&65.15\%\\
\SLNIwoNI \texttt{w/o MAS} &63.54\% \\
\SLNI \texttt{w/o MAS} &70.75\% \\
 \hline
{\tt No-Reg} &66.33\%\\
\SLNIwoNI &64.50\% \\
\SLNI  &70.94\% \\
\hline
\end{tabular}
\caption[ No tasks boundaries test case on Cifar-100.]{\label{tab:no_boundaries}  No tasks boundaries test case on Cifar-100. Top block, avg. acc  on each group of classes using each group model. Bottom block, avg. acc. on each group at the end of the training.}
\end{scriptsize}
\vspace*{-0.2cm}
\end{wraptable}
In the previous experiments, we  considered the standard task incremental setting as in~\cite{li2016learning,Zenke2017improved,aljundi2018memory,Serr2018Hard}, where at each time step we receive a task along with its training data and a new classification layer is initiated for the new task, if needed.
Here, we are  interested in a more realistic scenario where the data distribution shifts gradually without  hard task boundaries.
To test this setting, we use the Cifar-100 dataset. Instead of considering a set of 10 disjoint tasks each composed of 10 classes, as in the previous experiment (Section~\ref{sec:scl-experiments:cifarimagenet}), we now start by sampling with high probability $(2/3)$ from the first 10 classes and with low probability $(1/3)$ from the rest of the classes. 
We train the  network (same architecture as
in Section~\ref{sec:scl-experiments:cifarimagenet}) for a few epochs and then change the sampling probabilities to be high $(2/3)$ for classes $11-20$  and low $(1/3)$ for the remaining classes. This process is repeated until sampling with high probability  from the last 10 classes and low from the rest.  We use one shared classification layer throughout and estimate the importance weights and the neurons importance after each training step (before changing the sampling probabilities). 
We consider 6 variants: our \SLNI, the ablations \SLNIwoNI and without regularizer {\tt No-Reg}, as in Section~\ref{sec:scl-experiments:ablation}, as well each of these three trained without the MAS importance weight regularizer, denoted as \texttt{w/o MAS}.
Table~\ref{tab:no_boundaries} presents the accuracy {averaged over} the ten groups of ten classes,  using each group model (i.e.~the model trained when this group was sampled with  high probability) in the top block and the average accuracy on each of the ten groups at the end of the training (middle and bottom block).
We can deduce the following:  1) \SLNI improves the performance considerably (by more than 4\%) even without importance weight regularizer. 
2) In this scenario without hard task boundaries there is less forgetting than in the scenario with hard task boundaries
studied in Section~\ref{sec:scl-experiments:cifarimagenet} for Cifar (difference between rows in top block to corresponding rows in middle block). {As a result,} the improvement obtained by deploying the importance weight regularizer is moderate: 
at $70.75\%$, \SLNI \texttt{w/o MAS} is already better than {\tt No-Reg} reaching $66.33\%$. 3) While  \texttt{SLNI w/o MAS} improves the individual models performance ($72.14\%$ 
compared to $69.20\%$), it fails to improve the overall performance at the end of the sequence ($63.54\%$ compared to $65.15\%$), as important neurons are not excluded from the penalty and hence they are changed or inhibited leading to tasks interference and performance deterioration.  
\subsection{Comparison with the State of the Art}
\label{sec:scl-experiments:object}
To compare our proposed approach with the different state-of-the-art continual learning methods, we use a sequence of 8 different object recognition tasks, introduced  in Chapter~\ref{ch:MAS}. The sequence starts from AlexNet~\cite{krizhevsky2012imagenet} pretrained on ImageNet~\cite{ILSVRC15} as a base network, following the setting of Chapter~\ref{ch:MAS}. 
We compare against the following:
\noindent
{\em Learning without Forgetting}~\cite{li2016learning} ({\tt LwF}), {\em Encoder Based Lifelong Learning}~({\tt EBLL}) (see Chapter~\ref{ch:ebll}),  {\em Incremental Moment Matching}~\cite{lee2017overcoming} ({\tt IMM}), {\em Synaptic Intelligence}~\cite{Zenke2017improved} ({\tt SI}) and sequential finetuning ({\tt FineTune}),
in addition to the case of {\tt MAS} \cite{aljundi2018memory} alone,~i.e. our {\tt No-Reg}  before. Compared methods were run with the exact same setup as in Chapter~\ref{ch:MAS}. For our regularizer, we disable  dropout, since dropout encourages redundant activations which contradicts  our regularizer's role. Also, since the network is pretrained, the locality introduced in \SLNI may conflict with the already pretrained activations. For this reason, we also test \SLNI with randomly initialized fully connected layers. Our regularizer is applied with {\tt MAS} as a continual learning method.
Table~\ref{tbl:8tasks_obj} reports the average test accuracy at the end of the sequence achieved by each method. \SLNI improves even when starting from a pretrained network and disabling dropout. Surprisingly, even with randomly initialized fully connected layers, \SLNI improves $1.8\%$ over the state of the art using a fully pretrained network.  

\begin{table}[h!]
 \vspace*{-8pt}
\centering
\small
 \begin{tabular}{|l|l|}
\hline
Method&Average accuracy\\
\hline
{\tt Finetune}&32.67\%\\
{\tt LWF}~\cite{li2016learning}&49.49\%\\
 {\tt EBLL}& 50.29\%\\
{\tt IMM}~\cite{lee2017overcoming}&43.4\%\\
{\tt SI}~\cite{Zenke2017improved}&50.49\%\\
{\tt EWC}~\cite{kirkpatrick2017overcoming}&50.00\%\\
{\tt MAS}& 52.69\%\\
\SLNI -fc pretrained  &53.77\%\\
\SLNI -fc randomly initialized  & {\bf 54.50\%}\\
\hline
\end{tabular}
    \caption[8 tasks object recognition sequence.]{ \label{tbl:8tasks_obj} 8 tasks object recognition sequence. Average test accuracy per task after training the last task.}
    \vspace{-20pt}
\end{table}
\subsection{Spatial Locality Test}
\begin{figure}[h!]
\vspace*{-1.2cm}
    \centering
    \subfloat{{\includegraphics[width=0.495\textwidth]{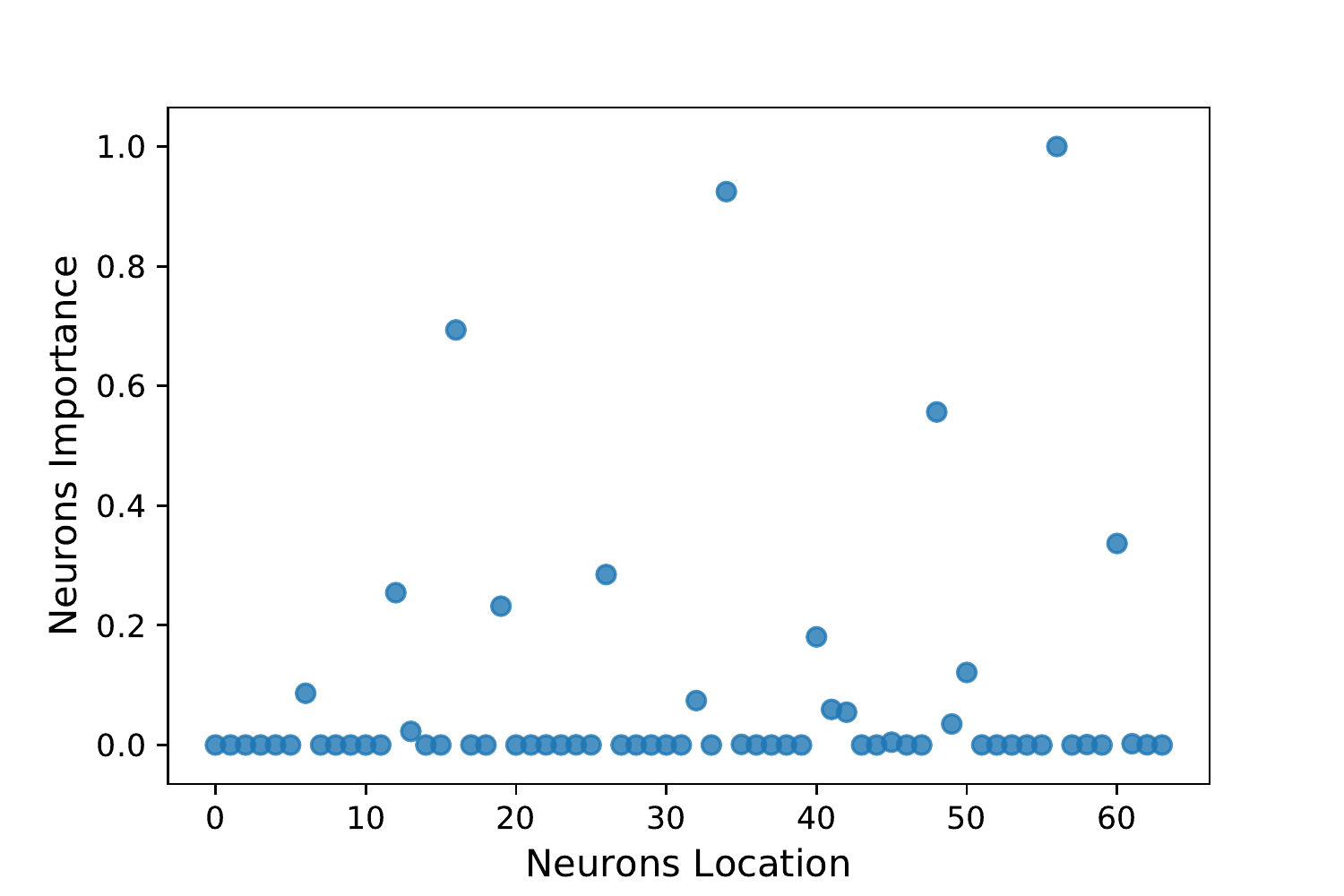} }}%
    \hfillx
    \subfloat{{\includegraphics[width=0.495\textwidth]{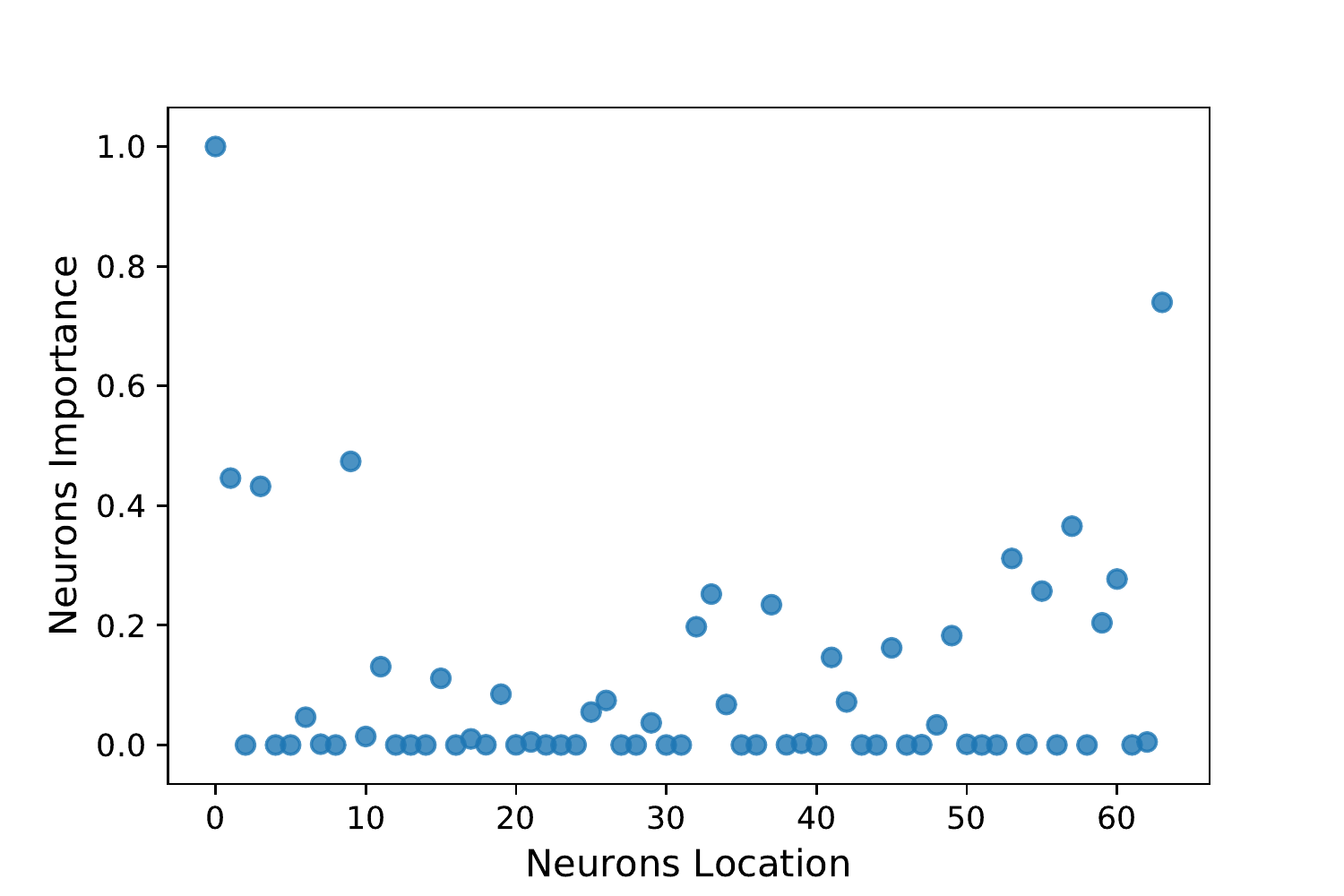} }}%
  
    \caption[First layer neuron importance after  learning the first task.]{\small First layer neuron importance after  learning the first task (blue). Left:  \SNI, Right:  \SLNI.  More active neurons are tolerated in  \SLNI.}
    \label{fig:scl-scale1}%
\end{figure}

\begin{figure}[h!]
\vspace*{-1.cm}
    \centering
    \subfloat{{\includegraphics[width=0.495\textwidth]{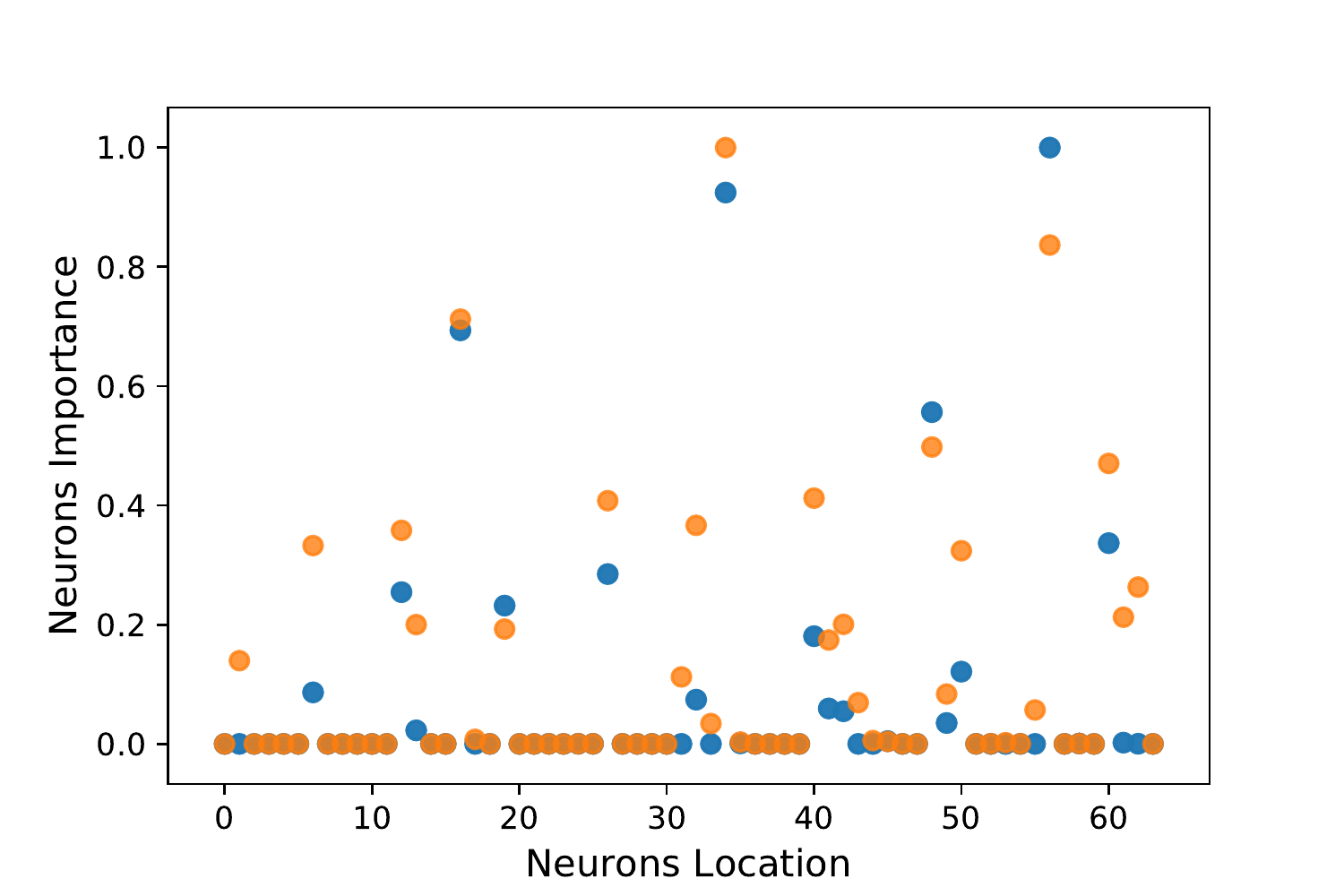} }}%
    \hfillx
    \subfloat{{\includegraphics[width=0.495\textwidth]{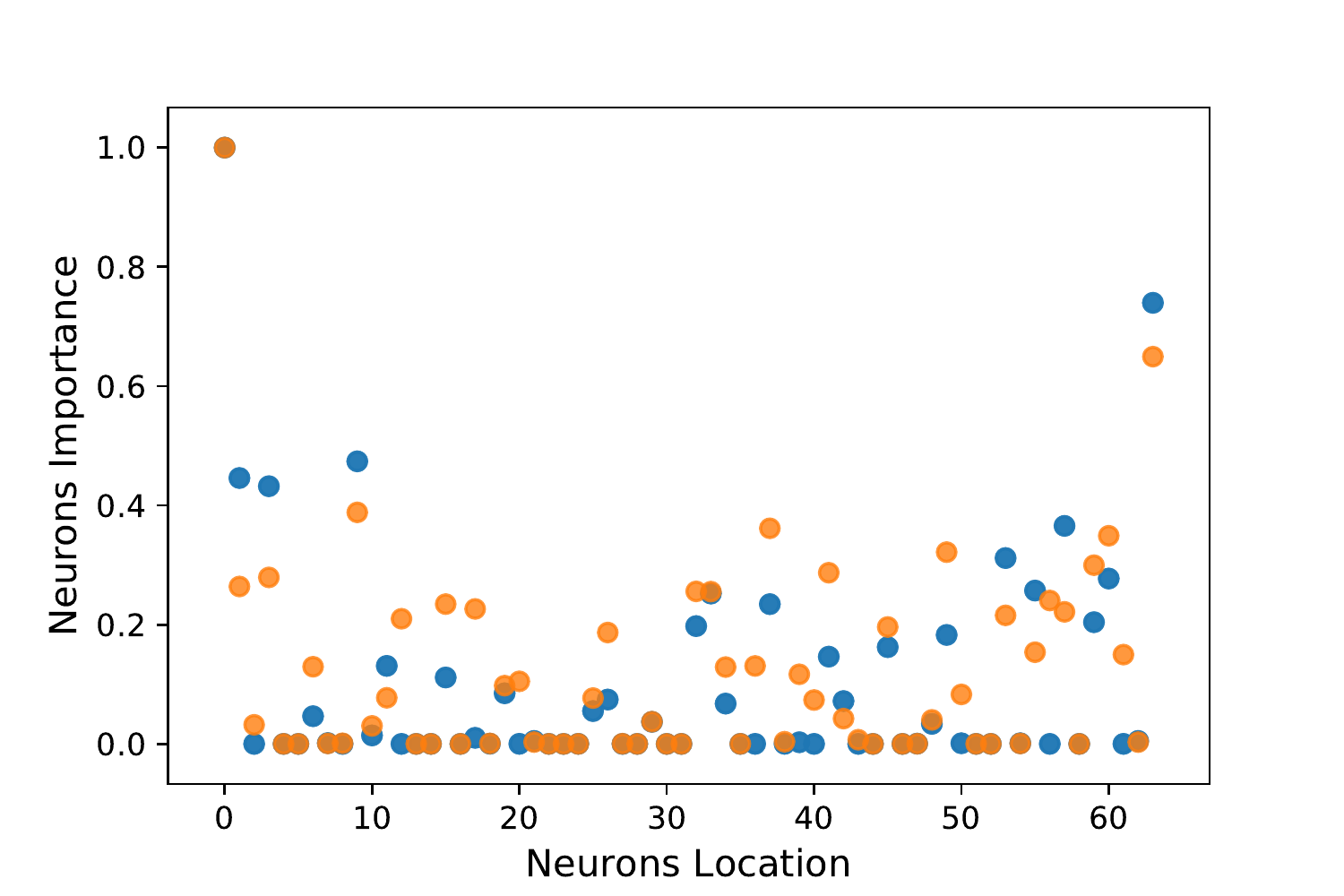} }}%
  
    \caption[First layer neuron importance  after learning the second task.]{\small First layer neuron importance  after learning the second task (orange), superimposed on Figure~\ref{fig:scl-scale1}. Left: \SNI, Right: \SLNI. \SLNI allows new neurons, especially those that were close neighbours to previous important neurons, to become active and to be used for the new task. \SNI penalizes all unimportant neurons equally. As a result, previous neurons are adapted for the new tasks and less new neurons are getting activated. }
    \label{fig:scl-scale2}%
\end{figure}

\begin{figure}[h!]
\vspace*{-1.cm}
    \centering
    \subfloat{{\includegraphics[width=0.495\textwidth]{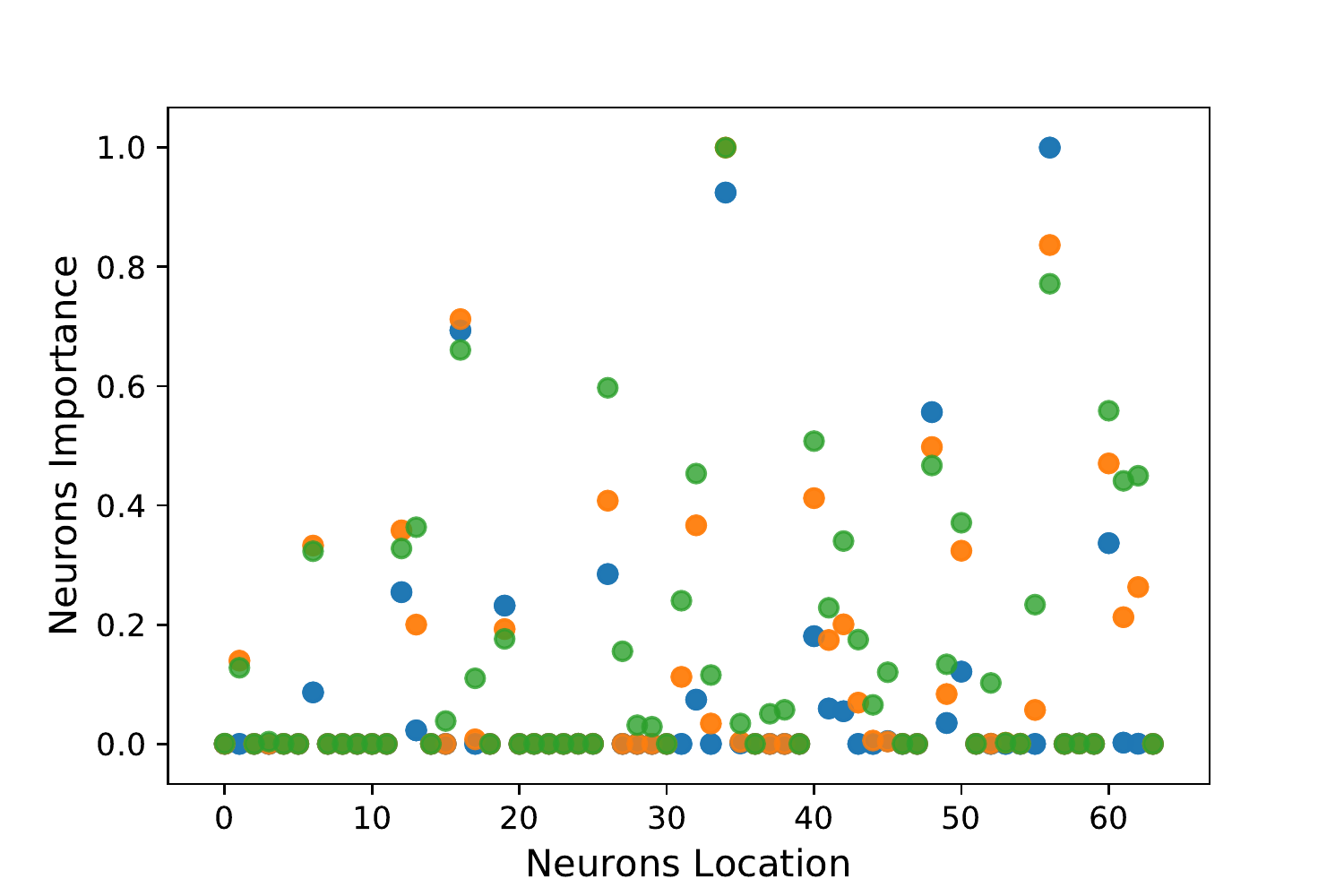} }}%
    \hfillx
    \subfloat{{\includegraphics[width=0.495\textwidth]{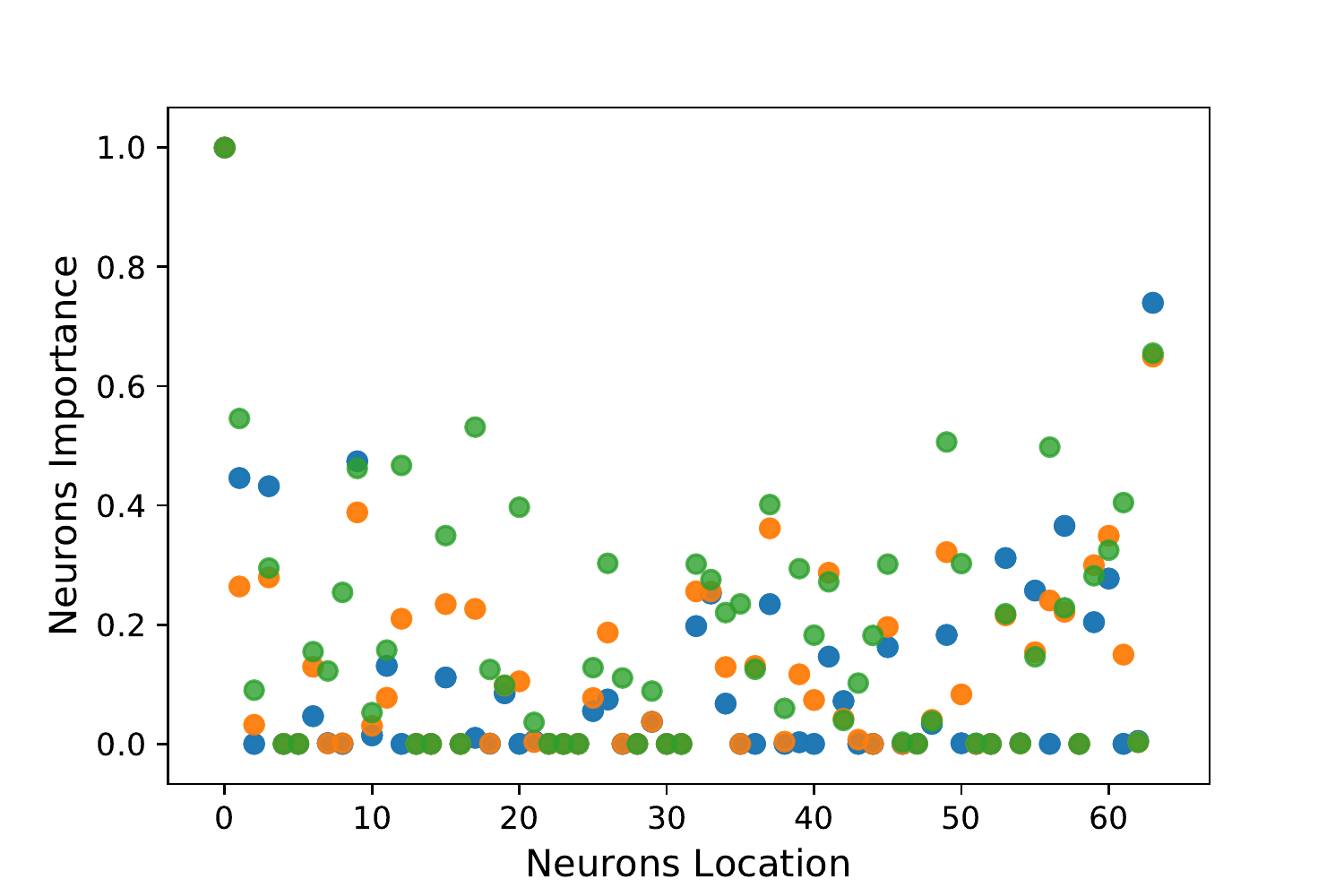} }}%
  
    \caption[First layer neuron importance after  learning the third task.]{\small First layer neuron importance after  learning the third task (green), superimposed on Figure~\ref{fig:scl-scale2}. Left: \SNI, Right:  \SLNI.  \SLNI allows previous neurons to be re-used for the third task. It avoids changing the previous important neurons by adding new neurons. For \SNI, very few neurons are newly deployed. The new task is learned mostly by adapting previous important neurons. }
    \label{fig:scl-scale3}%
\end{figure}

\begin{figure}[h!]
\vspace*{-1.2cm}
    \centering
    \subfloat{{\includegraphics[width=0.495\textwidth]{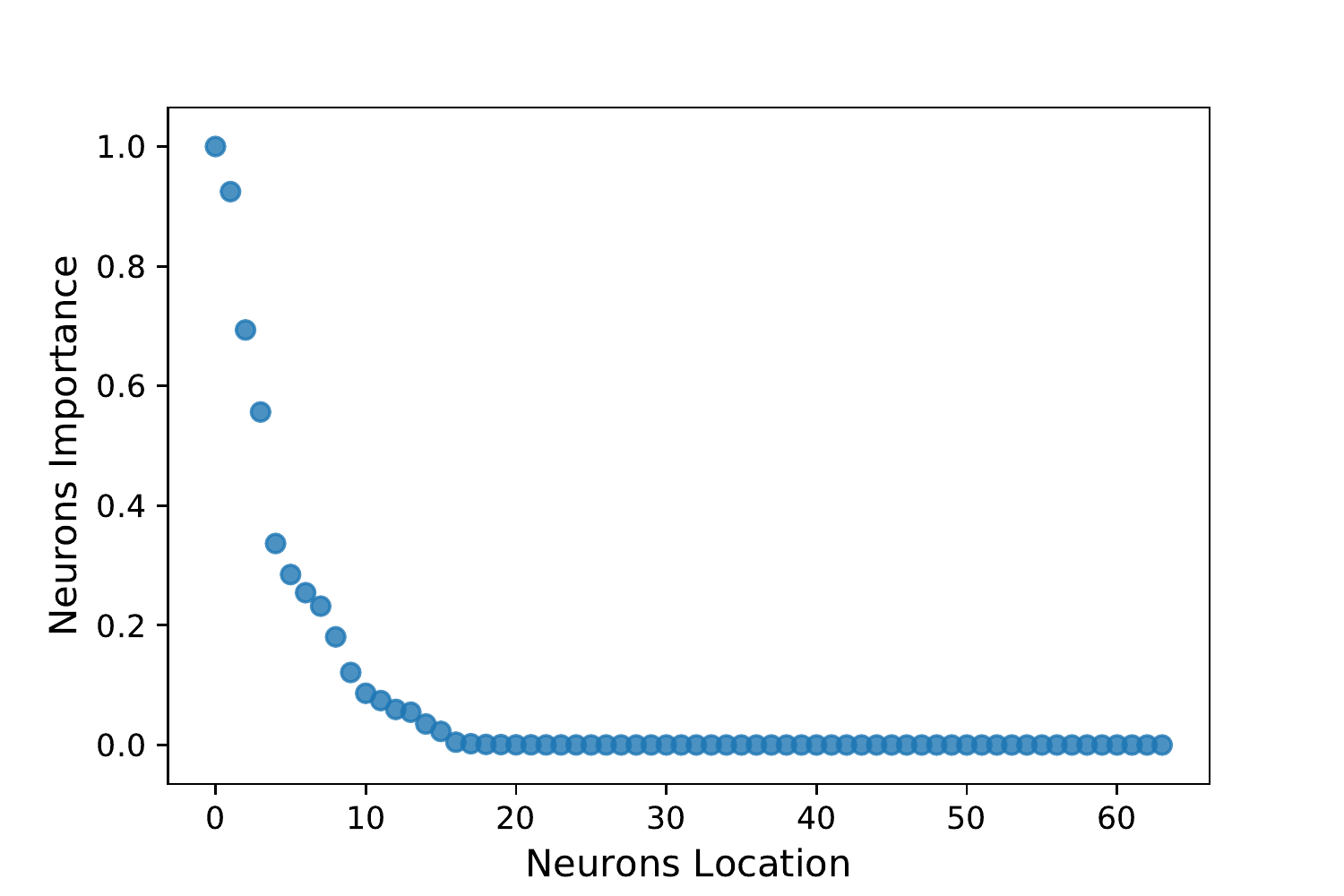} }}%
    \hfillx
    \subfloat{{\includegraphics[width=0.495\textwidth]{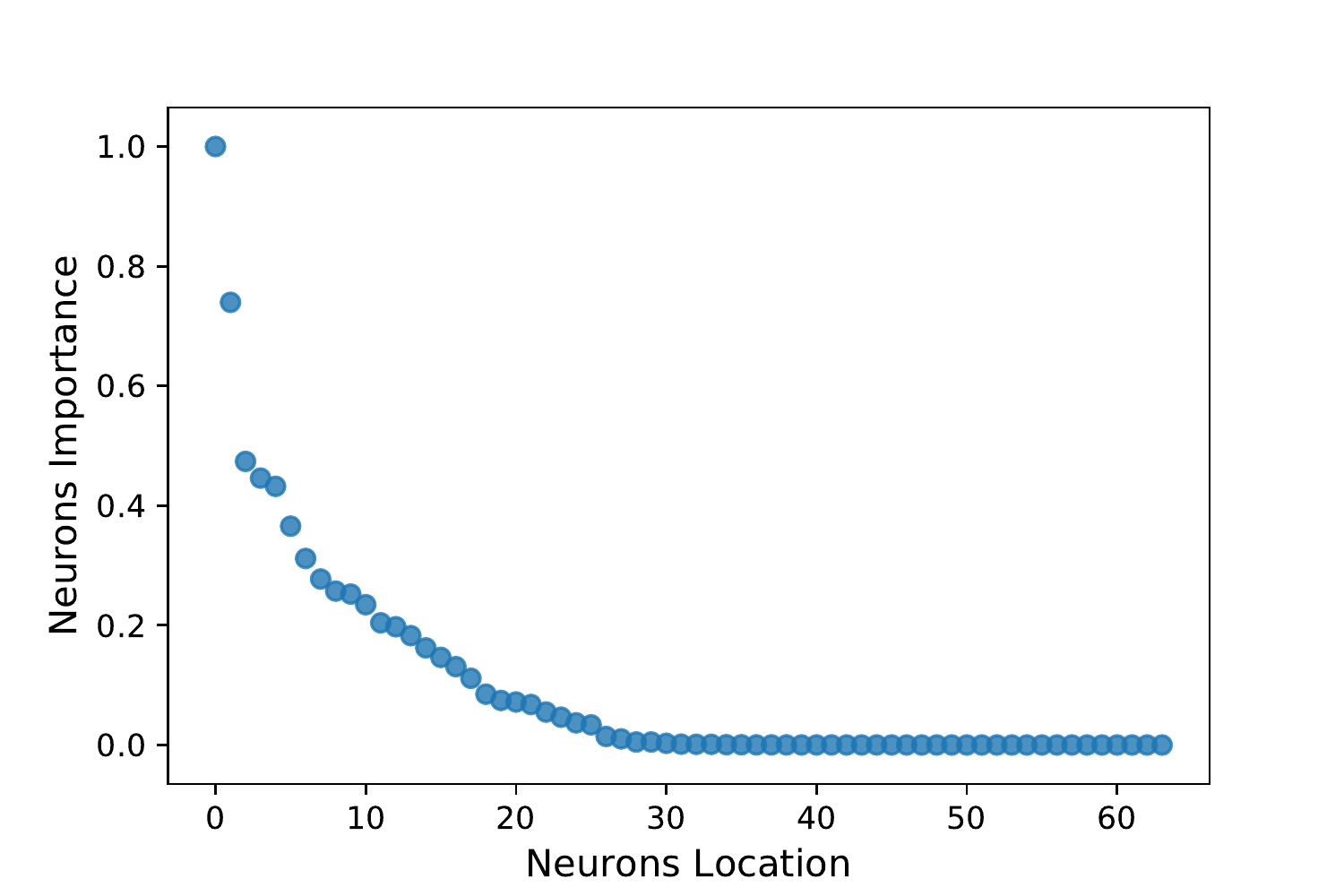} }}%
    \caption[ First layer neuron importance after  learning the first task, sorted in descending order according to the first task.]{\small First layer neuron importance after  learning the first task, sorted in descending order according to the first task neuron importance (blue). Left: \SNI, Right:  \SLNI.  More active neurons are tolerated in  \SLNI.}
    \label{fig:scl-scale1b}%
\end{figure}

\begin{figure}[h!]
\vspace*{-0.76cm}
    \centering
    \subfloat{{\includegraphics[width=0.495\textwidth]{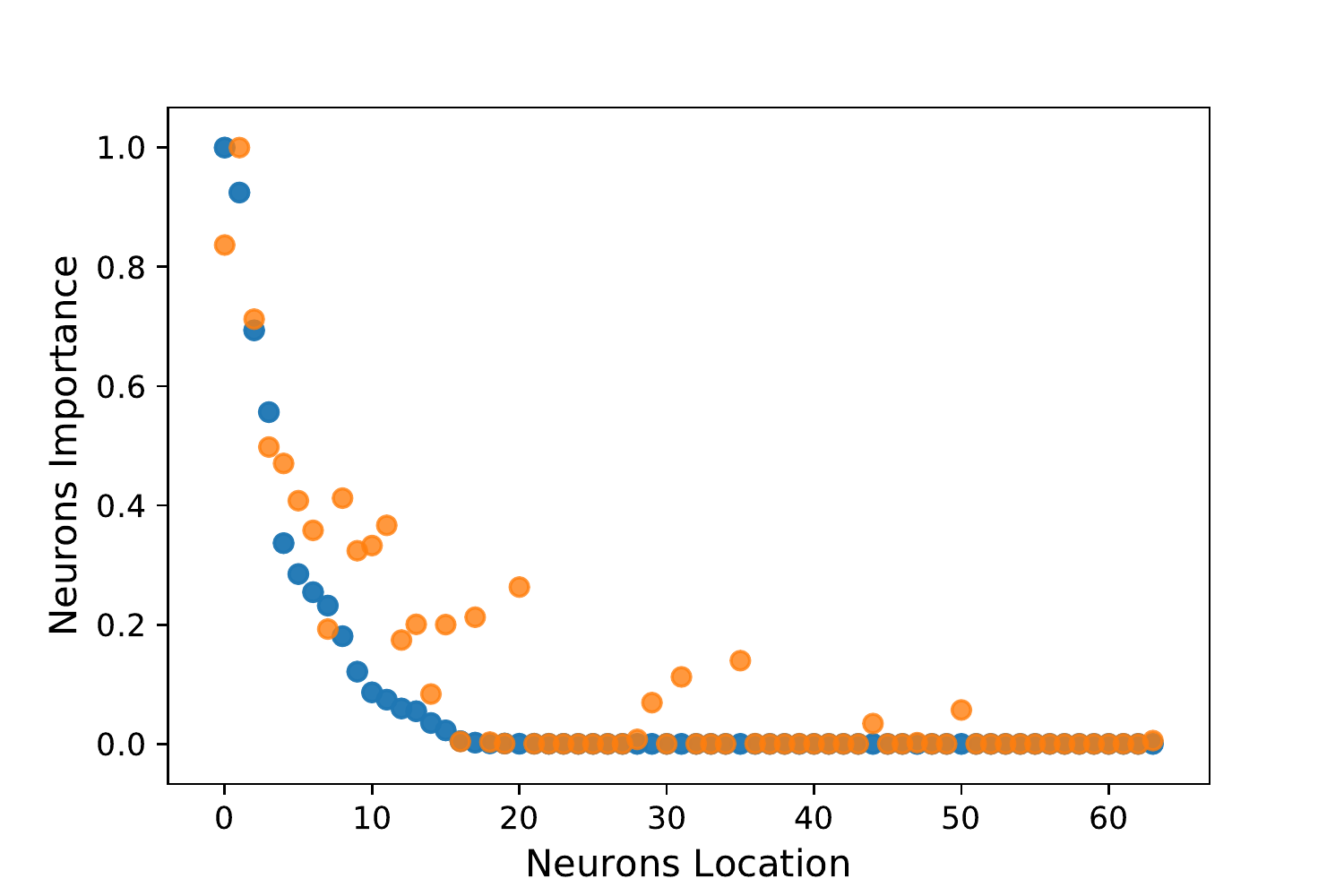} }}%
    \hfillx
    \subfloat{{\includegraphics[width=0.495\textwidth]{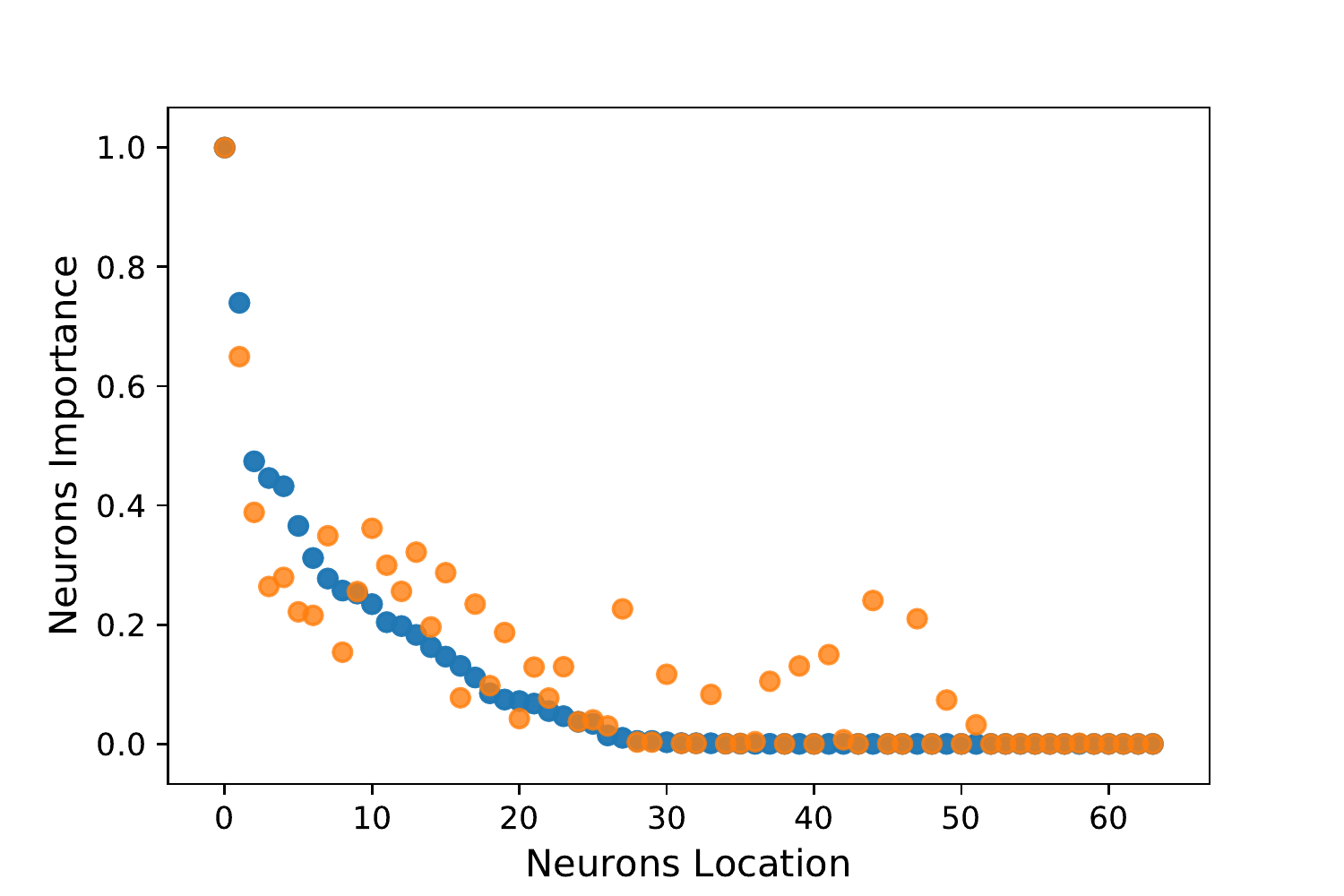} }}%
    \caption[ First layer neuron importance after  learning the second task, sorted in descending order according to the first task.]{\small First layer neuron importance  after learning the second task sorted in descending order according to the first task neuron importance (orange), superimposed on top of Figure~\ref{fig:scl-scale1b}. Left: \SNI, Right:  \SLNI.  \SLNI allows new neurons to become active and be used for the new task. \SNI penalizes all unimportant  neurons equally and hence more neurons are re-used then initiated for the first time. }
    \label{fig:scl-scale2b}%
\end{figure}

\begin{figure*}[h!]
\vspace*{-0.76cm}
    \centering
    \subfloat{{\includegraphics[width=0.495\textwidth]{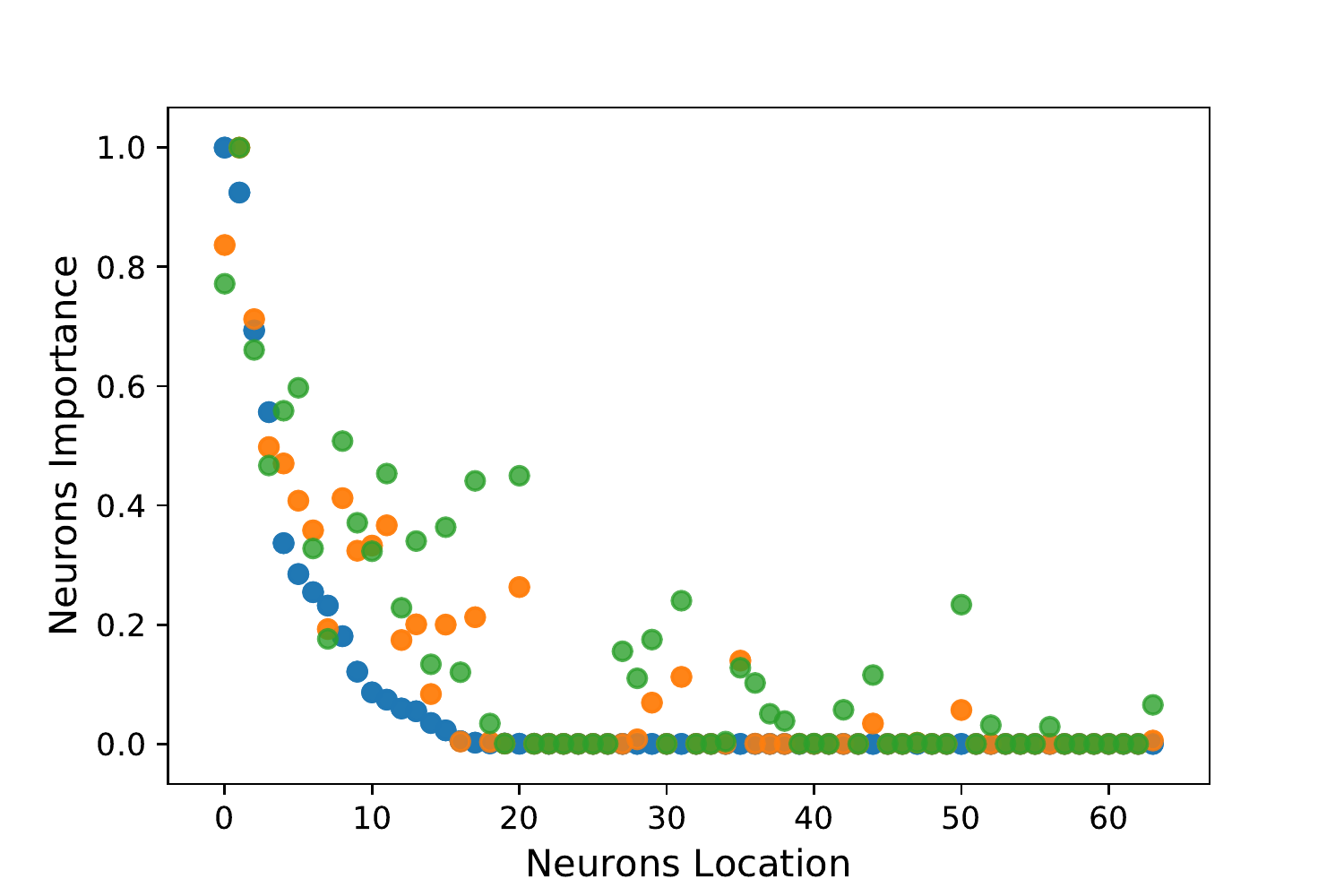} }}%
    \hfillx
    \subfloat{{\includegraphics[width=0.495\textwidth]{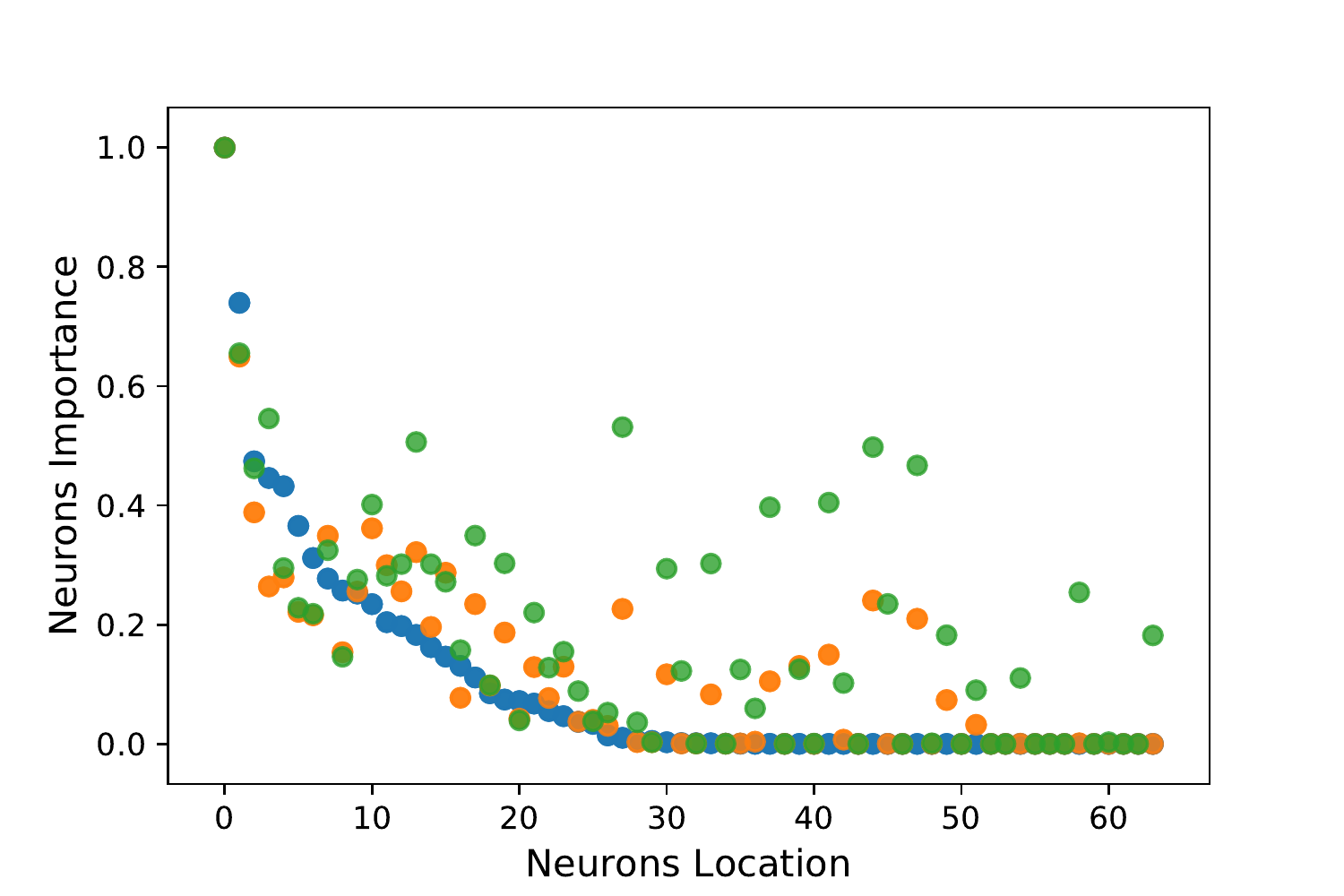} }}%
    \caption[ First layer neuron importance after  learning the third task, sorted in descending order according to the first task]{ \small First layer neuron importance after  learning the third task sorted in descending order according to the first task neuron importance (green), superimposed on top of Figure~\ref{fig:scl-scale2b}.  Left: \SNI, Right:  \SLNI.  \SLNI allows previous neurons to be re-used for the third task while activating new neurons to cope with the needs of the new task. For \SNI, very few neurons are newly deployed while most previous important neurons for previous tasks are re-adapted to learn the new task. }
    \label{fig:scl-scale3b}%
\end{figure*}

To avoid penalizing all the active neurons,  our \SLNI weights the correlation penalty between each two neurons based on their spatial distance using  a Gaussian function. We want to visualize the effect of this spatial locality on the neurons activity. To achieve this, we have used the first 3 tasks of the permuted MNIST sequence as a test case and visualized the neurons importance after each task. This is done using the network of hidden layer size $64$. 
Figure~\ref{fig:scl-scale1}, Figure~\ref{fig:scl-scale2} and Figure~\ref{fig:scl-scale3} show the neurons importance after each task. Figure~\ref{fig:scl-scale1b}, Figure~\ref{fig:scl-scale2b} and Figure~\ref{fig:scl-scale3b} show the neurons importance after each task sorted in descending order according to the first task neuron importance.  The left column is without locality, i.e. \SNI, and the right column is \SLNI. Blue represents the first task, orange the second task and green the third task.
When using \SLNI, inhibition is applied in a local manner allowing more active neurons which could potentially improve the representation power. When learning the second task, new neurons become important regardless of their closeness to first task important neurons as those neurons are excluded from the inhibition. As such, new neurons are becoming active as new tasks are learned. For \SNI on the other hand, all neural correlation is penalized in the first task and for later tasks.  Very few neurons are able to become active and important  due to the strong global inhibition, where previous neurons that are excluded from the inhibition are easier to be re-used.
\section{Summary}\label{sec:scl-conclusion}
In this chapter, we have studied the problem of continual learning using a network with fixed capacity 
-- a prerequisite for a scalable and computationally efficient solution.
A key insight of our approach is that in the context of continual learning (as opposed to other contexts where sparsity is imposed, such as network compression or avoiding overfitting), sparsity should be imposed at the level of the representation rather than at the level of the network parameters. 
Inspired by lateral  inhibition in the mammalian brain, we impose sparsity by means of a new regularizer that decorrelates nearby active neurons. 
We integrate this in a model which \emph{learns selflessly} a new task by leaving capacity for future tasks and at the same time \emph{avoids forgetting} previous tasks by taking into account neurons importance.
\clearpage

  
  
  


\cleardoublepage
  \graphicspath{{\chaptersdir/Online_Continual_Learning/image/}}%
  \chapter{Regularization based Online Continual Learning}\label{ch:reg_ocl}
Methods proposed in the literature towards continual  learning 
typically operate in a task incremental   setting~\cite{lee2017overcoming,li2016learning,Zenke2017improved,kirkpatrick2017overcoming,mallya2018packnet}.
A sequence of tasks is learned, one at a time, with all the data of the current task available but not of previous or future tasks.
Task boundaries and identities are known at all times.
This setup, however, is less likely   in various practical applications. 
Therefore we investigate how to transform 
task incremental setting to a more general online continual scenario.
We develop a system that keeps on learning over time in a streaming fashion,
with data distributions gradually changing and without the notion of separate tasks. 
To this end, we build on our importance weight regularizer, MAS~(Chapter~\ref{ch:MAS}), and show how it can be made online
by providing a protocol to decide i) when to update the importance weights, ii) which data to use to update them, and iii) how to accumulate the importance weights at each update step. 
Experimental results show the validity of the approach in the context of two applications: (self-)supervised learning of a face recognition model by watching soap series and learning a robot to avoid collisions. 

This  is a joint work with Klaas Kelchtermans. We have developed closely the approach. While I focused on the soap series experiments, Klaas focused on the experiments of the robot navigation. This work was published as an article in CVPR 2019~\cite{aljundi2018task}.
\section{Introduction}
In machine learning, one of the most basic paradigms 
is to clearly distinguish between a training and testing phase.
Once a model is trained and validated, it switches to a test mode: the model gets frozen and deployed for inference on 
previously unseen data, without ever making changes to the model parameters again. 
This setup 
assumes a ‘‘static'' world, 
with no distribution shifts over time.
Further, it assumes a static task specification, so no new requirements in terms of output (e.g. novel category labels) or new tasks added over time. 
Such strong division between training and testing makes it easier to develop novel machine learning algorithms, 
yet is also very restrictive.

Inspired by biological systems,  continual learning aims at breaking this strong barrier between the training and testing phase. As we have shown in previous chapters, many continual learning methods have been developed with the goal of training a shared model on one task at a time with  significant progress achieved. However, this specific
setup highly depends on knowing the task boundaries. 
These boundaries indicate 
good moments to consolidate knowledge, namely after learning a task. 
Moreover, data can be shuffled within a task so as to guarantee i.i.d. assumption. 
In an online setting, on the other hand, data needs to be processed in a
streaming fashion
and data distributions might change gradually.

In this chapter, we aim at overcoming this requirement of hard task boundaries. 
In particular, we investigate how methods proposed for task incremental setting can be generalized to an online setting. 
This requires a protocol to determine when to consolidate knowledge. 
Moreover, we investigate the effect of keeping a small buffer with difficult 
samples. 
For the latter, we take inspiration from the field of reinforcement learning, namely experience replay \cite{dqn2014mnih}, although using much smaller replay buffers, unlike very recent work of Rolnick et al.~\cite{rolnick17continualexpreplay}.

Task incremental setting
has mostly been studied for image classification~\cite{li2016learning,aljundi2016expert,lee2017overcoming,Zenke2017improved,rannen2017encoder}. 
Whenever the learner arrives at a new task, that is when learning on the previous task has converged, 
a standard procedure is to extend the output layer of the network with additional ``heads'' for   the new task's categories. 
Instead, the output of  the network considered here is fixed. 
In  the first application, learning to recognize faces, we cope with a varying number of categories by using an embedding rather than class predictions. 
In the second application, learning a lightweight robot to navigate without collisions, it is not the output labels that change over time but rather the environment. 
For both applications, data are processed in a streaming fashion. This is challenging, since the data are not i.i.d. distributed, causing samples within one batch to be unbalanced.

The contributions of this chapter are as follows: i) We were the first to extend the task incremental setting to free and unknown task boundaries in an online continual learning scenario; ii) We develop protocols to integrate our importance weight regularizer, MAS, in this online continual learning setting; iii) Our experiments on face recognition from T.V. series and on monocular collision avoidance prove the effectiveness of our 
method
in handling the distribution changes in the streaming data and stabilizing the online learning behavior, resulting in knowledge accumulation rather than catastrophic interference and improved performance in all the test cases.

In the following we discuss related work
(Section~\ref{sec:regonl-relatework}). We then describe our   online continual learning approach in Section~\ref{sec:regonl-methos}. We validate our system in  Section~\ref{sec:regonl-experiments} and end with discussion and conclusion in Section~\ref{sec:regonl-discussion}.

\section{Related Work}\label{sec:regonl-relatework}
\paragraph{Online Learning:}
As we have have explained in the introduction ( Chapter~\ref{ch:introduction}, Section~\ref{sec:intro-fields})  online learning addresses the learning problem over a stream of data instances sequentially. 

A first set of online learning algorithms consists of different techniques designed to learn a linear 
model~\cite{hazan2008adaptive,duchi2011adaptive,strehl2008online,hu2011online}. 
Online learning with kernels~\cite{kivinen2004online} extends this line of work to non-linear models, but the models remain shallow and their performance lags behind the modern deep neural networks.
Some recent works include \cite{sahoo2017online,ramasamy2018online},  start from a small network and then adapt the capacity by adding more neurons as new samples arrive. Typically,  online learning methods don't consider cases where the data generating distribution is changing overtime as in  continual learning. As such, they are  vulnerable to catastrophic forgetting once a distribution change occurs.

In terms of applications, the work of Pernici \etal~\cite{pernici17video, pernici2018memory} is similar to our first application scenario. They learn face identities in a self-supervised fashion via temporal consistency. They start from the VGG face detector and descriptor, and use a memory of detected faces. 
In contrast, we start from a much weaker pretrained model (not face-specific), and update the model parameters over time while they do not.

\paragraph{Continual Learning:}
   
While all continual learning methods to the date of development~(2018/2019) ~\cite{Zenke2017improved,kirkpatrick2017overcoming,aljundi2018memory,lee2017overcoming,yoon2018lifelong,li2016learning,rannen2017encoder,rebuffi2016icarl,sprechmann2018memory}
follow the task incremental learning setup, a special mention here goes to
Gradient Episodic Memory~(GEM)~\cite{lopez2017gradient}, as it moves a step forward towards the online setting. GEM assumes that the learner receives examples one by one but simplifies the scenario to locally i.i.d. drawn samples from a  distribution of a given task. It further assumes that a task identifier is given and used it to build an episodic memory of most recent samples for each seen  task. This memory is then exploited to prevent forgetting when learning new tasks. In our work, we use a very small buffer, around 100 samples, to  guarantee  better learning of hard samples. In the same time, this small buffer is deployed  to estimate the parameters importance once a stable phase of learning has been reached. 

The concept of  replay buffer  is often used in Deep Reinforcement Learning (DRL). 
However a crucial difference is that in both old and recent DRL works the replay buffer typically contains up to 1M samples corresponding to over 100 days of experience \cite{dqn2014mnih,rainbow17hessel}.
A common DRL technique, known as ``prioritized sweeping'', is to sample experiences with  large errors more often than others \cite{Moore1993}.
Our approach follows a similar strategy, however, to select which samples are going to  be stored in a small buffer. 

\section{Method}\label{sec:regonl-methos}
Our goal is to design a training method for task-free online continual learning. 
In this setting  where data is streaming and the distribution is shifting gradually, it is unclear whether current continual learning methods can be applied and how.
%

After studying  some of the existing methods, we find out that our importance weight regularizer (MAS), see Chapter~\ref{ch:MAS},  is indeed one of the most promising methods in this respect.
It enjoys the following favorable characteristics. 
\begin{enumerate}
\item {\em Constant memory}: it only stores an importance weight for each parameter in the network avoiding an increase in memory consumption over time.
\item {\em Problem agnostic}: it can be applied to any task and is not limited to classification. In particular, we can use it with an embedding as output, avoiding the need to add extra ``heads'' for new outputs over time.
\item  {\em Fast}: it only needs one backward pass to update the importance weights. During training, the gradients of the imposed penalty are simply the change that occurs on each parameter weighted by its importance. Therefore, the penalty gradients can be added locally and do not need 
 a backpropagation step.
 \item {\em Top performance}: MAS shows superior performance to other importance weight regularizers~\cite{hsu2018re}.

\end{enumerate}
In order to deploy MAS in an online continual learning scenario, we need to determine i) when to update the importance weights, ii) which data to use to update the importance weights, and iii) how to accumulate the importance weights at each update step.
We first introduce the considered online continual learning setup, then explain our training procedure under this setup.

\myparagraph{Setup:}
Online continual learning represents the general and desired setting of continual learning as we have explained in the introduction (Chapter~\ref{ch:introduction}, Section~\ref{par:general_ocl_settings}). Here is a gentle reminder: We assume an infinite stream of data and a supervisory or self-supervisory signal that is generated based on few consecutive samples. At each time step $t$, the system receives a few consecutive samples along with their generated labels $\{x_t,y_t\}$ drawn non i.i.d. from a current distribution $Q$. Moreover, the distribution $Q$ could itself experience sudden or gradual changes. The system is unaware of when these distribution changes are happening. The goal is to continually learn and update a function $f$ that minimizes the prediction errors on previously seen and future samples. In other words, it aims at continuously updating and accumulating knowledge.

We formulate the learning objective of an online system as follows.
Given an input model with parameters $\theta$, the system at each time step reduces
the empirical risk based on the recently received samples and a small buffer $\mathcal{B}=(X_\mathcal{B},Y_\mathcal{B})$ composed of updated hard samples. 
The  objective  function is then:
\begin{equation}\label{eq:regonl-online-objective}
J(\theta)= \ell( f(x_t;\theta),y_t)+\ell(f(X_{\mathcal{B}};\theta),Y_{\mathcal{B}})    
\end{equation}
Due to the strong non-i.i.d. conditions and the very low number of samples used for the gradient step, the system is vulnerable to catastrophic interference between recent samples and previous samples and faces difficulty in accumulating the knowledge over time.


As we explained in Chapter~\ref{ch:MAS}
in a traditional task incremental  setting, MAS estimates an importance weight for each network parameter after each training phase (task). The importance weights are estimated using training or test data from the task at hand. When learning a new task, changes to important parameters are penalized.


         


\myparagraph{When to update importance weights:}
In the case of a task incremental  setting where tasks have predefined boundaries, importance weights are updated after each task, when learning has converged. In the online case, the data is streaming without knowledge of a task's start or end (\ie when distribution shifts occur). We need a mechanism to determine when to update the importance weights. For this, we look at the surface of the loss function. 

By observing the loss, we can derive some information about the data presented to the system.
When the loss 
decreases, this indicates that the model has learned some meaningful new knowledge from those seen samples. Yet the loss does not systematically decrease all the time. When new samples are received that are harder or contain different concepts or input patterns than what was presented to the learner before, the loss may increase again. In these cases, the model has to update its knowledge, while minimally interfering with what has been learned previously. 

\begin{figure}[t]
\centering
\includegraphics[width=1\textwidth]{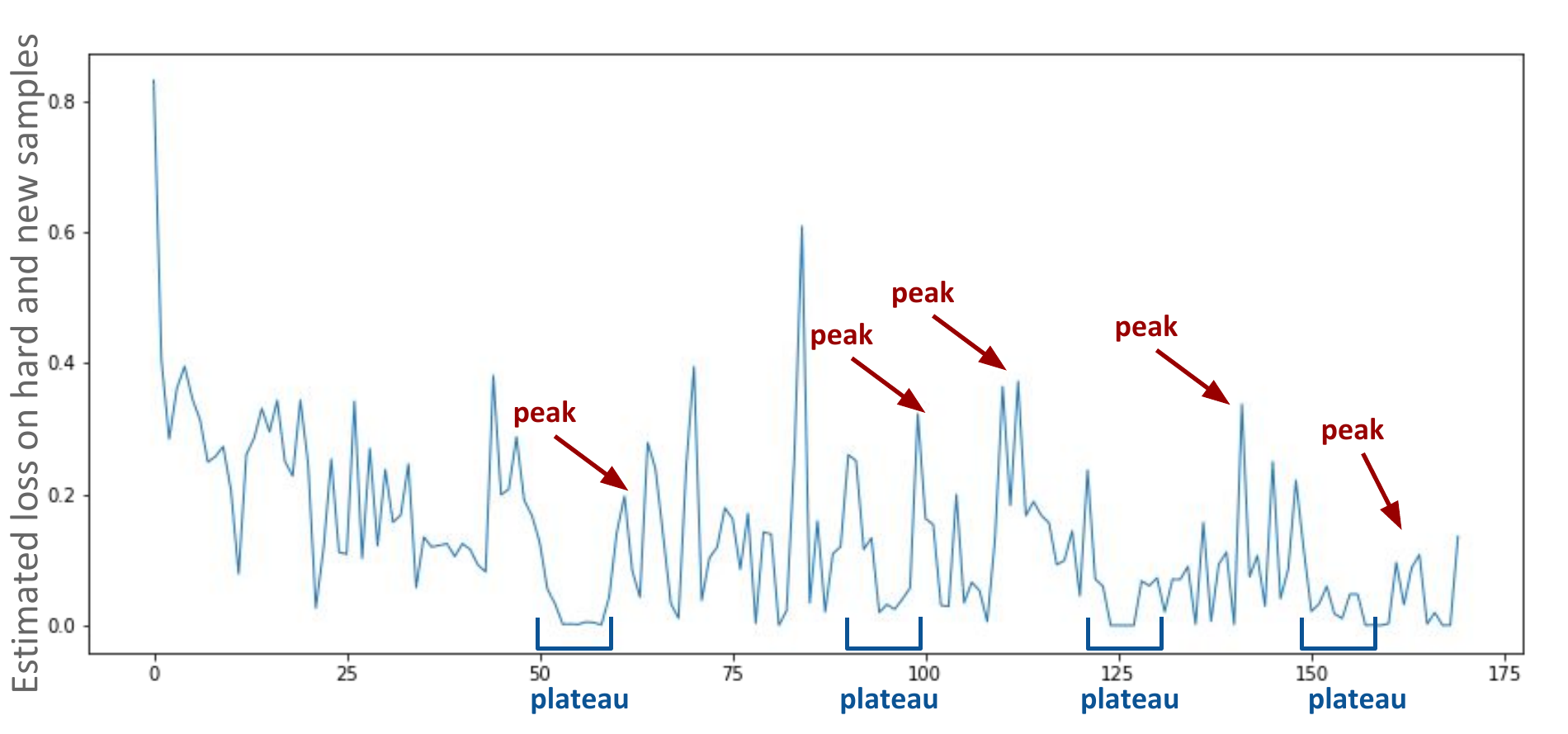}
  \caption[Figure shows ``plateaus'' and ``peaks'' in the loss surface, detected by our method.]{ By detecting ``plateaus'' and ``peaks'' in the loss surface our method decides when to 
  update the importance weights. Figure corresponds to the BBT experiment, see Section~\ref{sec:regonl-experiments:tvseries}; x-axis represents update steps}
  \label{fig:regonl-Loss_Surface}
  \end{figure}

We can conclude that plateaus ~(i.e. areas  with  consistent small values) in the loss function indicate stable learning regimes, where the model is confidently predicting the current labels, see Figure~\ref{fig:regonl-Loss_Surface}. Whenever the model is in such a stable area, it is a good time to  identify the parameters that are important for the currently acquired knowledge. When learning new, ``different'' samples the model will then be encouraged to preserve this knowledge.
This should allow the model to accumulate knowledge over time rather than replacing previously learned bits of knowledge. 
  
\myparagraph{Detecting plateaus in the loss surface:}
To detect these plateaus in the loss surface, we use a sliding window over consecutive losses during training. We monitor the mean and the variance of the losses in this window and trigger an importance weight update whenever they are both lower than a given threshold. 
We do not keep re-estimating importance weights: we only re-check for plateaus in the loss surface after observing a peak. Peaks are detected when the loss window  mean becomes higher than 85\% of a normal distribution estimated on the loss window of the previous plateau - that is when 
$\mu(L_{win})>\mu^{old}_L+\sigma^{old}_L$ where $\mu^{old}_L$ and $\sigma^{old}_L$
are the statistics of the previously detected plateau. This accounts for the continuous fluctuations in the loss function in the online learning and detects when significantly harder samples are observed.

\myparagraph{A small buffer with hard samples:} In a task incremental learning setup, importance weights are estimated on  the training/test data of a  task after training. This is not an option for online learning, as storing all the previous data violates the condition of our setup. On the other hand, using only the most recent sequence of samples would lead to misleading estimates 
as 
these few consecutive samples might not be representative and hence do not capture the acquired knowledge correctly.  We propose to use a small buffer of hard samples.
The  buffer of hard samples  is updated at each learning step by keeping the samples with highest loss among the new samples and the current buffer. This is important as previous samples cannot be revisited.  Hence, it also gives the system the possibility to re-process those hard samples and adjust its parameters towards better predictions. In addition, it provides a better estimate of the gradient step by averaging over the recent and hard samples. More importantly, the hard buffer  represents  better the previous history than the few recent samples, hence allows for a better identification of importance weights. Note that we  don’t consider the presence of outliers. In the context of online continual learning, the only assumption we make is that the label we receive for each training sample is valid. Handling outliers is left for further investigation.

\myparagraph{Accumulating importance weights:} As we frequently update the importance weights, simply adding the new estimated importance values to the previous ones would lead to very high values and exploding gradients. Instead, we maintain a cumulative moving average of the estimated importance weights. Note, one could deploy a decaying factor that allows replacing old knowledge in the long term. However, in our experiments a cumulative moving average showed more stable results. 

After updating the importance weights, the model continues the learning process while penalizing changes to parameters that have been identified as important so far. As such our final  objective function is: 
\begin{equation}
J(\theta)=
\ell(f(x_t;\theta),y_t)+\ell(f(X_\mathcal{B};\theta),Y_\mathcal{B})  +{\lambda} \sum_{i,j} \Omega_{ij}(\theta_{ij}-\theta^{*}_{ij})^2 
\end{equation}
where $\theta^{*}$ are the parameters values at the last importance weight update step.  Algorithm~\ref{regonl:algo} summarizes the different steps of the proposed continual learning system.
\begin{algorithm}[h!]
\caption{Online Continual Learning}\label{regonl:algo}
  \begin{algorithmic}[1]
  \footnotesize
    \State Input:$\delta_{\mu}, \delta_{\sigma}, \mathcal{S},\lambda$ \Comment{Loss window statistics, number of gradient steps.}
    \State Initialize: $\mathcal{B}=\{\},\mathcal{W} =\{\},\Omega=\vec{0}, \theta^*=\vec{0}, \mu^{old}_L=0, \sigma^{old}_L=0$, $\mathcal{P} =1$ 
 \MRepeat
    \State Receive M recent samples $X,Y$
    \For{$s$ in $\mathcal{S}$}
        \State ${J(\theta)}=\ell_{\theta}(X,Y)+\ell_\theta(X_{\mathcal{B}},Y_{\mathcal{B}})+{\lambda} \sum_{i,j} \Omega_{ij}(\theta_{ij}-\theta^{*}_{ij})^2$
                \If{ $s=1$ }
            \State  $\mathcal{W} \leftarrow \text{update}(\mathcal{W}, \mathcal{\ell_\theta}(X,Y),\mathcal{\ell_\theta}(X_\mathcal{B},Y_\mathcal{B}))$ \Comment{Update the loss window.}
        \EndIf
        \State $\theta \leftarrow$SGD($J(\theta))$

    \EndFor
    \If{$ \mathcal{P}$ \& $\mu(\mathcal{W})<\delta_\mu$  \&  $\sigma(\mathcal{W})<\delta_\sigma$} 
         \State $\Omega \leftarrow \text{update}(\Omega,\theta,(X_\mathcal{B},Y_\mathcal{B}))$
         \State $\theta^*\leftarrow \theta$
         \State $\mu^{old}_L=\mu(\mathcal{W}), \sigma^{old}_L=\sigma(\mathcal{W})$
         \State $\mathcal{W}=\{\}$, $\mathcal{P}=0$
    \EndIf
    \If{$\mu(\mathcal{W})>\mu^{old}_L+\sigma^{old}_L$}
        \State $\mathcal{P}=1$
    \EndIf
    \State $(X_\mathcal{B},Y_\mathcal{B})\leftarrow\text{update}((X_{\mathcal{B}},Y_{\mathcal{B}}),(X,Y), \ell_{\theta}(X,Y))$
  \EndRepeat
  \end{algorithmic}
\end{algorithm}
\section{Experiments}\label{sec:regonl-experiments}
As a proof-of-concept, we validate our proposed method on a simple synthetic experiment. 
Later, we evaluate the method on two applications with either weak or self-supervision. 
First, we learn actor identities from watching soap series. The second application is robot navigation. 
In both cases, data are streaming and online continual learning is a key factor.

\subsection{Synthetic Experiment}\label{sec:regonl-experiments:synthetic}
We want to validate our method on a controlled environment where the changes in the training data are well identified. For this reason, we constructed a binary classification problem with points in 4D in/out of the unit sphere. We consider a simple two tasks scenario where each task corresponds to a quadrant in the unit sphere. A simple hypothesis of a hyper plane separating in/out points of the first quadrant will be orthogonal to an optimal  plane solving the second quadrant task.  As a result, catastrophic forgetting is expected to happen if no mechanism is deployed to prevent changing the learned hypothesis drastically. 

\begin{figure}[t]
\centering
\includegraphics[width=0.9\textwidth]{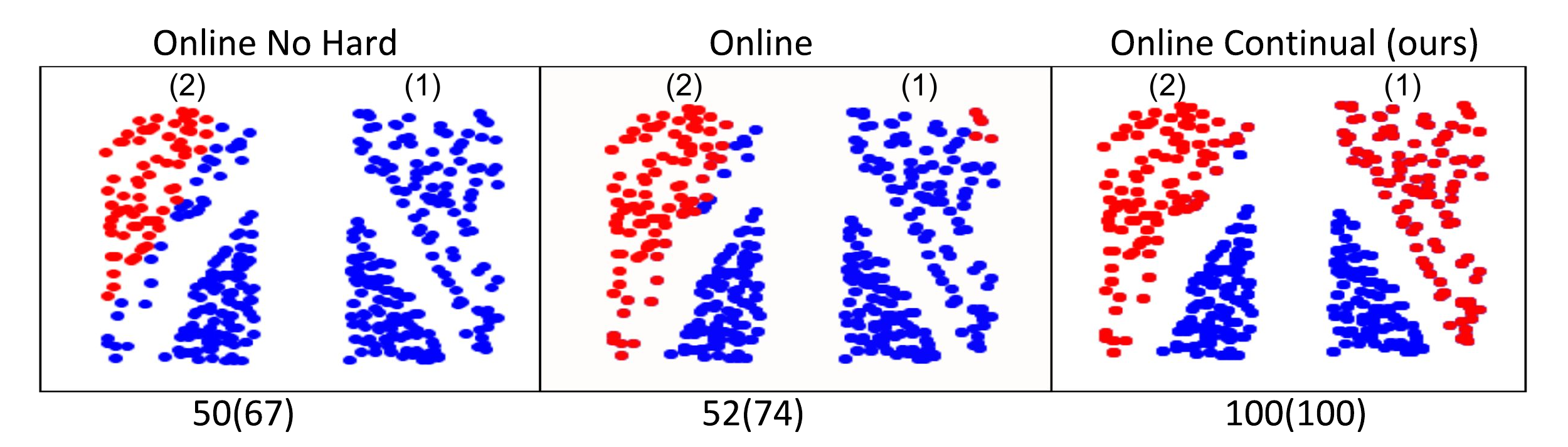}
\caption[Synthetic experiment]{ Synthetic experiment: Predictions on first~(right) and second~(left) quadrants after training on the second  quadrant. Test accuracy (\%) on the first quadrant (total test accuracy on both quadrants(\%)) shown beneath the figure.}
\label{fig:regonl-synthetic}
\end{figure}

 To validate the effect of our approach, coined ~{\tt Online Continual}, we  consider two baselines:
 \begin{itemize}
     \item {\tt Online} a model trained online using a hard buffer with the  objective of Equation~\ref{eq:regonl-online-objective}
     \item {\tt Online No Hard} the online trained model but without the aid of the hard buffer.
     
 \end{itemize}
All methods are trained online with data from the first quadrant and then task switches to the second quadrant still online without providing the compared methods with this shift information. 
Figure~\ref{fig:regonl-synthetic} depicts the methods predictions near the decision boundary in the two quadrants at the end of training the second quadrant. 

As expected the forgetting is sever and apparent in the baselines predictions. The hard buffer results in better learning (higher total test accuracy) but doesn't stand against forgetting alone. 
Our method {\tt Online Continual} with MAS deployed succeeds  in perfectly predicting the in/out points of the unit sphere from  both the first and the second quadrants.  Our method has updated the importance weights after learning a good hypothesis on first quadrant and used the importance weights to prevent forgetting when data shifted to the second quadrant. 

Having proven the effectiveness of our method on a simple synthetic experiment, we move to test more complex situations with streaming data and gradually shifting distributions.

\subsection{Continual Learning by Watching Soap Series}\label{sec:regonl-experiments:tvseries}
Here, we assume that an intelligent agent is watching soap series episodes  and learns to differentiate between the faces of the different actors. The agent is equipped with a face detector module that is detecting faces online and a multi-object face tracker. In the case of weak supervision, we assume there is an annotator telling the agent whether two consecutive tracks are of the same identity or not. For the self-supervised case, we use the fact that if two faces are detected in the same image then their tracks must belong to two different actors. In both cases, faces within a track considered to belong to the same actor.

\myparagraph{Setup:}
We start from  AlexNet~\cite{krizhevsky2014one} architecture with the convolutional layers pre-trained on ImageNet~\cite{krizhevsky2012imagenet} and the fully connected layers initialized randomly. The output layer is of size 100. Since the input consists of two tracks of two different identities, we use the triplet margin loss~\cite{balntas2016learning} which has been shown to work well in face recognition applications. This has the additional advantage that we don't need to know all the identities beforehand and new actors can be added as more episodes are watched. 

\begin{figure*}[t]
    \centering
    \includegraphics[width=\textwidth]{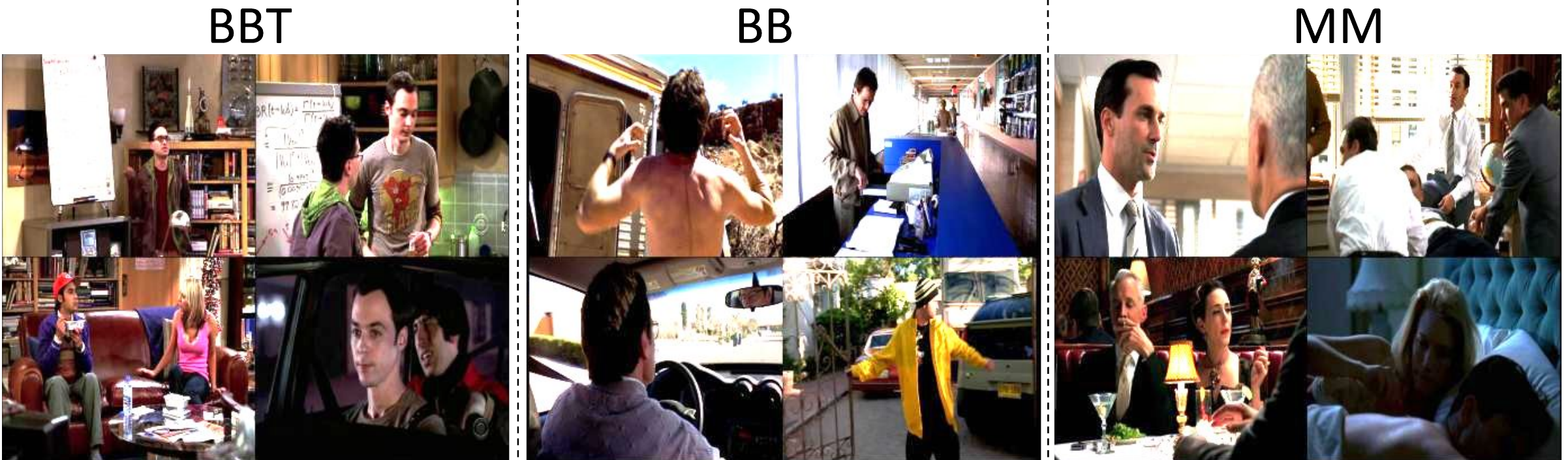}
    \caption{ Four example images for each soap series, from left to right: Big Bang Theory (BBT), Breaking Bad (BB) and Mad Men (MM).}
    \label{fig:regonl-soapseries}
\end{figure*}

\myparagraph{Dataset:}
We use the actor labelling dataset from~\cite{aljundi2016actor}, specifically 6 episodes of The Big Bang Theory (BBT), 4 episodes of Breaking Bad (BB), and one episode of Mad Men (MM)\footnote{Unfortunately, there was an issue with the labels for the other episodes of Mad Men, which prevented us from using these.}. Note that for BB and MM, the episodes were further split into a total of 22 and 5 chunks, respectively.
For each episode we use the frames, detected faces and tracks along with track labels from~\cite{aljundi2016actor}. 
Tracks are processed in chronological order, imitating the setting where tracks are extracted in an online fashion while watching the soap series. As a result, the data are clearly non-i.i.d..
Figure \ref{fig:regonl-soapseries} shows 4 example frames  for each of the different soap series: Big Bang Theory, Breaking Bad and Mad Men. 
These examples demonstrate the scene diversity and large variance in imaging conditions. Breaking Bad is more actor-centric with a majority of the frames showing only the main character.

For the supervised setup, every tenth/fifth track is held out as test data in BBT/BB respectively since the latter has more tracks, 339 tracks BBT compared to 3941 BB. All the other tracks are used for training.
As we only have one episode for MM, we decided not to use it for the supervised setup.

For the self-supervised setup, BB turned out to be unsuitable, given that it is an actor centric series with a large majority of the scenes focusing on one actor. To still have results on two series, we do report also on MM in this case, in spite of it being only one episode.
Further, the original tracks provided by~\cite{aljundi2016actor} were quite short (an average of 8/22 faces per track in BBT/MM).  Since this is problematic for the self-supervised setting, we use a simple heuristic based on the distance between the faces embedding (based on AlexNet pretrained on ImageNet) to merge adjacent tracks belonging to the same actor. Table~\ref{tab:regonl-tvstatistics} states the total number of training tracks and images in each dataset regarding both weak supervision and self supervision scenarios.

\myparagraph{Training:}
Whenever two tracks are encountered belonging to different actors, a training step is performed using the detected faces (one face every 5 frames). If the two tracks contain more than 100 faces, a random sampling step is performed. We use a hard buffer size of 100 triplets and a fixed loss window size of 5. A few gradient descent steps are performed at each training step (2-3 for the supervised setting, 10-15 for the self-supervised one). We use SGD optimizer with a learning rate of $10^{-4}$. Hyperparameters were set based on the first BBT episode, please refer to  Table \ref{tab:regonl-hyperparamsimul} for more details.

\myparagraph{Test:}
To test the accuracy of the trained model on recognizing the actors in the soap series, we use 5 templates of each actor selected from different episodes. 
We then compute the Euclidean distance of each test face to the templates, based on the learned representation, and assign the input face to the identity of the template that is closest. 

\begin{table}[h]
    \centering
    \begin{tabular}{|r|l|l|l|}
    \hline
        T.V. series&self-supervision&$\#$ of tracks&$\#$  of training images\\ \hline
        BBT &No & 339& 2,633\\ \hline
        BBT & Yes & 223&588 \\ \hline
        BB & No&1,387& 5,190 \\ \hline
        MM & Yes&144 & 2,763\\ \hline
    \end{tabular}
    \caption{Statistics of the deployed T.V. series datasets in both supervision cases.}
    \label{tab:regonl-tvstatistics}
\end{table}
\begin{table}[h]
    \centering
    \resizebox{\textwidth}{!}{
    \begin{tabular}{|r|l|l|l|}
         \hline
        & Soap Series (Sec.~\ref{sec:regonl-experiments:tvseries} ) &  Corridor ( Sec.~\ref{sec:regonl-experiments:collisionavoidance} ) &  Turtlebot  ( Sec.~\ref{sec:regonl-experiments:realworld} ) \\
         \hline
         Architecture & AlexNet & Tiny v2 & Tiny v2 \\
         Initialization & ImageNet & random & random \\
         Learning rate & 0.0001 & 0.01 & 0.01 \\
         Optimizer & SGD & SGD & SGD \\
         Hard Buffer Size $|\mathcal{B}|$ & 100 & 40 & 30 \\
         Regularization Weight ($\lambda$) & 100 & 0.5 & 0.5 \\
          Mean Loss Threshold$\delta_{\mu}$& 0.3 & 0.5 & 0.5 \\
          Variance Loss Threshold $\delta_{\sigma}$ & 0.1 & 0.1 & 0.02 \\
         Loss Window Length $|\mathcal{W}|$ & 5 & 5 & 5 \\
         \hline
    \end{tabular}}
    \caption{Hyperparameters used in the different experiments of this chapter.}
    \label{tab:regonl-hyperparamsimul}
\end{table}

\myparagraph{Baselines:}
To estimate the benefit of our system, {\tt Online Continual}, we compare it against the following baselines:
\begin{enumerate}
    \item {\tt Initial} : the pretrained model, \ie before training on any of the episodes. 
    \item {\tt Online} : a model trained in the explained online setting but without our importance weight regularizer, MAS.
   
    \item {\tt Online Joint} : a model trained online, again without MAS regularization, but with shuffled tracks across episodes to obtain i.i.d. samples.
    \item {\tt Offline Joint} : a model that differs from Online Joint  by going multiple epochs over the whole data. This stands as an upper bound.
  
\begin{figure*}[t]
    \centering
    \subfloat[]{\includegraphics[width=0.5\textwidth]{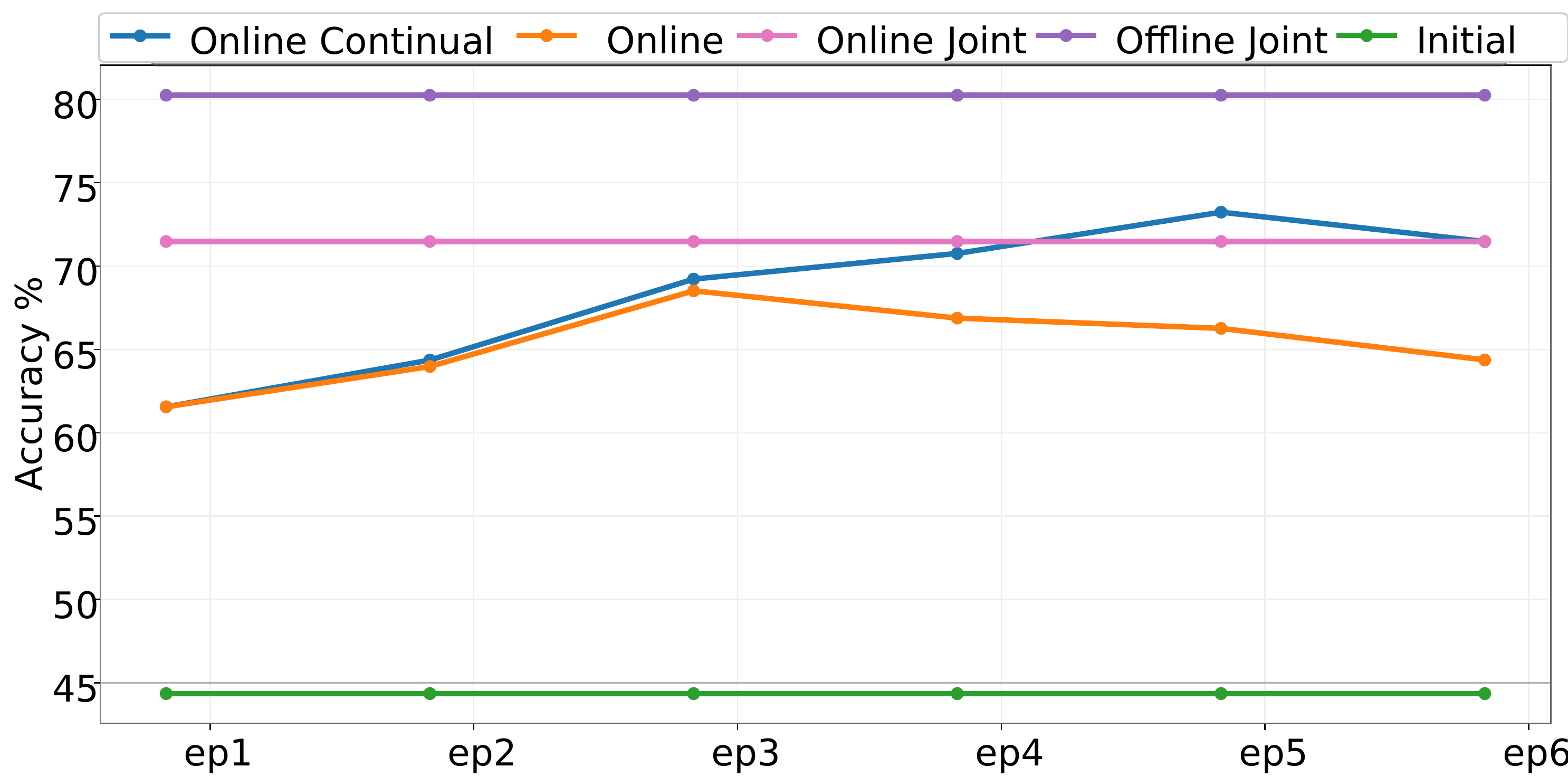} }%
     \hfillx
    \subfloat[]{\includegraphics[width=0.5\textwidth]{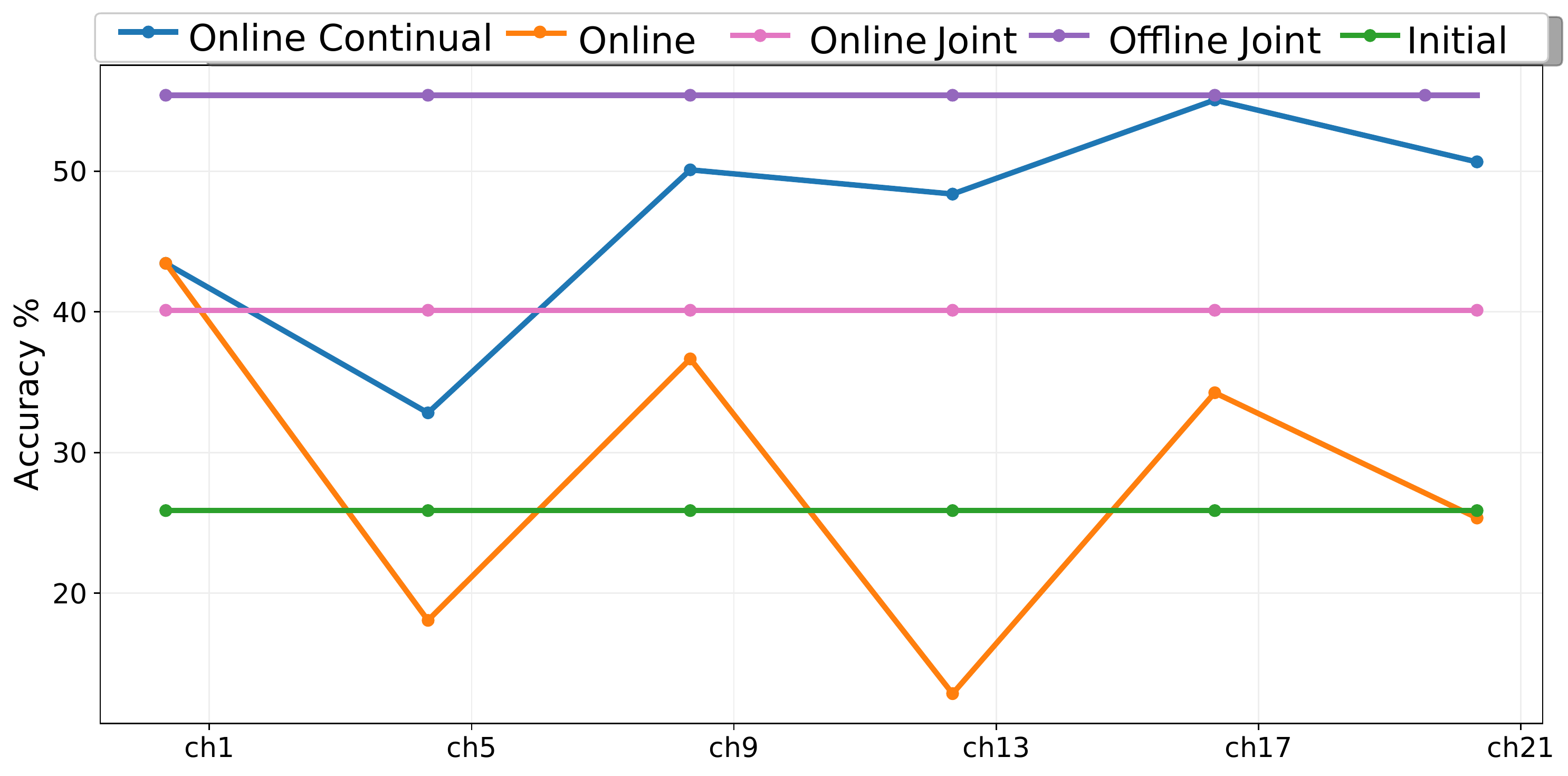}} 
         \caption[Weak supervision results]{  Accuracy on test data at the end of each episode for BBT (a) and after chunks 1,5,9,13,17 and 21 of BB (b), under weak supervision.}
         \label{fig:regonl-soapseries_super}%
\end{figure*}
\end{enumerate}
\begin{figure*}%
    \centering
    \subfloat[]{\includegraphics[width=0.5\textwidth]{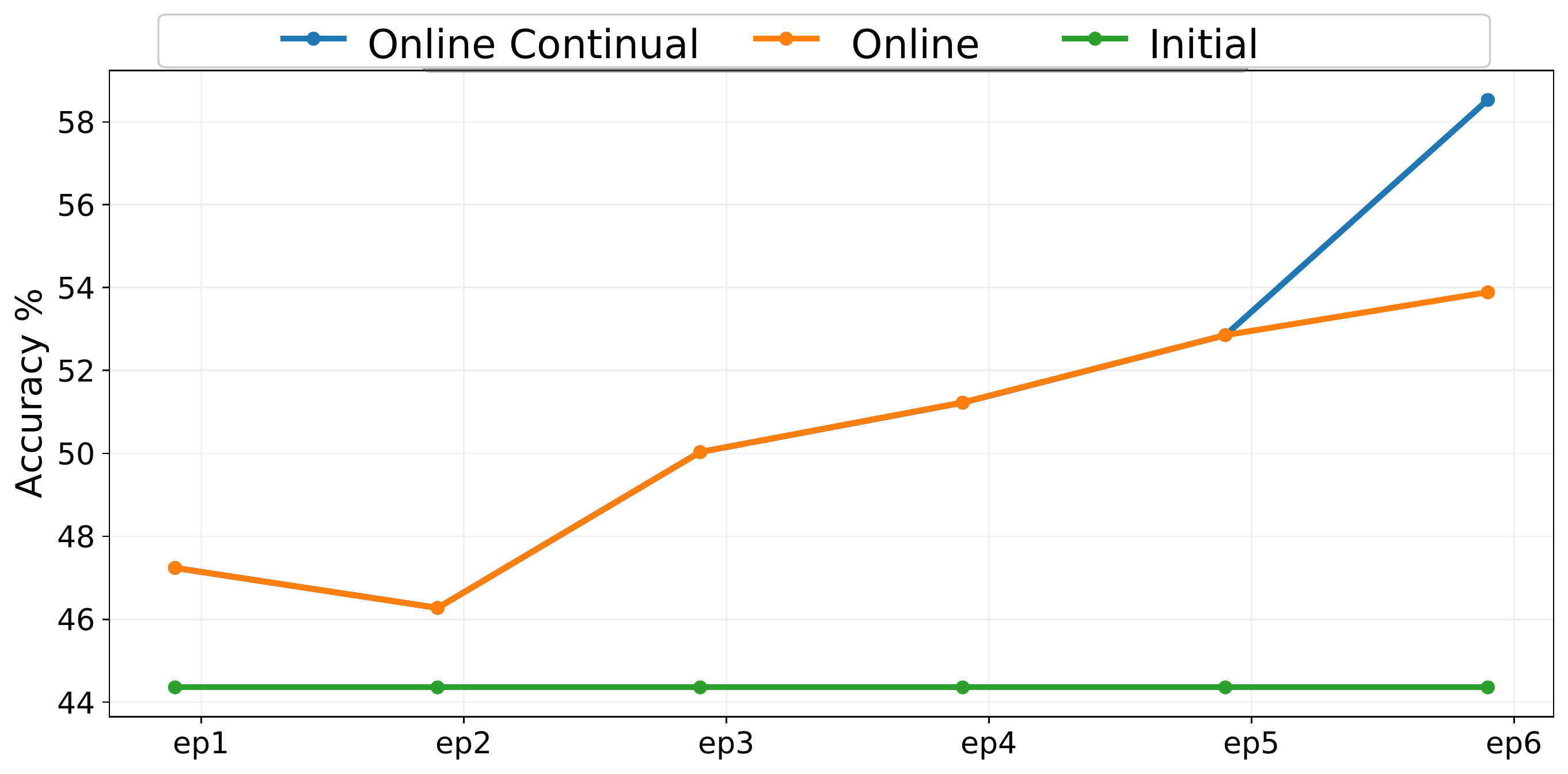} }%
    \hfillx
    \subfloat[]{\includegraphics[width=0.495\textwidth]{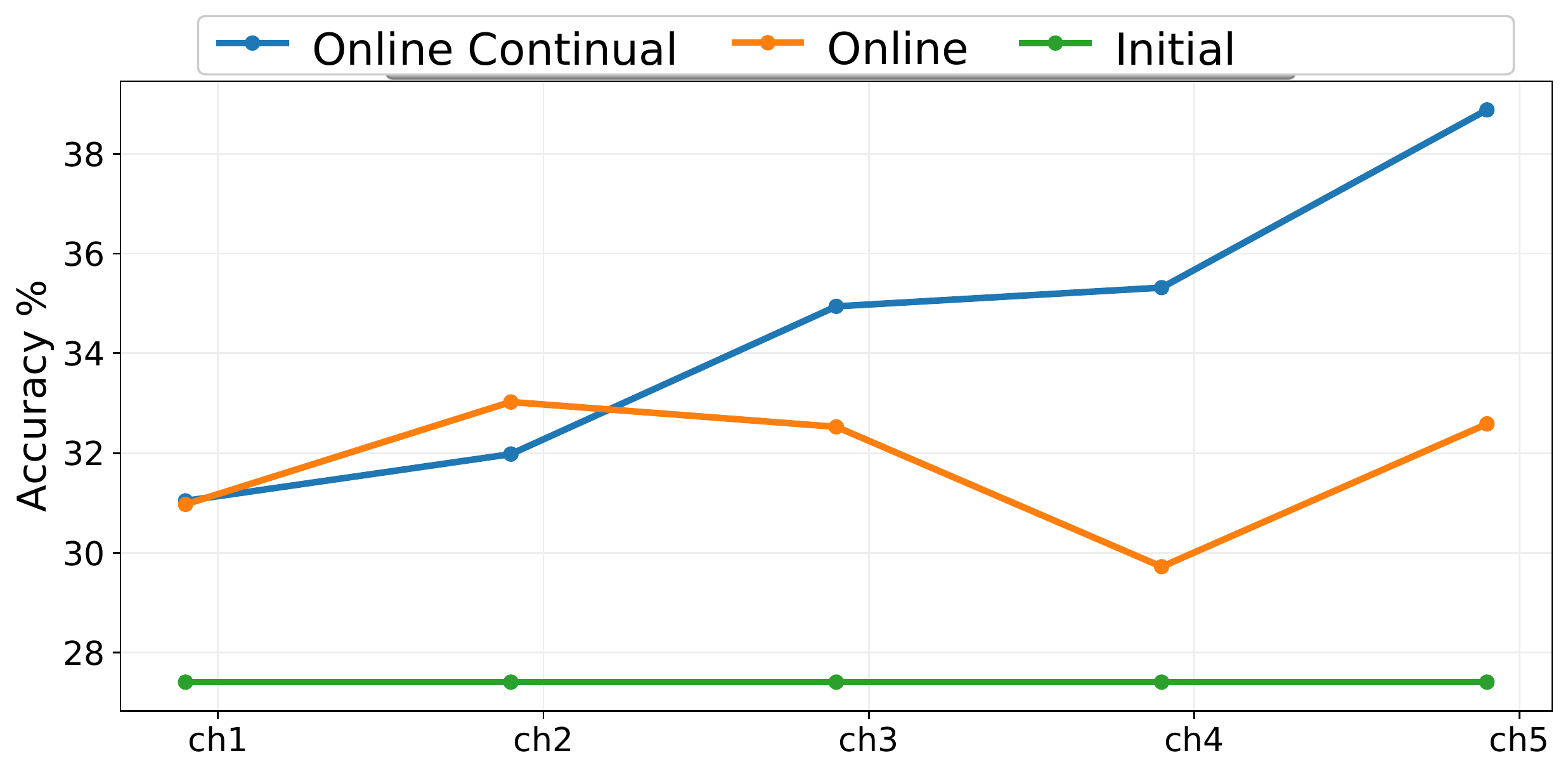} }%
    \caption[ Self-supervision results]{ Self-supervised setting: accuracy on all faces of BBT after each episode (a) and of MM after each of the 5 chunks (b). }
    \label{fig:regonl-selfsupervised}%
\end{figure*}

\subsubsection{Weak Supervision Results}
Figure~\ref{fig:regonl-soapseries_super} (a) shows the actor recognition accuracy evaluated on all the test data of BBT, at the end of each episode.
Initially,  {\tt Online}  (orange) 
obtains an increase of $20\%$ in accuracy compared to the initial model. Yet it fails to continue accumulating knowledge and improving the accuracy as training continues.
After the third episode, the overall accuracy starts to decay, probably because the knowledge learned from these new episodes interferes with what was learned previously. In contrast, our {\tt Online Continual} learning system (blue) continues to improve its performance and achieves at the end of the 6 episodes an  accuracy that matches the accuracy of the model trained with shuffled data under the i.i.d. condition ({\tt Online Joint}, pink). {\tt Offline Joint} (purple) with multiple revisits to the shuffled data achieves the top performance. Note that this is only $8\%$ higher than our continual learning system trained under the online and changing distribution condition. 

Figure~\ref{fig:regonl-soapseries_super} (b) shows the accuracy on all the test data of BB, after each 4 chunks while learning the 4 episodes.
Clearly this  series is much harder than BBT. Most of the shots are outdoor and under varied lighting conditions, as also noted in~\cite{aljundi2016actor}. This corresponds to large distribution changes within and between episodes.  Here,  {\tt Online} (orange) fails to increase the performance after the first episode. Its accuracy notably fluctuates, probably depending on how (un)related the recently seen data is to the rest of the series. Again, our {\tt Online Continual} learning system (blue) succeeds in improving and accumulating knowledge -- up to a $100\%$ improvement over  {\tt Online}. Like  {\tt Online}, its performance drops at times, yet the drops are dampened significantly, allowing the model to keep on learning over time. 
Surprisingly, it even outperforms  {\tt Online Joint} baseline (pink) and comes close to  {\tt Offline Joint } upper bound (purple) that only reaches this accuracy after ten revisits to the training data.

\subsubsection{Self Supervision Results}
Next we move to the case of learning with self-supervision.
This scenario reflects the ideal case where continual learning becomes most interesting. 
Remember that, as a clue for self-supervision, we use the fact that multiple tracks appearing in the same image should have different identities. 
We use the six episodes of BBT, although only the first and the sixth episodes actually have a good number of tracks with two persons appearing in one image. 
Figure~\ref{fig:regonl-selfsupervised}~(a) shows the accuracy on all the episodes after learning each episode. Note how  {\tt Online} learning baseline (orange) continues to improve slightly as more episodes are watched. It's only when we get to the last episode, with a larger number of useful tracks, 
that our {\tt Continual Online}  (blue) starts to outperform  {\tt Online} baseline.

Figure~\ref{fig:regonl-selfsupervised}~(b) shows the recognition accuracy on the first episode of Mad Men after each chunk. Similar to the previous experiments our {\tt Online Continual} (blue) succeeds in improving the performance and accumulating the knowledge. We conclude that the ability of our system to learn continually is clearly shown, both for weak and self-supervised scenarios. 
\subsubsection{Ablation Study}
\begin{figure*}%
\vspace*{-0.4cm}
    \centering
 {\includegraphics[width=0.8\textwidth]{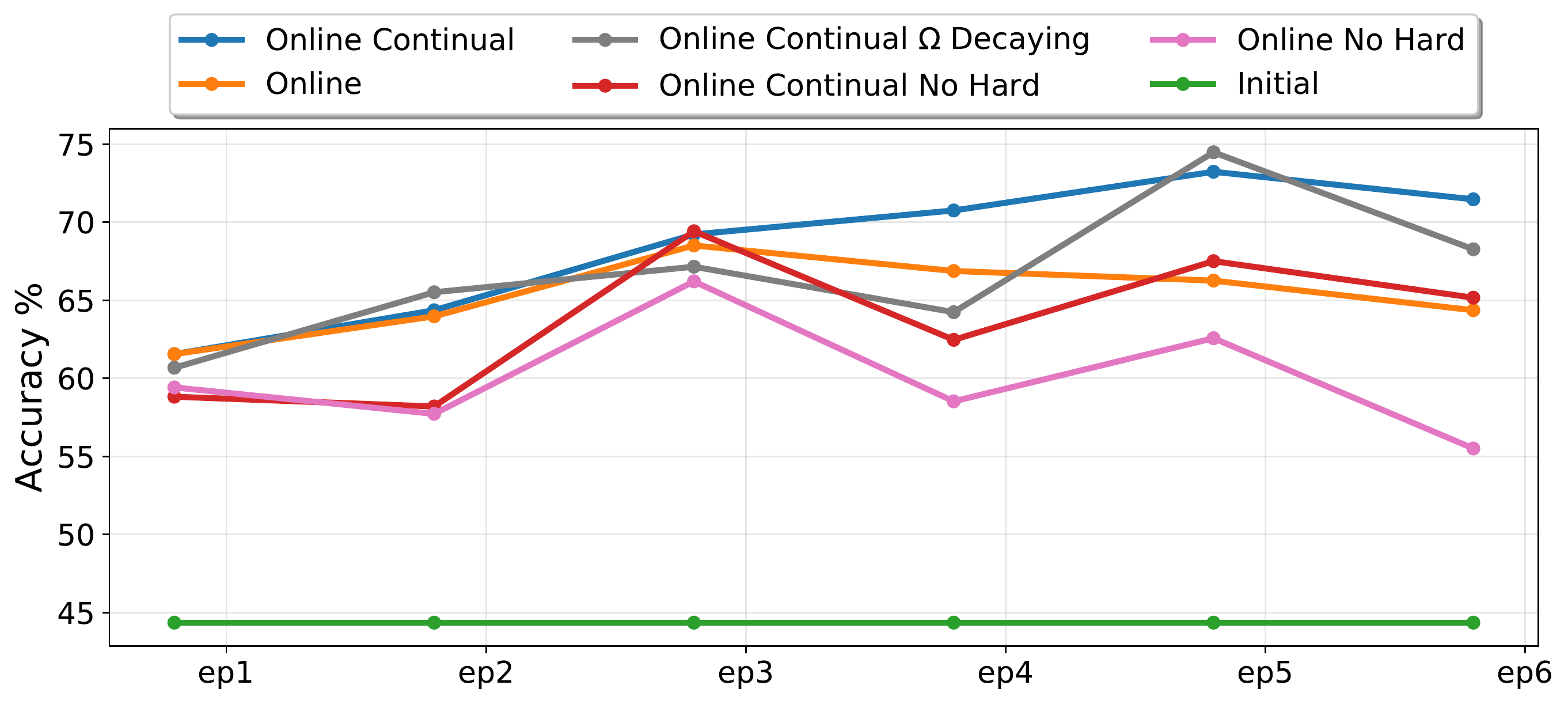}}    \caption[A study on the importance of the hard buffer and the cumulative $\Omega$ average versus a decaying $\Omega$.]{  \label{fig:regonl-soapextra} A study on the importance of the hard buffer and the cumulative $\Omega$ average versus a decaying $\Omega$, figure shows the test accuracy after each episode of BBT.}
   
\end{figure*}

Next we perform an ablation study to evaluate the impact of two components of our system. The first factor is the hard buffer used for stabilizing the online training and for updating the importance weights. The second factor is the mechanism for accumulating importance weights across updates. In our system we use a cumulative moving average, which gives all the estimated importance weights the same weight. An alternative is to deploy a decaying average. This reduces the impact of old importance weights in favor of the newest ones. To this end, one can set $\Omega_{\mathcal{T}}=(\Omega_{\mathcal{T}-1} +\Omega^*)/2$ where $\Omega^*$ are the currently estimated importance weights. Figure~\ref{fig:regonl-soapextra} shows the accuracy on all the test data of BBT after each episode achieved by the different variants. The hard buffer clearly improves the performance of both the {\tt Online} and {\tt Online Continual}. The buffer, even if small, gives the learner a chance to re-pass over hard samples and to adjust its gradients for a better estimate of the parameters update step. Additionally, it allows a better estimate of the importance weights used in our {\tt Online Continual} learning system.  
The decaying average for the importance weights update, leads to more fluctuations due to the higher impact of more recent importance estimates. 
This allows more forgetting and more bias towards the recent estimate that could be unrepresentative to the overall test data.
  
\begin{figure*}%
    \centering
{\includegraphics[width=0.7\textwidth]{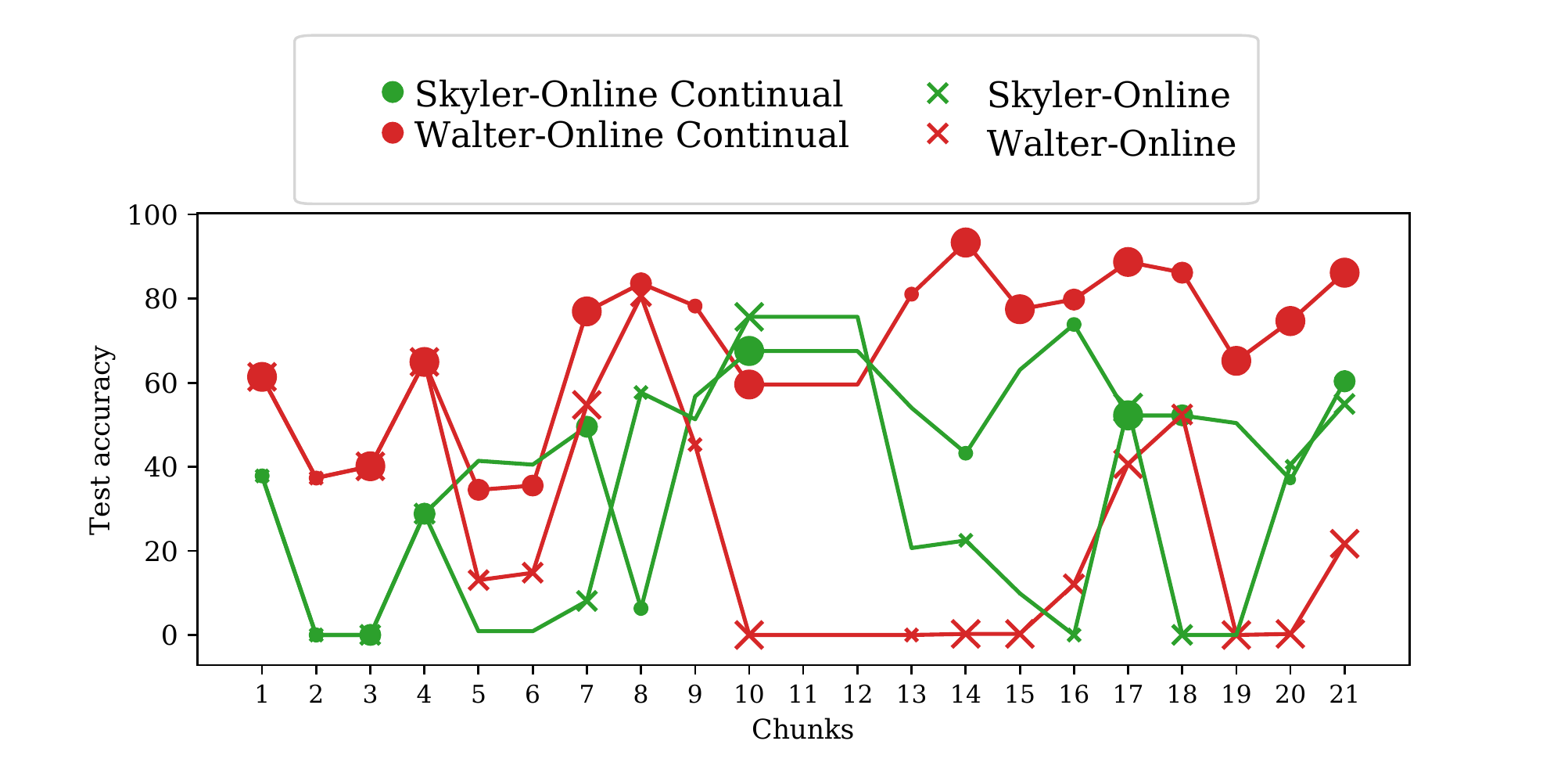}}

    \caption[A study on the actors recognition during the course of training.]{ \label{fig:regonl-actorstudy} A study on the actors recognition during the course of training, figure shows two main actors test accuracy after each chunk in BB.}

  \vspace*{-0.4cm}
\end{figure*}

\myparagraph{Relationship between samples and recognition performance during training:}
To show how the predictions on the seen actors change over the online training time,  we plotted the accuracies per actor after each chunk (for the two most frequent characters of BB, to avoid overloading the figure), see Figure\ref{fig:regonl-actorstudy}. Marker size indicates the actor's frequency in a chunk, no marker indicates zero appearance. Low frequency in a chunk typically causes the accuracy of {\tt Online}  to drop while our method is more stable. For example the recognition accuracy of ``Skyler'' drops significantly after learning chunks 13 to 16 for the {\tt Online } baseline. This is due to the very rare appearance of this actor in these chunks. Note that our {\tt Online Continual} maintains relatively high accuracy on recognizing this actor (Skyler) after these chunks. It is also worth mentioning that the performance of {\tt Online} keeps fluctuating during the learning sequence while ours {\tt Online Continual} tend to show more stable behavior after few chunks have been learned. This indicates that the model has accumulated a sufficient knowledge to recognize these actors in the upcoming sequences.

\subsection{Monocular Collision Avoidance}\label{sec:regonl-experiments:collisionavoidance}
Collision avoidance is the task in which a robot navigates through the environment aimlessly while avoiding obstacles. We train a neural network to perform this task based on a single input image.
Training is done with self-supervision where a simple heuristic based on extra sensors, serves as a supervising expert.
The  neural network learns to imitate the expert's control,  cloning its behavior.
Learning to navigate into different environments requires collecting huge amount of training data and as in the case of other machine learning tasks, new environments can always be visited. 
 As such, continually learning to navigate in varying environments using generated supervisory signals 
 is an excellent setup for deploying our system.

\myparagraph{Architecture:}
A neural network model takes a 128x128 RGB frame as input and outputs three discrete steering directions.
The architecture consists of 2 convolutional and 2 fully-connected layers with ReLU-activations. 
Differently from the soap series experiments,  training starts from a  randomly  initialized network. SGD optimizer  with  cross-entropy loss is used. 
More details on architecture and hyper-parameters are reported in Table~\ref{tab:regonl-hyperparamsimul}

\begin{figure}[h!]
    \centering
    \includegraphics[width=1.\textwidth]{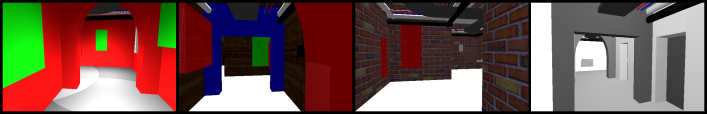}
    \caption[Example views in the corridor sequence corresponding to environments A, B, C and D]{ Example views in the corridor sequence corresponding to environments A, B, C and D, depicted from left to right.}
    \label{fig:regonl-corridors}
    \vspace*{-0.5cm}
\end{figure}

\myparagraph{Simulation:}
The experiment is done in a Gazebo simulated environment with the Hector Quadrotor model.
The expert is a heuristic reading scans from a Laser Rangefinder mounted on the drone and turning towards the direction with the furthest depth.
The demonstration of the expert follows a sequence of four different corridors, referred to as A,B,C and D. 
The environments differ in texture, obstacles and turns, as shown in Figure~\ref{fig:regonl-corridors}.

\myparagraph{Training and Evaluation}
Every 10 frames a training step  occurs, minimizing the  training objective. For each model, three networks are trained with different seeds.  

\begin{figure*}[h!]

    \centering
    \includegraphics[width=1\textwidth]{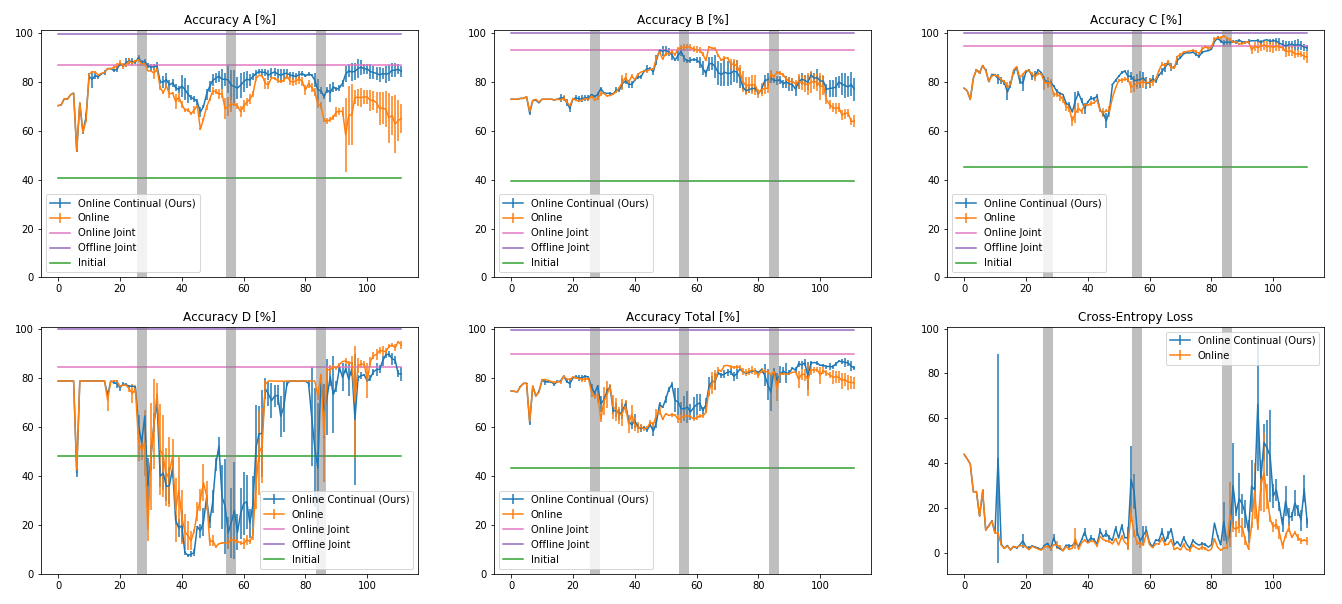}
    \caption[Training accuracies on each corridor during  learning  the  (A,B,C,D) sequence]{ Training accuracies on each corridor during  learning  the  (A,B,C,D) sequence. Gray vertical lines indicate the transition to a new environment. The lower right figure shows the cross-entropy loss on the recently received samples. The accuracies of the baselines are added as horizontal lines. Note that the offline joint training accuracy is closer to $100\%$}
    \label{fig:regonl-corridoraccuracies}
     \vspace*{-0.3cm}
\end{figure*}

We consider the same baselines as in the soap series experiments:{\tt Initial}, {\tt Online},  {\tt Online Joint}, {\tt Offline Joint } in addition to our system {\tt Online Continual}. Note that {\tt Initial} is a randomly initialized model here. Figure~\ref{fig:regonl-corridoraccuracies} shows the accuracies achieved during  training on  the different environments by the compared methods. 
The accuracy of both {\tt Online} and {\tt Online Continual}  increases in environments where it is currently learning, with grey bars indicating environment changing. However, {\tt Online}  tends to forget the early environments while training on new environments.
Especially in environment D, the  forgetting of earlier environments A and B is outspoken.
The cross-entropy loss in environment D rises for all models, indicating a significant change in the data generating distribution.

\begin{figure}[h!]
    \centering
    \includegraphics[width=0.7\textwidth]{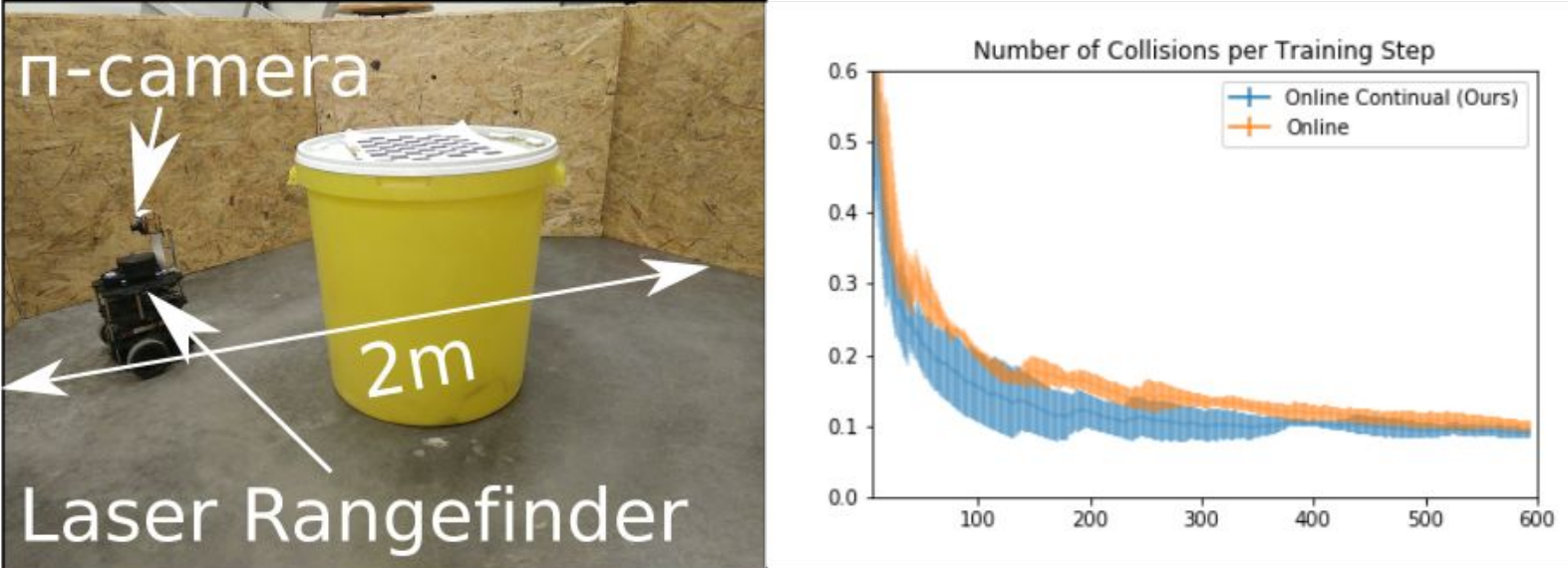}
     \caption[Number of collisions per training step in real-world online and on-policy setup.]{ Left: Real-world online setup. Right: Number of collisions per training step. Our system  speeds up the training, rapidly achieving lower number of collisions.}
    \label{fig:regonl-realworld}
    \vspace*{-0.8cm}
\end{figure}




\subsection{Proof of Concept in the Real World}\label{sec:regonl-experiments:realworld}
In a final experiment, we deploy our online continual learning system on a turtlebot in a small arena in our lab, see Figure \ref{fig:regonl-realworld}. The model is on-policy pretrained offline in a similar simulated environment. On-policy refers to the model being in control during training instead of the expert.
In the previous experiments, our system has proven to be advantageous when  experiencing changing environments. In this setup we show that our system   is also  beneficial  during on-policy training within one environment. 
Again, an expert based on the Laser Rangefinder is providing a self-supervisory signal.
On-policy learning tends to be more difficult as the data contains a lot of ``dummy'' samples when the model visits irrelevant states.
This data inefficiency causes the model to learn slower and possibly forget along the way.
For example, if the model collides on the left side, the recent signals teach the model to turn right more often.
However, after crossing the arena and bumping on the right side, one still wants the model to remember its mistakes made earlier. 
As such, preserving acquired knowledge over time is crucial for on-policy online learning.
In Figure~\ref{fig:regonl-realworld}, we show the number of collisions per step over time with error bars taken over three different models.
Clearly our system, {\tt Online Continual}, helps the model to learn faster, with the number of collisions dropping more rapidly  than  {\tt Online} alone.
\section{Discussion {\&} Summary }\label{sec:regonl-discussion}

Importance weight regularization appears most effective in online training scenarios when learning environments experience  changes while it has  a stabilizing effect on the training of one environment.  In some cases however, our system tends to slow down the adaptation to newly seen data. Especially when the new data is much more informative or representative than the old, it initially has a negative effect on the training. 
In other words, pure online learning is faster to adapt to new changes but therefore also inherently less stable. Ultimately, whether the stabilizing effect of continual learning is advantageous or not, depends on the time scale of the changes in the data.

While in this work we focus on a setting where the network architecture remains fixed, and no new outputs or tasks are added over time, in the next chapter we study  settings with larger shifts. 
For instance in a class-incremental setting, an extra output unit could be added to the network each time a new category label appears. 





In conclusion, we pushed the limits of current task incremental setting towards online task-free continual learning. We assume an infinite stream of input data, containing changes in the input distribution both gradual and sudden. 
Our protocol deploys  the importance weight regularization in the  online setting by deciding when, how and on what data to perform importance weight updates.
Its effectiveness is validated successfully for both supervised and self-supervised learning. More specifically, by using our continual learning method, we demonstrate an improvement of stability and performance over the studied baselines in applications like learning face identities from watching T.V. series and robotic collision avoidance.

\cleardoublepage
  \graphicspath{{\chaptersdir/Rehearsal_OCL/image/}}%
  \chapter{Replay based Online Continual Learning}\label{ch:reh_ocl}
An ideal continual learning agent learns online with a non-stationary and never-ending stream of data. Regularization based methods tend to suffer from gradual forgetting as no historical samples are re-examined. In general, replay based methods usually store a buffer from the previous data for the purpose of rehearsal.  Rehearsal of historical samples allows refreshing old memories in long and largely shifting training sequences. Previous works often depend on task boundary and i.i.d. assumptions to properly select samples for the replay buffer. In this work, aiming for a pure online continual learning, we formulate sample selection as a constraint reduction problem based on the constrained optimization view of continual learning. The goal is to select a fixed subset of constraints that best approximate the feasible region defined by the original constraints. We show that it is equivalent to maximizing the diversity of samples in the replay buffer with parameters gradient as the feature. We further develop a greedy alternative that is cheap and efficient. The advantage of the proposed method is demonstrated by comparing to  alternatives under the online continual learning setting. Further comparisons are made against state of the art methods that rely on task boundaries,  showing comparable or even better results for our method.

This work was developed during a research stay at Montreal Institute for Learning Algorithms, MILA. This work  was published as an article in  NeurIPS 2019~\cite{Aljundi2019continualNoT}.
\section{Introduction}
\label{Introduction}
Online continual learning represents a more challenging yet more realistic setting than the task incremental  setting.  The difficulty comes from the non i.i.d.'ness of the data stream and the unannounced distribution shifts, i.e. task boundaries. 
Most of the works in the major families of continual learning methods use the relaxed task incremental assumption: the data are streamed one task at a time, with different distributions for each task, while keeping the independent and identically distributed (i.i.d.) assumption and performing offline training within each task. Consequently, they are not directly applicable to the more general setting where data are streamed online with neither i.i.d. assumption nor task boundary information. The parameter isolation family explicitly associates neurons with different tasks, with the consequence that task boundaries are mandatory during both training and testing. Due to the dependency on task boundaries during test, this family of methods tilts more towards multi-task learning than continual learning. Both prior-focused  and replay-based methods rely on task boundaries to update the prior or to select a representative set of samples but they have the potential to be adapted to the general setting.

Indeed, in the previous chapter we developed a protocol for deploying our  importance weight regularizer, MAS, in the general online continual learning setting. We, however, were aware that a regularization method alone might fail in situations where large distribution shifts are encountered as in incremental classification. 
This is mainly because the  penalty term is a regularizer  rather than a hard constraint. The parameter gradually drifts away from the feasible regions of previous tasks, when there is a never ending stream of data. Additionally, the learned decision boundaries evolve during the learning of new classes and optimal parameters for previous classes become outdated.  Therefore,  it is often necessary to hybridize the prior-focused approach with the replay-based methods for better results \cite{nguyen2017variational,farquhar2018towards}. In the previous chapter, we added a  small hard buffer to soften the strict online learning scheme but it was rather a very small buffer resembling a short  term memory.

The replay-based approach stores the information in the example space either directly in a replay buffer or in a generative model. When learning new data, old examples are reproduced from the replay buffer or generative model, which is used for rehearsal/retraining or used as constraints for the current learning. 

In this chapter, we develop strategies to populate the replay buffer under the most general condition where no assumptions are made about the online data stream.

Our contributions are as follows: 
\begin{itemize}
    \item We formulate replay buffer population as a constraint selection problem and formalize it as a solid angle minimization problem.
    \item We propose a surrogate objective for it and empirically verify that the surrogate objective aligns with the goal of solid angle minimization.
    \item As a cheap alternative for large sample selection, we propose a greedy algorithm that is as efficient as reservoir sampling yet immune to an imbalanced data stream.
    \item  We compare our method to different selection strategies and show the ability of our solutions to always select a subset of samples that best represents the previous history.
    \item We perform experiments on continual learning benchmarks and show that our method is on par with, or better than, the previous methods, yet, requiring no i.i.d. assumptions or task boundaries.

\end{itemize}

In the remainder of this chapter, we discuss the closely related works in Section~\ref{sec:rehonl-relatedwork}. We present the constraint optimization view of continual learning in Section~\ref{sec:rehonl-constrainedopt} and propose our constraint based sample selection method in 
Section~\ref{sec:rehonl-constraint_selection_without_task_boundaries}. In Section ~\ref{sec:rehonl-onlineselection}, we show how to deploy the sample selection in the online setting and present an in-exact greedy alternative for cheap and fast 
selection. Section~\ref{sec:rehonl-exp}  studies our sample selection methods in different online continual learning scenarios and prove the effectiveness of our approach. Finally, Section~\ref{sec:rehonl-conclusion} concludes the chapter.

\section{Related Work}\label{sec:rehonl-relatedwork}

Recent works that use a replay buffer for continual learning include iCaRL \cite{rebuffi2016icarl} and GEM \cite{lopez2017gradient}, both of which allocate memory to store a core-set of examples from each task. These methods still require task boundaries in order to divide the storage resource evenly to each task. There are also a few previous works that deal with the situation where task boundary and i.i.d. assumption is not available. For example, reservoir sampling has been employed in \cite{chaudhry2019continual,isele2018selective} so that the data distribution in the replay buffer follows the data distribution that has already been seen. The problem of reservoir sampling is that the minor modes in the distribution with small probability mass may fail to be represented in the replay buffer. As a remedy to this problem, coverage maximization is also proposed in \cite{isele2018selective}. It intends to keep diverse samples in the replay buffer using Euclidean distance as a difference measure.
While the Euclidean distance may be enough for low dimensional data, it could be uninformative when the data lies in a structured manifold embedded in a high dimensional space. In contrast to previous works, we start from the constrained optimization formulation of continual learning, and show that the data selection for the replay buffer is effectively a constraint reduction problem.

\section{Continual Learning as Constrained Optimization}\label{sec:rehonl-constrainedopt}

We consider the supervised learning problem with an online stream of data where one or a few pairs of examples $(x,y)$ are received at a time. The data stream is non-stationary with no assumption on the distribution such as the i.i.d. hypothesis. Our goal is to optimize the loss on the current example(s) without increasing the losses on the previously learned examples.
\subsection{Problem Formulation}
We formulate our goal as the following constrained optimization problem. Without loss of generality, we assume the examples are observed one at a time.
\begin{align}
\label{eqn:rehonl-objective_ocl}
\theta^t=&\>\underset{\theta}{\text{argmin}}\>\ell(f(x_t;\theta), y_t) \\
\text{s.t.} \>\ell(f(x_i;\theta), y_i) & \leq \ell(f(x_i;\theta^{t-1}), y_i);\>\forall{i}\in{[0\twodots{t-1}]} \nonumber
\end{align}

$f(.;\theta)$ is a model parameterized by $\theta$ and $\ell$ is the loss function. $t$ is the index of the current example and $i$ indexes the previous examples.

As suggested by Gradient episodic memory (GEM)~\cite{lopez2017gradient}, the original constraints can be rephrased to the constraints in the gradient space:
\begin{align}
\label{eqn:rehonl-constraint_grad_space}
    \langle{g,g_i}\rangle&=\left\langle{\frac{\partial\ell(f(x_t;\theta),y_t)}{\partial\theta},\frac{\partial\ell(f(x_i;\theta),y_i)}{\partial\theta}}\right\rangle\geq{0};
\end{align}
However, the number of constraints in the above optimization problem increases linearly with the number of previous examples. The required computation and storage resource for an exact solution of the above problem will increase indefinitely with time. It is thus more desirable to solve the above problem approximately with a fixed computation and storage budget. In practice, a \textit{replay buffer} $\mathcal{M}$ limited to $M$ memory slots is often used to keep the previous examples. The constraints are thus only active for $(x_i,y_i)\in{\mathcal{M}}$. How to populate the replay buffer then becomes a crucial research problem.

GEM~\cite{lopez2017gradient} assumes access to task boundaries and an i.i.d. distribution within each task episode. It divides the memory budget evenly among the tasks, i.e. $m=M/T$ slots are allocated for each task, where $T$ is the number of tasks. The last $m$ examples from each task are kept in the memory. This has clear limitations when the task boundaries are not available or when the i.i.d. assumption is not satisfied. In this work, we consider the problem of how to populate the replay buffer in a more general setting where the above assumptions are not met.

\subsection{Sample Selection as Constraint Reduction}
\label{sec:rehonl-constraint_selection_without_task_boundaries}
Motivated by Equation~\ref{eqn:rehonl-objective_ocl}, we set our goal to selecting $M$ examples so that the feasible region formed by the corresponding reduced constraints is close to the feasible region of the original problem. We first convert the original constraints in \ref{eqn:rehonl-constraint_grad_space} to the corresponding feasible region:
\begin{equation}
\label{eqn:rehonl-feasible}
C=\bigcap_{i\in{[0\twodots{}t-1]}}\{g|\langle{g,g_i}\rangle\geq{0}\}
\end{equation}

Geometrically, $C$ is the intersection of the half spaces described by $\langle{g,g_i}\rangle\geq{0}$, which forms a polyhedral convex cone. The relaxed feasible region corresponding to the replay buffer is:

\begin{equation}
\label{eqn:rehonl-relaxed_feasible}
\Tilde{C}=\bigcap_{g_i\in{\mathcal{M}}}\{g|\langle{g,g_i}\rangle\geq{0}\}   
\end{equation}

\begin{wrapfigure}{r}{0.35\textwidth}
\vspace*{-0.8cm}
\begin{center}
\includegraphics[width=0.35\textwidth]{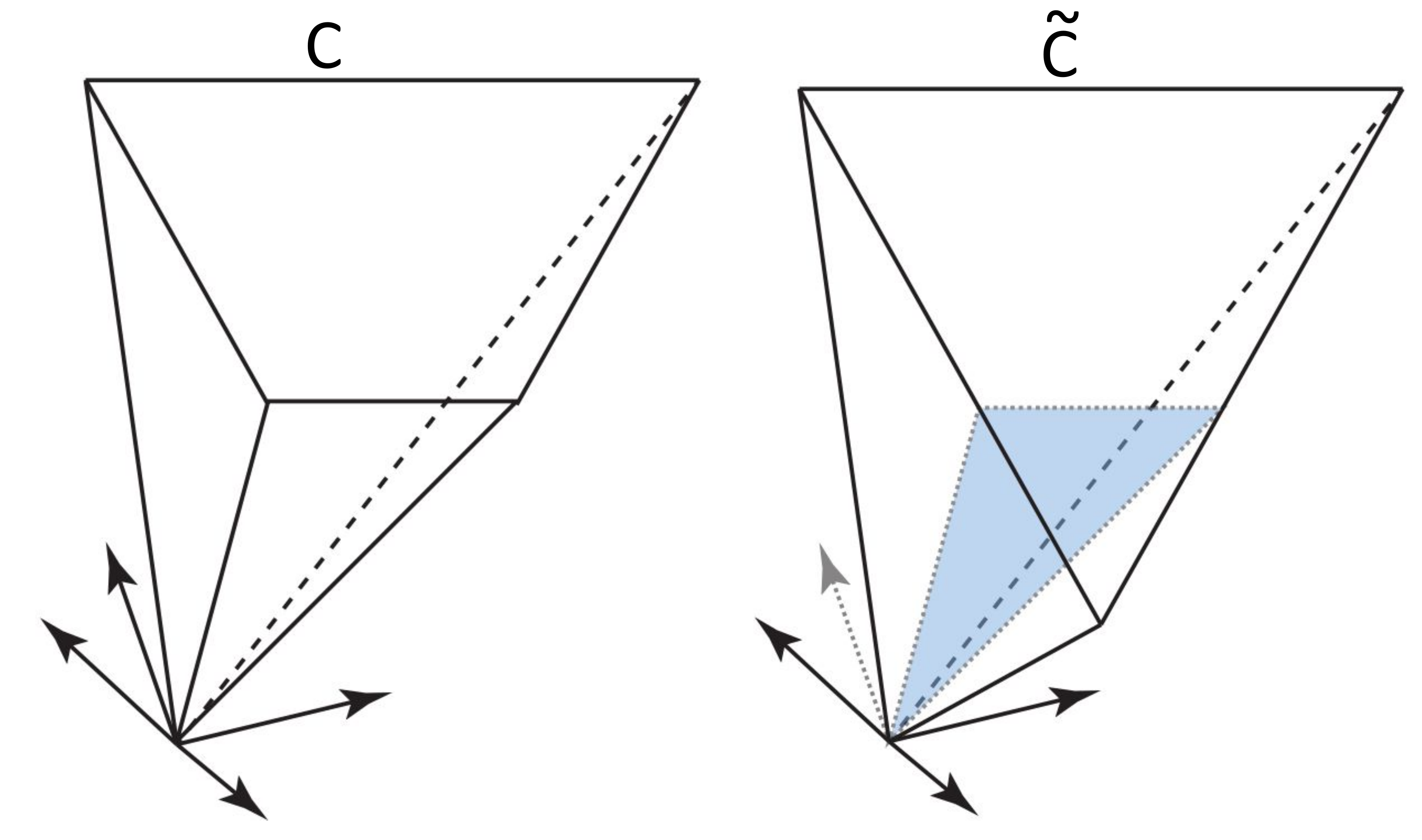}
\end{center}
\vspace*{-0.5cm}
\caption[Feasible region (polyhedral cone) before and after constraint selection.]{  \label{fig:rehonl-constraint_selection} Feasible region (polyhedral cone) before and after constraint selection. The selected constraints (excluding the blue one) are chosen to best approximate the original feasible region.}
\vspace*{-0.8cm}
\end{wrapfigure}

For best approximation of the original feasible region, we require $\Tilde{C}$ to be as close to $C$ as possible. It is easy to see that $C\subset{\Tilde{C}}$ because $\mathcal{M}\subset{[g_0\twodots g_{t-1}]}$. We illustrate the relation between $C$ and $\Tilde{C}$ in Figure \ref{fig:rehonl-constraint_selection}.

On the left, $C$ is represented while the blue hyperplane on the right corresponds to a constraint that has been removed. Therefore, $\Tilde{C}$ (on the right) is larger than $C$ for the inclusion partial order. As we want $\Tilde{C}$ to be as close to $C$ as possible, we actually want the 
``smallest'' $\Tilde{C}$, where ``small'' here remains to be defined, as the inclusion order is not a complete order. A good measure of the size of a convex cone is its solid angle defined as the intersection between the cone and the unit sphere.
\begin{equation}
    \label{eqn:rehonl-min_full_angle}
    \mathrm{minimize}_{\mathcal{M}}\> \lambda_{d-1}\left(\mathcal{S}_{d-1} \cap \bigcap_{g_i\in{\mathcal{M}}}\{g|\langle{g,g_i}\rangle\geq{0}\}\right)
\end{equation}

where $d$ denotes the dimension of the space, $\mathcal{S}_{d-1}$ the unit sphere in this space, and $\lambda_{d-1}$ the Lebesgue measure~\footnote{Lebesgue measure is a standard way of assigning a measure to subsets of n-dimensional Euclidean space extending the notions of length for intervals, area or volume in higher dimensional space of elementary geometrical sets, such as cubes or rectangles, to more complex sets~\cite{ wiki:Lebesgue}} in dimension $d-1$.
Therefore, solving \ref{eqn:rehonl-min_full_angle} would achieve our goal.
Note that, in practice, the number of constraints and thus the number of gradients is likely to be way smaller than the dimension of the gradient, which means that the feasible space can be seen as the Cartesian product between its own intersection with $span\left(\mathcal{M}\right)$ and the orthogonal subspace of $span\left(\mathcal{M}\right)$.
That being said, we can actually reduce our interest to the size of the solid angle in the $M$-dimensional space $span\left(\mathcal{M}\right)$, as in \ref{eqn:rehonl-min_reduced_angle}.
\begin{equation}
    \label{eqn:rehonl-min_reduced_angle}
    \mathrm{minimize}_{\mathcal{M}}\> \lambda_{M-1}\left(S_{M-1}^{span{\left(\mathcal{M}\right)}} \cap \bigcap_{g_i\in{\mathcal{M}}}\{g|\langle{g,g_i}\rangle\geq{0}\}\right)
\end{equation}
where $S_{M-1}^{span{\left(\mathcal{M}\right)}}$ denotes the unit sphere in $span{\left(\mathcal{M}\right)}$.
Note that even if the sub-spaces $span{\left(\mathcal{M}\right)}$ are different from each other, they all have the same dimension as $M$, which is fixed, hence comparing their $\lambda_{M}$-measure makes sense.
However, this objective is hard to minimize since the formula of the solid angle is complex, as shown in \cite{ribando2006measuring} and \cite{beck2009positivity}. Therefore, we propose, in the next section, a surrogate to this objective that is easier to deal with.

\subsection{An Empirical Surrogate to Feasible Region Minimization}\label{sec:rehonl-surrogate}

\begin{figure}[t]

    	\begin{minipage}[b]{0.4\textwidth}
    		\centering
    		  {\includegraphics[width=1\textwidth]{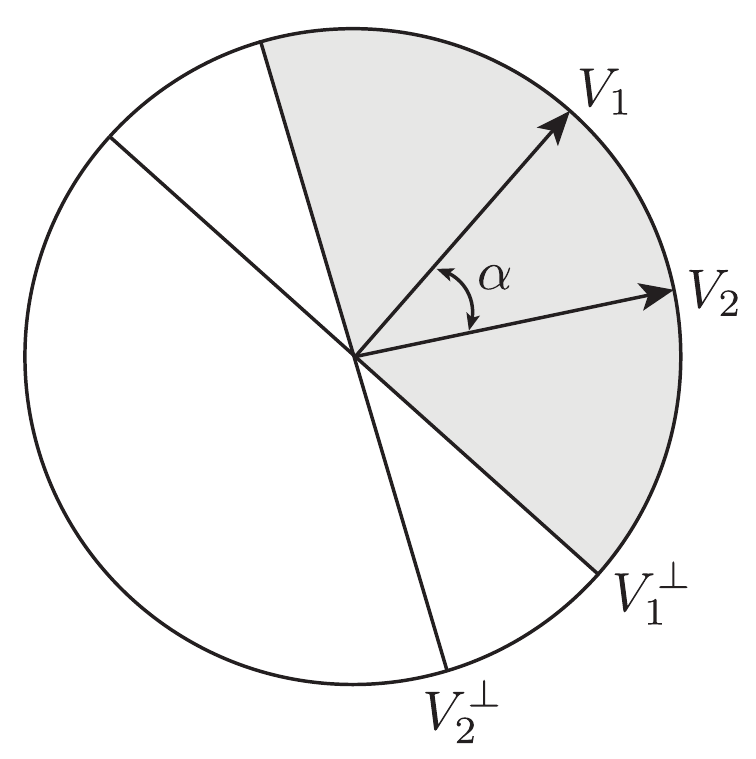}}
    		\caption{ \label{fig:rehonl-dual_angle} Relation between angle formed by two vectors ($\alpha$) and the associated feasible set (grey region)}
    	\end{minipage}
    	 \hfillx
    	    	\begin{minipage}[b]{0.4\textwidth}
    		\centering
    		{\includegraphics[width=1\textwidth]{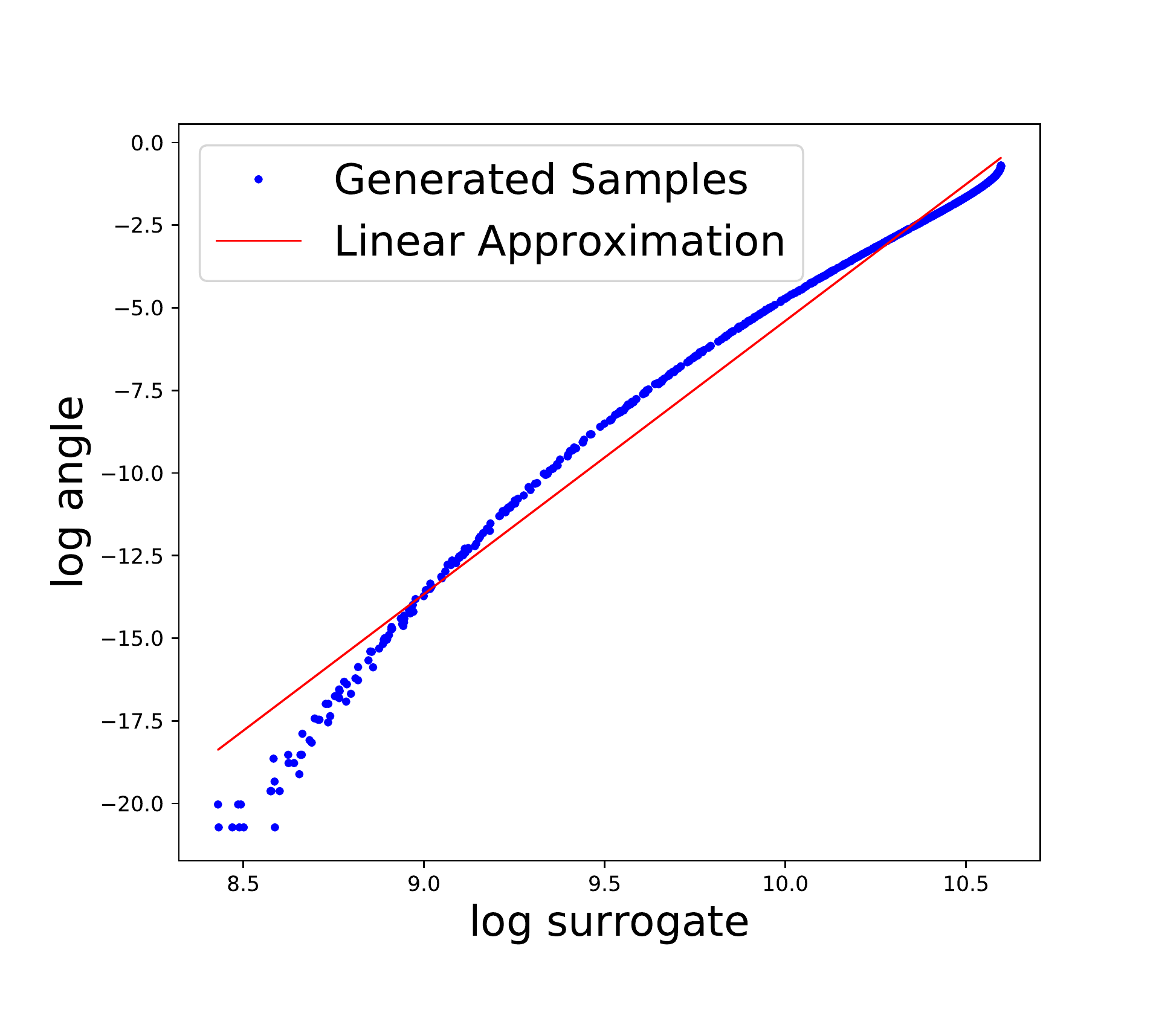}}
    	 \caption[Correlation between solid angle and our proposed surrogate in 200D log scale. ]{ \label{fig:rehonl-correlation_data} Correlation between solid angle and our proposed surrogate in 200D log scale. Note that we only need monotonicity for our objective to hold.  }
    	\end{minipage}
    \end{figure}

Intuitively, to decrease the feasible set, one must increase the angles between each pair of gradients.
Indeed, this is exact in 2D as shown in Figure \ref{fig:rehonl-dual_angle}.
Based on this observation, we propose the surrogate in Equation~\ref{eqn:rehonl-surrogate}. 
\begin{align}
\label{eqn:rehonl-surrogate}
\mathrm{minimize}_{\mathcal{M}}\>\sum_{i,j\in{\mathcal{M}}}\frac{\langle{g_i,g_j}\rangle}{\|g_i\|\|g_j\|} \\
s.t.\>\mathcal{M}\subset{[0\twodots{}t-1]};\>|\mathcal{M}|=M \nonumber
\end{align}

We empirically studied the relationship between the solid angle and the surrogate function in higher dimensional space using randomly sampled vectors as the gradients. 

We faced the problem of no tractable formula for the solid angle, so we estimated it with a sampling method.
The estimated angle is an average of Bernoulli random variables with variance then bounded by $\frac{1}{4N}$ where $N$ is the number of these Bernoulli variables.
By taking $N = 10^9$, we reach an asymptotic confidence interval of length around $10^{-4}$.

Given a set of sampled vectors, the surrogate value is computed analytically and the solid angle is estimated using Monte Carlo approximation of Equation~\ref{eqn:rehonl-min_reduced_angle}.
The results are presented in Figure \ref{fig:rehonl-correlation_data}, which shows a monotonic relation between the solid angle and our surrogate.

\subsection{Keeping Diverse Samples in the Buffer}
Intuitively, keeping diverse samples in the replay buffer could be an efficient way to use the memory budget. It is worth noting that minimizing Equation~\ref{eqn:rehonl-surrogate} is equivalent to maximizing  the variance of the gradient direction as is shown in Equation~\ref{eqn:rehonl-varianceInterpretation}.

\begin{align}
\label{eqn:rehonl-varianceInterpretation}
\mathrm{Var}_{\mathcal{M}}\left[\frac{g}{\|g\|}\right]  =& \frac{1}{M} \sum_{k\in{\mathcal{M}}} \left|\left| \frac{g}{\|g\|} \right|\right|^2 -\left|\left| \frac{1}{M} \sum_{k\in{\mathcal{M}}} \frac{g}{\|g\|} \right|\right|^2 \\
=&  1  - \frac{1}{M^2} \sum_{i,j\in{\mathcal{M}}}\frac{\langle{g_i,g_j}\rangle}{\|g_i\|\|g_j\|}\nonumber
\end{align}

This brings up a new interpretation of the surrogate, which is maximizing the diversity of samples in the replay buffer using the parameter gradient as the feature. Note how this is different from maximizing the variance directly on the samples or on the hidden representations, we argue that the parameter gradient could be a better option given its root in Equation \ref{eqn:rehonl-objective_ocl}. This is also verified with experiments, Section ~\ref{sec:rehonl-sampleselectioncomparison}.

\subsection{Online Sample Selection}\label{sec:rehonl-onlineselection}
\subsubsection{Online sample selection with Integer Quadratic Programming (IQP).}
We assume an infinite input stream of data where at each time a new sample(s) is received. From this stream we keep a fixed buffer of size $M$ to be used as a representative of the previous samples. To reduce the computational burden, we use a ``recent'' buffer in which we store the incoming examples. Once this is full, we perform selection on the union of the replay buffer and the ``recent'' buffer and replace the samples in the replay buffer with the selection. To perform the selection of $M$  samples, we solve Equation~\ref{eqn:rehonl-surrogate} as an integer quadratic programming problem.

We first normalize the gradients: $G=\frac{\langle{g_i,g_j}\rangle}{\|g_i\|\|g_j\|}$ and find a selection vector $S$ that minimizes the following:
\begin{align*}
\underset{S}{\text{minimize}}&\quad \frac{1}{2} S^TGS \\
s.t.\quad& \mathbf{1}^T.S=M\\
& s_m \in \{0,1\}\quad \forall  s_m  \in  S
\end{align*}

where  $\mathbf{1}$ is a ones vector of the same size as $S$. Selected samples will correspond to values of 1 in $S$.
The exact procedures of our IQP based online sample selection are described in Algorithm \ref{algo}. While this is reasonable in the case of a small buffer size, we observed a big overhead when utilizing larger buffers which is likely the case in practical scenarios. The overhead comes from both i) the need to estimate the gradient of each sample in the buffer and the recent buffer and ii) from solving the quadratic problem that is polynomial w.r.t. the size of the buffer. Since this might limit the scalability of our approach, we suggest an alternative greedy method. 
 
\subsubsection{An in-exact greedy alternative.}
We propose an alternative greedy method based on a heuristic, which could achieve the same goal of keeping diverse examples in the replay buffer, but is much cheaper than performing integer quadratic programming. The key idea is to maintain a score for each sample in the replay buffer. The score is computed by the maximal cosine similarity of the current sample with a fixed number of other random samples in the buffer. When there are two samples similar to each other in the buffer, their scores are more likely to be larger than the others. 
In the beginning when the buffer is not full, we add incoming samples along with their score to the replay buffer. Once the buffer is full, we randomly select samples from the replay buffer as the candidate to be replaced. We use the normalized score as the probability of this selection. The score of the candidate is then compared to the score of the new sample to determine whether the replacement should happen or not.

 More formally, denote the score as $\mathcal{C}_i$ for sample $i$ in the buffer. Sample $i$ is selected as a candidate to be replaced with probability $P(i)=\mathcal{C}_i/\sum_j{\mathcal{C}_j}$. The replacement is a Bernoulli event that happens with probability $\mathcal{C}_i/(c+\mathcal{C}_i)$ where $\mathcal{C}_i$ is the score of the candidate and $c$ is the score of the new data. We can apply the same procedure for each example when a batch of new data is received.
Algorithm~\ref{algo_greedy} describes the main steps of our gradient based greedy sample selection procedure.

  \adjustbox{valign=t*}{

  \begin{minipage}{\textwidth}
 
    \begin{minipage}[b]{0.48\textwidth}
\begin{algorithm}[H]
\footnotesize
\caption{IQP Sample Selection}\label{algo}
  \begin{algorithmic}[1]
  \State Input:   $M_r$, $M_b$
  
  \Function{SelectSamples}{$\mathcal{M}$, $M$}
  \State $\hat{\mathcal{M}}\leftarrow\mathrm{argmin}_{\hat{\mathcal{M}}}\>\sum_{i,j\in{\hat{\mathcal{M}}}}\frac{\langle{g_i,g_j}\rangle}{\|g_i\|\|g_j\|}$
  \State s.t. $\hat{\mathcal{M}}\subset{\mathcal{M}};\>|\hat{\mathcal{M}}|=M$
  \State \textbf{return} $\hat{\mathcal{M}}$
  \EndFunction
  \State Initialize: $\mathcal{M}_r$, $\mathcal{M}_b$
  \State Receive: $(x,y)$ \Comment{one or few consecutive examples}
  \State $\theta \leftarrow$ Update($x$, $y$, $\mathcal{M}_b;\theta$)
  \State $\mathcal{M}_r\leftarrow\mathcal{M}_r\cup\{(x,y)\}$
  \If {len($\mathcal{M}_r$) $>M_r$} 
  \State $\mathcal{M}_b\leftarrow\mathcal{M}_b \cup \mathcal{M}_r$
   \State $\mathcal{M}_r\leftarrow \{ \}$
  \If {len($\mathcal{M}_b$) $>M_b$} 
    \State $\mathcal{M}_b\leftarrow$SelectSamples($\mathcal{M}_b$,$M_b$)
    
  \EndIf
\EndIf
\end{algorithmic}
\end{algorithm}
\end{minipage}
    \hfillx
    \begin{minipage}[b]{0.5\textwidth}
\begin{algorithm}[H]
\footnotesize
\caption{Greedy Sample Selection}\label{algo_greedy}
  \begin{algorithmic}[1]
  \State Input:  $n$,  $M$ \Comment{Number of selected random samples to compare against, buffer size. }
  \State Initialize: $\mathcal{M}$, $\mathcal{C}$
  \State Receive: $(x, y)$
   \State $\theta \leftarrow $Update($x$, $y$, $\mathcal{M};\theta$)
  \State $X, Y\leftarrow$ RandomSubset($\mathcal{M}$, n)
  \State $g\leftarrow\nabla{\ell_\theta(x, y)}$; $G\leftarrow\nabla_\theta\ell(X,Y)$
  \State $c=\mathrm{max}_i(\frac{\langle{g,G_i}\rangle}{\|g\|\|G_i\|})+1$ \Comment{make the score positive}
  \If {len($\mathcal{M}$) $>=M$}
  \If {$c<1$} \Comment{cosine similarity < 0}
  \State $i\sim{P(i)=\mathcal{C}_i/\sum_j{\mathcal{C}_j}}$
  \State $r\sim{\mathrm{uniform}(0,1)}$
   \If {$r<\mathcal{C}_i/(\mathcal{C}_i+c)$}
    \State $\mathcal{M}_i\leftarrow(x,y)$; $\mathcal{C}_i\leftarrow{c}$
    \EndIf
\EndIf
\Else
\State   $\mathcal{M}\leftarrow\mathcal{M}\cup\{(x,y)\}$; $\mathcal{C}\cup\{c\}$
  \EndIf

\end{algorithmic}
\end{algorithm}
  \end{minipage}
 \end{minipage}}
 
 \vspace{1cm}
 
 \subsection{Constraint vs Regularization}
 Projecting the gradient of the new sample(s) exactly into the feasible region is computationally very expensive especially when using a large buffer. A usual work around for constrained optimization is to convert the constraints to a soft regularization loss. In our case, this is equivalent to performing rehearsal on the buffer.  
 Note that~\cite{chaudhry2018efficient} suggests to constrain only with one random gradient direction from the buffer as a cheap alternative that works equally well to constraining with the gradients of the  previous tasks. It was later shown by the same authors~\cite{chaudhry2019continual} that rehearsal on the buffer achieves a  competitive performance. In our method, we do rehearsal  while  we evaluate both rehearsal and constrained optimization on a small subset of disjoint MNIST  and show comparable results.
 
\section{Experiments}\label{sec:rehonl-exp}
This section serves to validate our approach and show its effectiveness in dealing with continual learning problems where task boundaries are not available.

\myparagraph{Datasets}
\label{sec:rehonl-datasets}

\textbf{Disjoint MNIST}:  MNIST dataset  divided into 5 tasks based on the labels with  two labels in each task. 
We use 1k examples per task for training and report results on all test examples.

\textbf{Permuted MNIST}:  We perform 10 unique permutations on the pixels of the MNIST images. The permutations result in 10 different tasks with same distributions of labels but different distributions of the input images. 
Following \cite{lopez2017gradient}, each of the tasks in permuted MNIST contains only 1k training examples. The test set for this dataset is the union of the MNIST test set with all different permutations.

\textbf{Disjoint Cifar-10}: Similar to disjoint MNIST, the dataset is split into 5 tasks according to the labels, with two labels in each task. As this is harder than MNIST, we use a total of 10k training examples with 2k examples per task.

In all experiments, we use a fixed batch size of 10 samples and perform few iterations over a batch (1-5). Note that this is different from multiple epochs over the whole data. For disjoint MNIST, we report results using different buffer sizes in table~\ref{tab:rehonl-mnistsplit_selection}. For permuted MNIST results are reported using buffer size 300 while for disjoint Cifar-10 we couldn't get sensible performance for the studied methods with buffer size smaller than 1k. All results are averaged over 3 different random seeds.\\

\myparagraph{Models}

Following~\cite{lopez2017gradient}, for disjoint and permuted MNIST we use a two-layer neural network with 100 neurons each while for Cifar-10 we use ResNet18.  Note that we employ a shared head in the incremental classification experiments, which is much more challenging than the multi-head used in~\cite{lopez2017gradient}.

\subsection{Comparison with Sample Selection Baselines}~\label{sec:rehonl-sampleselectioncomparison}
We want to study the buffer population in the context of the online continual learning setting when no task information is present and no assumption on the data generating distribution is made. Since most existing works assume knowledge of task boundaries, we decide to deploy 3 baselines along with our two proposed methods. Given a fixed buffer size $M$ we compare the following:

\textbf{Random} ({\tt Rand}):
Whenever a new batch is received, it joins the buffer. When the buffer is full, we randomly select  samples to keep  size $M$ from the new batch and samples already in buffer.

\textbf{Online Clustering}:
A possible way to keep diverse samples in the buffer is online clustering with the goal of selecting a set of $M$ centroids. This can be done either in {the feature space } or in  {the gradient space}. In the feature space we consider a  selection baseline, denoted as Feature based Sample Selection~({\tt FSS-Clust}),  where we use as a metric the distance between the samples features, here the last layer before classification. For the selection baseline in the gradient space, denoted as Gradient based Sample Selection ({\tt GSS-Clust}),   we consider as a metric the Euclidean distance between the normalized gradients. We adapted the doubling algorithm for incremental clustering described in~\cite{charikar2004incremental}.  

\textbf{IQP Gradient based Sample Selection ({\tt GSS-IQP})}:
Our surrogate to select  samples that minimize the feasible region described in Equation~\ref{eqn:rehonl-surrogate} and solved as an  integer quadratic programming problem. Due to the cost of computation we report our {\tt GSS-IQP} on permuted MNIST and disjoint MNIST only.

\textbf{Gradient based Greedy Sample Selection ({\tt GSS-Greedy})}:
Our greedy selection variant detailed in Algorithm~\ref{algo_greedy}. Note that differently from 
previous selection strategies, it doesn't require re-processing all the recent and buffer samples to perform the selection  which is a huge gain in the online streaming setting.

  \begin{table*}[h!]
\centering
\begin{tabular}{ | l |  c | c | c | }
\hline
\diagbox{Method}{Buffer Size} &  300 & 400 & 500  \\  \hline 
{\tt Rand} &37.5  & 45.9  &57.9  \\  \hline
{\tt GSS-IQP(ours)}&75.9 & 82.1& 84.1\\ \hline
{\tt GSS-Clust}&   75.7& 81.4& 83.9  \\ \hline
{\tt FSS-Clust}&  75.8&80.6 &83.4   \\ \hline
{\tt GSS-Greedy(ours)}&  {\bf 82.6}& {\bf 84.6} & {\bf 84.8}  \\ \hline
\end{tabular}
\caption{\label{tab:rehonl-mnistsplit_selection} Average test accuracy in \% of  sample selection methods  on disjoint MNIST with different buffer sizes. }
\end{table*}
\begin{table*}[h!]
\centering
\centering
\tabcolsep=0.11cm
\footnotesize

\resizebox{\textwidth}{!}{
\begin{tabular}{ | l | c| c | c | c |c|c|c|c|c|c|c| }
\hline
Method &  T1 & T2 & T3 & T4  & T5 & T6 & T7 & T8 & T9 & T10 & Avg. \\  \hline 
{\tt Rand} &66.29&58.82&67.9&59.93&71.45&71.31&79.57&80.3&84.3&{\bf 86.9}& 72.7 \\  \hline 
{\tt GSS-IQP(ours)}  &74.1& 69.73 & 70.77&70.5&73.34&78.6&81.8&81.8&{\bf 86.4}&85.45& 77.3  \\ \hline
{\tt GSS-Clust} &75.4&{\bf 76.66} &{\bf 77.89}& {\bf 73.56} &{\bf 80.7}&{\bf 80.83}&{\bf 82.4}&{\bf 83.53}&83.8&84.75& {\bf 79.96}  \\ \hline
{\tt FSS-Clust}  &82.2&71.34&76.9&70.5&70.56&74.9&77.68&79.56&82.7&85.3& 77.8  \\ \hline
{\tt GSS-Greedy(ours)}&{\bf 83.35}&70.84&72.48&70.5&72.8&73.75&79.86&80.45&82.56&84.8& 77.3 \\ \hline
\end{tabular}}
\caption[Comparison of different selection strategies on permuted MNIST benchmark.]{\label{tab:rehonl-permutedmnist_selection}  Comparison of different selection strategies on permuted MNIST benchmark, buffer size 300. Table shows test accuracies in $\%$ on each task at the end of the training sequence.}

\end{table*}

  \begin{table*}[h!]
  \centering
  
\begin{tabular}{ | l | c| c | c | c |c|c| }
\hline
Method &  T1 & T2 & T3 & T4  &T5 & Avg. \\  \hline 
{\tt Rand} & 0 & 0.49 & 5.68 & {\bf 52.18}& 84.96 & 28.6 \\  \hline 
{\tt GSS-Clust}  & 0 &5.91 &15.91& 12.62&78.14&22.5 \\ \hline
{\tt FSS-Clust}  &  0.17 & 0.82&5.42& 38.12&{\bf 87.90}&26.7   \\ \hline
{\tt GSS-Greedy(ours)} & {\bf42.36}&{\bf 14.61}&{\bf 13.60}&{19.30}&77.83&{\bf 33.56}  \\ \hline
\end{tabular}
\caption[Comparison of different selection strategies on disjoint Cifar-10 benchmark.]{\label{tab:rehonl-cifar10split_selection}  Comparison of different selection strategies on disjoint Cifar-10 benchmark, buffer size 1k. Table shows test accuracies  in $\%$ on each task at the end of the training sequence.}
  \end{table*}
\subsection{Performance of Sample Selection Methods}

Tables~\ref{tab:rehonl-mnistsplit_selection},~\ref{tab:rehonl-permutedmnist_selection},~\ref{tab:rehonl-cifar10split_selection} report the test accuracy on each task at the end of the data stream for disjoint MNIST, permuted MNIST, disjoint Cifar-10 respectively. First of all, the accuracies reported in the tables might appear lower than state of the art numbers. This is due to the strict online setting, the use  of a shared head and more importantly the no use of   task boundary. In contrast, all previous works assume availability of task boundary either at training or both at training and testing. The performance of the random baseline {\tt Rand} clearly indicates the difficulty of this setting. It can be seen that both of our selection methods exhibit stable and good performance over the different buffer sizes on different benchmarks. Notably, the gradient based clustering {\tt GSS-Clust} performs comparably and even favorably on permuted MNIST to the feature clustering {\tt FSS-Clust} suggesting the effectiveness of  a gradient based metric in the continual learning setting. Surprisingly, {\tt GSS-Greedy} performs on par and even better than the other selection strategies especially on disjoint Cifar-10 indicating   not only a cheap but a strong sample selection strategy.

\subsection{Performance under Blurry Task Boundary}
 An interesting setting  is the scenario where there are no clear task boundaries in the data stream. As we mentioned in the introduction,  such situation can happen in practice. We start by blurring the task boundaries in disjoint Cifar-10 benchmark. For each task in the dataset, we keep the majority of the examples while we randomly swap a small percentage of the examples with other tasks. A larger swap percentage corresponds to more blurry boundaries. This is a similar setting to what we   used in Chapter~\ref{ch:ssl}, Section~\ref{sec:scl-experiments:noboundaries}. We keep 90\% of the data for each task, and introduce 10\% of data from the other tasks.  We make comparisons to the other studied selection methods. Since tasks are not disjoint, forgetting is not as severe as in the case of complete disjoint tasks. Hence, we use a buffer of 500 samples and train on 1k samples per task which allows us to run our {\tt GSS-IQP} more smoothly.

Table~\ref{tab:rehonl-cifar10_blurry_split_selection} reports the accuracy of each  task at the end of the sequence. Both of our  methods perform better than other selection strategies.
   \begin{table*}[h]
  \centering
\begin{tabular}{ | l | c| c | c | c |c|c| }
\hline
Method &  T1 & T2 & T3 & T4  &T5 & Avg. \\  \hline 
{\tt Rand} &0&3.45&9.85&{\bf 54.67}&78.76&29.0  \\  \hline 
{\tt GSS-IQP(ours)} &9.38&{\bf 11.33}&{\bf 17.05}&{ 30.84} &{\bf 79.53}&{\bf 29.6}  \\ \hline
{\tt GSS-Clust}  &2.43&16.75&9.09&20.71&77.98&25.0 \\ \hline
{\tt FSS-Clust}  &  2.95&05.09&6.06&38.16&78.14&26.0  \\ \hline
{\tt GSS-Greedy(ours)} &{\bf 34.2}&11.14&14.96&20.25&67.5&{\bf 29.6} \\ \hline
\end{tabular}
\caption{\label{tab:rehonl-cifar10_blurry_split_selection}  Comparison of different selection strategies on disjoint Cifar-10 with blurry task boundary, buffer size 500. Table shows test accuracies  in $\%$ on each task at the end of the training sequence.}
  \end{table*}
  
    \begin{table*}[t]
  \centering
\begin{tabular}{ | l | c| c | c | c |c|c| }
\hline
Method &  T1 & T2 & T3 & T4  &T5 & Avg. \\  \hline 
{\tt GSS-IQP(Constrained)} &90.0&70.0&45.13&88.77&86.08&76.26  \\ \hline
{\tt GSS-IQP(Rehearsal)} &81.5&69.47&46.96&69.80&88.0&71.3  \\ \hline
\end{tabular}
  \caption[Comparison between  Rehearsal and Constrained optimization with our GSS-IQP method on disjoint MNIST and  buffer size 100.]{\label{tab:rehonl-mnist_constrained}  Comparison between  Rehearsal and Constrained optimization with our GSS-IQP method on disjoint MNIST and  buffer size 100. Table shows test accuracies in $\%$ on each task at the end of the training sequence.}
  \end{table*}
  
      \begin{table*}[h]
  \centering
\begin{tabular}{ | l | c| c | c | c |c|c| }
\hline
Method &  T1 & T2 & T3 & T4  &T5 & Avg. \\  \hline 
{\tt GSS-IQP(Constrained)} &95.0&83.0&68.7&87.6&82.4&83.4  \\ \hline
{\tt GSS-IQP(Rehearsal)} &94.6&83.89&50.6&77.0&88.67&78.9  \\ \hline
\end{tabular}
  \caption[Comparison between  Rehearsal and Constrained optimization with our GSS-IQP method on disjoint MNIST and  buffer size 200. ]{\label{tab:rehonl-mnist_constrained_b200}  Comparison between  Rehearsal and Constrained optimization with our GSS-IQP method on disjoint MNIST and  buffer size 200. Table shows test accuracies in $\%$ on each task at the end of the training sequence.}
  \end{table*}  
 \subsection{Constrained Optimization Compared to Rehearsal}
 By the end of Section~\ref{sec:rehonl-onlineselection}, we have elaborated on the computational complexity of the constrained optimization, which renders then infeasible with large buffers. That's mainly because at each learning step,  gradient of each sample in the buffer needs to be estimated and then the new sample gradient needs to be projected onto the feasible region determined by all the samples gradients. As an alternative, we perform rehearsal on the buffer.   Here, we want to compare the performance of the two update strategies, the constrained optimization~{\tt GSS-IQP(Constrained)} and the rehearsal~{\tt GSS-IQP(Rehearsal)}. We consider disjoint MNIST benchmark and use 200 training samples per task.
 Table~\ref{tab:rehonl-mnist_constrained} reports the test accuracy on each task achieved by each strategy at the end of the training when using a buffer of size 100 while table~\ref{tab:rehonl-mnist_constrained_b200} reports the accuracy with 200 buffer size.  ~{\tt GSS-IQP(Constrained)} improves over {\tt GSS-IQP(Rehearsal)} with a margin of $3-5\%$ but requires a long time to train as it scales polynomialy with the number of samples in the buffer apart from the need to compute the gradient of each buffer sample at each training step. {\tt GSS-IQP(Rehearsal)} with larger buffer is less computational and yields similar results, comparing  {\tt GSS-IQP(Rehearsal)} with a buffer of size 200 ($78.9\% $) and {\tt GSS-IQP(Constrained)} with a buffer of size 100  $(76.26\%)$.

  \subsection{Comparison with Reservoir Sampling}
Reservoir sampling~\cite{vitter1985random} is a simple replacement strategy to fill the memory buffer when the task boundaries are unknown based on the underlying assumption that the overall data stream is i.i.d. distributed. It would work well when each of the tasks has a similar number of examples. However, it could lose the information on the under-represented tasks if some of the tasks have significantly fewer examples than the others. In this chapter, we study and propose algorithms to sample from an imbalanced stream of data.  Our strategy has no assumption on the data stream distribution, hence it could be less affected by imbalanced data, which is often encountered in practice. 

We test this scenario on  disjoint MNIST. We modify the data stream to settings where one of the tasks has an order of magnitude more samples than the rest, generating 5 different sequences where the  first sequence has $2000$ samples of the first task and $200$ from each other task, the second sequence has $2000$ samples from the second task and $200$ from others, and same strategy applies to the rest of the sequences. Table~\ref{tab:rehonl-reserv_selection} reports the average accuracy at the end of each sequence over 3 runs with 300 samples as buffer size.
It can be clearly seen that our selection strategies outperform reservoir sampling. Our improvement reaches $15\%$. While reservoir sampling has been introduced by~\cite{chaudhry2018efficient} as a cheap powerful strategy to populate the buffer samples, here we show that the effectiveness of such strategy cannot be taken for granted.

Having shown the robustness of our selection strategies both {\tt GSS-IQP} and {\tt GSS-Greedy}, we move now to compare with state of the art replay-based methods that allocate a separate buffer per task and only play samples from previous tasks during the learning of others. 
 \begin{table*}[t]

\footnotesize
\centering
\begin{tabular}{ | l | c| c | c | c |c|c| }
\hline
Method &  Seq1 & Seq2 & Seq3 & Seq4  &Seq5 & Avg. \\  \hline
{\tt GSS-IQP(ours)}&{\bf 75.9}  &  76.2&79.06 &76.6& 74.7& 76.49 \\ \hline
{\tt GSS-Greedy(ours)} &71.2 &{\bf 78.5} &{\bf 81.5} &{\bf 79.5}& {\bf 79.1}&{\bf 77.96}  \\ \hline
{\tt Reservoir} & 63.7 & 69.4 & 66.8 & 69.1 & 76.6 & 69.12 \\  \hline
\end{tabular}
\caption[Comparison with reservoir sampling on different imbalanced data sequences from disjoint MNIST, buffer size 300.]{\label{tab:rehonl-reserv_selection}  Comparison with reservoir sampling on different imbalanced data sequences from disjoint MNIST, buffer size 300. Table shows average test accuracies  in $\%$  on each sequence  achieved at its end.}
\end{table*}

\subsection{Comparison with State of the Art Task Aware Methods}
Our method ignores any tasks information  which places us at a disadvantage because the methods that we compare to utilize the task boundaries as an extra information. In spite of this disadvantage, we show that our method performs similarly on these datasets.
We consider the following methods:

{\tt Online} is a model trained online on the stream of data without any mechanism to prevent forgetting.\\
{\tt Online Joint} is the Online baseline trained on an i.i.d. stream of the data.\\
{\tt Offline Joint} is a model trained offline for multiple epochs with i.i.d. sampled batches and serves as an upper bond.\\
{\tt GEM}~\cite{lopez2017gradient} stores a fixed amount of random examples per task and uses them to provide constraints when learning new examples.\\
{\tt iCaRL}~\cite{rebuffi2016icarl} follows an incremental classification setting. It also stores a fixed number of examples per class but uses them to rehearse the network when learning new information.

For ours, we report both  {\tt GSS-IQP} and  {\tt GSS-Greedy} on permuted and disjoint MNIST and only {\tt GSS-Greedy} on disjoint Cifar-10 due to the computational burden. 
Since  we perform multiple iterations over a given batch still in the online setting, we treat the number of iterations  as a hyper parameter for  {\tt GEM}  and {\tt iCaRL}. We found that {\tt GEM} performance constantly deteriorates with multiple iterations while {\tt iCaRL} improves.
\begin{figure*}[t]
    \centering
    \subfloat[\footnotesize  Disjoint MNIST]{\label{fig:rehonl-eve_plot_mnistsplit_200}\includegraphics[height=0.15\textheight, width=0.43\textwidth]{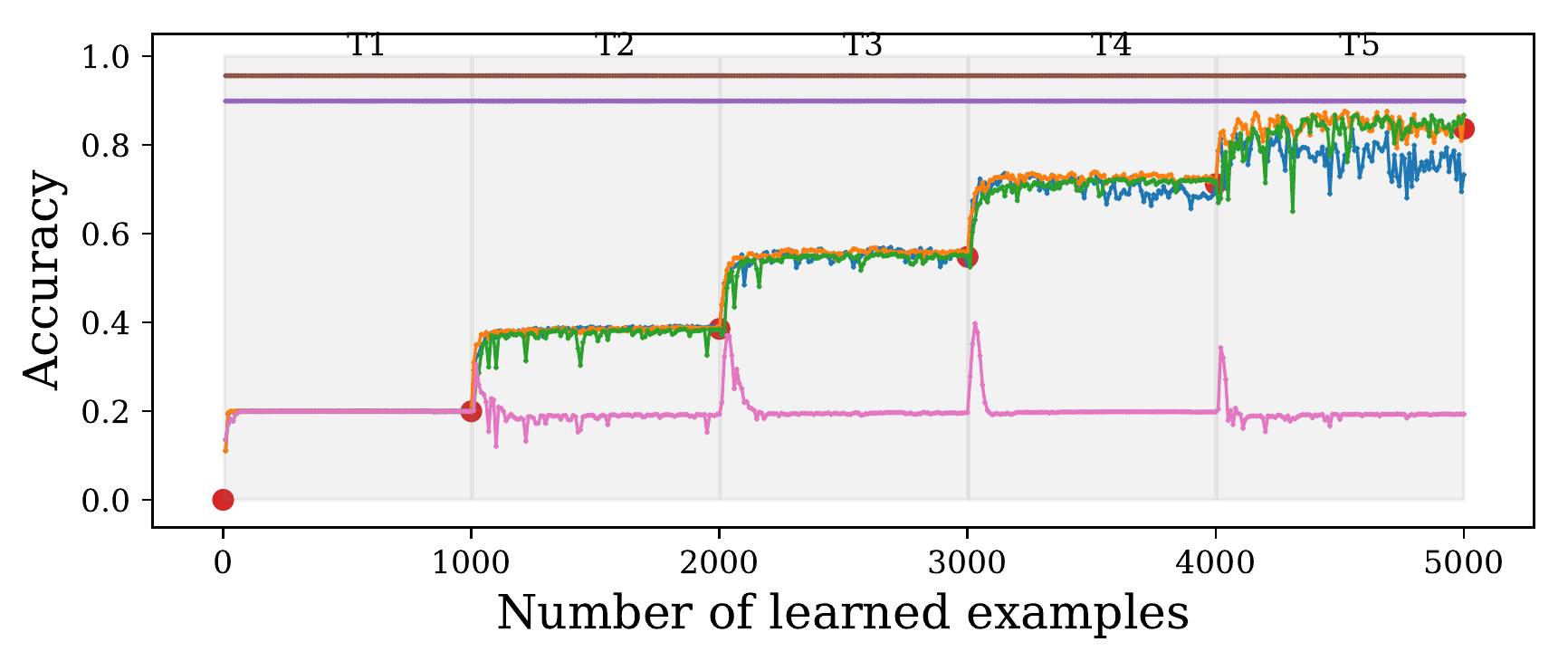}}
    \subfloat[\footnotesize  Permuted MNIST]{\label{fig:rehonl-eve_plot_mnistpermuted_200}\includegraphics[height=0.15\textheight, width=0.45\textwidth]{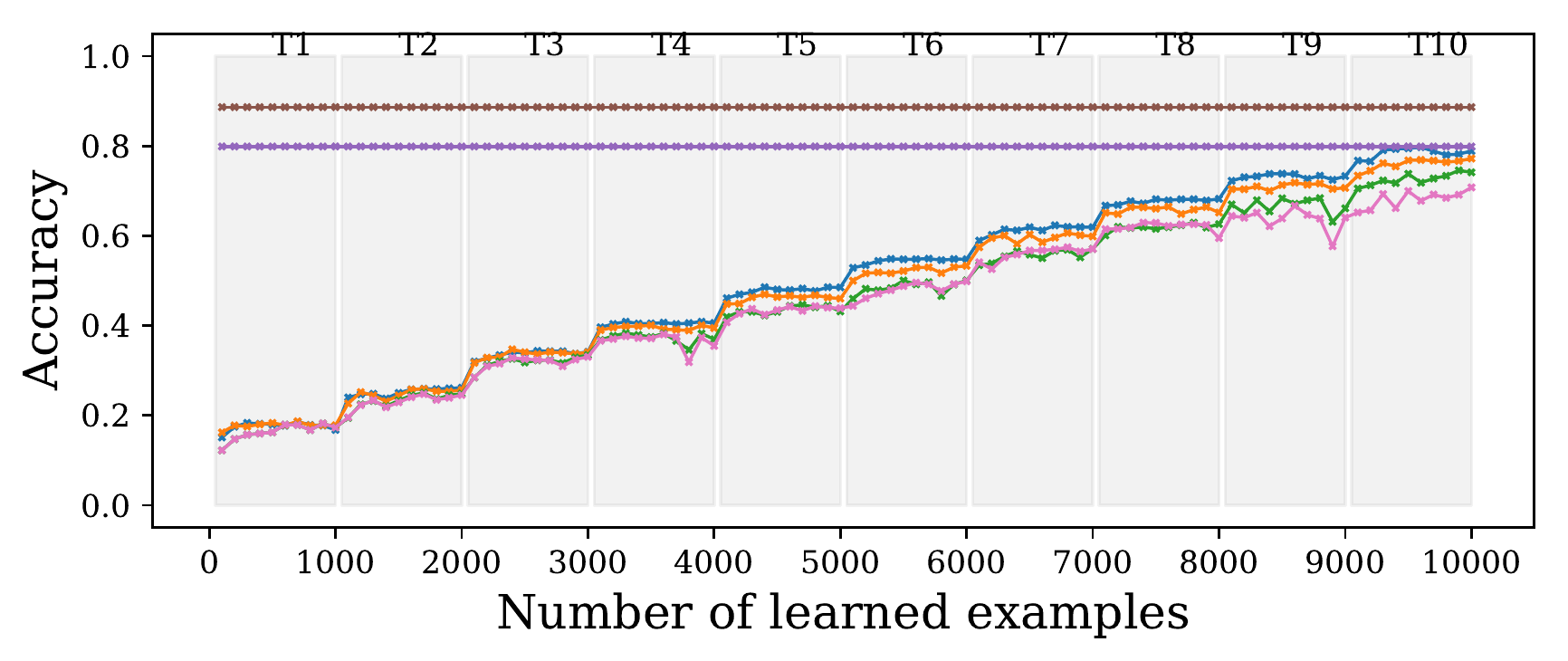}}
    \hfillx
    \subfloat{\includegraphics[height=0.15\textheight, width=0.13\textwidth]{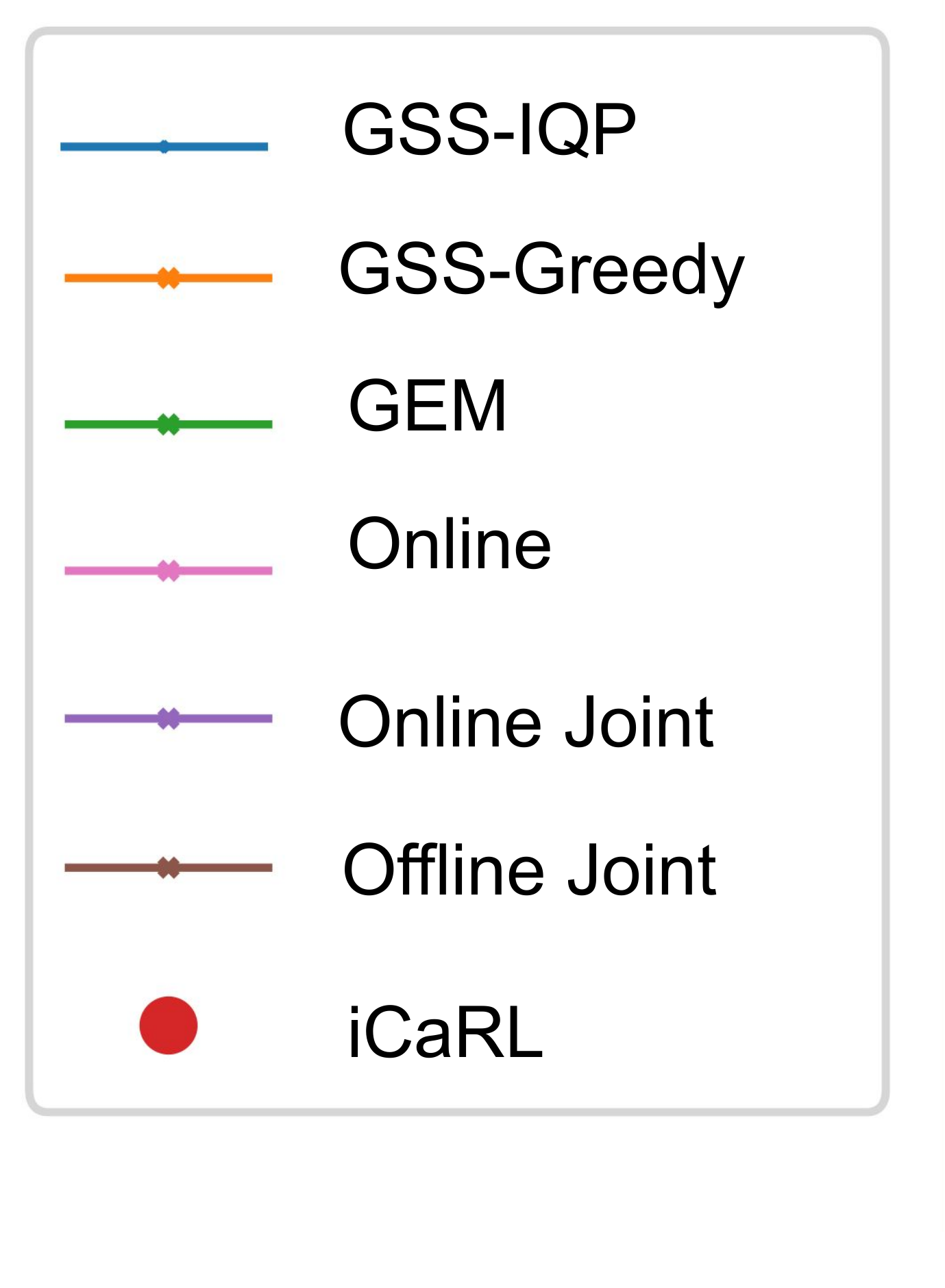}}
    \caption[ Comparison with state of the art task aware replay methods  on disjoint MNIST and permuted MNIST.]{ Comparison with state of the art task aware replay methods  on disjoint MNIST and permuted MNIST, buffer size 300. Figures show test accuracy in $\%$.}
    \label{fig:rehonl-standard_setup}
\end{figure*}
Figure~\ref{fig:rehonl-eve_plot_mnistsplit_200} shows the test accuracy on disjoint MNIST which is evaluated during the training procedure at an interval of 100 training examples with 300 buffer size. 
For the i.i.d. baselines, we only show the achieved performance at the end of the training. For {\tt iCaRL}, we only show the accuracy at the end of each task because {\tt iCaRL} uses the selected exemplars for prediction that only happens at the end of each task. We observe that both variants of our method have a very similar learning curve to GEM except the few last iterations where  {\tt GSS-IQP}  performance  slightly drops.

Figure~\ref{fig:rehonl-eve_plot_mnistpermuted_200} compares our methods with the baselines and {\tt GEM} on the permuted MNIST dataset. Note that {\tt iCaRL} is not included, as it is designed only for incremental classification. From the good performance of the {\tt Online} baseline it is apparent that permuted MNIST has less interference between the different tasks. Our  two variants perform better than GEM and get close  to {\tt Online Joint} performance.

\begin{figure*}[h!]
  \centering
\includegraphics[ width=0.7\textwidth]{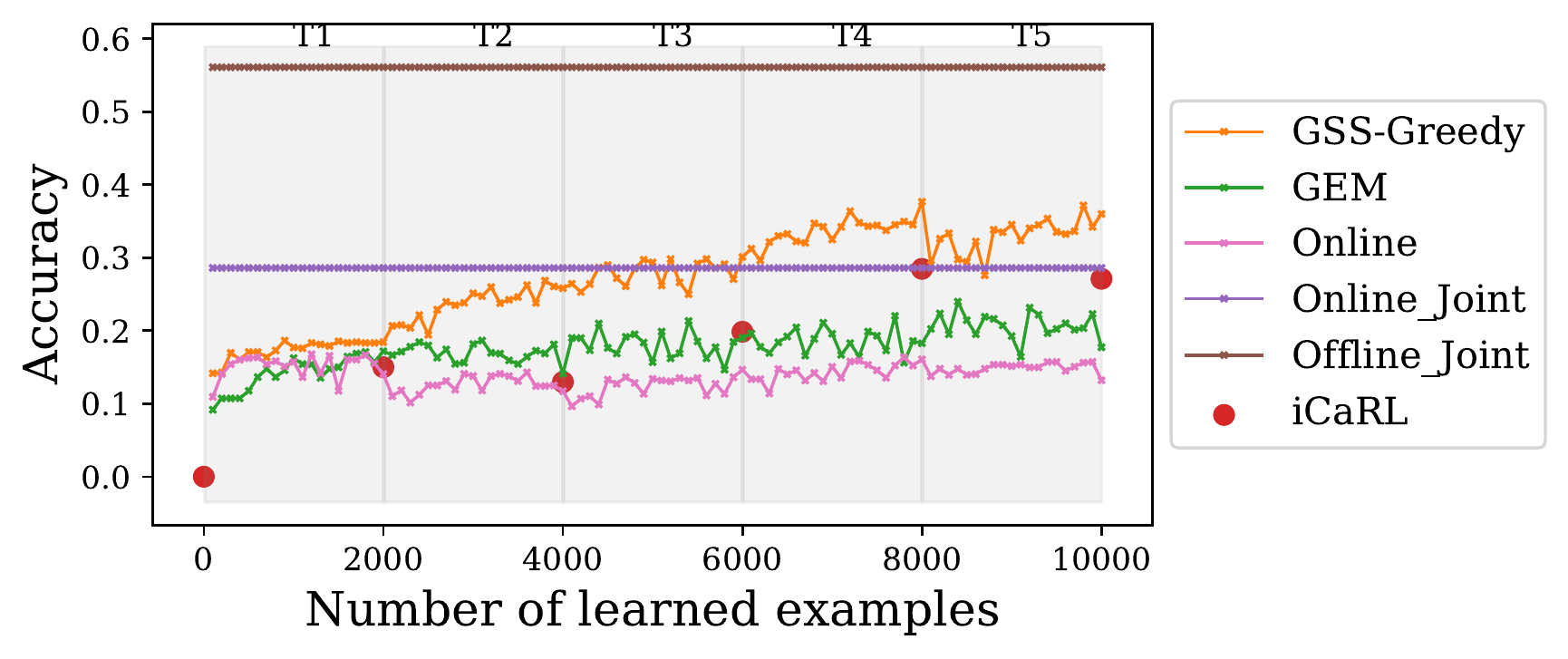}
   \caption[Comparison with state of the art task aware replay methods on disjoint Cifar-10.]{\label{fig:rehonl-eve_plot_cifar}  Comparison with state of the art task aware replay methods on disjoint Cifar-10, buffer size 1k. Figure show test accuracy in $\%$.}
   \end{figure*}

Figure~\ref{fig:rehonl-eve_plot_cifar} shows the accuracy on  disjoint Cifar-10 evaluated during the training procedure at an interval of 100 training examples. %
 {\tt GSS\_Greedy} shows better performance than {\tt GEM} and {\tt iCaRL}, and it even achieves a better average test performance at the end of the sequence than {\tt Online Joint}. We found that {\tt GEM} suffers  more forgetting on  previous tasks while {\tt iCaRL} shows lower performance on the last task. Note that our setting is much harder than offline  tasks training used in iCaRL~\cite{rebuffi2016icarl}.

\section{Summary}\label{sec:rehonl-conclusion}

In this chapter, we studied online continual learning on a never ending stream of data sampled from largely shifting distribution   as  in the case of incremental classification with no task boundaries and shared output layer.
We prove that in the online continual learning setting we can smartly select a finite number of samples to be representative of all previously seen data without knowing task boundaries. We aim for samples diversity in the gradient space and introduce a greedy selection approach that is efficient and constantly outperforming other selection strategies. We still perform as well as algorithms that use the knowledge of task boundaries to select the representative examples. Moreover, our selection strategy gives us advantage under the settings where the task boundaries are blurry or data are imbalanced.

\cleardoublepage

%



%
  \graphicspath{{\chaptersdir/conclusion/image/}}%
  
\chapter{Conclusion}\label{ch:conclusion}
In this chapter, we  first summarize the main contributions of the work done during the course of this PhD. We  then discuss  the advances within the continual learning field in the past few years while shedding light on the impact and limitations of the methods proposed in this thesis. We  conclude this thesis by pointing at some promising research directions towards realizing continual learning. 
 \section{Summary of Contributions}
While the catastrophic forgetting problem has been studied since the early development of neural networks,  continual learning is an emerging field that only recently started to have a well acknowledged definition. During the first year of my PhD, I was working on the topic of domain adaptation and had presented two methods, not described in this manuscript. The first one proposed an on the fly domain adaptation method~\cite{aljundi2016lightweight} and the other was more  of a practical domain adaptation method for actor recognition in T.V. series~\cite{aljundi2016actor}.  Domain adaptation is limited to two tasks/domains and unidirectional. After a period of extensive research, we started working on the concept of incremental  experts learning  on different domains extending transfer learning to  an unlimited sequence of domains. 
The idea  evolved to the work described in Chapter~\ref{ch:expertgate}. By designing a sequentially learned gate, we showed  expert level performance on each domain maintaining a deployment time close  to that of one model, without assuming any access to data from previous domains. Our work, Expert Gate, was among the first works to propose a solution for inferring the task ID at test time given unseen data. This has a great advantage as it overcomes the need for an extra dependence on   human intervention in the continual learning cycle. Most of the subsequent works on avoiding catastrophic forgetting and on continual learning  require an imaginary task oracle at test time. It is worth mentioning that  Expert Gate was successfully deployed in industry on cases with more than  80 domains  learned in a sequence. During the last development stages of Expert Gate, a closely related method~\cite{rusu2016progressive} was proposed by a  team from Deep Mind. It grows a neural network by connecting a new  module for each new task. 
Concurrently, Learning without Forgetting~\cite{li2016learning} (LwF) proposed to deploy the knowledge distillation loss for learning a sequence of tasks using one shared neural network model. In Expert Gate, we have shown that LwF is vulnerable
to forgetting when there is a significant distribution shift between tasks but beneficial when related tasks are being learned. As such we have implemented  LwF as a method for knowledge transfer between domains.

Expert Gate provides  expert level performance at the expense of storing all the previously learned models. However, as explained in Chapter~\ref{ch:ebll}, many scenarios emerge where there is a need to learn the different upcoming tasks using a single model. Starting from the same concept of sequentially learning  light shallow autoencoders, we have proposed a solution for incremental task learning using one shared model. The autoencoders are used here to preserve the important features learned for the previous tasks during the learning of other tasks. In Chapter~\ref{ch:ebll} we have presented our Encoder Based Lifelong Learning work~( EBLL ) and show that it is advantageous over LwF mostly when the learned tasks are not largely related. LwF and EBLL use the data of the new task as a proxy for the previous tasks data, hence they are sensitive to big distribution shifts. 

Another line of methods had emerged~\cite{kirkpatrick2017overcoming,Zenke2017improved}: prior-focused methods that use the knowledge acquired from the previous tasks as a prior when learning the new tasks. An $\ell_2$ regularization is used to penalize the changes on the  important parameters. These methods are problem agnostic and can have a constant memory consumption~\cite{Zenke2017improved} which are very desired criteria in continual learning. However, the important parameters are estimated  on all the training data of a given task and then remained fixed for the rest of an agent lifetime, continual learning methods must be adaptive. In Chapter~\ref{ch:MAS}, we have argued that given an unlimited sequence of tasks and a fixed model capacity; it is impossible to preserve all the previous knowledge while still being able to learn new tasks. Instead of gradually forgetting equally all the previous knowledge, we proposed to learn the important bits of previous knowledge, those that are frequently used at deployment time.  We further prove that Hebbian learning can be seen as a local version of our method. We have shown the ability of our method, MAS, to learn what is important not to forget based on a specific test setting using any existing data. This brings into realization two additional desired characteristics of continual learning,  adaptive and graceful forgetting. Our method also showed  at the time of development, state of the art performance on the standard task incremental  setting.

At the time of developing MAS, our aim was to move the task incremental setting towards a smoother continual learning.  We thus proposed a new benchmark for continual learning. Instead of learning tasks as different as birds and digits, we proposed to learn  facts, <Subject, Object, Predicate> triplets. This eliminates the multi-head setting that hides most of the difficulties of continual learning, since each task has a separate unshared head. This work was published as a conference paper at ACCV2018~\cite{Elhoseiny2018Exploring} but not detailed in this manuscript.

Overcoming catastrophic forgetting in the absence of the previous tasks data and given a fixed model capacity, relies mostly on identifying the important features/parameters   for a previous task and preserving them during the learning process. As we showed in Chapter~\ref{ch:ssl}, an important aspect is that   the optimization process of a given task should be aware of future upcoming tasks to be learned. This plays the role of not utilizing all the model capacity but rather accounting for the future tasks. We have shed the light on the importance of sparsity in continual learning and  studied different regularization criteria and activation functions in the context of continual learning. Inspired by  neural inhibition in the neural biology,  we proposed a sparsity imposing regularizer that shows  larger rates of sparsity  compared to other regularizers. Our  regularizer improved significantly the continual learning performance on different studied datasets.

All the previous works, including ours, focus on the milder task incremental setting that maintains the offline training of each task and relies on knowing the task boundaries to process the acquired knowledge. While catastrophic forgetting seemed to be less of a problem given the new advances, many of the real applications seemed far  fetched under the task incremental setting. In Chapter~\ref{ch:reg_ocl}, we proposed a protocol for deploying our importance weight regularizer, MAS, in the online continual learning setting, removing any dependence on the task boundaries and performing online training on streaming data.   
We have shown improved learning behaviour and less forgetting when learning from streaming data in gradually changing environments such as recognizing actors in TV series or learning to avoid obstacles.

Online learning from a non stationary distribution is particularly challenging. The inferred decision boundaries will change  during the learning process and  previously acquired knowledge might soon become outdated  as new information is processed. Solving continual learning without storing any sample from the previous history  seems extremely hard in settings with largely shifting distributions.
In the last chapter (~\ref{ch:reh_ocl})
we have studied online continual learning using a buffer of historical samples from the previously seen data. We formulate  online continual learning as a constrained optimization problem and propose a gradient based sample selection procedure. In spite of not relying on any task information for selecting the stored samples, we have shown improved performance on different continual learning scenarios. 

Our works on online continual learning were the first to explore the challenging yet more realistic online continual learning setting. Only very recently new research works have started to explore this setting~\cite{he2019task,lomonaco2019continual}. 
 
The samples stored from the previous history are usually replayed randomly while learning new samples. While not explained in this manuscript, we have lately studied replay strategies during online learning and propose a method that searches for what samples would be the most interfered  given an estimated parameters update. This work was joint with researchers from MILA following my research visit. Its  outcome   was accepted as an article in  NEURIPS 2019 ( Conference on Neural Information Processing Systems). 

Finally, during the past last year, I have worked with  colleges from our group on a continual learning survey. We have  proposed two benchmarks for continual learning and evaluated the different existing families of continual learning methods. This was done while studying various effects such as tasks ordering and regularization.  This survey is under review with the IEEE Transactions on Pattern Analysis and Machine Intelligence (TPAMI) journal.

\section{Discussion and Future Research Directions}
In the introduction, Chapter~\ref{ch:introduction}, we listed  10 desired characteristics of continual learning, namely, constant memory, no task boundaries, online learning, forward transfer, backward transfer, problem agnostic, adaptive, no test time oracle, task revisiting and graceful forgetting. This list of  characteristics has evolved during the course of this PhD work. While the main target is to solve the stability/plasticity dilemma, meeting  some if not all the desired characteristics becomes crucial for the realization of a continual learning system. 

In Chapter~\ref{ch:expertgate}, we tolerated a linear increase in memory in order to obtain expert level performance on all tasks. There seems to be a better compromise that can be achieved. For example, instead of learning a model per task, the initial model can be divided into various modules, each specialized in a different task. A possible solution is to identify the important parameters for a given problem and isolate them when learning to solve different problems. The sequentially learned gates would be deployed to forward the data through the model in a similar manner to~\cite{DBLP:journals/corr/ShazeerMMDLHD17}. Here, we intentionally state that  specialization should apply to a ``problem'' indicating a group of similar tasks.   Within each of these modules, identified for a given problem, consolidation mechanisms as what we proposed in Chapter~\ref{ch:MAS} can be imposed to ensure that the newly encoded knowledge doesn't catastrophically interfere with what has previously been acquired. While catastrophic forgetting must be overcome, graceful forgetting should be allowed to insure the adaptivity of the system.   

It seems that complex biological systems have hierarchical organization~\cite{10.3389/neuro.11.037.2009}, or the so called modules within modules suggesting that neural modular composition might be necessary for achieving real continual learning of different tasks and problems. 
An attractive  approach that can be coupled with continual learning is meta learning. A possible direction is that besides the parameters being trained on the received tasks,  a meta model would  control the modularization  of the information to reduce the interference and increase the sharing. This is trained via  a  meta objective optimizing the forward/backward transfer gained  during the training cycle.
 
Another important aspect is the deployment of memory in a continual learning system. Achieving online continual learning involves continuous and fast adaptation in addition to a change of previous beliefs. This can't be realized without replaying previous memories. In fact, there seems to be  evidence for experience replay  between  hippocampus and neocortex in order to accommodate the recent events while preserving the old memories~\cite{honey2017switching}.
This can be achieved in continual learning by relying on a buffer sampled from the previous history as we proposed in Chapter~\ref{ch:reh_ocl} or via generative modelling. A generator can be trained to capture the data generating distribution and acts as connectionist  memory~\cite{ramapuram2017lifelong, lavda2018continual}. However, generative modelling is extremely hard to train online especially on a changing data distribution. This leaves big room for future research.

Overall, we believe that the contributions of this PhD  target different aspects of continual learning and  further investigation on how to combine the proposed techniques is necessary to move a step forward towards solving the continual learning problem.
\cleardoublepage

%




\backmatter

\includebibliography
\bibliographystyle{acm}
\bibliography{allpapers}

\includecv{curriculum}

\includepublications{publications}

\makebackcoverXII

\end{document}